\documentclass{article}
\usepackage{spconf,amsmath,graphicx}

\def\wplus{$\mathcal{W}^+$ }
\def\wstar{$\mathcal{W}^{\star}$ }
\def\wplussp{$\mathcal{W}^+$}
\def\wstarsp{$\mathcal{W}^{\star}$}

\usepackage{svg}
\usepackage{amsmath}
\usepackage{bm}
\usepackage{caption}
\usepackage{subcaption}
\usepackage{mwe}
\usepackage{graphicx}
\usepackage{rotating}
\usepackage{multirow}
\usepackage{array}
\usepackage{booktabs}
\usepackage{pifont}%

\usepackage{xcolor}
\usepackage{xparse}

\usepackage[normalem]{ulem}
\usepackage{ifthen}
\newboolean{final}
\setboolean{final}{false}

\newcommand{\FontFormatter}[2]{%
    \textcolor[rgb]{#1}{\textbf{#2}}%
}

\NewDocumentCommand{\TextFormatter}{ommm}{%
    \FontFormatter{#4}{%
        \IfNoValueTF{#1}{%
            $\scriptstyle \overset{#3}{++}$ #2%
        }{%
            \sout{#1} $\scriptstyle \xrightarrow{#3}$ #2%
        }%
    }%
}

\NewDocumentCommand{\TextFormatterFinal}{ommm}{#2}

\NewDocumentCommand{\pierre}{om}{%
    \TextFormatter[#1]{#2}{Pierre}{0.5, 0.0, 0.0}%
}

\NewDocumentCommand{\ms}{om}{%
    \TextFormatter[#1]{#2}{ }{0.0, 0.5, 0.0}%
}

\NewDocumentCommand{\bharath}{om}{%
\TextFormatter[#1]{#2}{Bharath}{0.0, 0.0, 1.0}%
}

\NewDocumentCommand{\xu}{om}{%
\TextFormatter[#1]{#2}{}{0.5, 0.0, 1.0}%
}

\usepackage{titlesec} %
\titlespacing*{\subsection}{0pt}{0.5\baselineskip}{\baselineskip}
\titlespacing*{\section}{0pt}{0.5\baselineskip}{\baselineskip}

\def \figwidth {0.14\textwidth}

\title{SEMANTIC UNFOLDING OF STYLEGAN LATENT SPACE}
\name{Mustafa Shukor, Xu Yao, Bharath Bushan Damodaran, Pierre Hellier}
\address{InterDigital, Inc., France}
\begin{document}
\maketitle
\begin{abstract}
Generative adversarial networks (GANs) have proven to be surprisingly efficient for image editing by inverting and manipulating the latent code corresponding to an input real image. This editing property emerges from the disentangled nature of the latent space. In this paper, we identify that the facial attribute disentanglement is not optimal, thus facial editing relying on linear attribute separation is flawed. We thus propose to improve semantic disentanglement with supervision. Our method consists in learning a proxy latent representation using normalizing flows, and we show that this leads to a more efficient space for face image editing.
\end{abstract}
\section{Introduction}
\begin{figure}[htbp]
     \centering
     \includegraphics[width=0.4\textwidth]{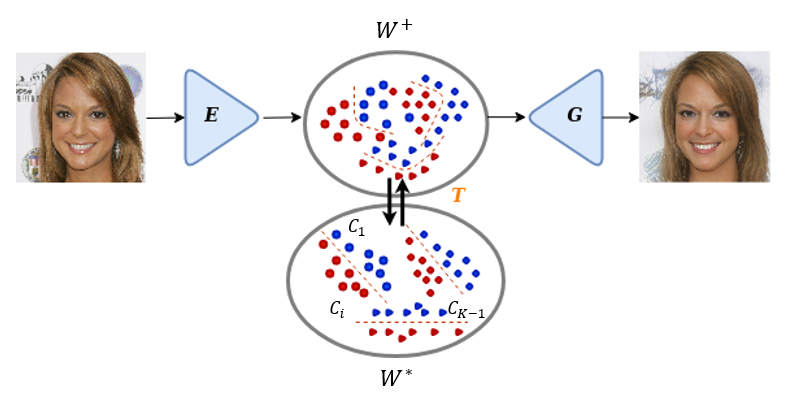}
    \caption{Illustration of our proposed approach. $E$ and $G$ are the StyleGAN2 encoder and generator, while $C=\{C_i\}_{i=0 \dots K-1}$ is a set of facial attribute classifiers. In this paper, only the transformation $T$ is trained, parametrized as a normalizing flow. Compared to the original StyleGAN2 latent space \wplus, our learned proxy latent space \wstar leads to a better attribute disentanglement and linear separability with hyperplanes. Hence, facial attribute editing is more efficient in the new learned representation \wstar.}
    \label{fig:bmvc2021}
\end{figure}

GANs \cite{gan} have shown tremendous success in generating high quality realistic images that are indistinguishable from real ones. Yet, several open problems regarding these generative models still exist, namely: image generation control, latent space understanding and attributes disentanglement. All these features are important for the generation and editing of high quality images.
Recently, many improvements have been proposed to the original GAN architecture, which led to unprecedented image quality. In particular, the state-of-the-art method StyleGAN \cite{stylegan, styleganimporveing} has been improved, leading to image editing methods using the latent representation. Such methods build upon the inversion of StyleGAN (i.e., retrieving the latent representation that explains the observed data) and the manipulation of the latent code for semantic image editing. 
Specifically, recent state of the art methods \cite{shen2020interpreting, shen2020interfacegan} assume that the attributes are disentangled and can be separated by hyperplanes, which enables interpolation and attributes manipulation. InterFaceGAN \cite{shen2020interfacegan} computes an editing direction for each facial attribute, orthogonal to the linear classification boundary for this attribute. We claim that the hyperplane classification boundary assumption is inaccurate. In addition, the attributes are not perfectly disentangled, which can further explain why these approaches do not lead to perfect attribute manipulation.
To solve this problem, the first solution is retraining the GAN with explicit constraints such as attribute disentanglement. However, it is known that the training of GANs is hard and computationally expensive. We propose in this paper an alternative approach without retraining. Specifically, we will focus on the aforementioned properties and learn a bijective transformation (\emph{i.e.}, Normalizing Flows) from the original latent space (\emph{i.e.}, \wplussp) to a new proxy latent space (\wstarsp). In \wstar,  the facial attributes are linearly separable and disentangled. The choice of a bijective transformation allows to benefit from the pretrained StyleGAN2 generative capabilities. Figure \ref{fig:bmvc2021} illustrates the proposed approach. The main benefit is to take off-the-shelf GAN and incorporate additional supervision, considering the problem to solve. In our case, since we focus on facial image editing, we enforce explicitly facial attribute disentanglement. 
\textbf{Our contributions are the following}; 1) We propose to learn a derived latent representation where supervision is used to explicitly disentangle  the facial attributes. 2) We propose to learn this proxy latent representation using normalizing flows, which can be applied to any pretrained GAN while preserving the generative capability of the original GAN. 3) We show experimentally that the desired properties are indeed enforced in this new latent representation, leading to a more efficient image manipulation.
The rest of the paper is organised as follows: section~\ref{sec:related works} presents related works on GANs. Our proposed method is detailed in section~\ref{sec:method} and section~\ref{sec:expe} presents the experimental results. Finally, conclusions are drawn in section~\ref{sec:conclusion}.
\label{sec:intro}
\section{Related work}

\noindent \textbf{GANs generation and inversion} \quad
GANs \cite{gan} are one type of generative models that are trained adversarially to generate complex data distributions. Several improvements have been proposed to improve GANs architecture \cite{radford2015unsupervised}, loss function \cite{wgan} and its training \cite{karras2017progressive}. Recently, StyleGAN \cite{stylegan, styleganimporveing} was introduced as the state of the art in high-resolution image generation, especially for human faces and allows better control of the generation process mainly due to its disentangled latent space.\\
There are different approaches to project/invert the image in the latent space of GANs. Optimization based approaches give the best reconstruction, although, it takes several minutes for each image to project \cite{creswell2018inverting, abdal2019image2stylegan}. To overcome this limitation, encoding based approaches seem an alternative solution \cite{creswell2018inverting, guan2020collaborative, ghfeatxu2020generative, wei2021simplebase}, where an encoder is trained to project the image in the latent space. These approaches give close performance to the optimization based ones while being much faster. The last approaches are hybrid, which combine the two previous ones \cite{zhu2020domain}.

\noindent \textbf{Image Editing} \quad Several methods have been proposed to leverage StyleGAN for image editing once a real image has been inverted in the latent representation of StyleGAN. \cite{collins2020editing, abdal2020image2stylegan++} propose a local editing framework based on interpolation in the latent space. GANSpace \cite{ganspace} applies a PCA in the latent space to find the editing directions. Others learn non-linear \cite{yao2021latent, abdal2020styleflow} transformations in the latent space to edit the image. In \cite{shen2020interfacegan, shen2020interpreting}, they edit the images following the normal to the hyperplanes that separate the attributes regions.

\label{sec:related works}
\section{Method}
In this section, we explain how to learn a proxy latent space (named \wstarsp) where the attributes are disentangled and linearly separable. In addition, we detail how other properties, useful for image editing, can be satisfied. 

Let us assume that a pretrained StyleGAN generator $G$ exists, that considers a latent code $w \in \mathcal{W}^{+}$ and generates a high resolution image $I$ (i.e., 1024 x 1024). To process real images, let us also assume that there exists a pretrained encoder $E$ that embeds any image in \wplus such that $G(E(I)) \simeq I$. The latent space of StyleGAN was trained for one main purpose: the generated images should be indistinguishable from natural images. Hence, no additional supervision on facial attribute disentanglement has been used during training. 

The main objective of our work is to construct a proxy latent representation, where additional properties useful for image editing can be enforced. To do so, we consider the state-of-the-art method InterfaceGAN \cite{shen2020interfacegan} that edits an image by mapping the latent code in a direction orthogonal to the classification hyperplane of the considered attribute. Hence, to improve this method, we aim at enforcing linear separability of facial attributes, maximization of attributes classification margins and regularization of other attributes.

To do so, we construct a proxy representation $w^{\star}$ by learning a bijective transformation $T: \mathcal{W}^{+} \rightarrow \mathcal{W}^{\star}$ that maps a latent code $w \in \mathcal{W}^{+}$ to $w^{\star} \in \mathcal{W}^{\star}$. $T$ is a Normalizing Flows (NFs) model and can be inverted explicitly.  Although, the transformation $T$ is modelled as a NF, it is noted that our work only requires the bijectivity, as such, we did not impose the prior distribution in the proxy latent space as we are not interested in the density estimation. The following section describes how the transformation $T$ is learned.

\noindent \textbf{Linear Separation of the Facial Attributes} \quad Our main objective is to learn $T$ so that the linear classification is optimal for each attribute. Let us consider a pre-trained set of $K$ attribute classifier $C_i:$ \wstar $\rightarrow \{0, 1\}, i \in [0,\dots,K-1]$, where $K$ is the number of facial attributes labeled in the image dataset. To maximize linear separability, the objective is to minimize for each sample $w$ the following loss:
\begin{equation}
\label{eq:att}
    \mathcal{L}_a = - \sum_{i=0}^{K-1} y_{i}\log(C_i(T(w)) + (1-y_i)\log(1-C_i(T(w))),
\end{equation}
Where $y_{i} \in \{0, 1\}$ is the label of the sample $w$ corresponding to the $i^{th}$ attribute. In \eqref{eq:att}, the classifiers are fixed and only $T$ is optimized as we are interested in obtaining a linear separation for each attribute. Theoretically, any linear classifier could be used. However, since a form of linear separation already exists in \wplussp, we choose to pretrain the linear classifiers first in \wplus and fix it while optimizing for $T$. This provides some regularization so that $T$ only focuses on improving the pre-trained classifier. In addition, it helps to converge faster.
\noindent \textbf{Classification margin} \quad Since in InterfaceGAN, the editing direction is orthogonal to the hyperplane, and the editing magnitude is chosen according to the distance to the classification hyperplane, it is desirable to maximize the classification margin between the positive and negative region of each attribute. Thus we optimize the following large margin loss for each latent sample $w$:
\begin{equation}
\label{eq:large_margin_loss}
    \mathcal{L}_{lm} = \sum_{i=0}^{K-1} -  m_i \cdot |\bm{w}^T\bm{d}_i|  +  (1-m_i) \cdot |\bm{w}^T\bm{d}_i | 
\end{equation}
Where $m_i=1$ (respectively $m_i=0$) when the latent code is correctly classified for the $i^{th}$ attribute (resp. wrongly classified), and $d_i$ is the normal to the classification hyperplane of the $i^{th}$ attribute.

\noindent \textbf{Regularization for Image Editing} \quad Specifically for image editing, it is desirable to preserve all other attributes when editing a specific one. We thus force the transformation $T$ to preserve all the other attributes when mapping the latent code back to \wplus:
\begin{equation}
    \label{eq:disto_class}
    \mathcal{L}_{ap} = \sum_{i=0}^{K-1} \sum_{j \neq i} \| C_i(w) - C_i(\hat{w})  \|
\end{equation}
Where $\hat{w}$ is the edited version of $w$ mapped back to \wplus. The edit is done by moving the code $T(w) = w^*$ along the editing direction corresponding to the attribute $j$ chosen randomly for each batch.
Finally, the total loss is written as follows: $\mathcal{L}_{W^{\star}} = \mathcal{L}_a + \lambda_{lm} \mathcal{L}_{lm}+  \lambda_{ap} \mathcal{L}_{ap}$.
where $\lambda_{lm}$ and $\lambda_{ap}$ are the weights of the respective loss terms.

\label{sec:method}
\section{Experiments}
\label{sec:expe}
In this section we evaluate the properties of our proposed proxy latent space \wstarsp. First, we detail the implementation details, next we describe the quantitative metrics evaluating the degree of linear classification and disentanglement. Finally, we present the qualitative results for image editing.

\noindent \textbf{Implementation Details}
We used a pretrained StyleGAN2 ($G$) on FFHQ dataset \cite{stylegan}. The images are encoded in \wplus using a pretrained StyleGAN2 encoder ($E$) \cite{encodinginstyle} (the parameters of the generator and the encoder remain fixed in all the experiments). The latent vector dimension in \wplus and \wstar is $18\times 512$.
Celeba-HQ \cite{karras2017progressive} was used and consists of 30000 high quality images (\emph{i.e.},  $1024 \times 1024$) of faces where each image has annotation for $K=40$ attributes. A single layer MLP model for each attribute ($C_i$) is used as linear classifier which is pretrained in \wplussp. 
For the NF model, Real NVP \cite{realnvp} was used without batch normalization. Each coupling layer consists of $3$ fully connected (FC) layers for the translation function and $3$ FC for the scale one with LeakyReLU as hidden activation and Tanh as output one.
For all the experiments, Adam optimizer was used with $\beta_1=0.9$ and $\beta_2=0.999$, learning rate=$1e-4$, $\lambda_{ap}=0.1$ and the batch size=$8$.
\begin{figure*}[htbp]
\small\addtolength{\tabcolsep}{-5pt}
\renewcommand{\arraystretch}{-1}
\centering
\begin{tabular}{p{0.25cm}ccccccc}
\centering
 &
 \textbf{Original} &
 \textbf{Inverted} &
 \textbf{Makeup} &
 \textbf{Male} &
 \textbf{Mustache} &
 \textbf{Chubby} &
 \textbf{Lipstick} \\
 \begin{turn}{90} \hspace{0.5cm} \wplus\end{turn} &
 \includegraphics[width=\figwidth]{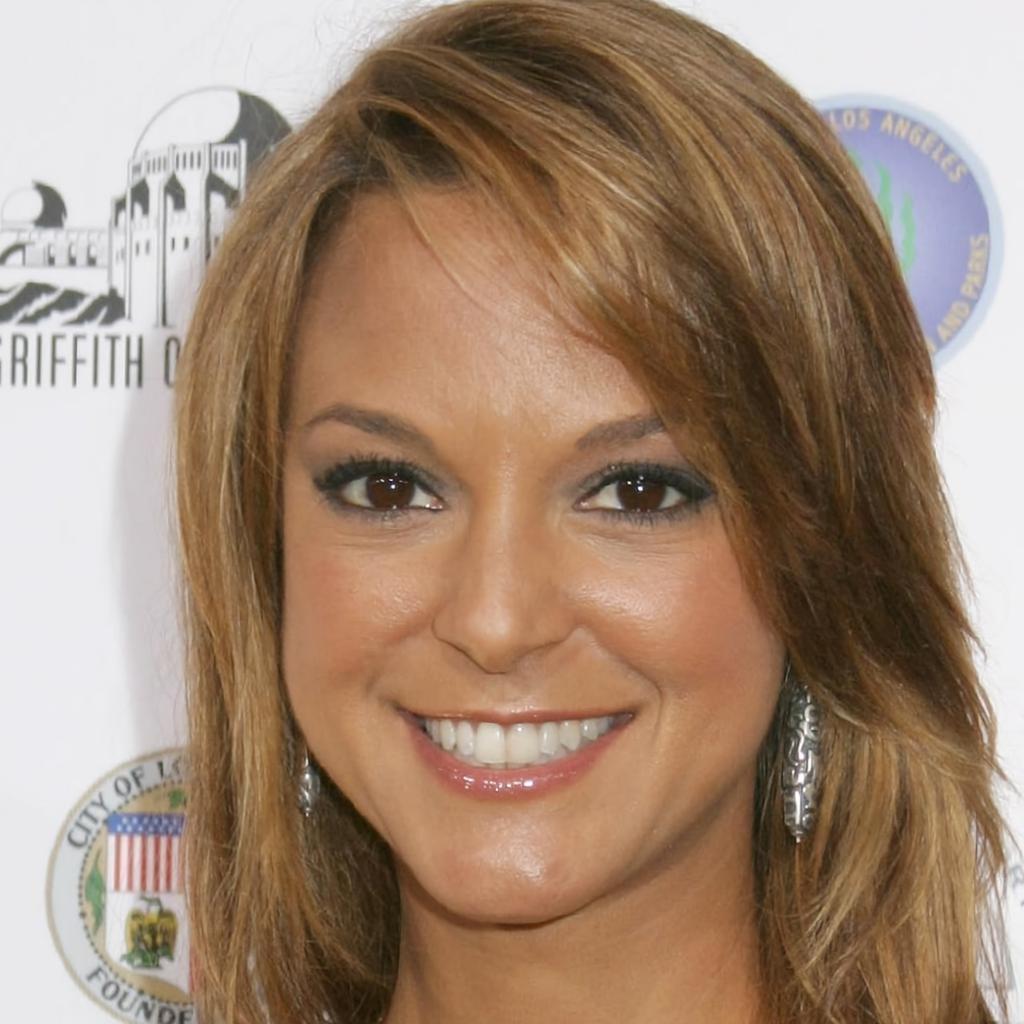} & 
 \includegraphics[width=\figwidth]{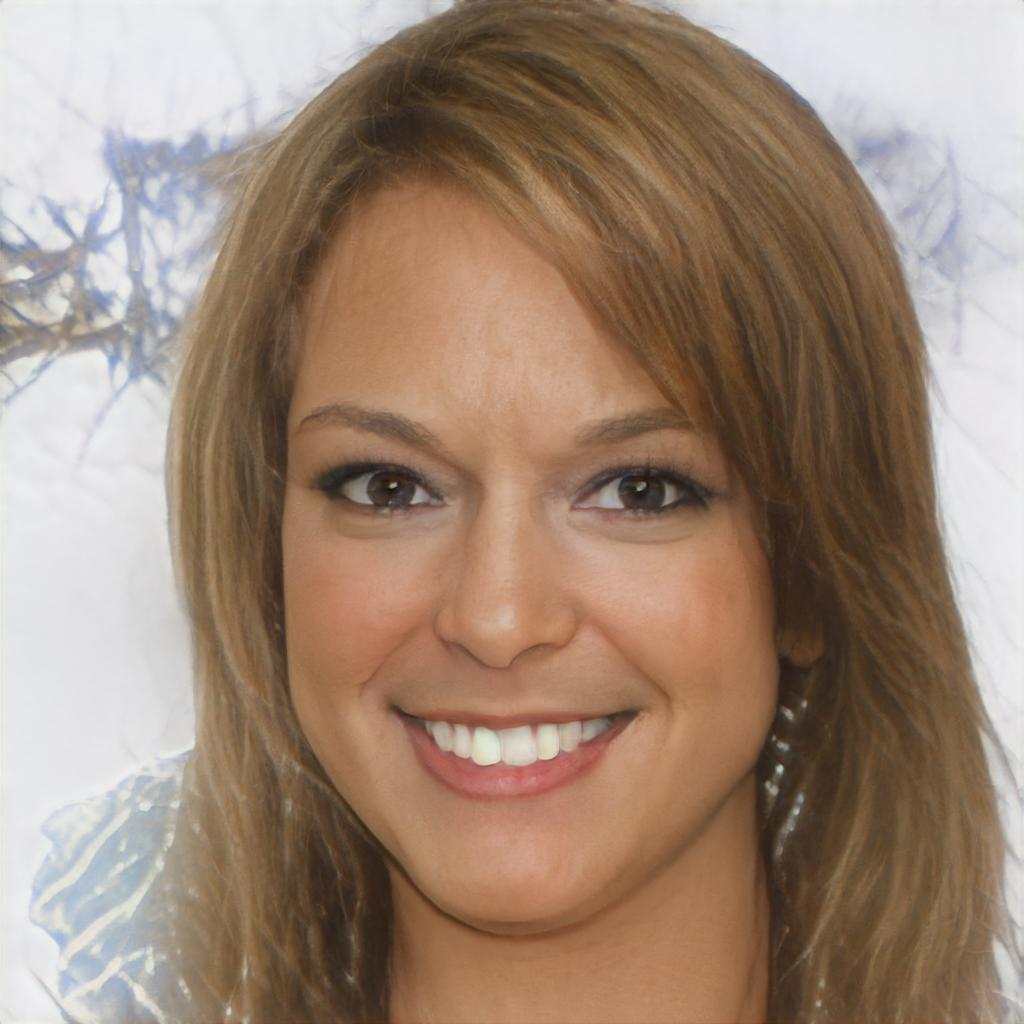} &
 \includegraphics[width=\figwidth]{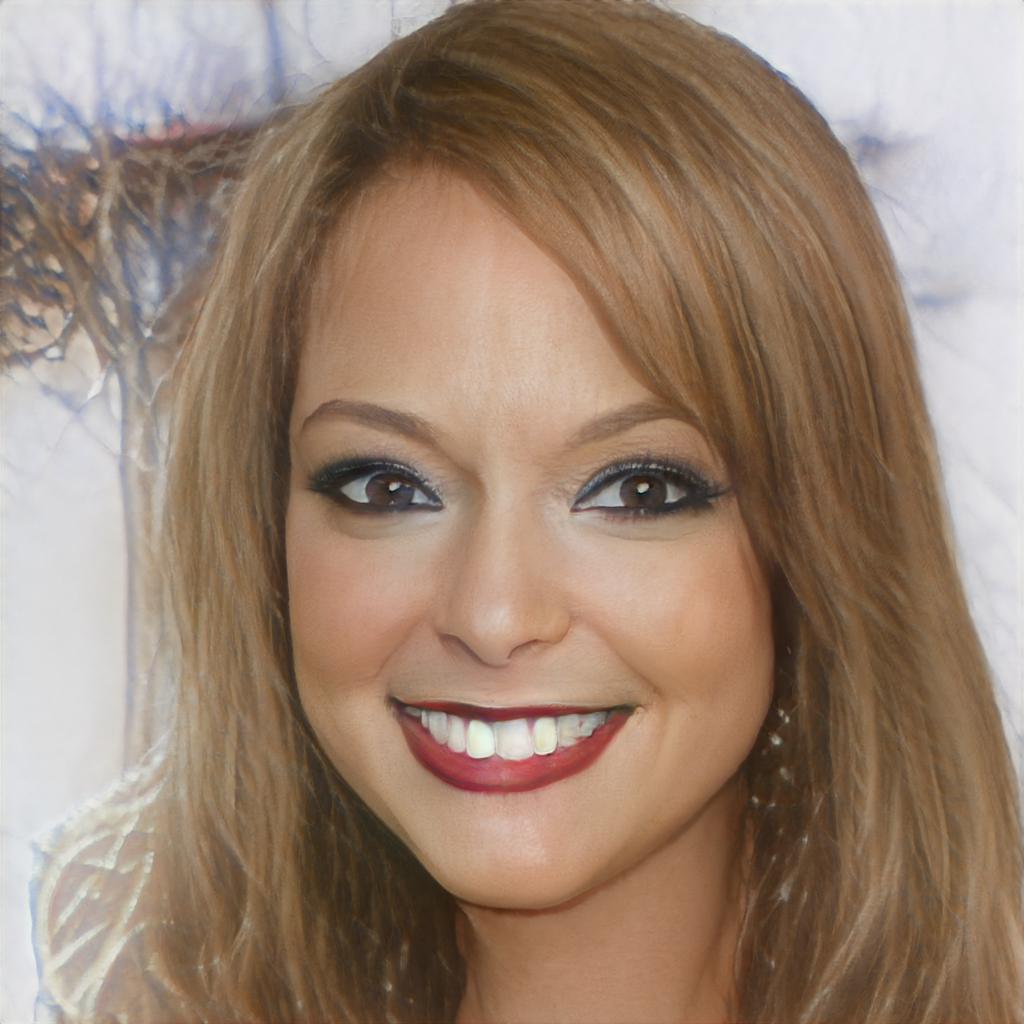} &
 \includegraphics[width=\figwidth]{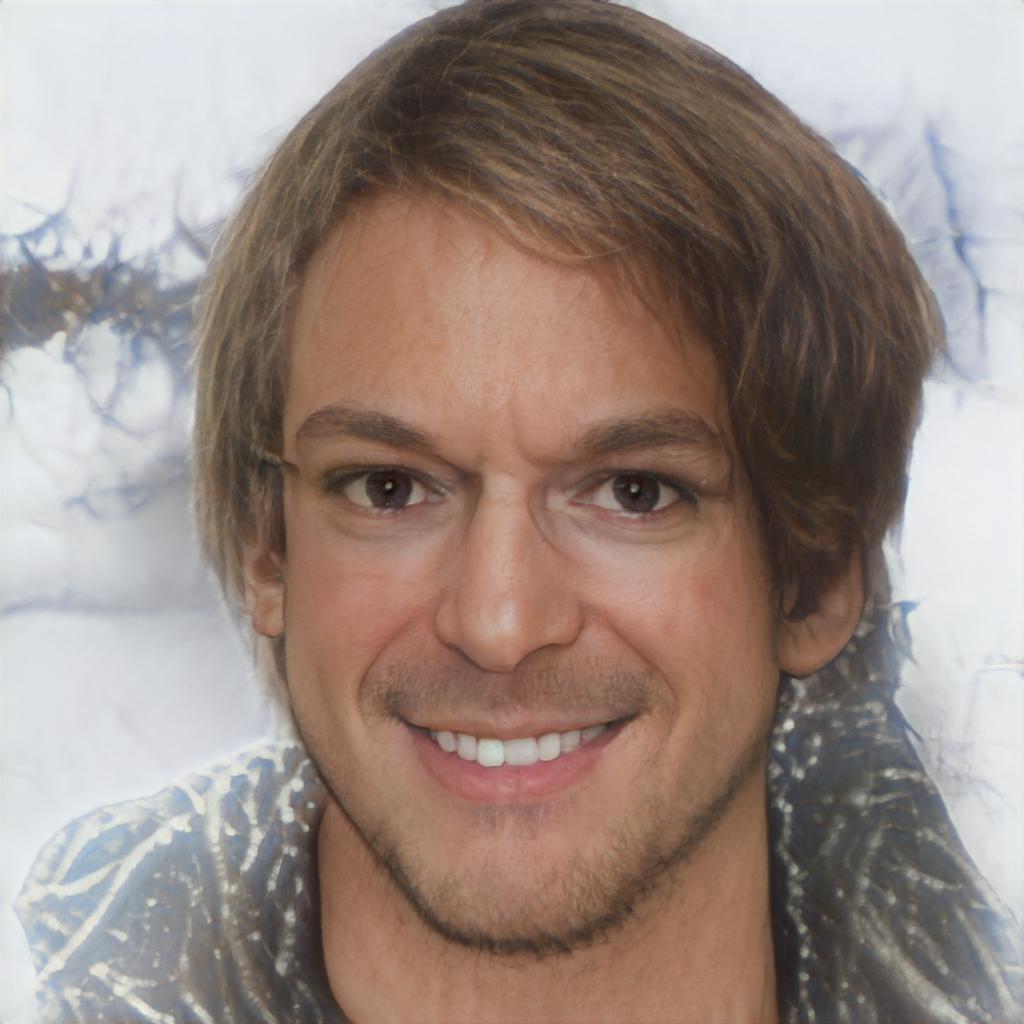} &
 \includegraphics[width=\figwidth]{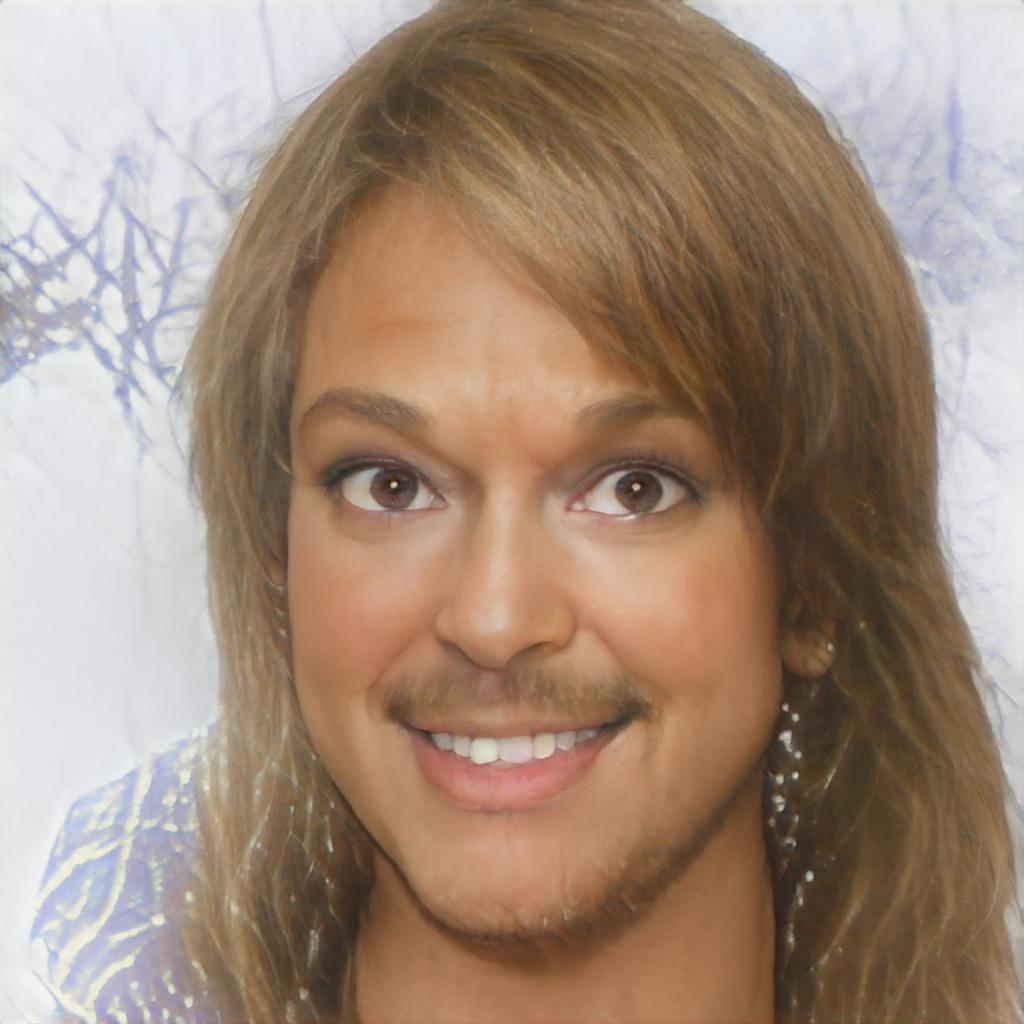} &
 \includegraphics[width=\figwidth]{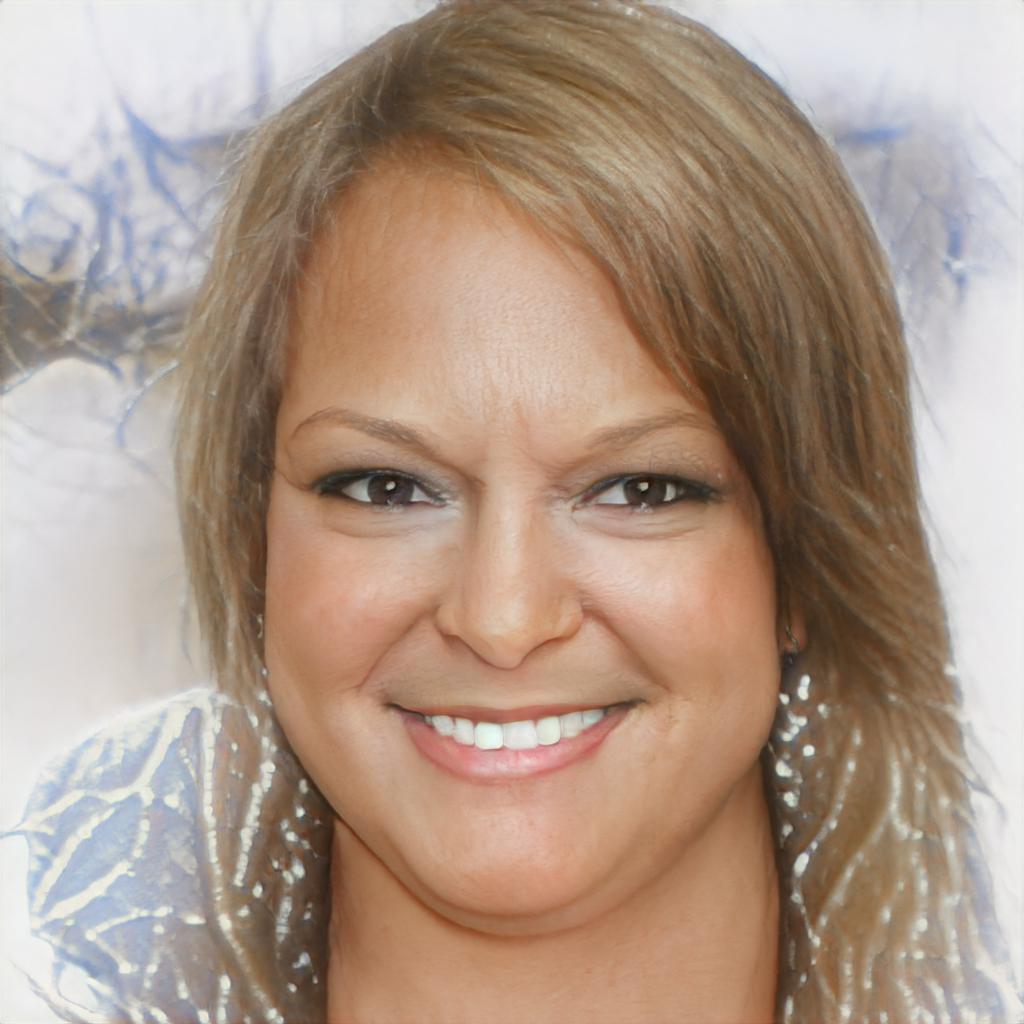} &
 \includegraphics[width=\figwidth]{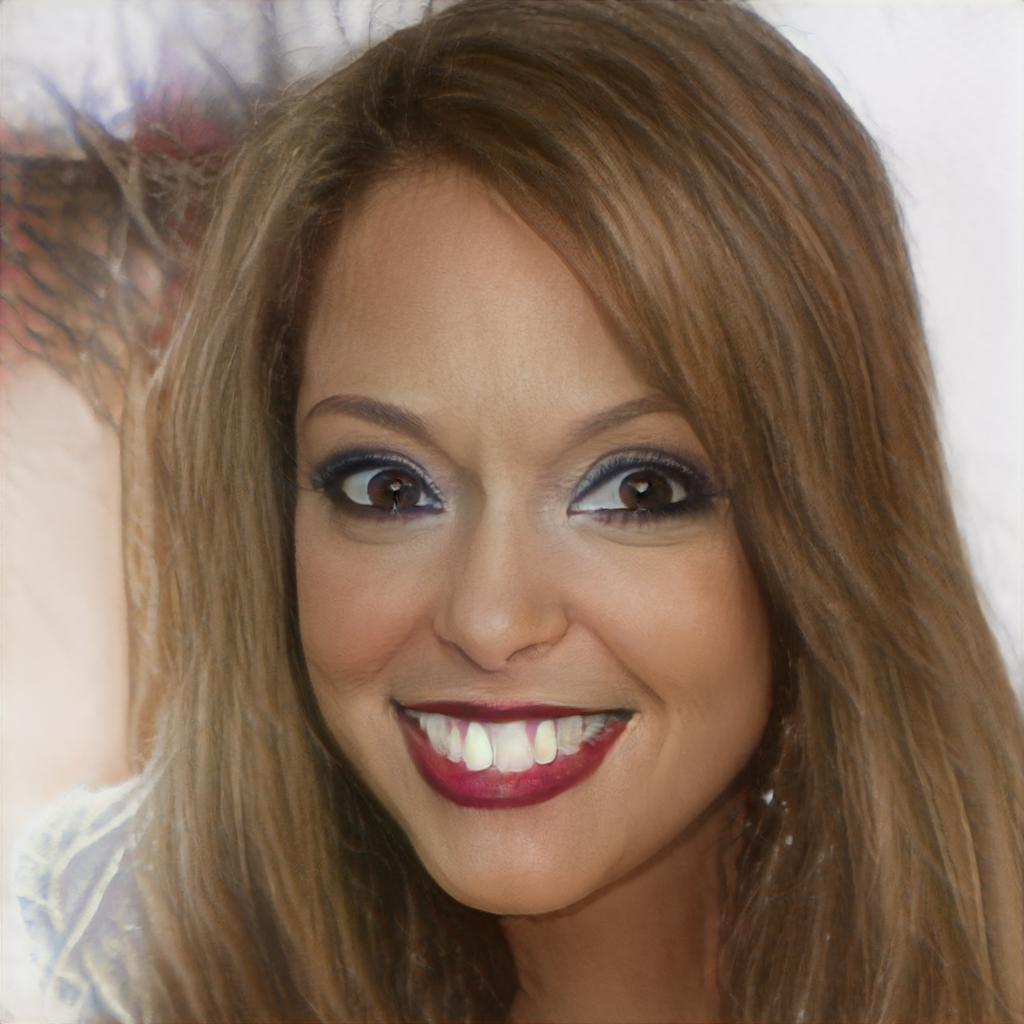} \\
 \begin{turn}{90} \hspace{0.5cm} \wstar (ours) \end{turn} &
 \includegraphics[width=\figwidth]{images/original/06002.jpg} & 
 \includegraphics[width=\figwidth]{images/inversion/18_orig_img_1.jpg} &
 \includegraphics[width=\figwidth, ]{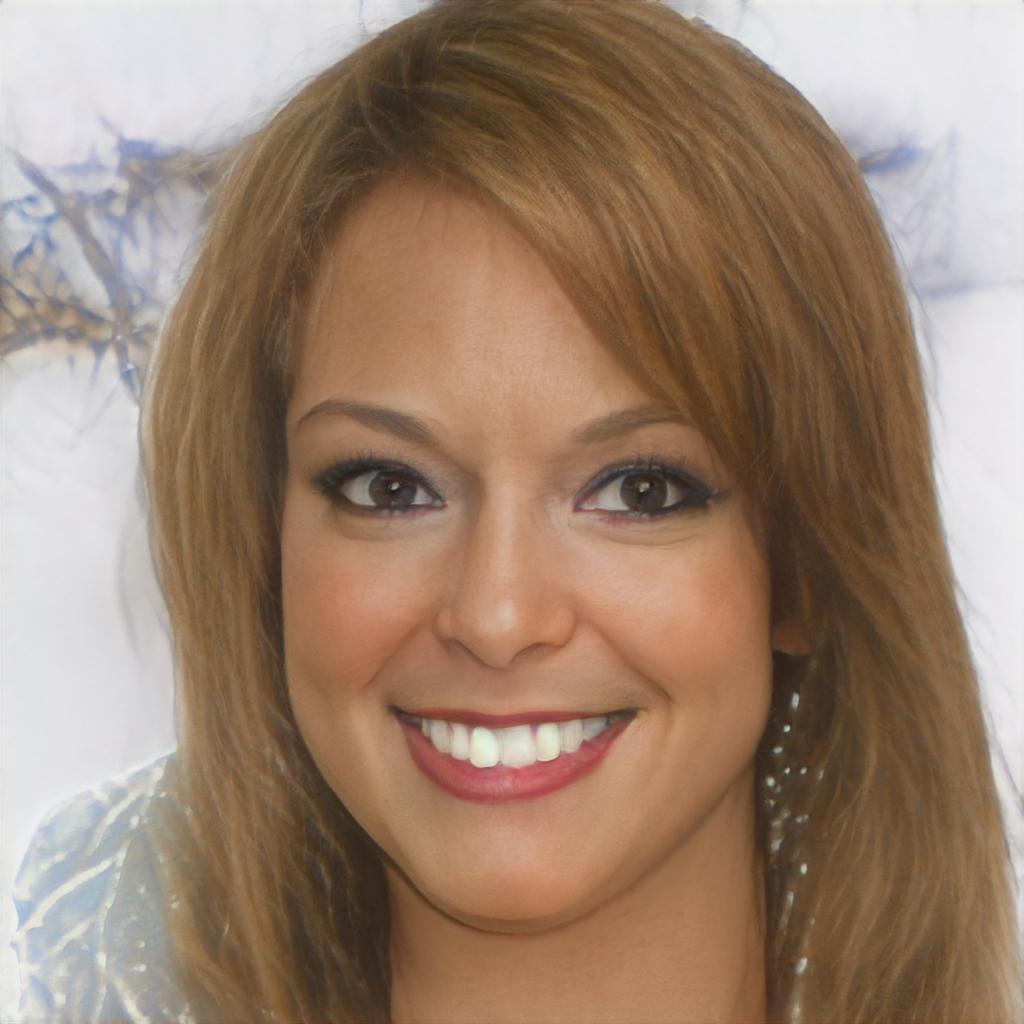} &
 \includegraphics[width=\figwidth, ]{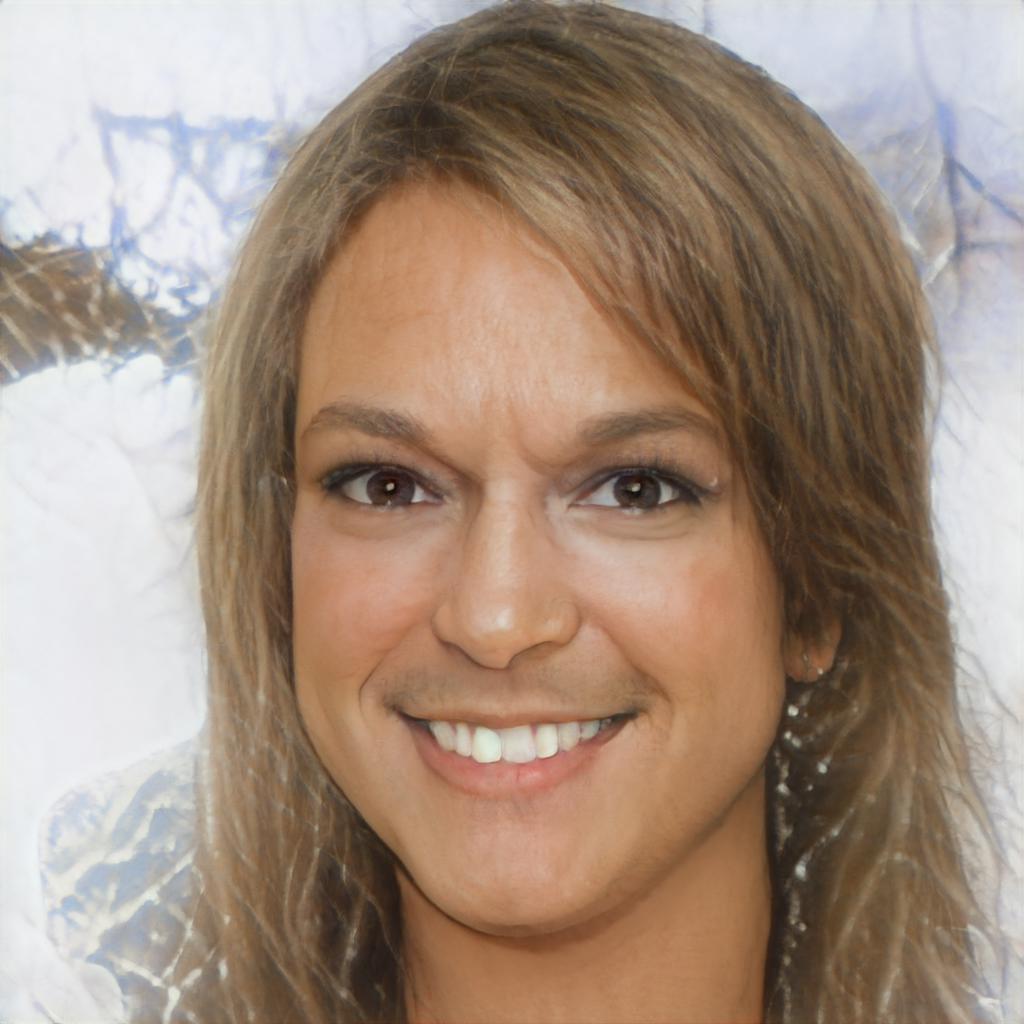} &
 \includegraphics[width=\figwidth, ]{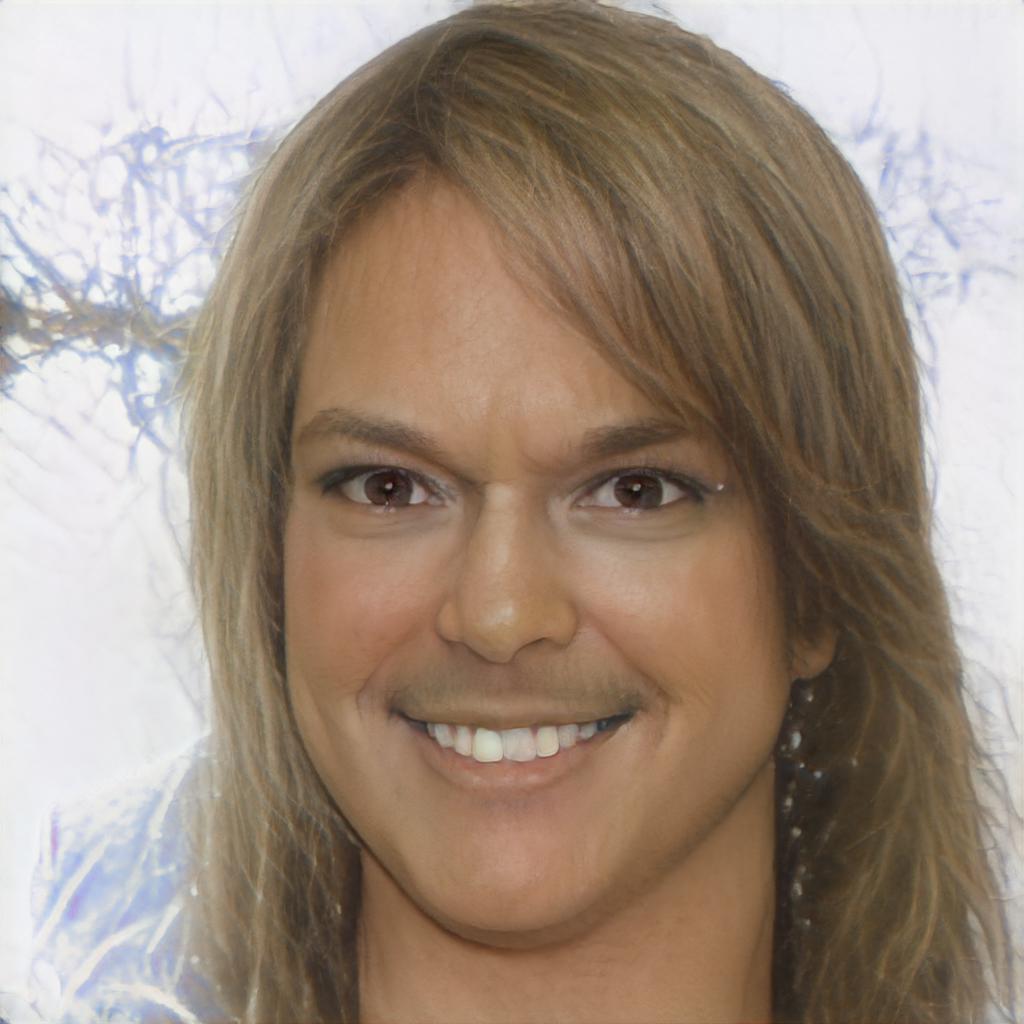} &
 \includegraphics[width=\figwidth, ]{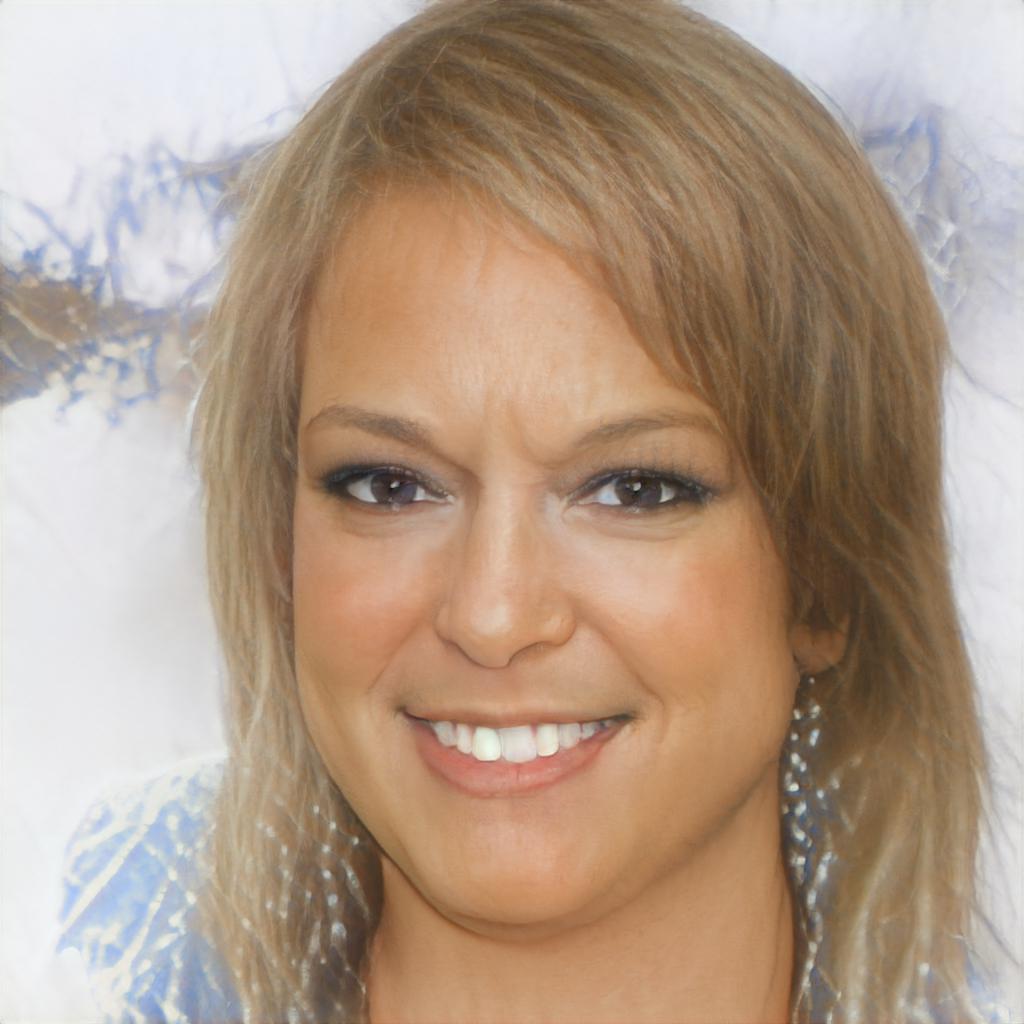} &
 \includegraphics[width=\figwidth, ]{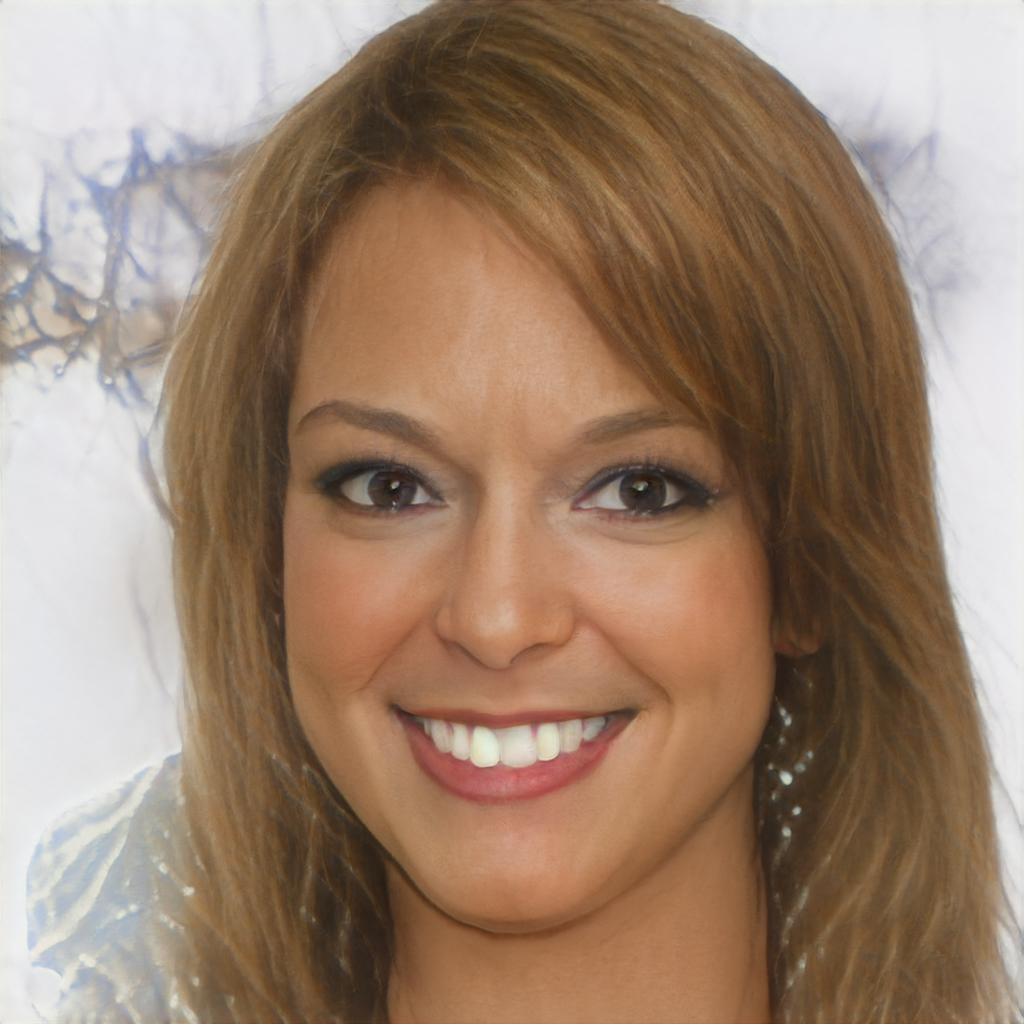} \\
  \begin{turn}{90} \hspace{0.5cm} \wplus\end{turn} &
 \includegraphics[width=\figwidth]{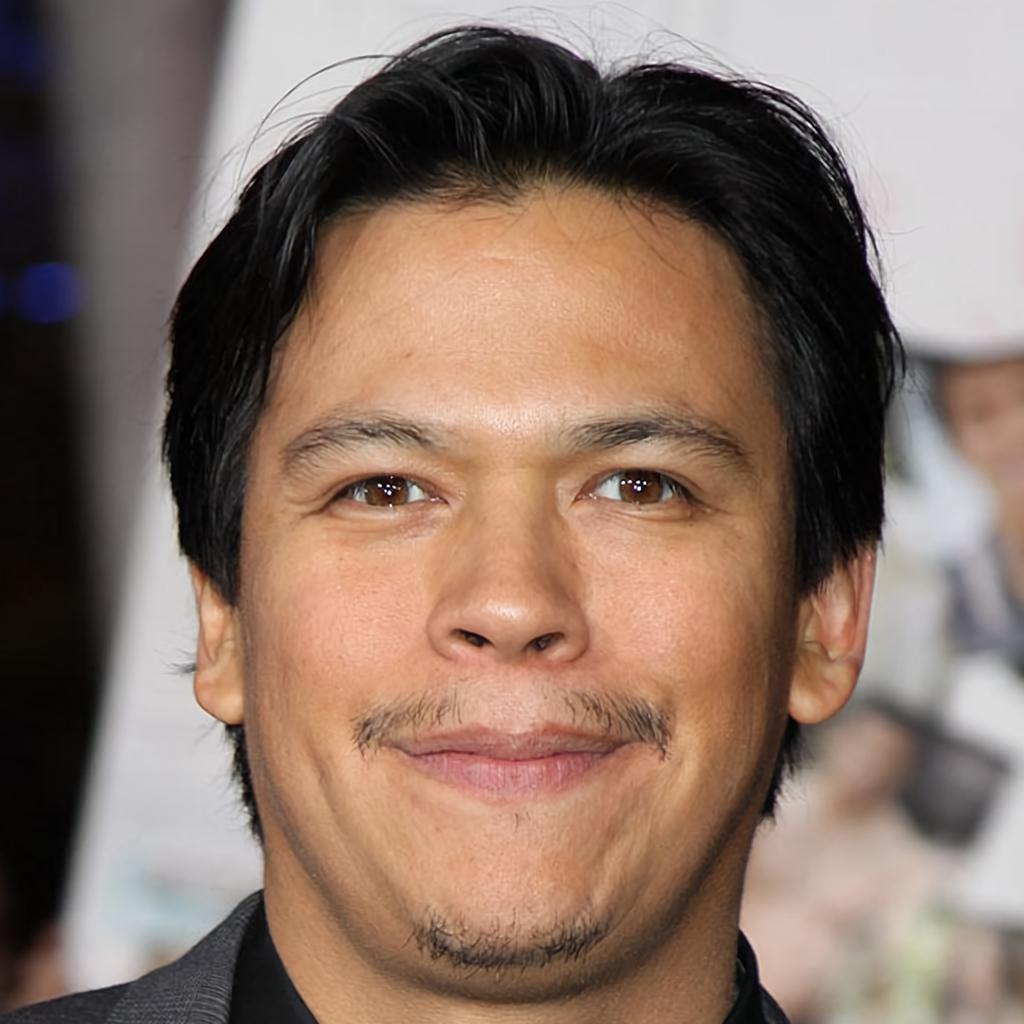} & 
 \includegraphics[width=\figwidth]{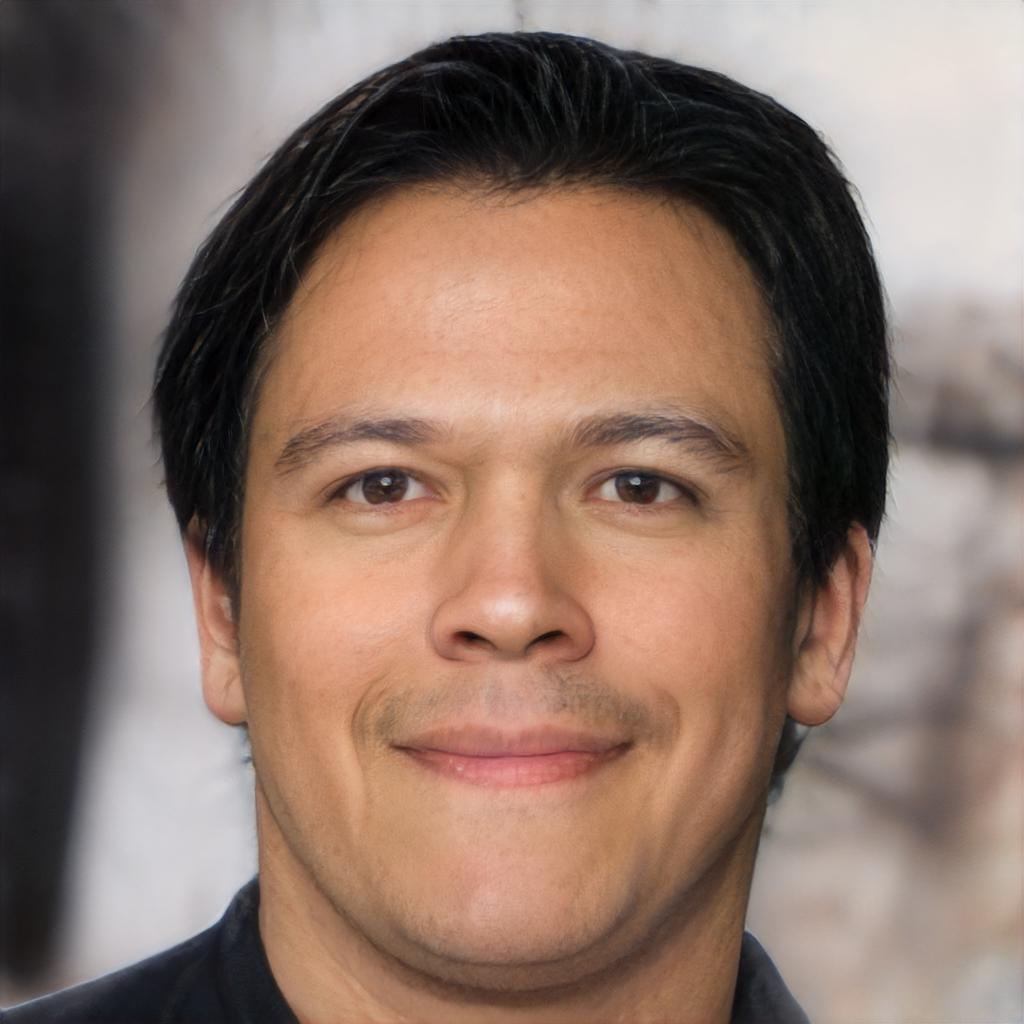} &
 \includegraphics[width=\figwidth]{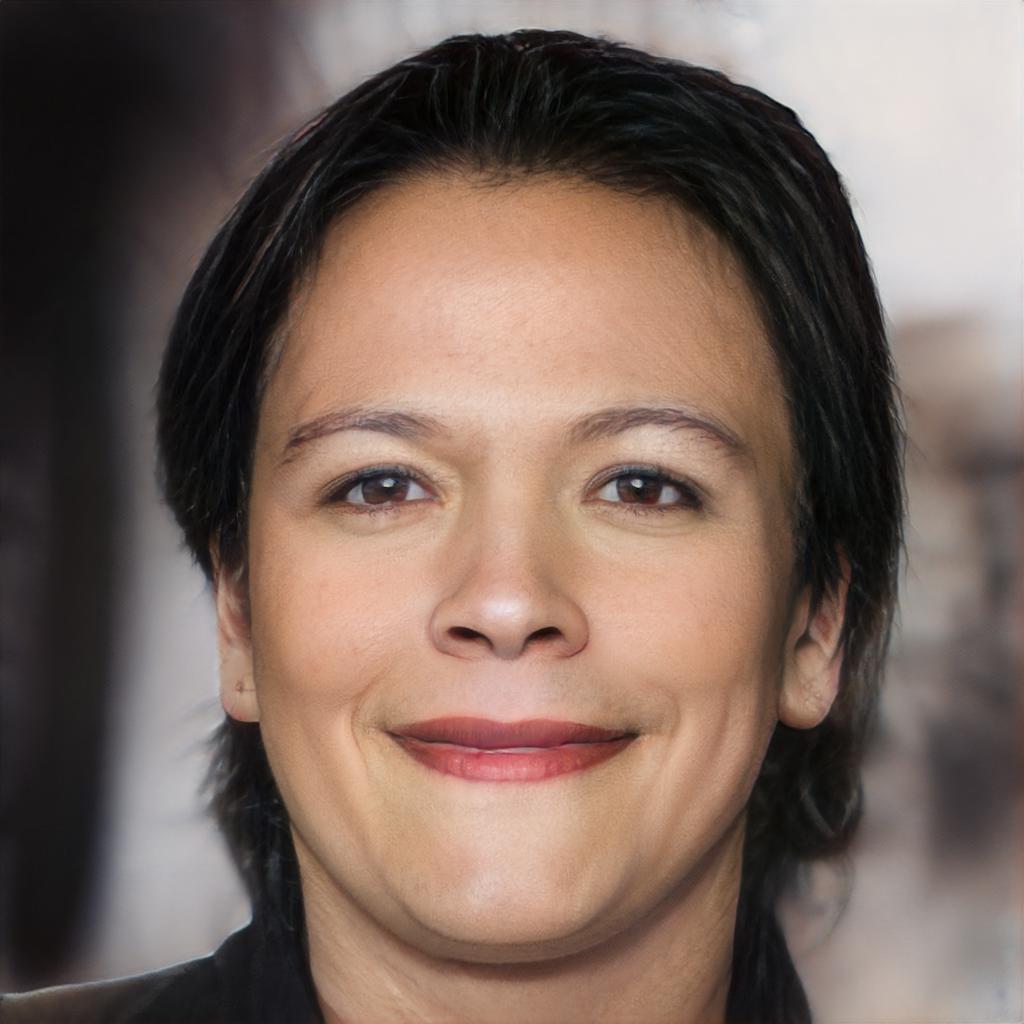} &
 \includegraphics[width=\figwidth]{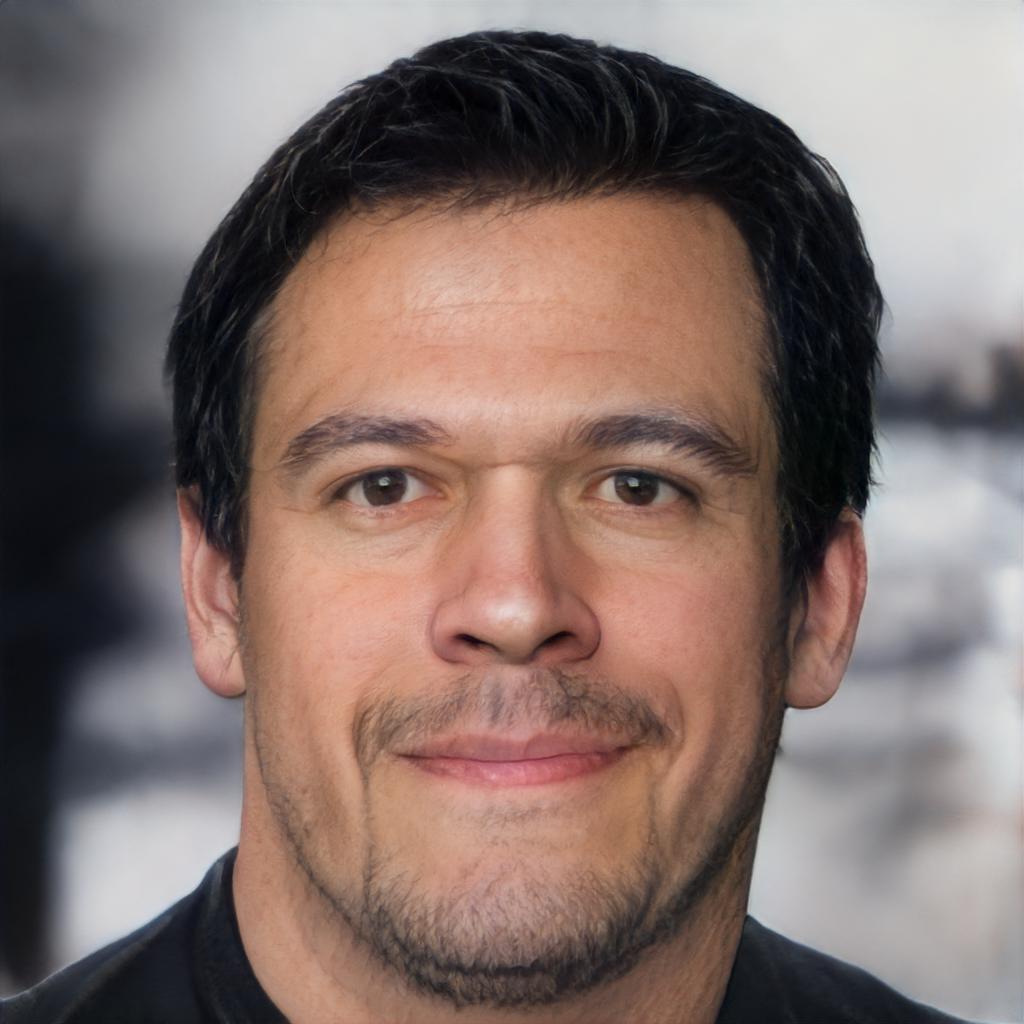} &
 \includegraphics[width=\figwidth]{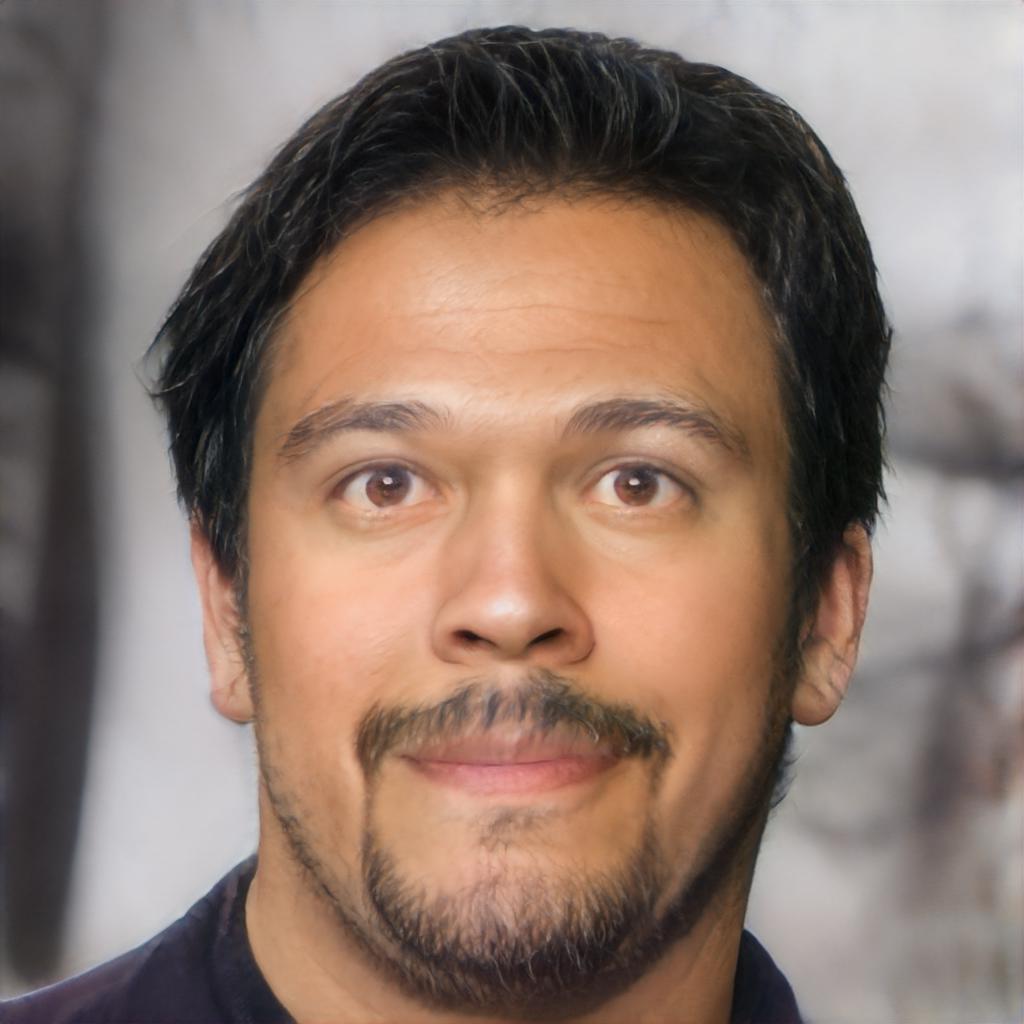} &
 \includegraphics[width=\figwidth]{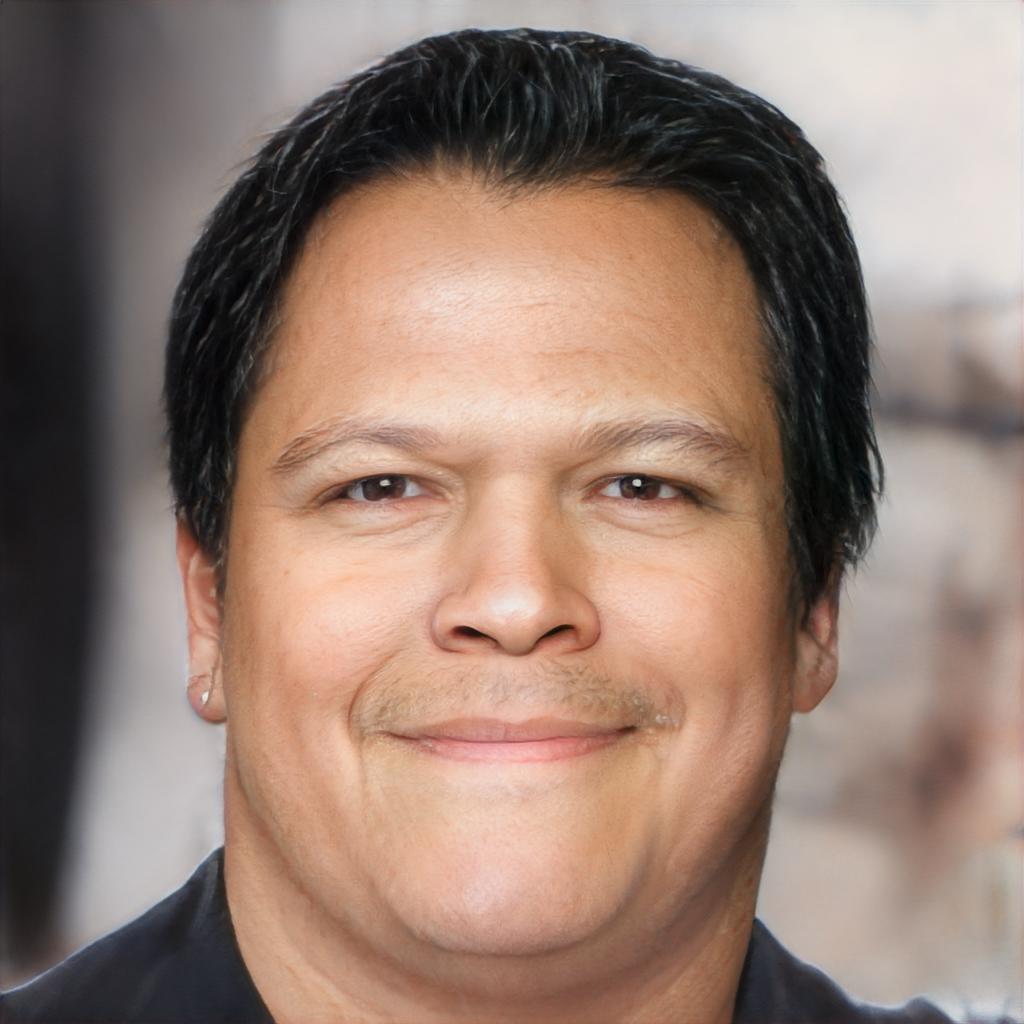} &
 \includegraphics[width=\figwidth]{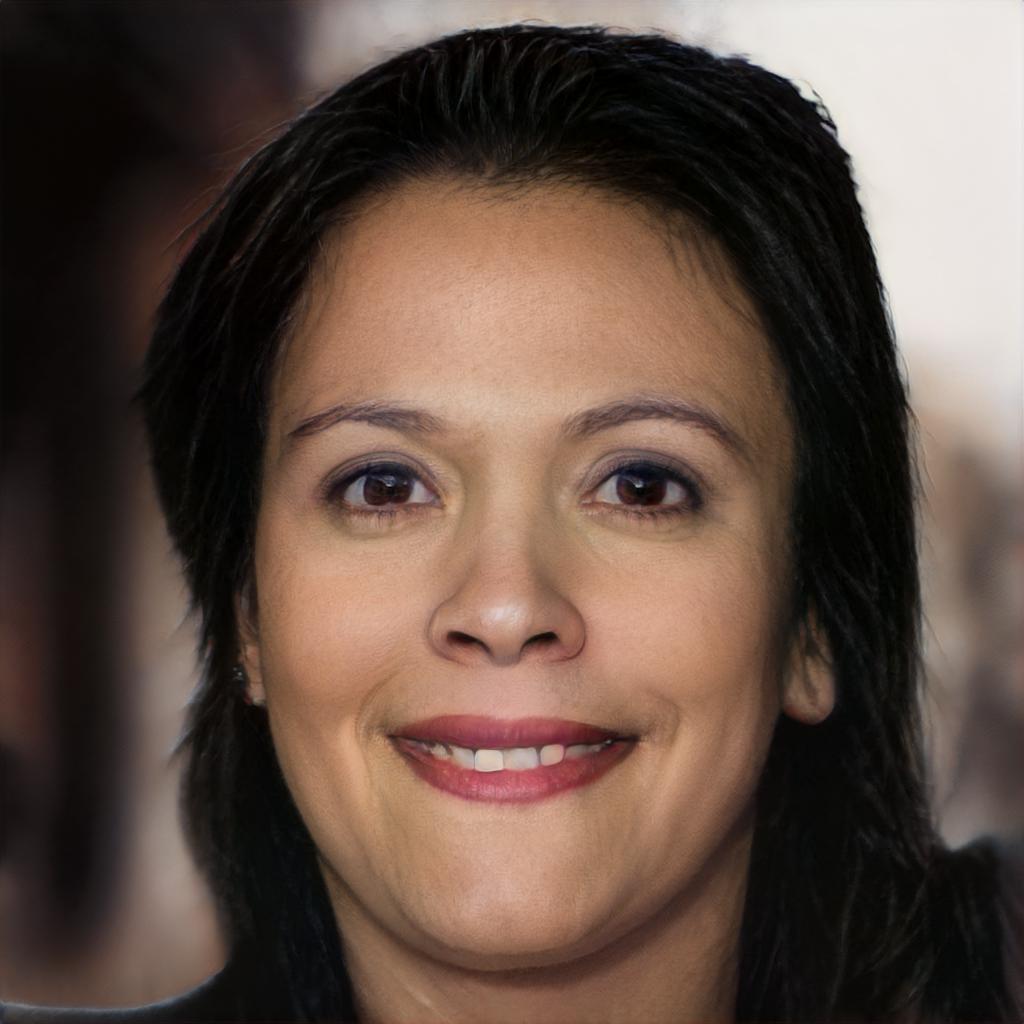} \\
  \begin{turn}{90} \hspace{0.5cm} \wstar (ours) \end{turn} &
 \includegraphics[width=\figwidth]{images/original/06004.jpg} & 
 \includegraphics[width=\figwidth]{images/inversion/18_orig_img_3.jpg} &
 \includegraphics[width=\figwidth]{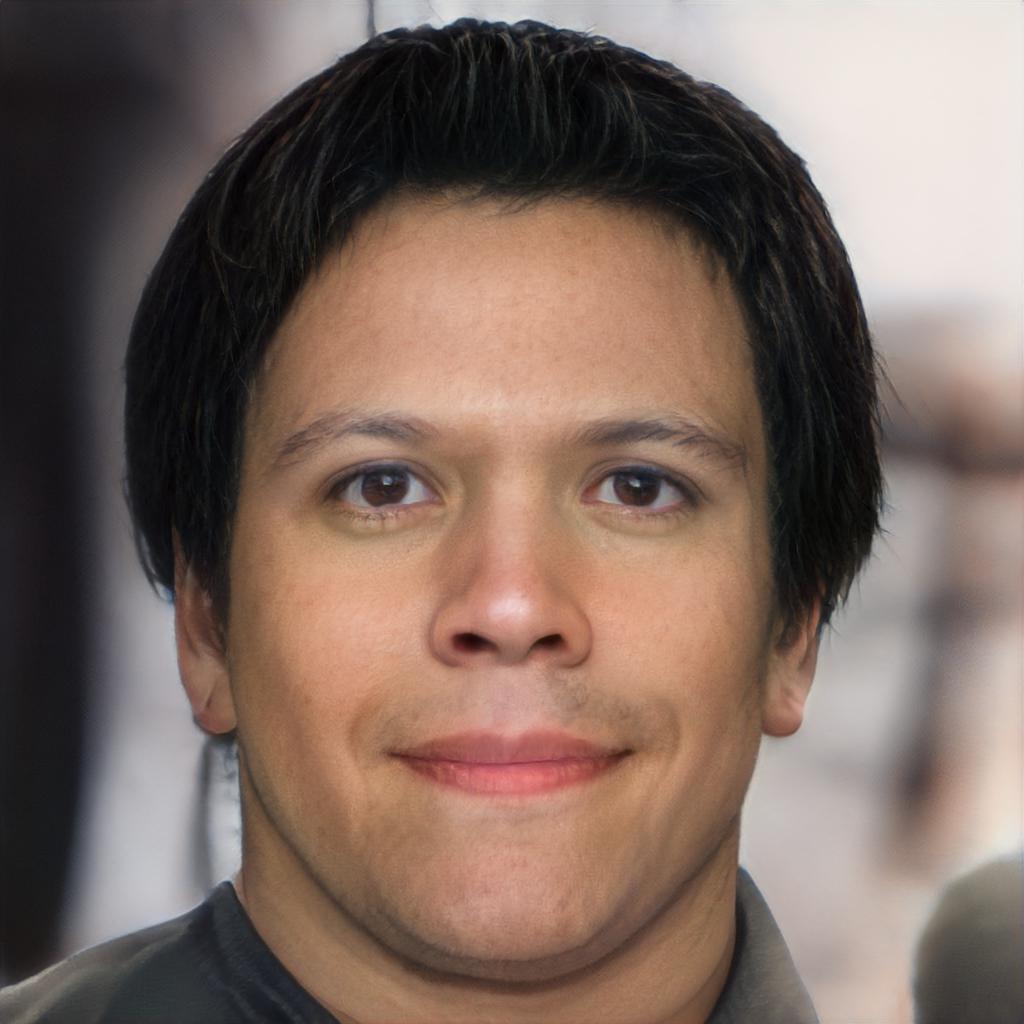} &
 \includegraphics[width=\figwidth]{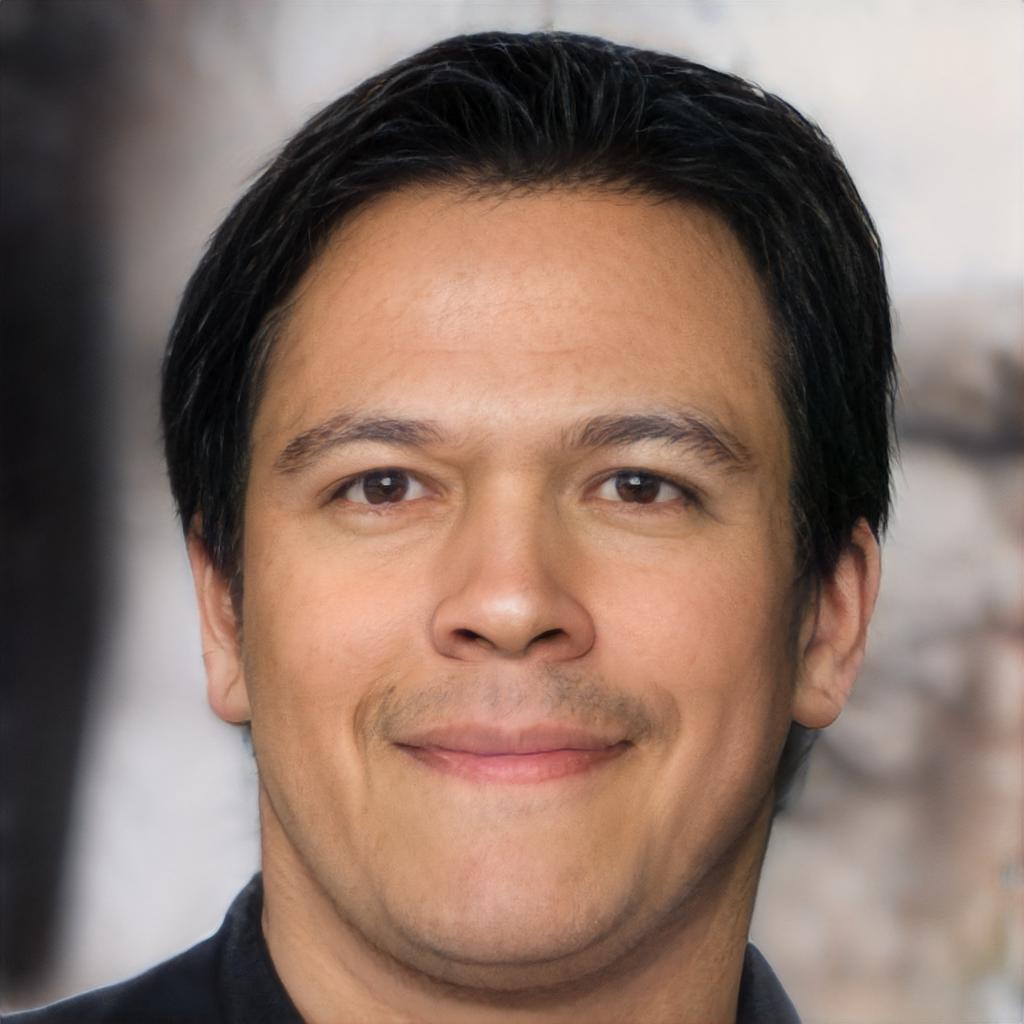} &
 \includegraphics[width=\figwidth]{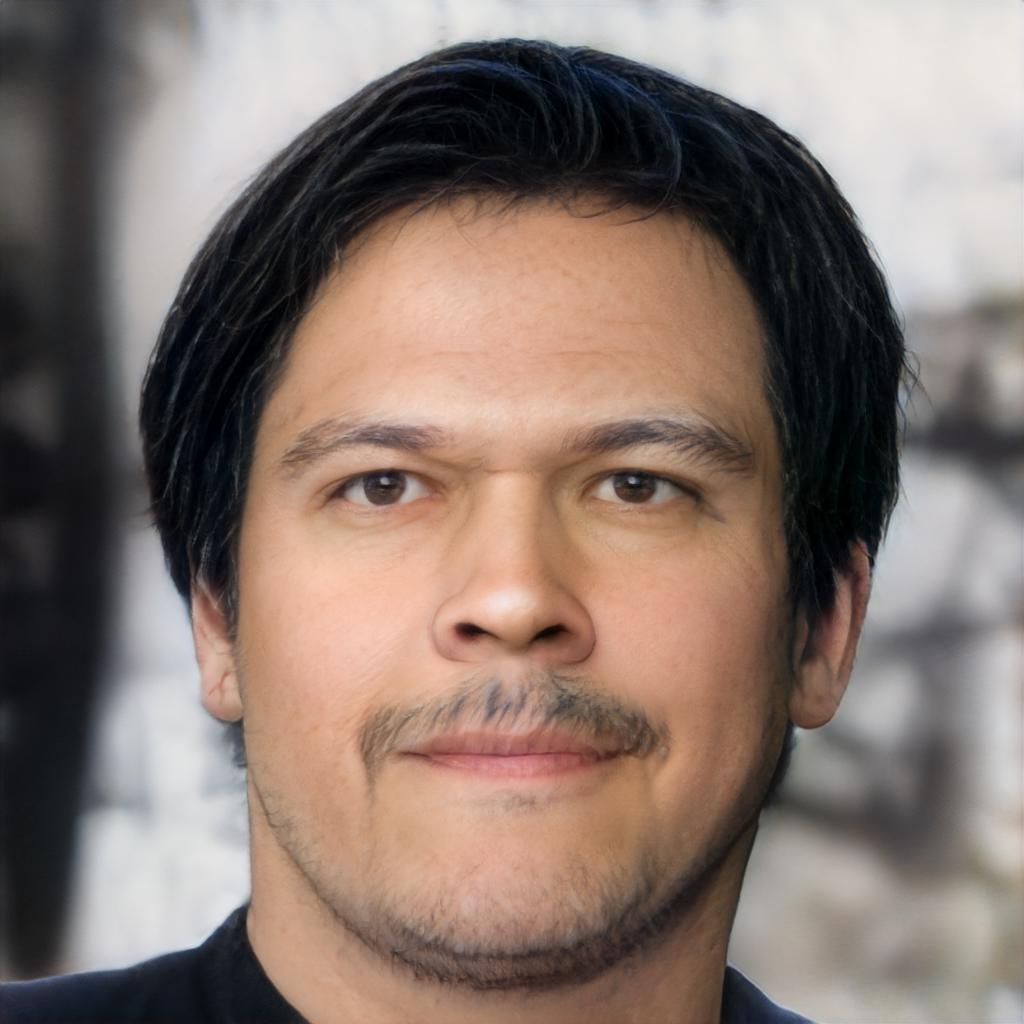} &
 \includegraphics[width=\figwidth]{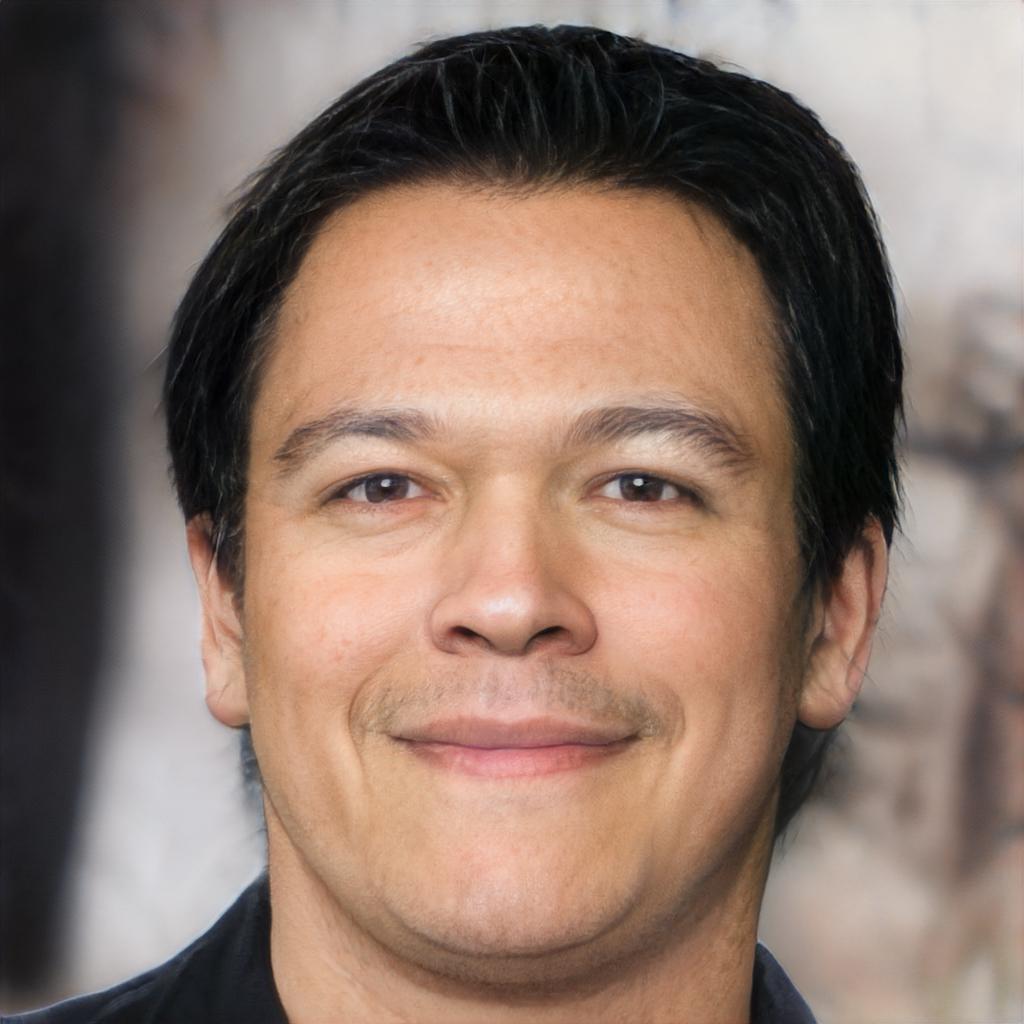} &
 \includegraphics[width=\figwidth]{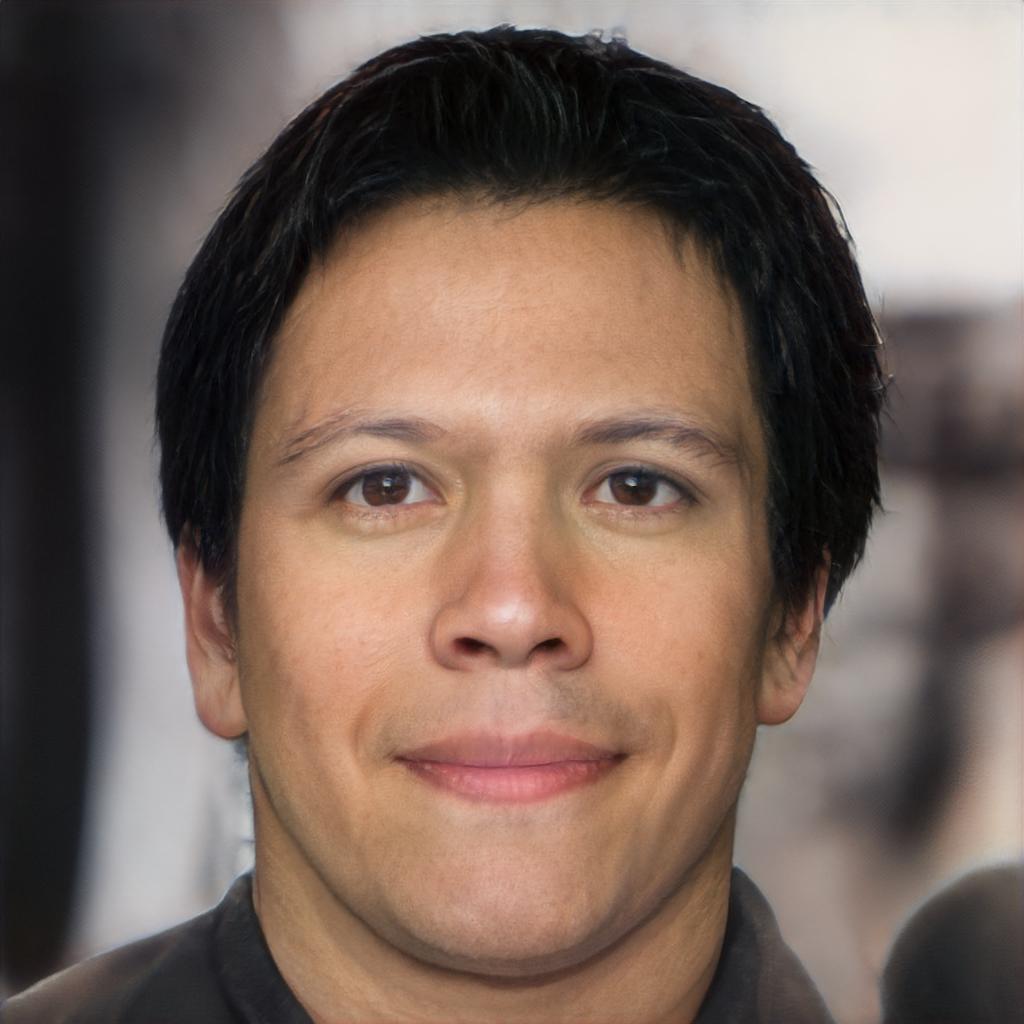} \\
 \begin{turn}{90} \hspace{0.5cm} \wplus \end{turn} &
 \includegraphics[width=\figwidth]{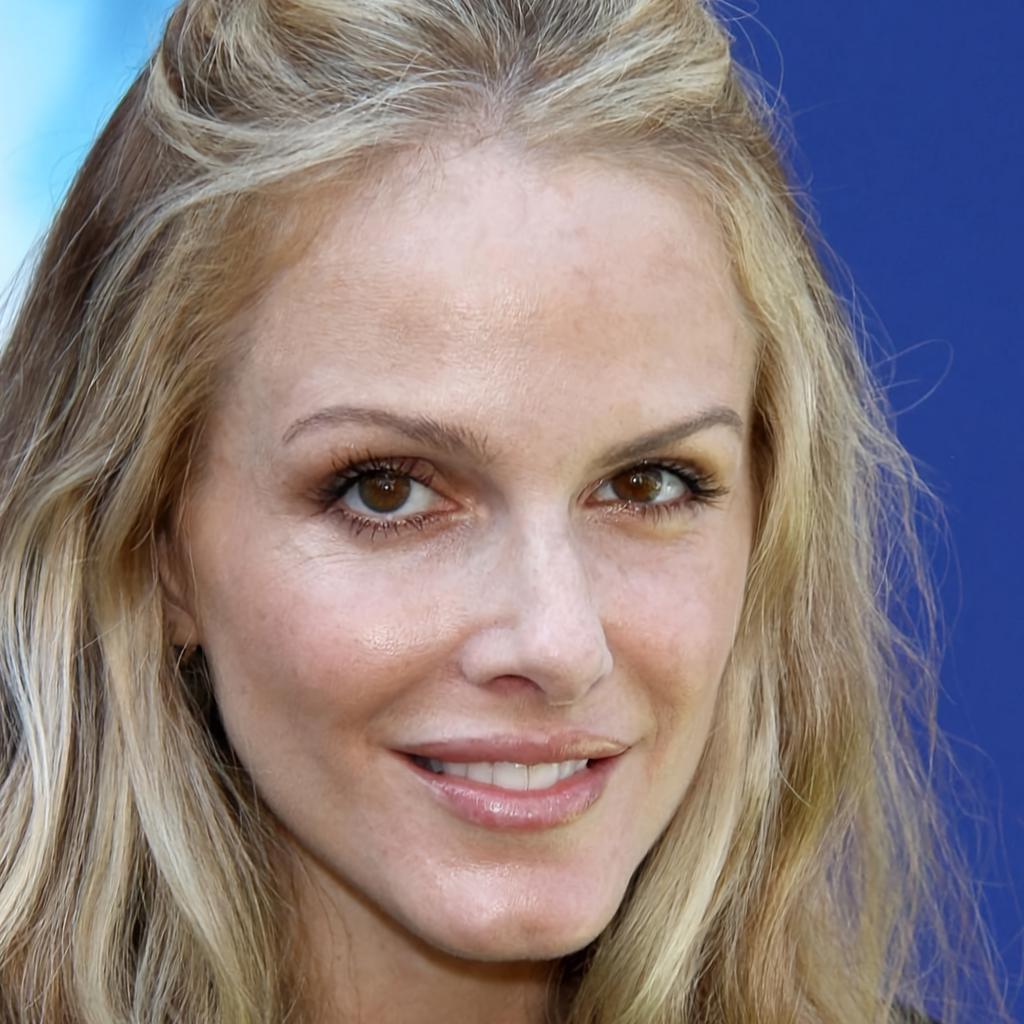} & 
 \includegraphics[width=\figwidth]{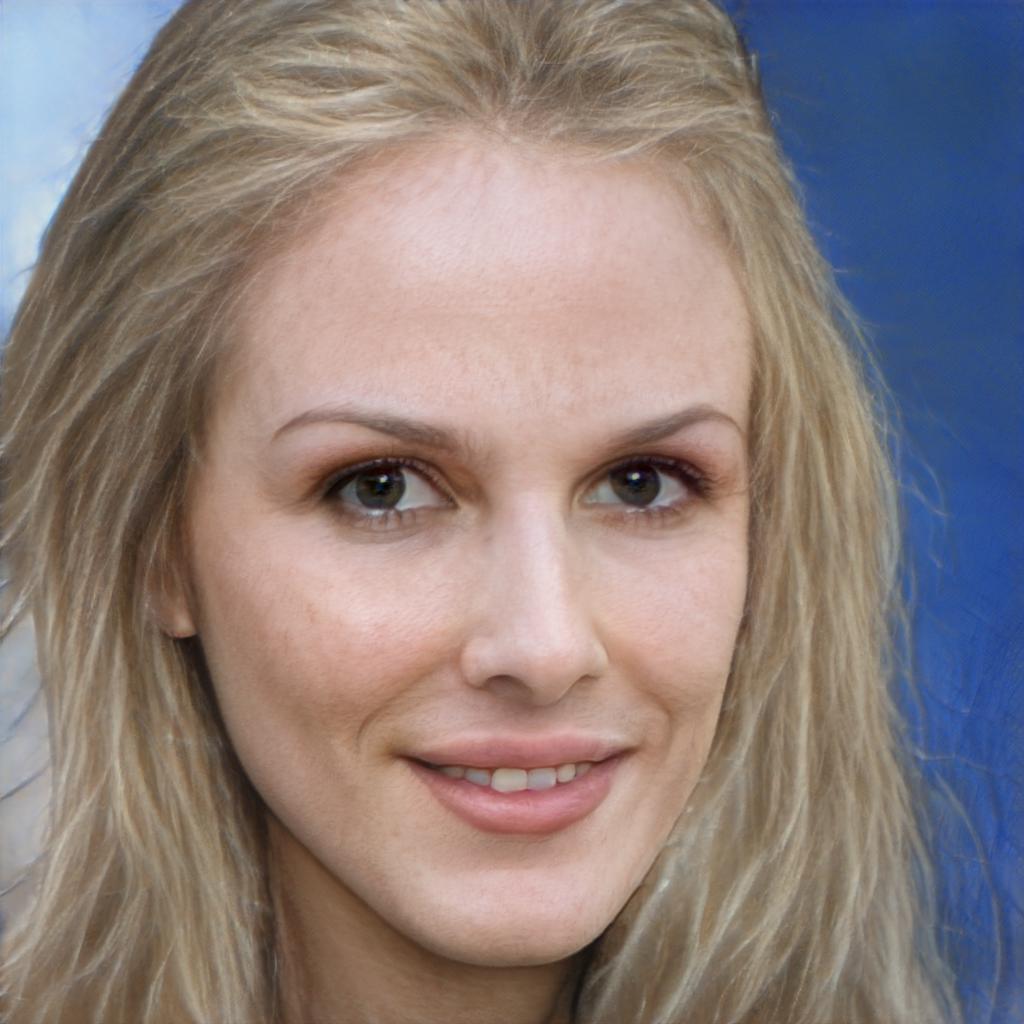} &
 \includegraphics[width=\figwidth]{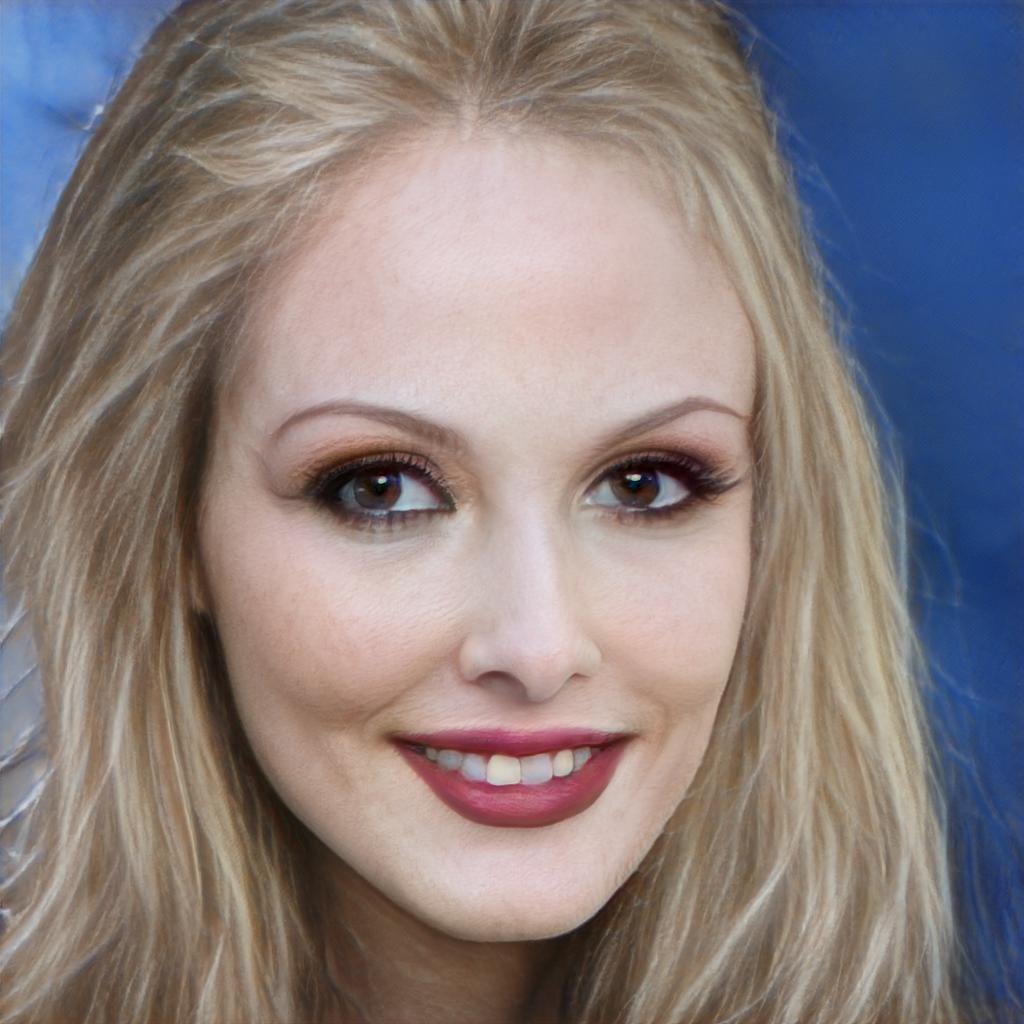} &
 \includegraphics[width=\figwidth]{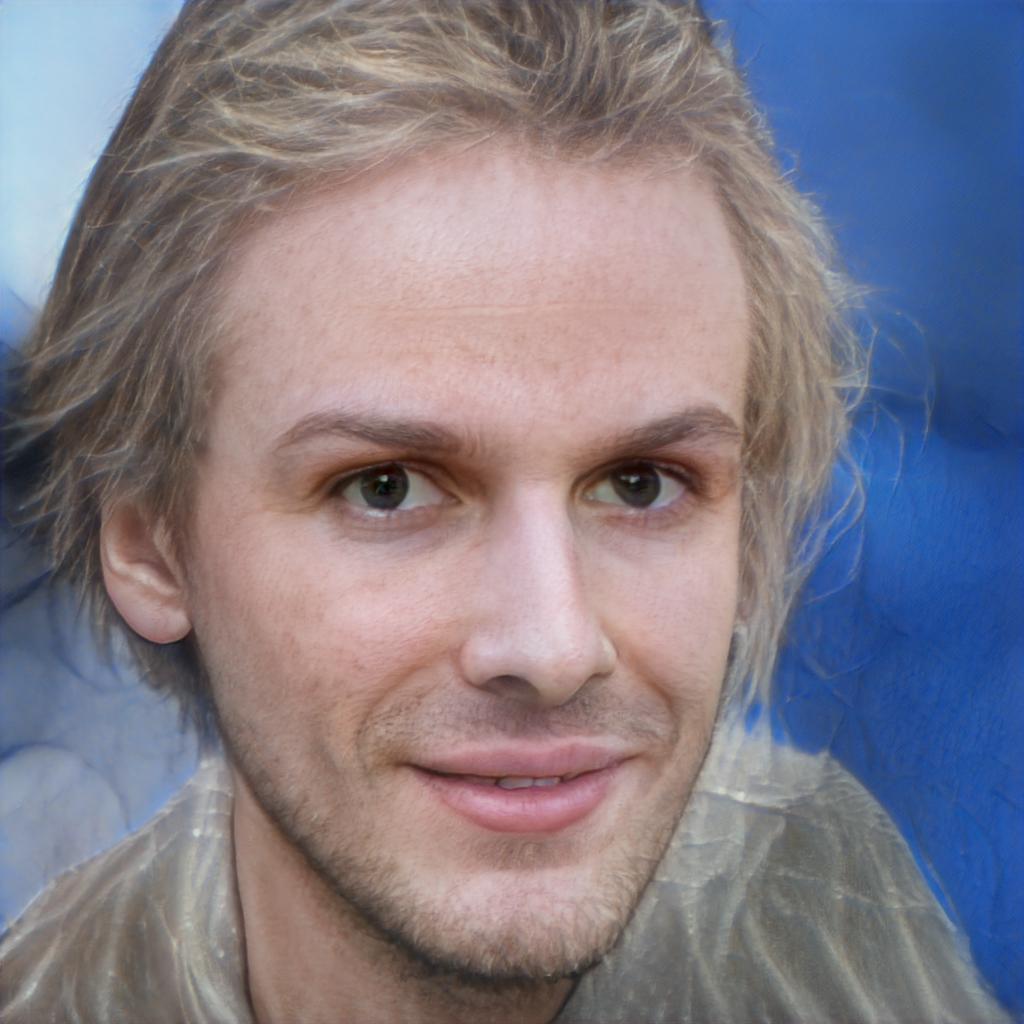} &
 \includegraphics[width=\figwidth]{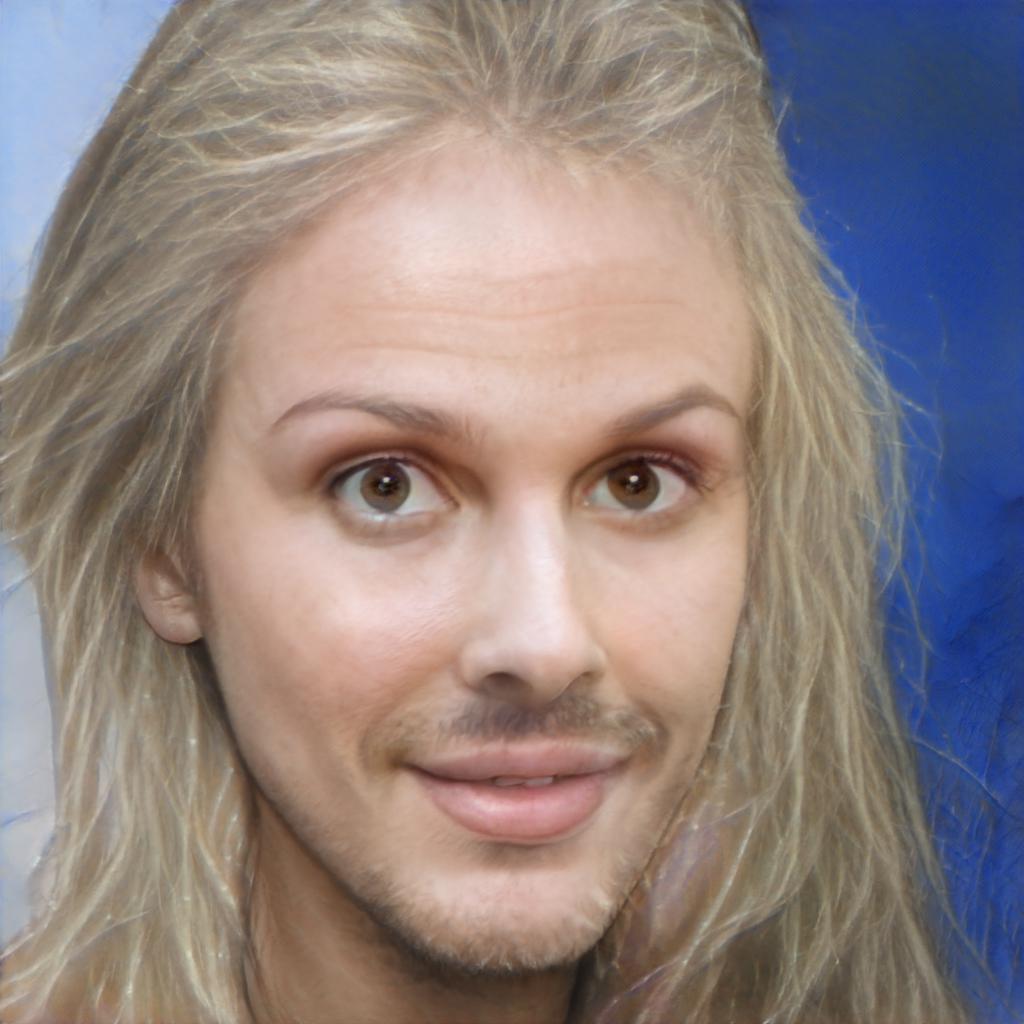} &
 \includegraphics[width=\figwidth]{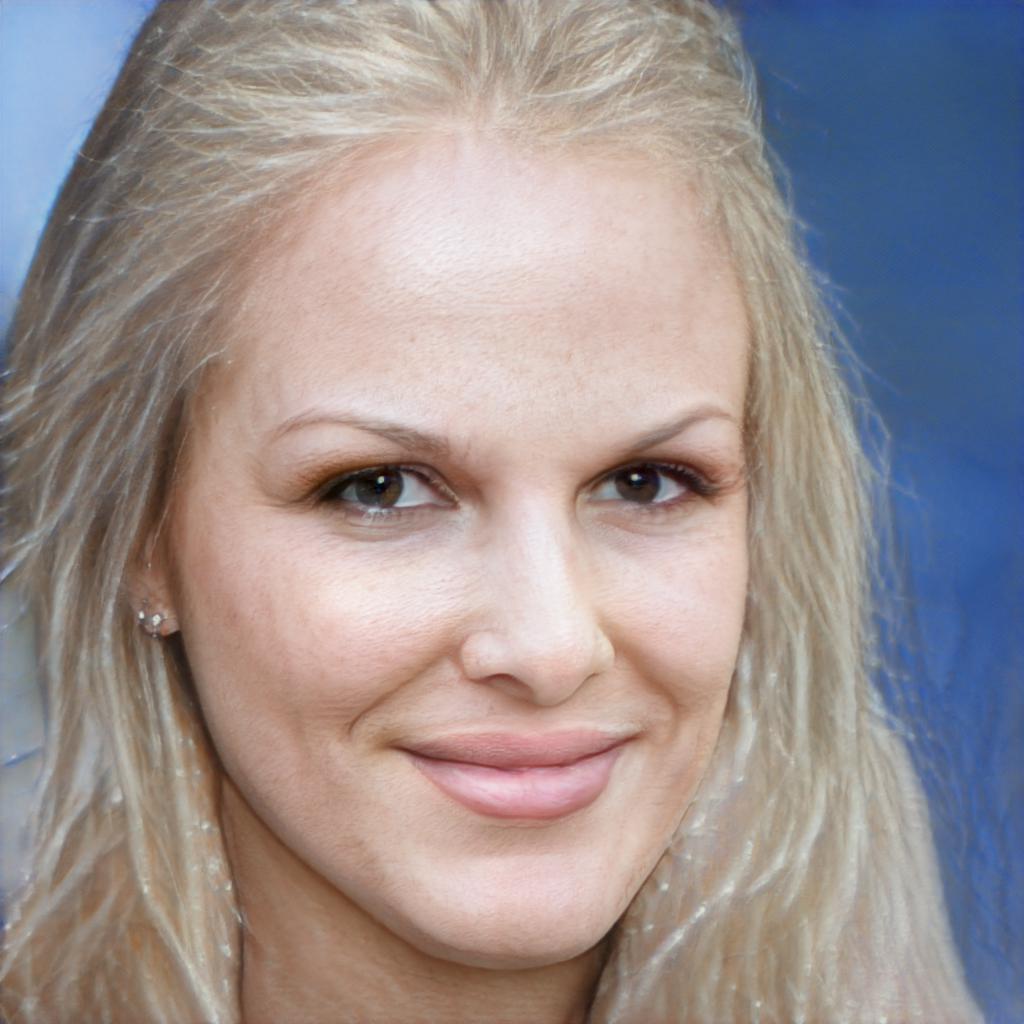} &
 \includegraphics[width=\figwidth]{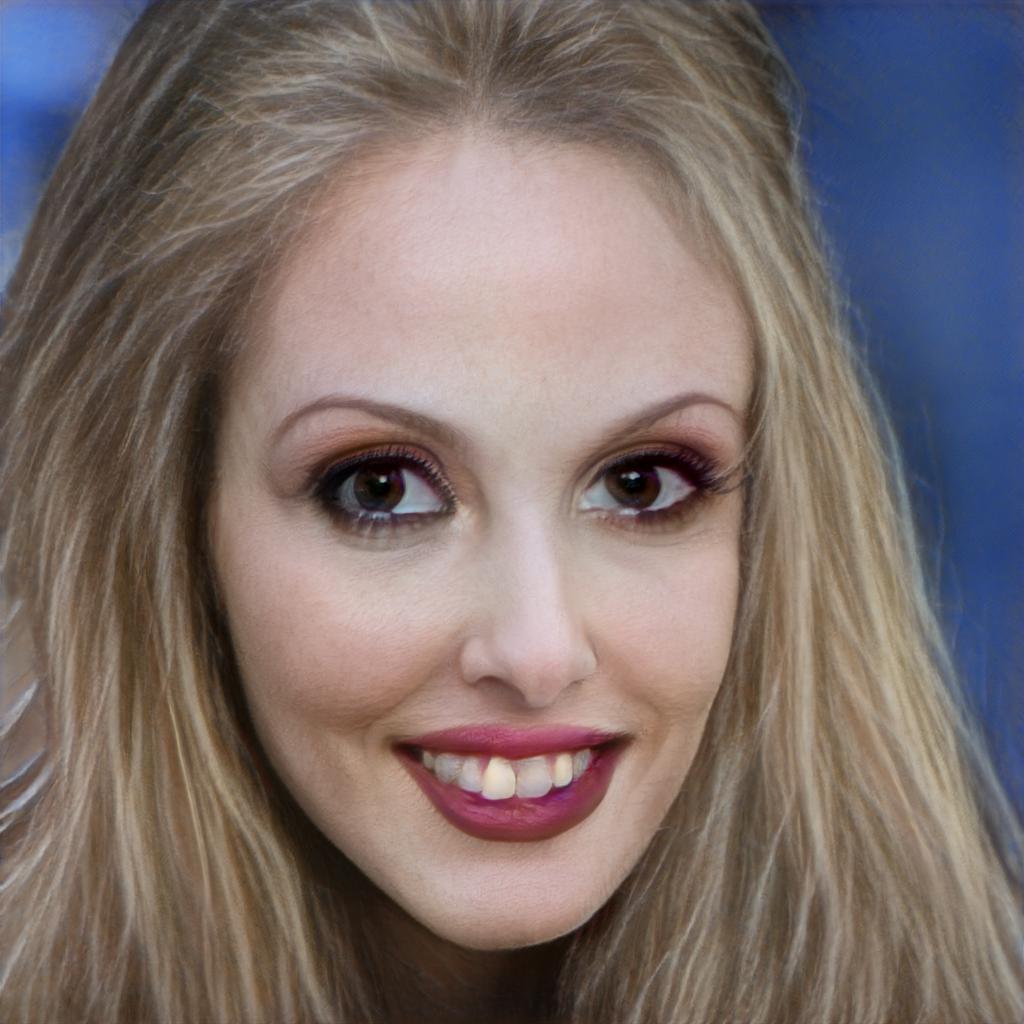} \\
 \begin{turn}{90} \hspace{0.5cm} \wstar (ours) \end{turn} &
 \includegraphics[width=\figwidth]{images/original/06011.jpg} & 
 \includegraphics[width=\figwidth]{images/inversion/18_orig_img_10.jpg} &
 \includegraphics[width=\figwidth]{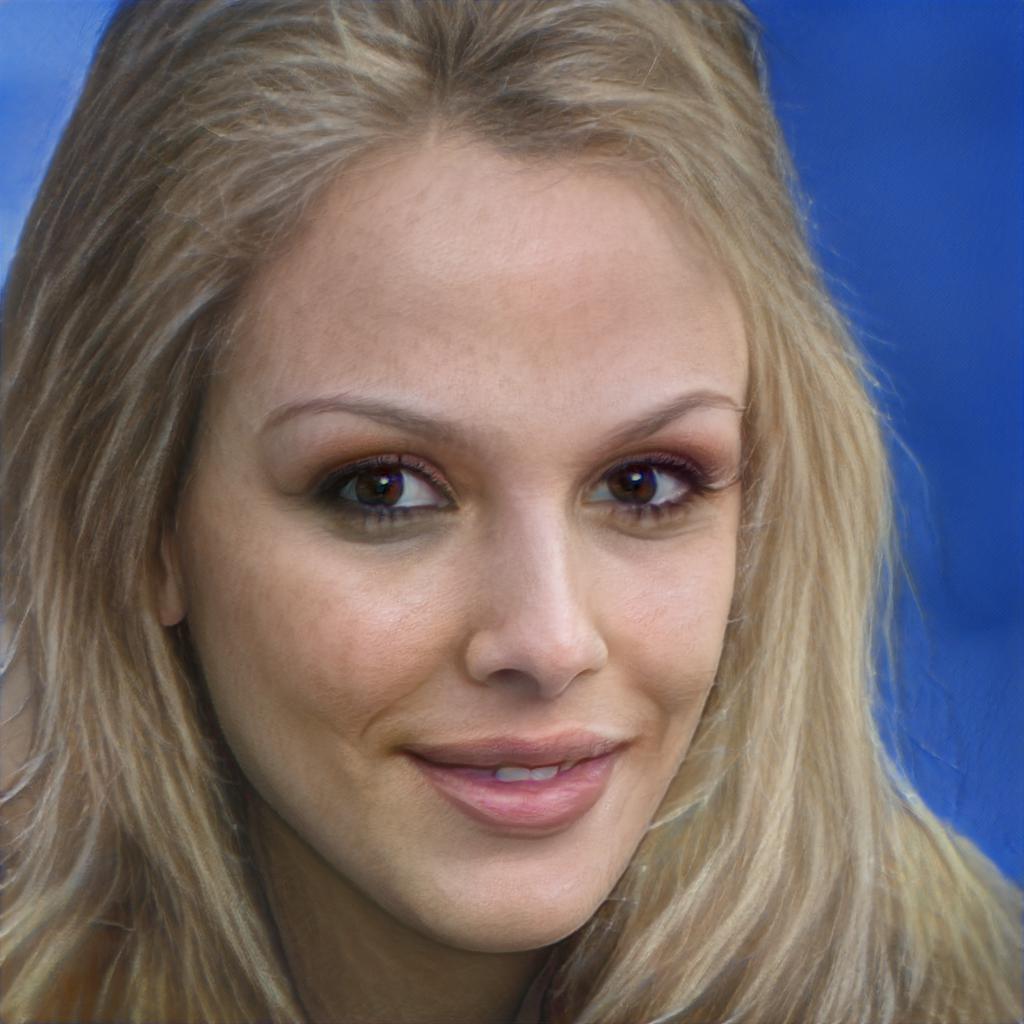} &
 \includegraphics[width=\figwidth]{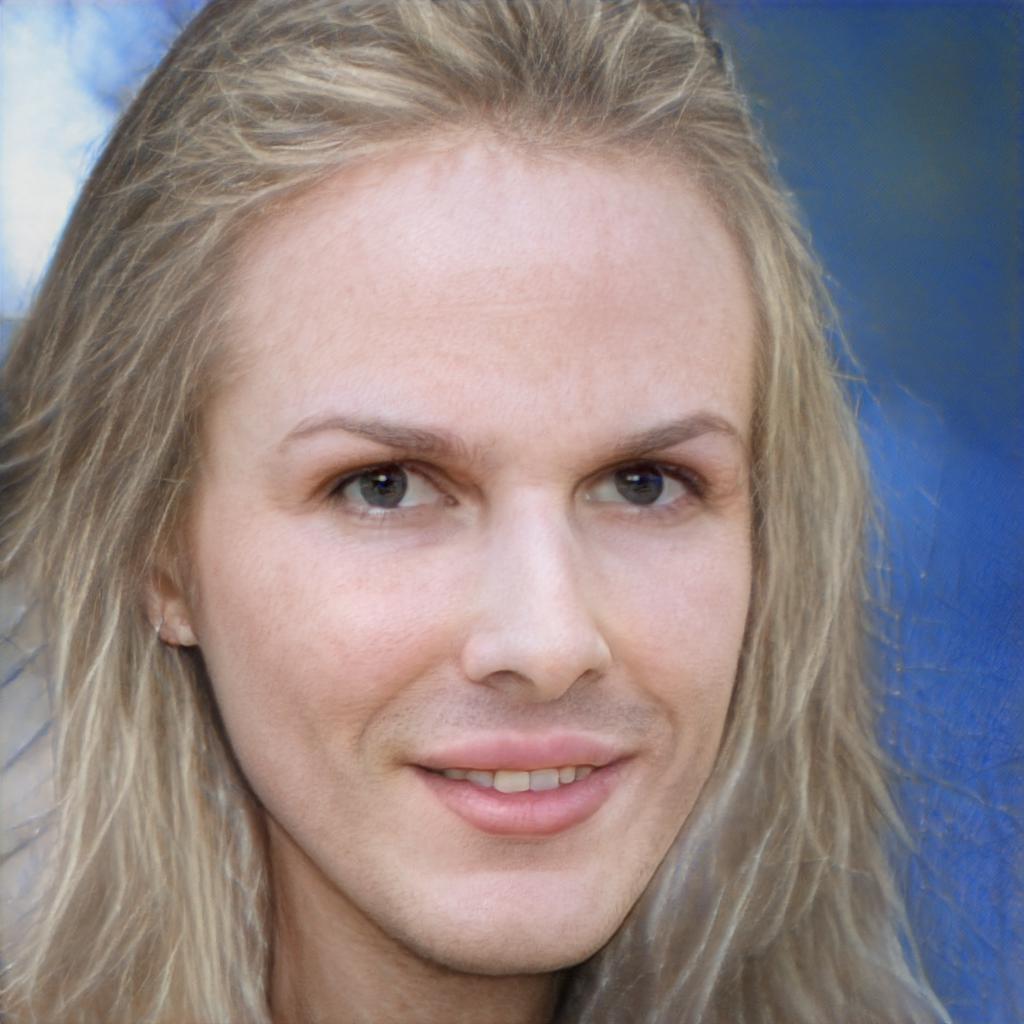} &
 \includegraphics[width=\figwidth]{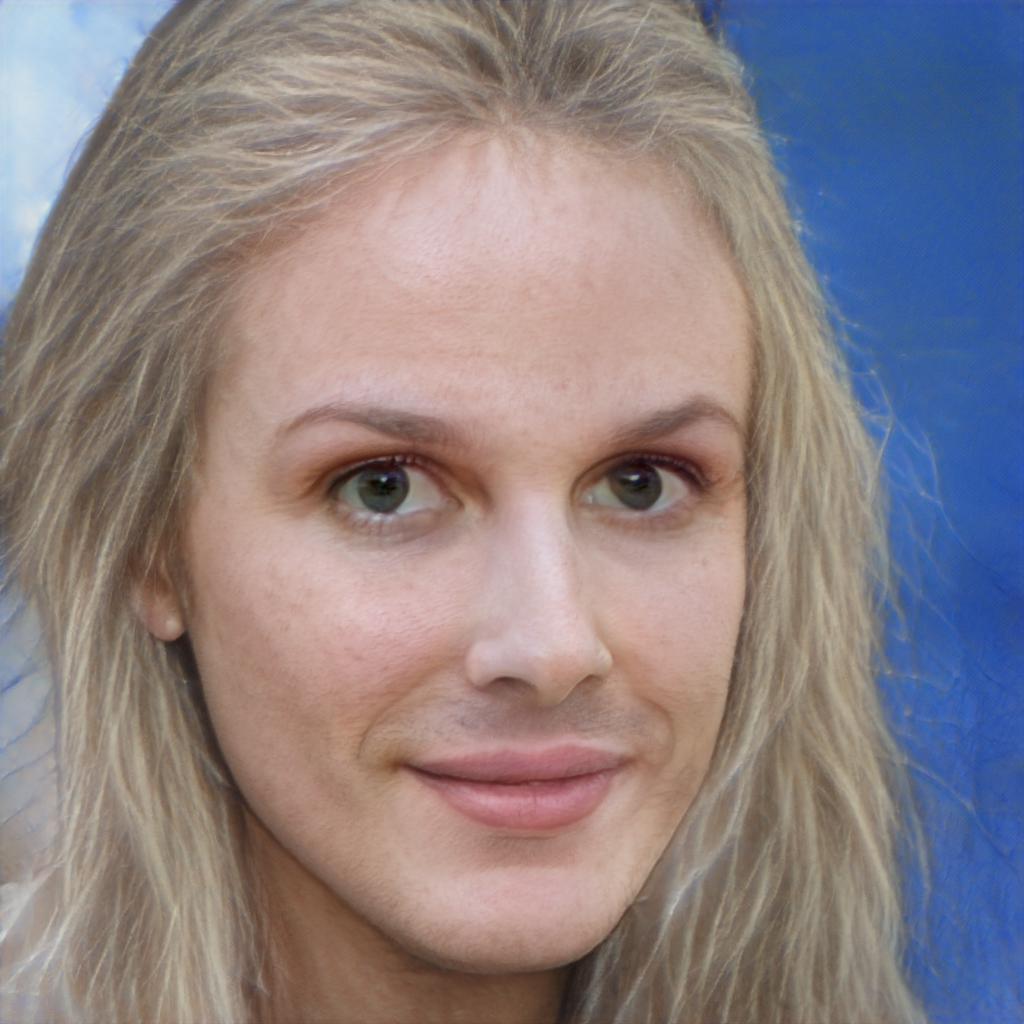} &
 \includegraphics[width=\figwidth]{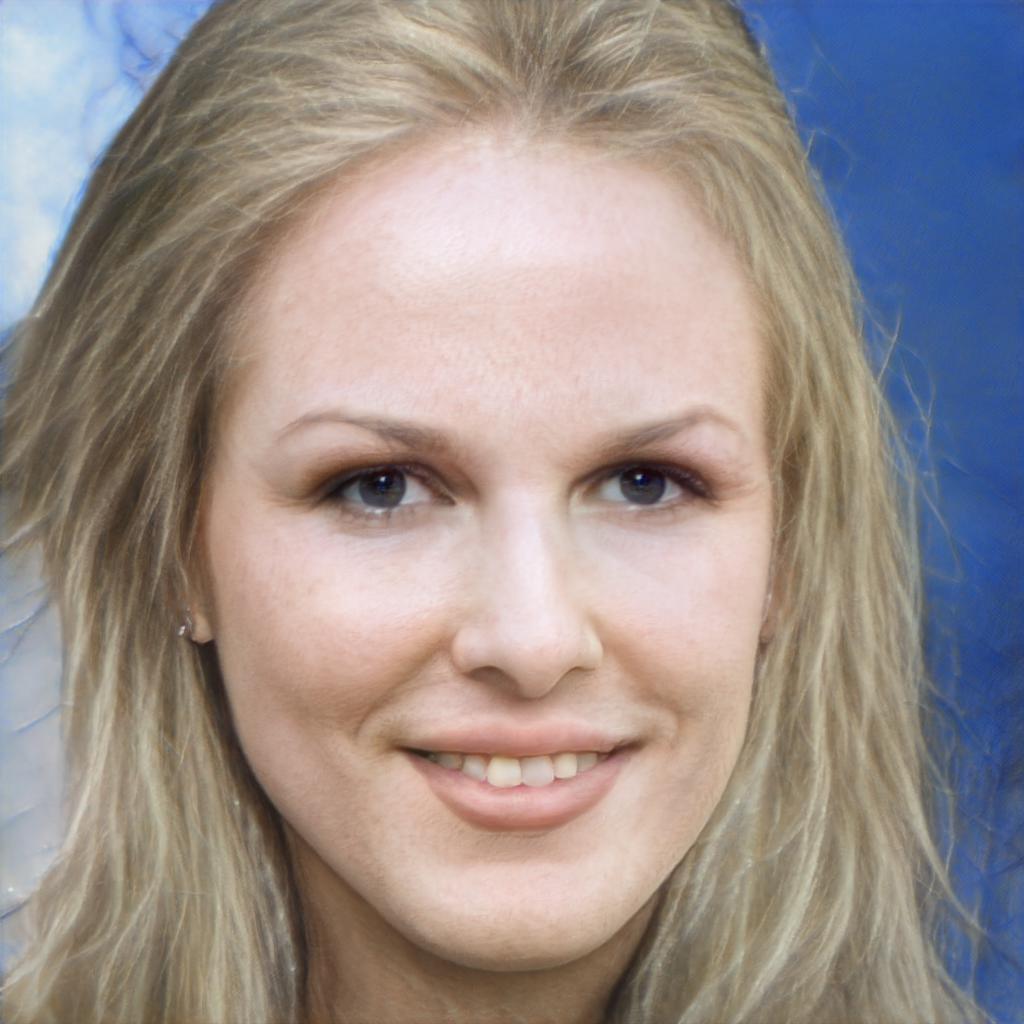} &
 \includegraphics[width=\figwidth]{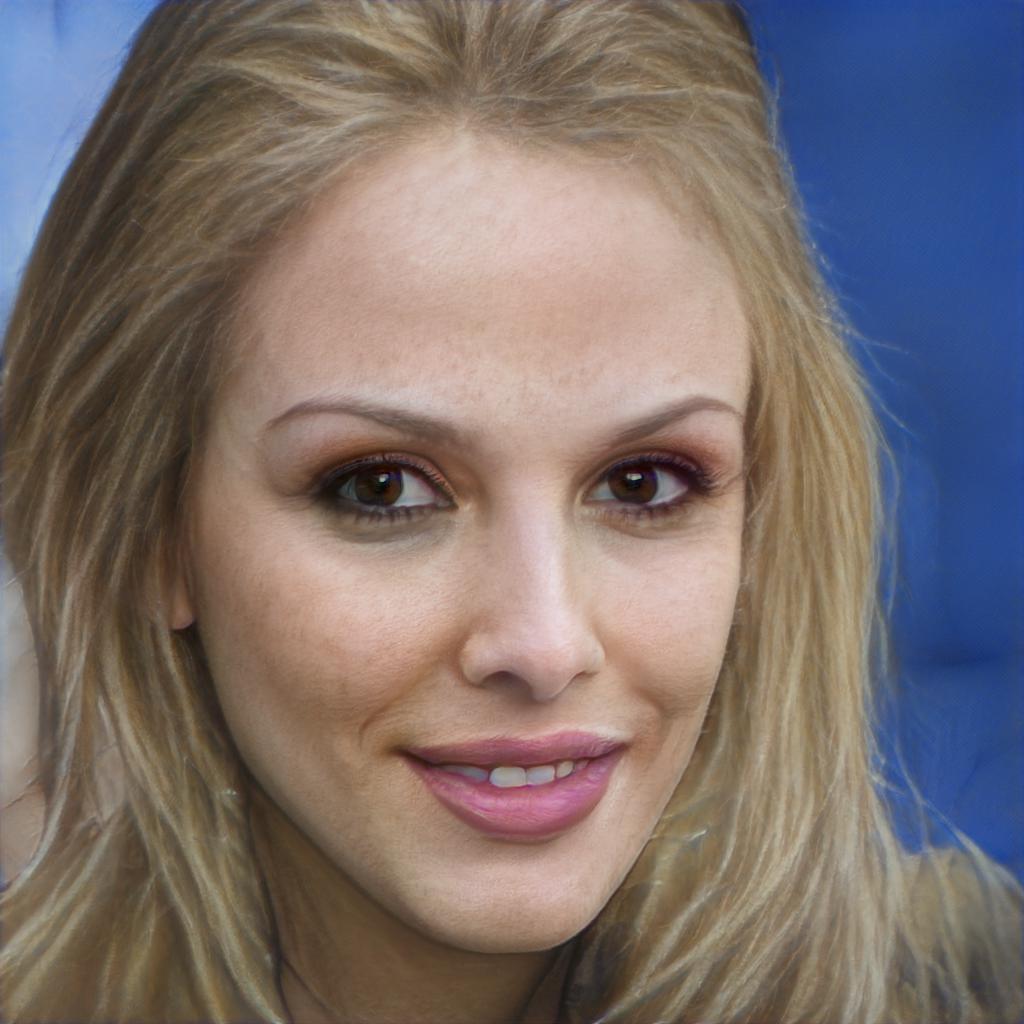} 
\end{tabular}
\caption{\small{Image editing at resolution $1024^2$ using InterFaceGAN \cite{shen2020interfacegan} in \wplus and \wstarsp. Input images (first row) are projected in the latent space of StyleGAN2 (second row), then the latent codes are moved in the direction that corresponds to changing one facial attribute (rows $2-6$). The editing is better in terms of attributes disentanglement in the new learned space \wstarsp (second line of each input) compared to the editing done in the original latent space \wplussp. Our method also leads to a better identity preservation.}}
\label{fig:att_inter_results}
\end{figure*}

\noindent \textbf{Quantitative metrics}
To assess the linear separation and disentanglement for attributes, we use classification accuracy and DCI \cite{Eastwood2018AFF} metrics.

\noindent \textit{Classification Accuracy:}
An SVM was trained from scratch for each attribute on $15000$ latent codes in the corresponding space. In \wplus, the latent codes were obtained after encoding the images of Celeba-HQ using the pretrained encoder, and in \wstar were obtained after mapping the encoded codes using the trained NF model $T$. Among the available latent codes, we split $80\%$ for training and the rest of them for validation. The accuracy measures are reported in table \ref{tab:res_dist_att} using minimum (Min Acc), maximum classification accuracy (Max Acc) among the attributes and average accuracy (Avg Acc).

\noindent \textit{DCI \cite{Eastwood2018AFF}:} DCI is used to assess disentanglement; Disentanglement (D) quantifies how much each dimension captures at most one attribute, Completeness (C) quantifies how much each attribute is captured by a single dimension and Informativeness (I) quantifies how much informative the latent code is for the attributes which is simply the classification error.
We used $40$ Lasso regressors with $\alpha=0.05$ from scikit-learn library \cite{pedregosa2011scikit}, trained on $2000$ samples from Celeba-HQ encoded using the pretrained encoder. Results are presented in table \ref{tab:res_dist_att}. These results show that the performance of a linear classifier is improved in \wstarsp. The DCI metrics are also clearly improved in \wstarsp.
\begin{table}[htbp]
\caption{The quantitative assessment of the attribute's linear separation, disentanglement in \wplus and \wstarsp. The $\uparrow$ ($\downarrow$) indicates the higher (lower) values are better and the best results in \textbf{bold}. The first part indicates that the min, max and average classification accuracy (over all $40$ facial attributes) are improved in \wstarsp. The second part shows that disentanglement is also improved in \wstarsp.}
\begin{center}
\small\addtolength{\tabcolsep}{-3pt}
\begin{tabular}{lccc|ccc}
\toprule
& \multicolumn{3}{c}{\textbf{Linear separation}} & \multicolumn{3}{c}{\textbf{Disentanglement}} \\
Space & Min acc $\uparrow$ & Max acc $\uparrow$ & Mean acc $\uparrow$ & D $\uparrow$ & C $\uparrow$ & I $\downarrow$ \\
\hline
\wplus & 0.635 & 0.979 & 0.834 & 0.67 & 0.53 & 0.33 \\
\wstarsp & \textbf{0.796} & \textbf{0.991} & \textbf{0.916} & \textbf{0.73} & \textbf{0.58} & \textbf{0.30}\\
\bottomrule
\end{tabular}
\end{center}
\label{tab:res_dist_att}
\end{table}
\vspace{-0.5cm}

\noindent \textbf{Image Editing}
Here we qualitatively demonstrate the benefits of the new proxy space for the image editing task. InterFaceGAN \cite{shen2020interfacegan} was retrained to manipulate the attributes of a given real image in both \wplus and \wstarsp. %
InterFaceGAN assumes that the positive and negative examples of each attribute are linearly separable, and the editing direction is simply the normal to the classification hyperplane. These normal directions were obtained after training an SVM for each attribute in both spaces. The edited images were generated after editing the latent codes in \wplus or \wstarsp before feeding them to the StyleGAN2 generator. For \wstarsp, the edited latent codes were mapped back to \wplus before feeding them to the generator.

\noindent \textbf{Results:} Figure \ref{fig:att_inter_results} shows the editing results on $5$ attributes in \wstarsp and \wplus and it shows that the editing results are visually better in \wstarsp than in \wplussp. In particular, we see the following observation in \wplussp: gender is still entangled with adding Makeup and Lipstick (3rd row where the male gender is changed to female), changing the gender to Male is entangled with adding Beard and the Hair (column 4), and adding Mustache is entangled with gender (row 5).
While in \wstarsp these attributes are better disentangled and the identity is better preserved. Finally, it is clear that we still obtain high quality images even if we did not retrain the generator and the editing is not done in \wplussp.
\label{sec:experiments}
\section{Conclusion}
\label{sec:conclusion}
We have presented in this paper a general framework to enforce additional properties to the latent space of generative models, without the burden of retraining the GAN. Specifically, we train a bijective transformation from the extended space of StyleGAN2 (\wplussp) to a proxy space (\wstarsp), where the facial attributes are disentangled. This method enables to enforce additional supervision on top of any pre-trained GAN. We have validated our approach by showing better facial editing results. In future work, additional properties could be considered as well (e.g., pose preservation), as well as other types of models (\emph{e.g.}, VAEs, GANs).

\clearpage
\newpage
\bibliographystyle{IEEEbib}
\bibliography{main}

\end{document}


\maketitle
The supplementary material is organized as follows: In section \ref{appendix:linear_sep}, we visualize the linear separation of the attributes in the latent space.  Finally, section \ref{appendix:ablation} details the ablation study.

\begin{figure*}[t]
\setlength\tabcolsep{2pt}%
\centering
\begin{tabular}{p{0.25cm}ccccccc}
\centering
 &
 \textbf{Original} &
 \textbf{Inverted} &
 \textbf{Makeup} &
 \textbf{Male} &
 \textbf{Mustache} &
 \textbf{Chubby} &
 \textbf{Lipstick} &
 \begin{turn}{90} \hspace{0.5cm} \wplus\end{turn} &
 \includegraphics[width=0.125\textwidth]{images/original/06002.jpg} & 
 \includegraphics[width=0.125\textwidth]{images/inversion/18_orig_img_1.jpg} &
 \includegraphics[width=0.125\textwidth]{images/wplus/18_img_3.jpg} &
 \includegraphics[width=0.125\textwidth]{images/wplus/20_img_3.jpg} &
 \includegraphics[width=0.125\textwidth]{images/wplus/22_img_3.jpg} &
 \includegraphics[width=0.125\textwidth]{images/wplus/13_img_3.jpg} &
 \includegraphics[width=0.125\textwidth]{images/wplus/36_img_3.jpg} &
 \begin{turn}{90} \hspace{0.5cm} $\mathcal{W}^{\star}_{ID}$ \end{turn} &
 \includegraphics[width=0.125\textwidth]{images/original/06002.jpg} & 
 \includegraphics[width=0.125\textwidth]{images/inversion/18_orig_img_1.jpg} &
 \includegraphics[width=0.125\textwidth, ]{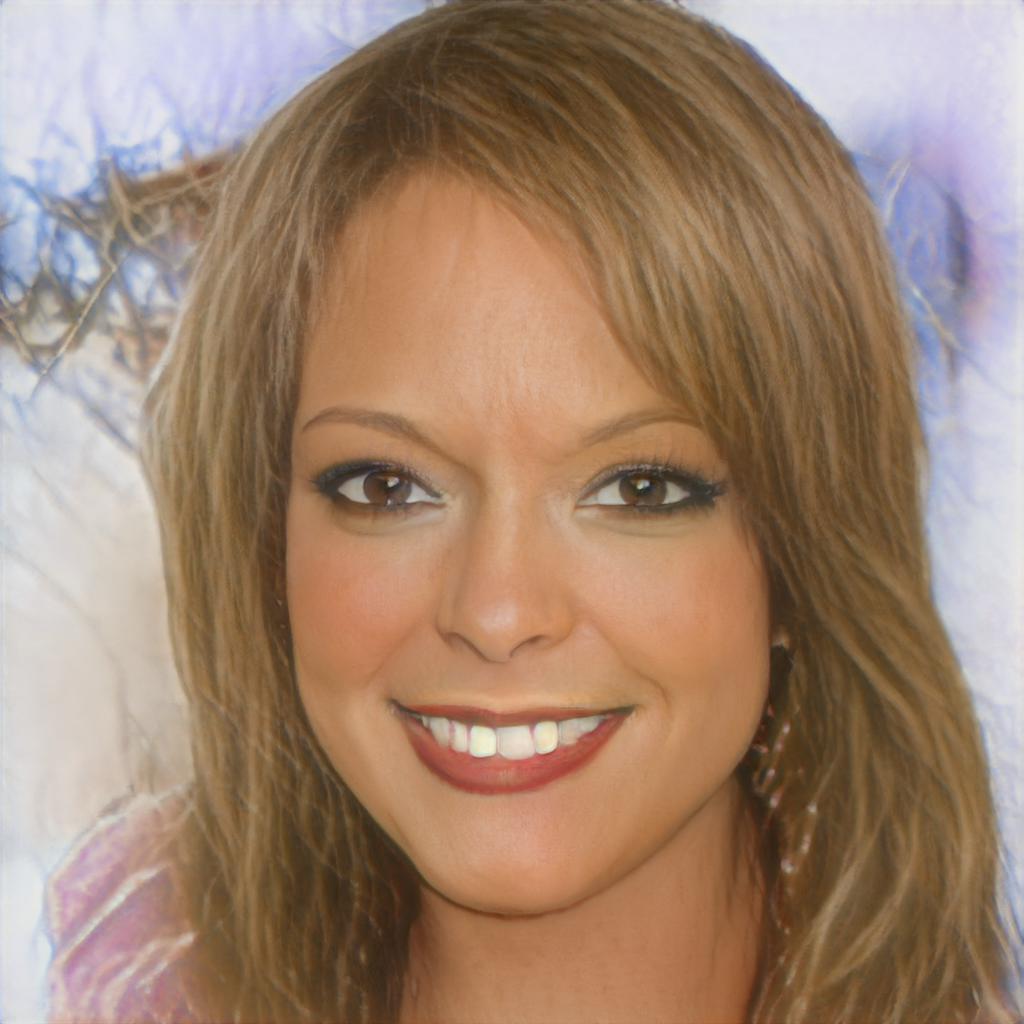} &
 \includegraphics[width=0.125\textwidth, ]{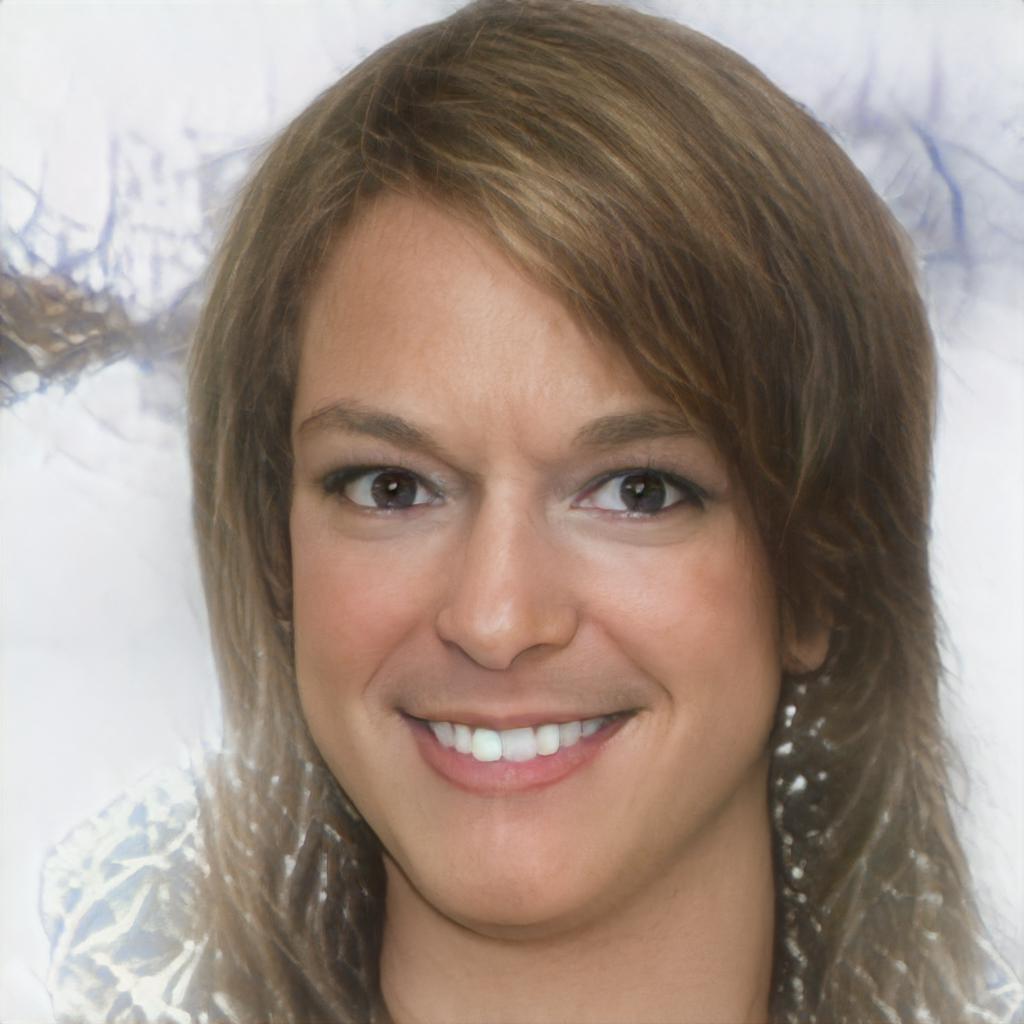} &
 \includegraphics[width=0.125\textwidth, ]{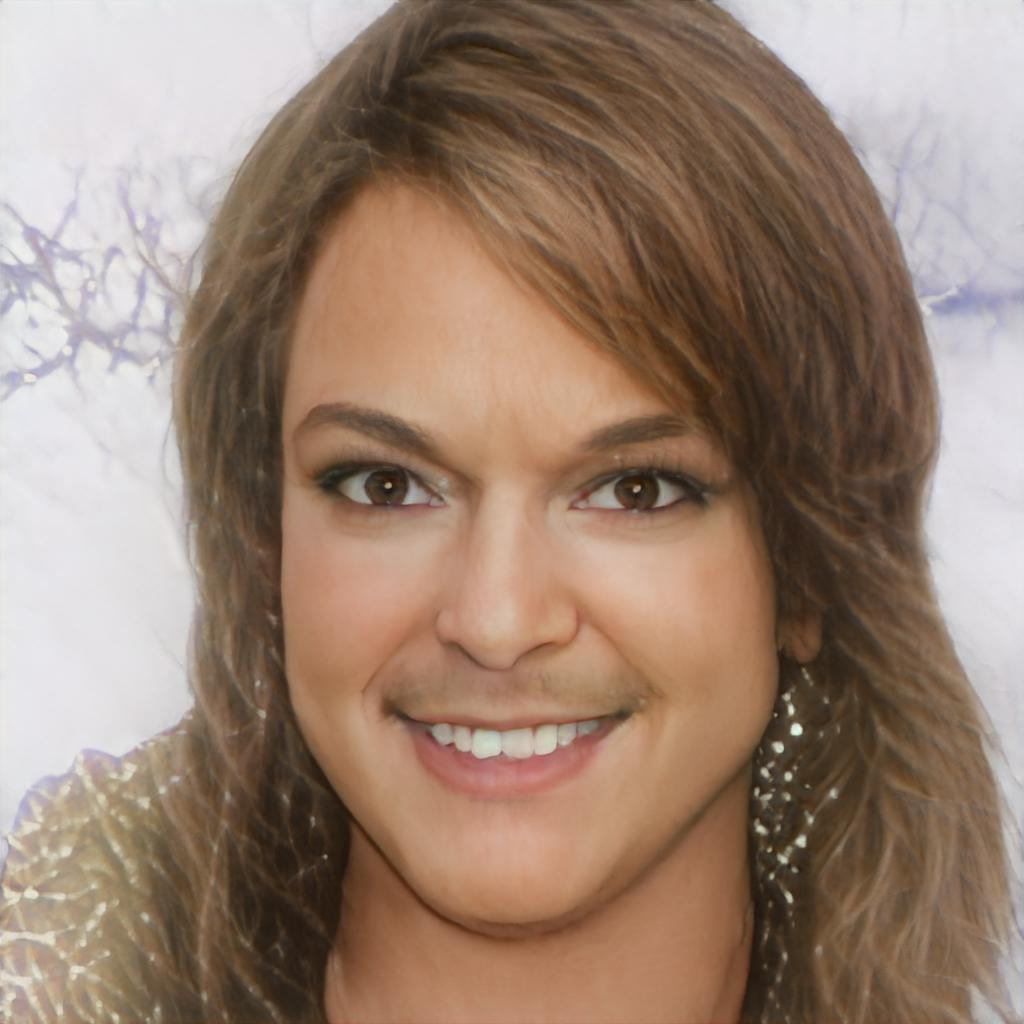} &
 \includegraphics[width=0.125\textwidth, ]{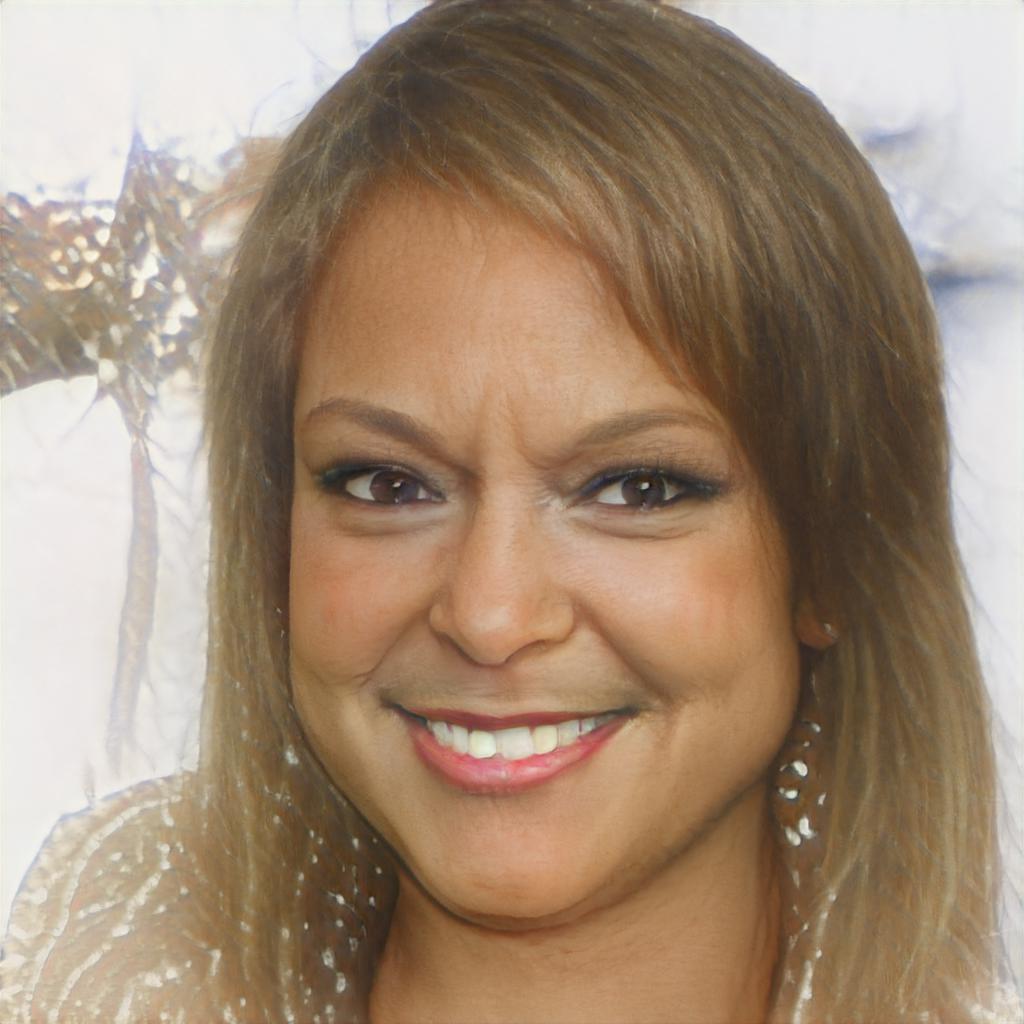} &
 \includegraphics[width=0.125\textwidth, ]{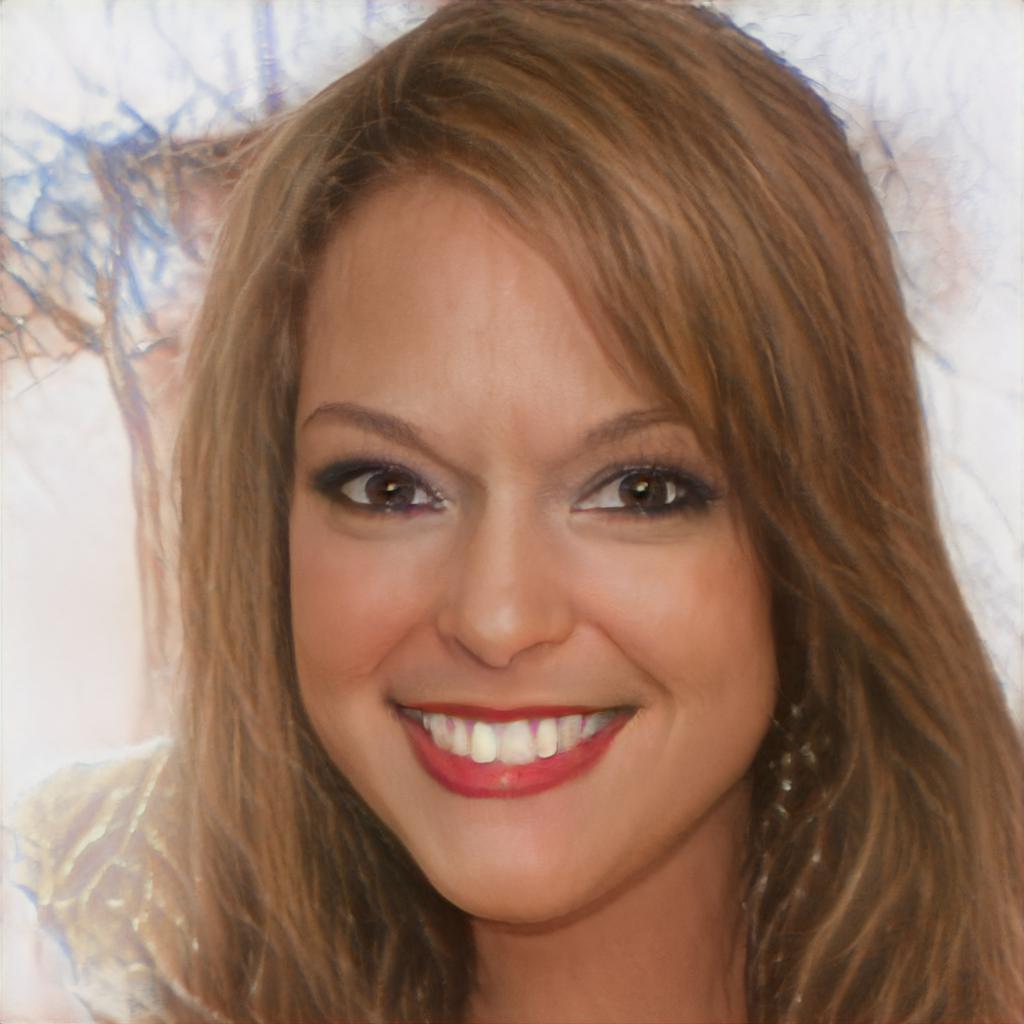} &
 \midrule
 \begin{turn}{90} \hspace{0.5cm} \wplus\end{turn} &
 \includegraphics[width=0.125\textwidth]{images/original/06004.jpg} & 
 \includegraphics[width=0.125\textwidth, ]{images/inversion/18_orig_img_3.jpg} &
 \includegraphics[width=0.125\textwidth, ]{images/wplus/18_img_5.jpg} &
 \includegraphics[width=0.125\textwidth, ]{images/wplus/20_img_5.jpg} &
 \includegraphics[width=0.125\textwidth, ]{images/wplus/22_img_5.jpg} &
 \includegraphics[width=0.125\textwidth, ]{images/wplus/13_img_5.jpg} &
 \includegraphics[width=0.125\textwidth]{images/wplus/36_img_5.jpg} &
 \begin{turn}{90} \hspace{0.5cm} $\mathcal{W}^{\star}_{ID}$ \end{turn} &
 \includegraphics[width=0.125\textwidth]{images/original/06004.jpg} & 
 \includegraphics[width=0.125\textwidth, ]{images/inversion/18_orig_img_3.jpg} &
 \includegraphics[width=0.125\textwidth, ]{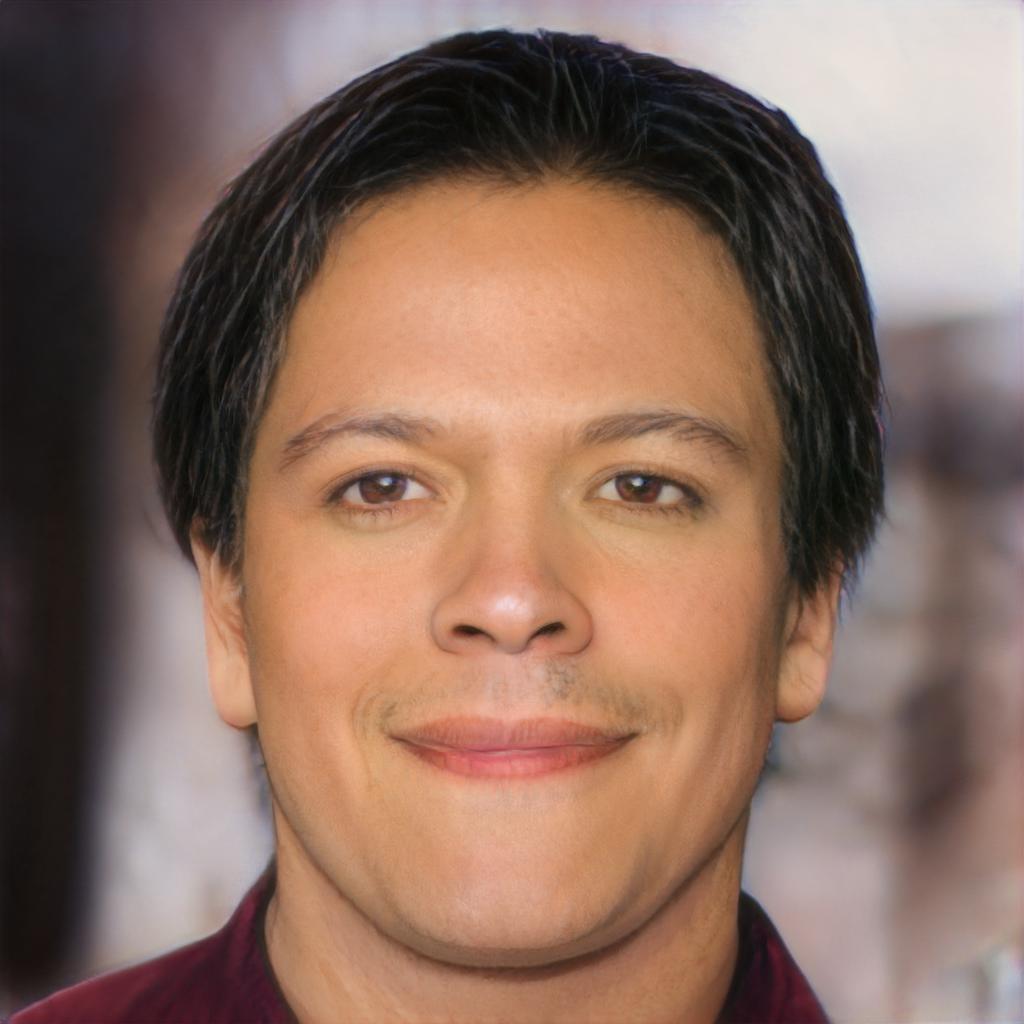} &
 \includegraphics[width=0.125\textwidth, ]{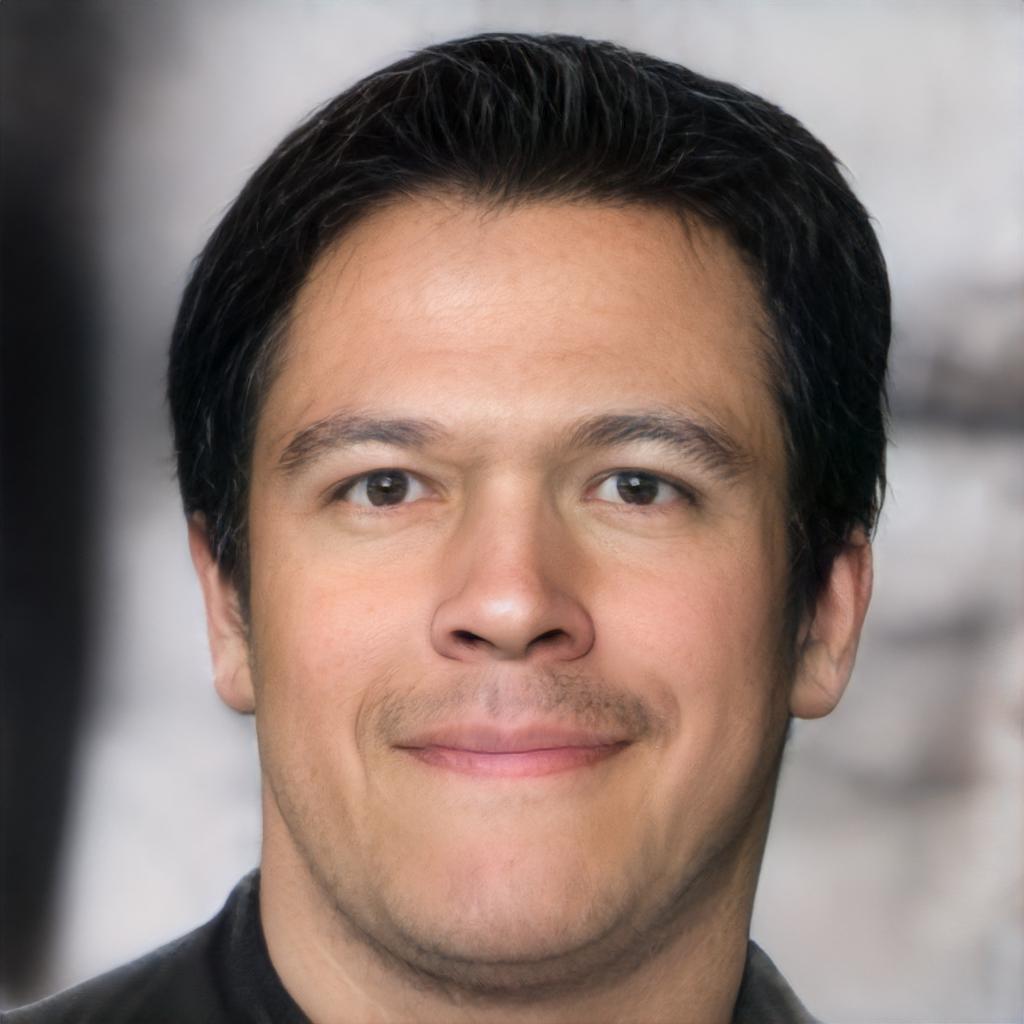} &
 \includegraphics[width=0.125\textwidth, ]{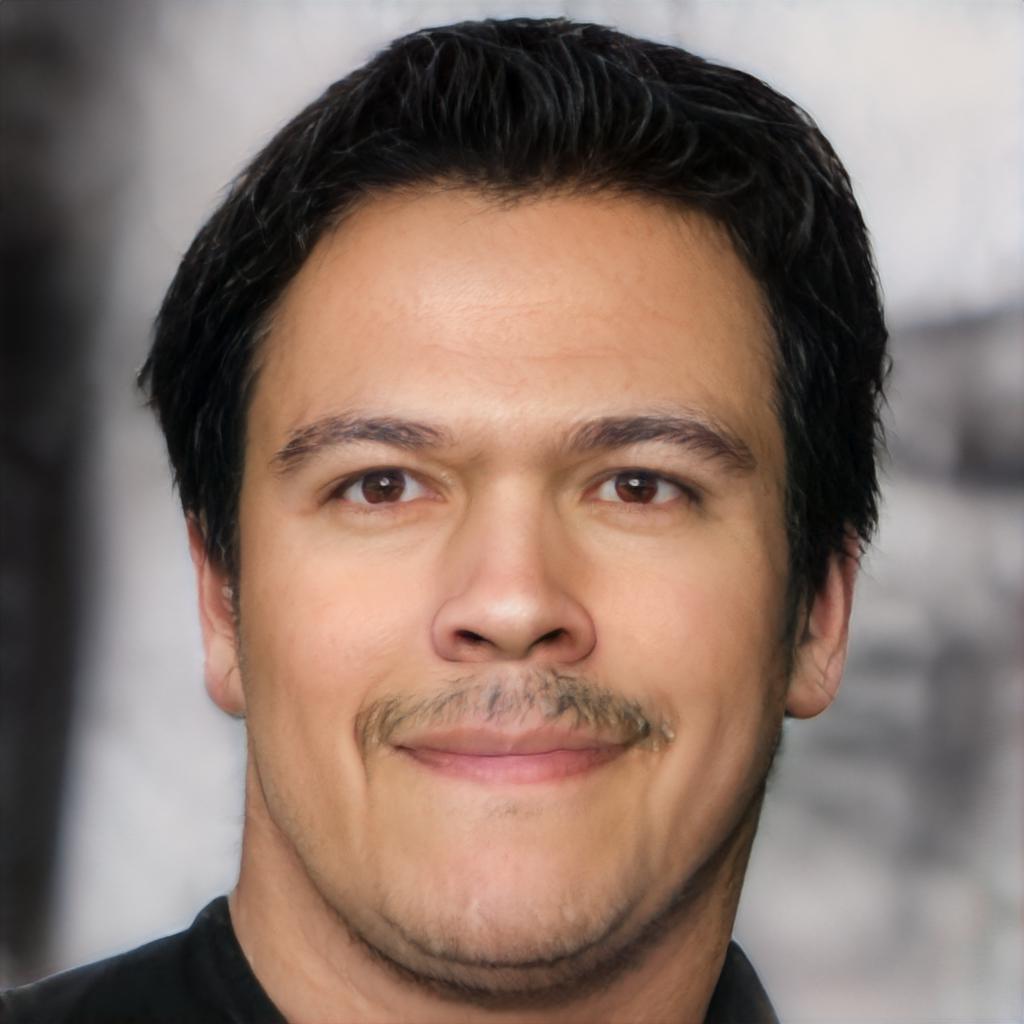} &
 \includegraphics[width=0.125\textwidth, ]{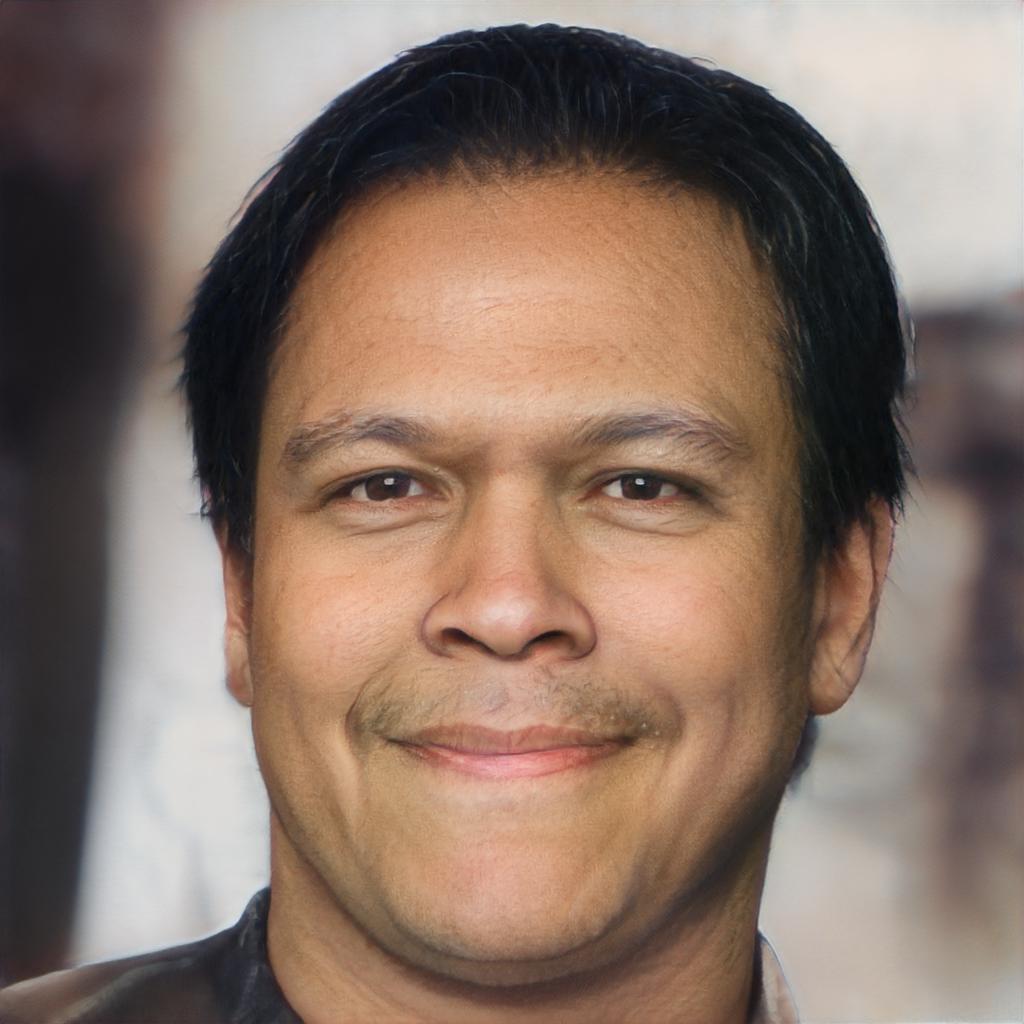} &
 \includegraphics[width=0.125\textwidth, ]{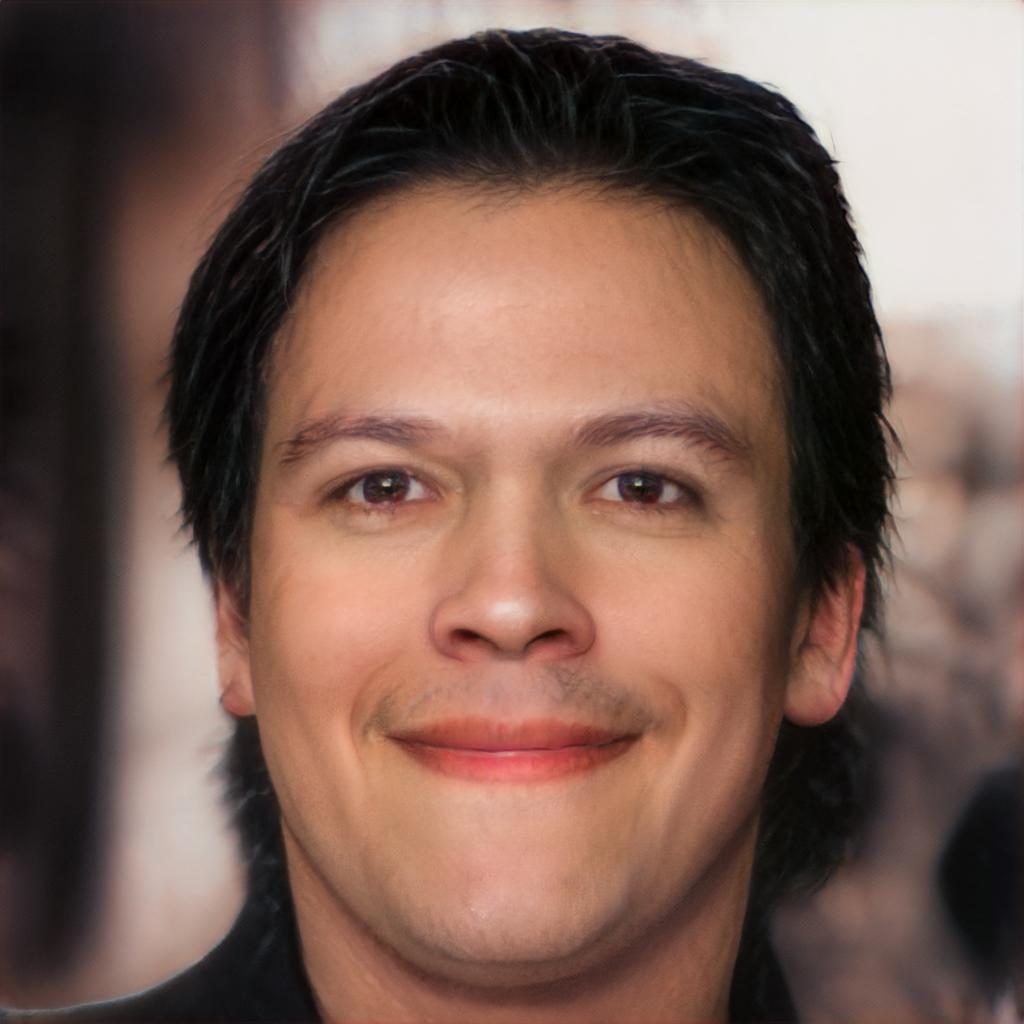} &
 \midrule
 \begin{turn}{90} \hspace{0.5cm} \wplus \end{turn} &
 \includegraphics[width=0.125\textwidth, ]{images/original/06011.jpg} & 
 \includegraphics[width=0.125\textwidth, ]{images/inversion/18_orig_img_10.jpg} &
 \includegraphics[width=0.125\textwidth, ]{images/wplus/18_img_12.jpg} &
 \includegraphics[width=0.125\textwidth, ]{images/wplus/20_img_12.jpg} &
 \includegraphics[width=0.125\textwidth, ]{images/wplus/22_img_12.jpg} &
 \includegraphics[width=0.125\textwidth, ]{images/wplus/13_img_12.jpg} &
 \includegraphics[width=0.125\textwidth]{images/wplus/36_img_12.jpg} &
 \begin{turn}{90} \hspace{0.5cm} $\mathcal{W}^{\star}_{ID}$ \end{turn} &
 \includegraphics[width=0.125\textwidth, ]{images/original/06011.jpg} & 
 \includegraphics[width=0.125\textwidth, ]{images/inversion/18_orig_img_10.jpg} &
 \includegraphics[width=0.125\textwidth,  ]{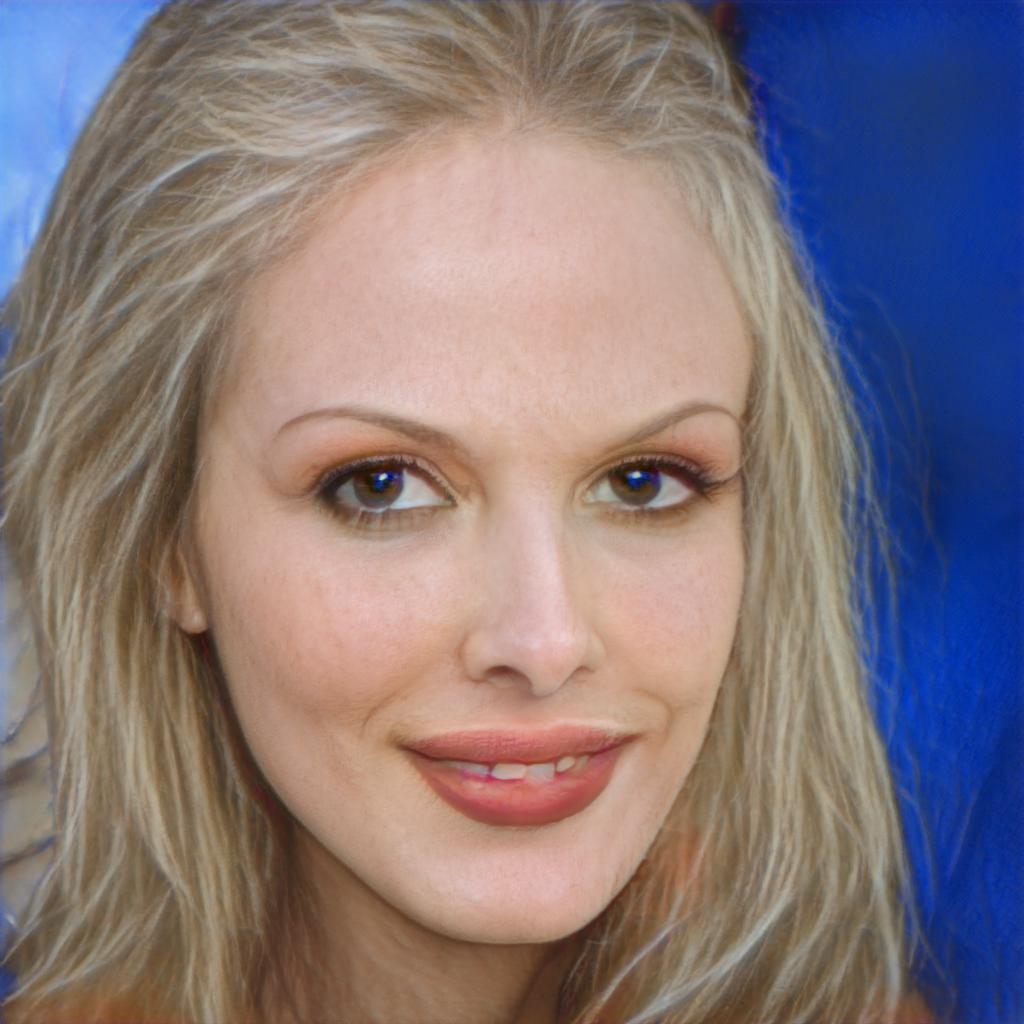} &
 \includegraphics[width=0.125\textwidth, ]{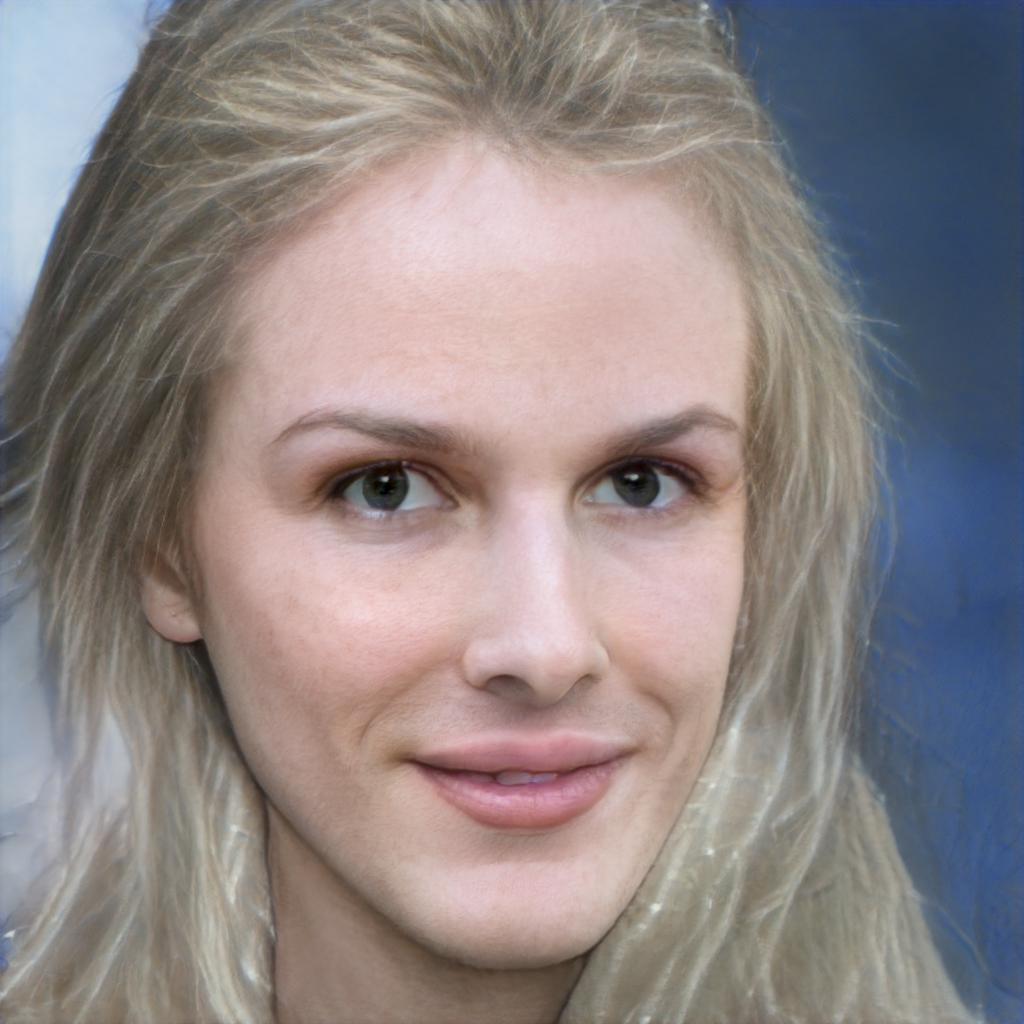} &
 \includegraphics[width=0.125\textwidth, ]{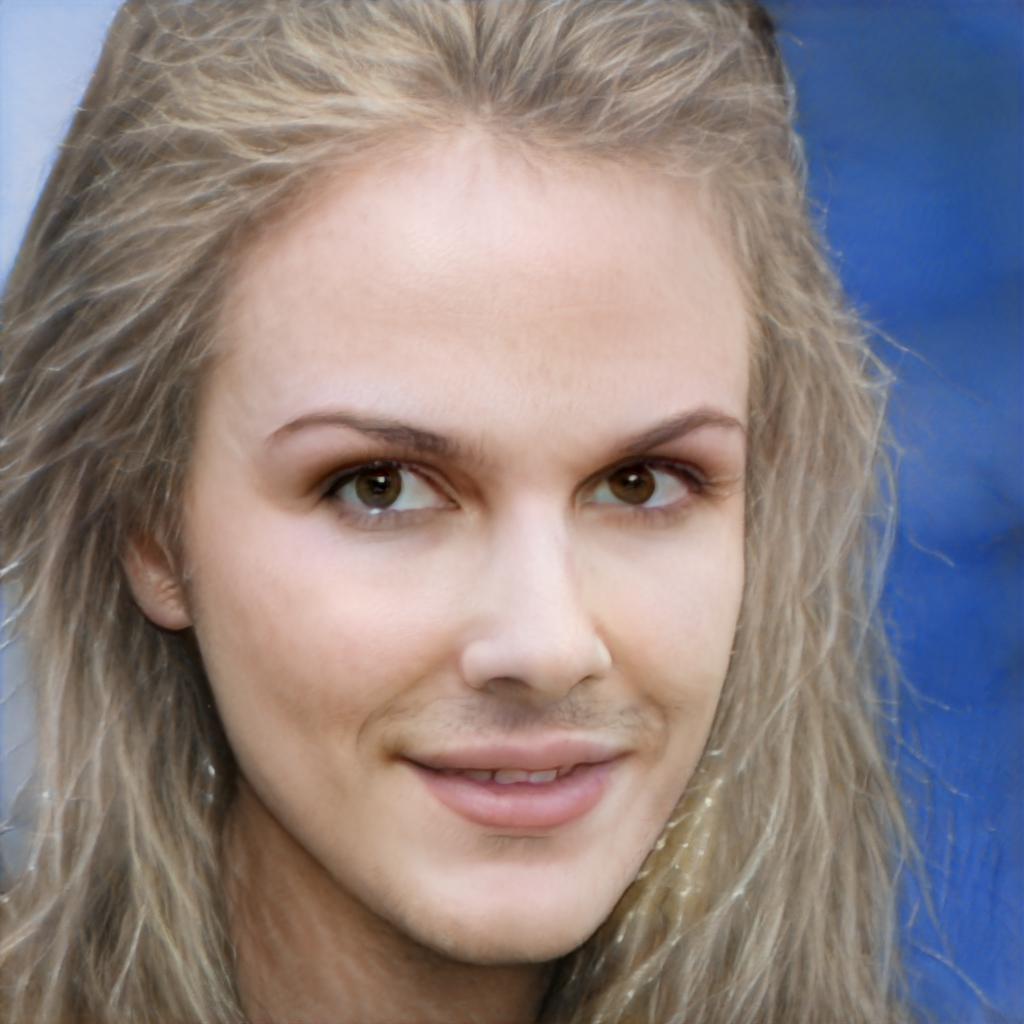} &
 \includegraphics[width=0.125\textwidth, ]{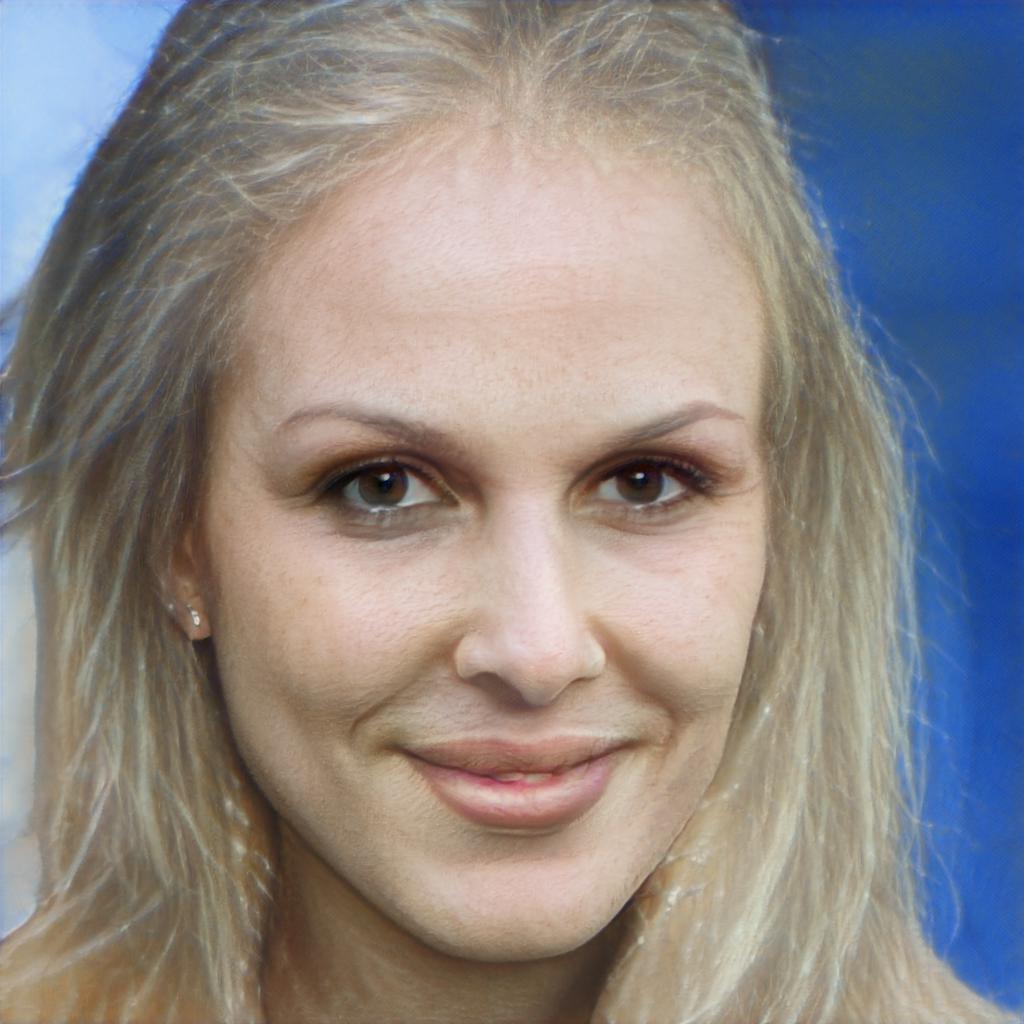} &
 \includegraphics[width=0.125\textwidth, ]{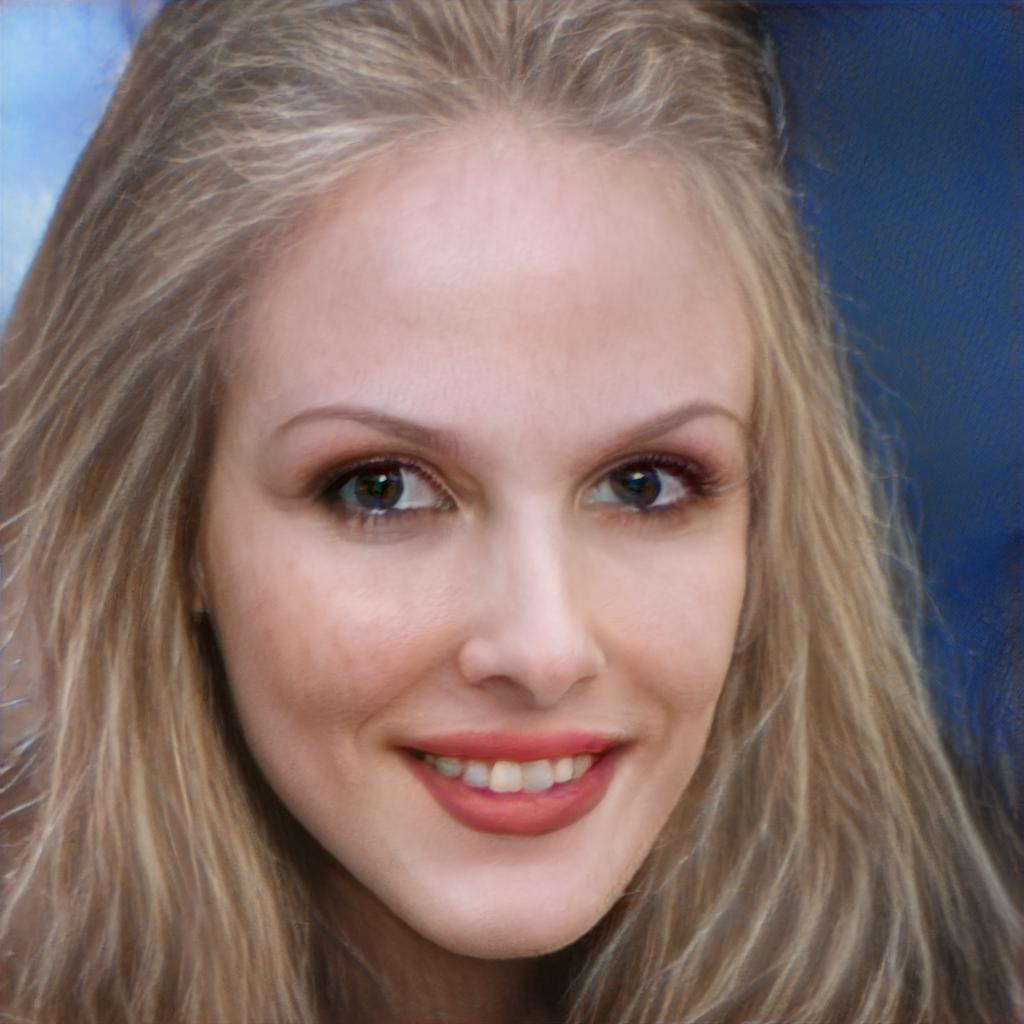} &
 \\
\end{tabular}
\caption{Image editing at resolution $1024^2$ using InterFaceGAN in \wplus  and $\mathcal{W}^{\star}_{ID}$: The images are projected in the latent space of StyleGAN2, then the latent codes are moved in the direction that corresponds to changing one facial attribute. The editing is better in terms of attributes disentanglement in the new learned space ($\mathcal{W}^{\star}_{ID}$) compared to the editing done in the original latent space (\wplussp). Our method also leads to a better identity preservation.}
\label{fig:att_inter_results}
\end{figure*}\\

\begin{figure*}[t]
\setlength\tabcolsep{2pt}%
\centering
\begin{tabular}{cc|cc|cc}
\centering
 \textbf{Original} &
 \textbf{Inverted} &
 \multicolumn{2}{c|}{Makeup} &
 \multicolumn{2}{c}{Old} & 

 \includegraphics[width=0.15\textwidth]{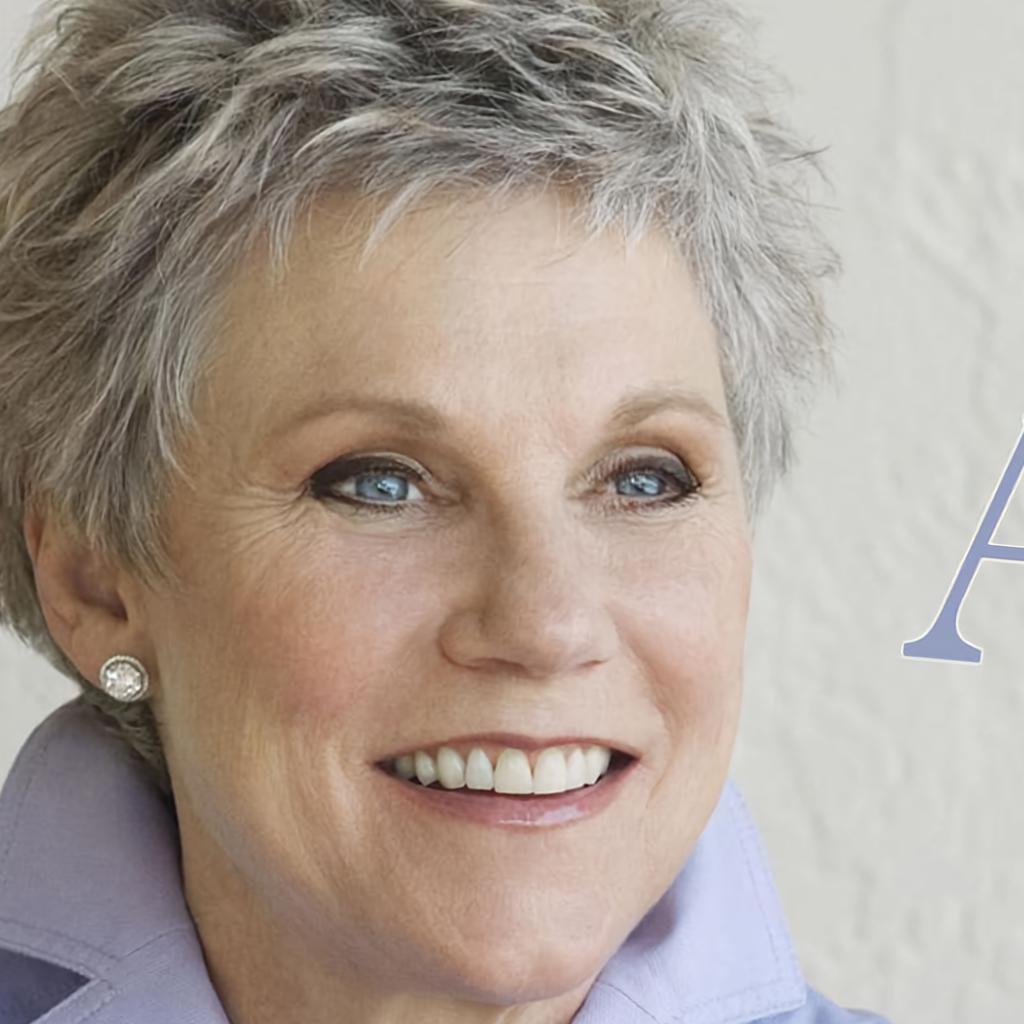} & 
 \includegraphics[width=0.15\textwidth]{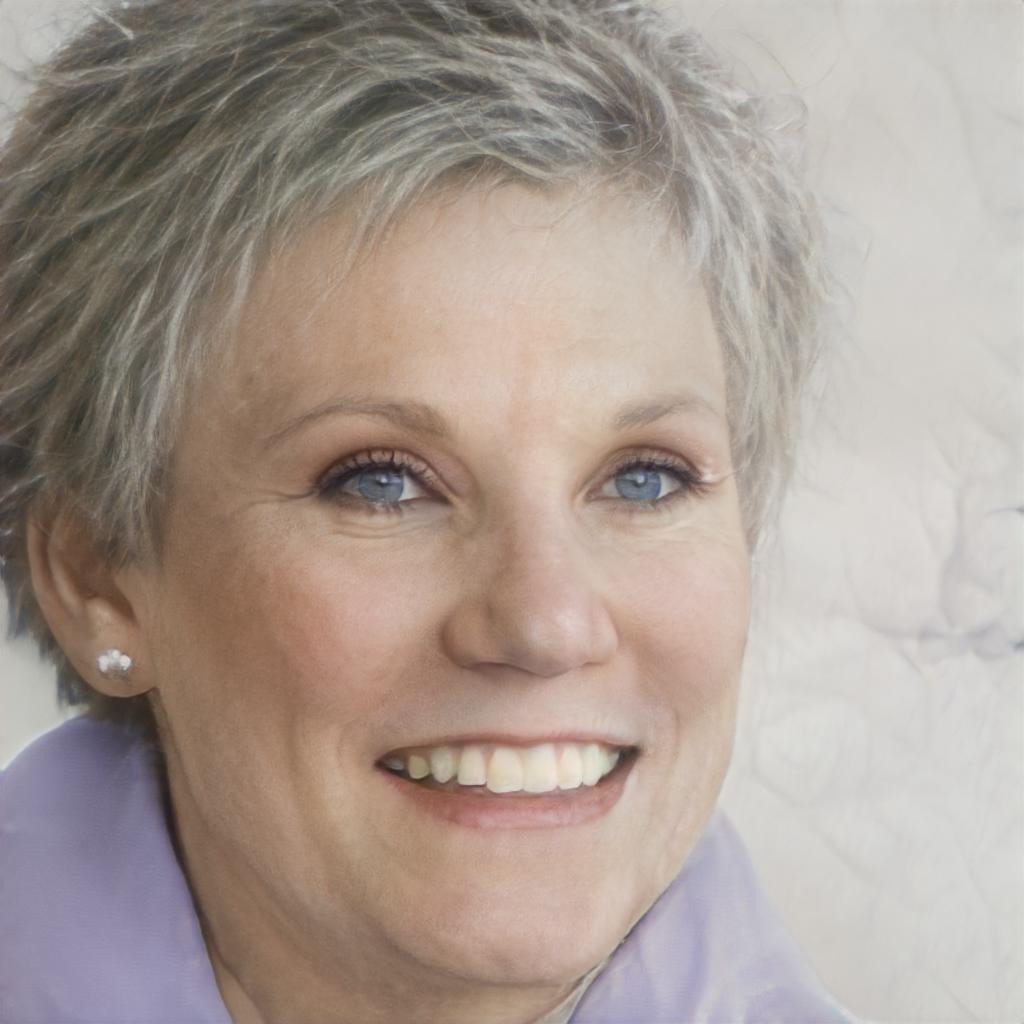} &
 \includegraphics[width=0.15\textwidth]{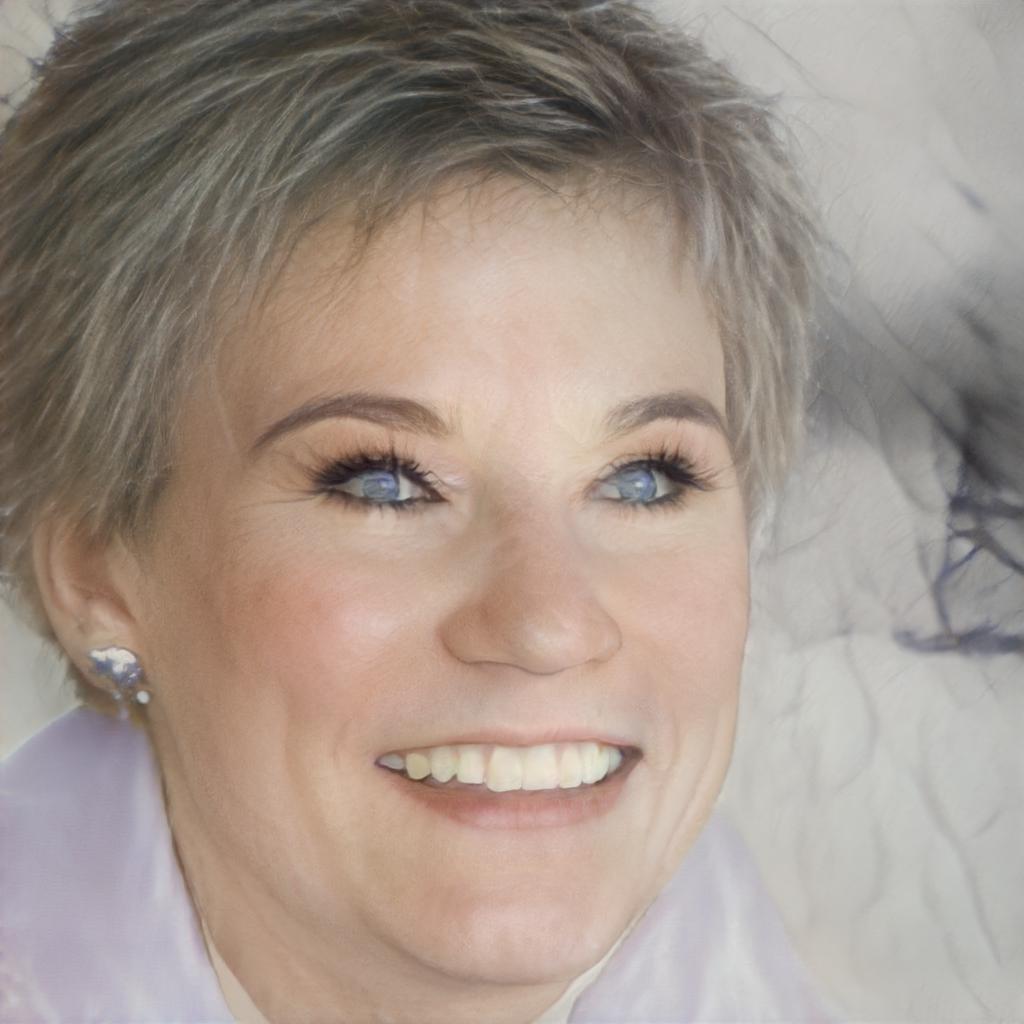} &
 \includegraphics[width=0.15\textwidth]{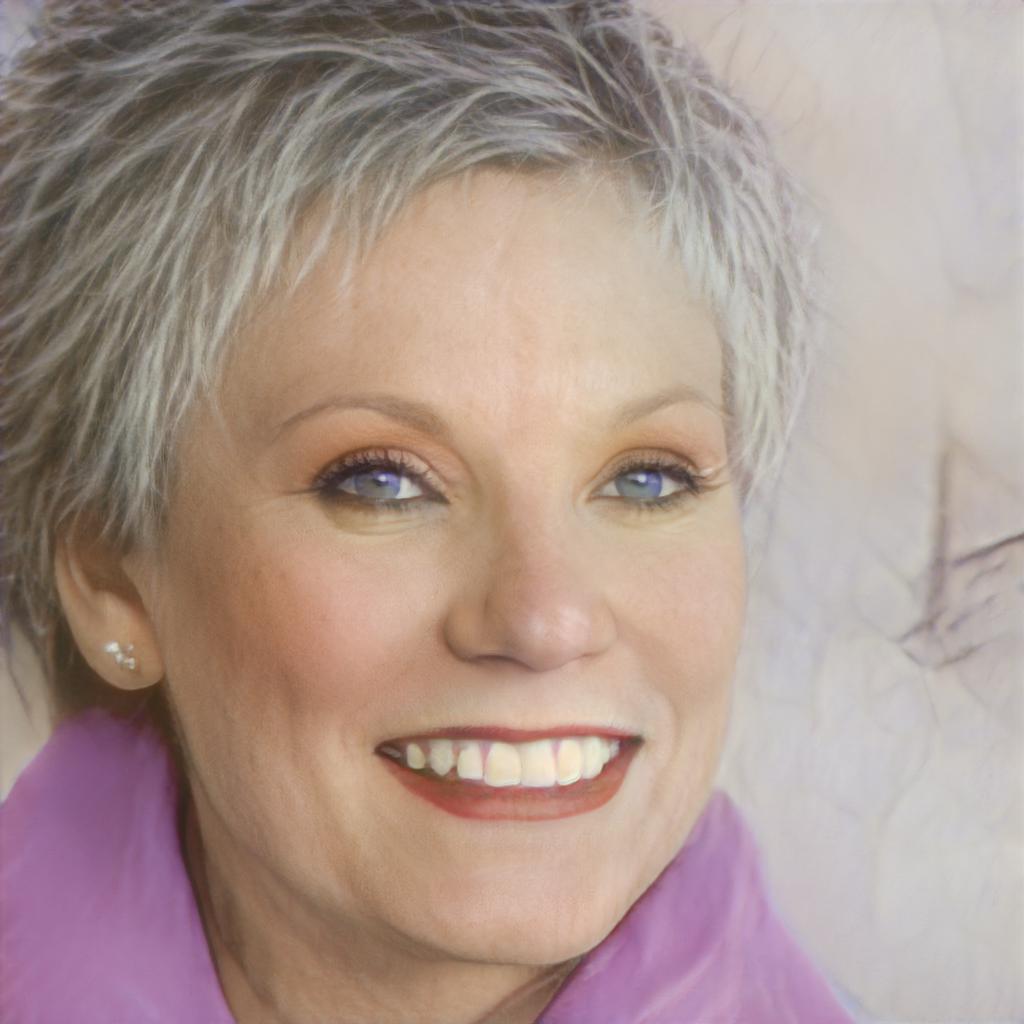} &
 \includegraphics[width=0.15\textwidth]{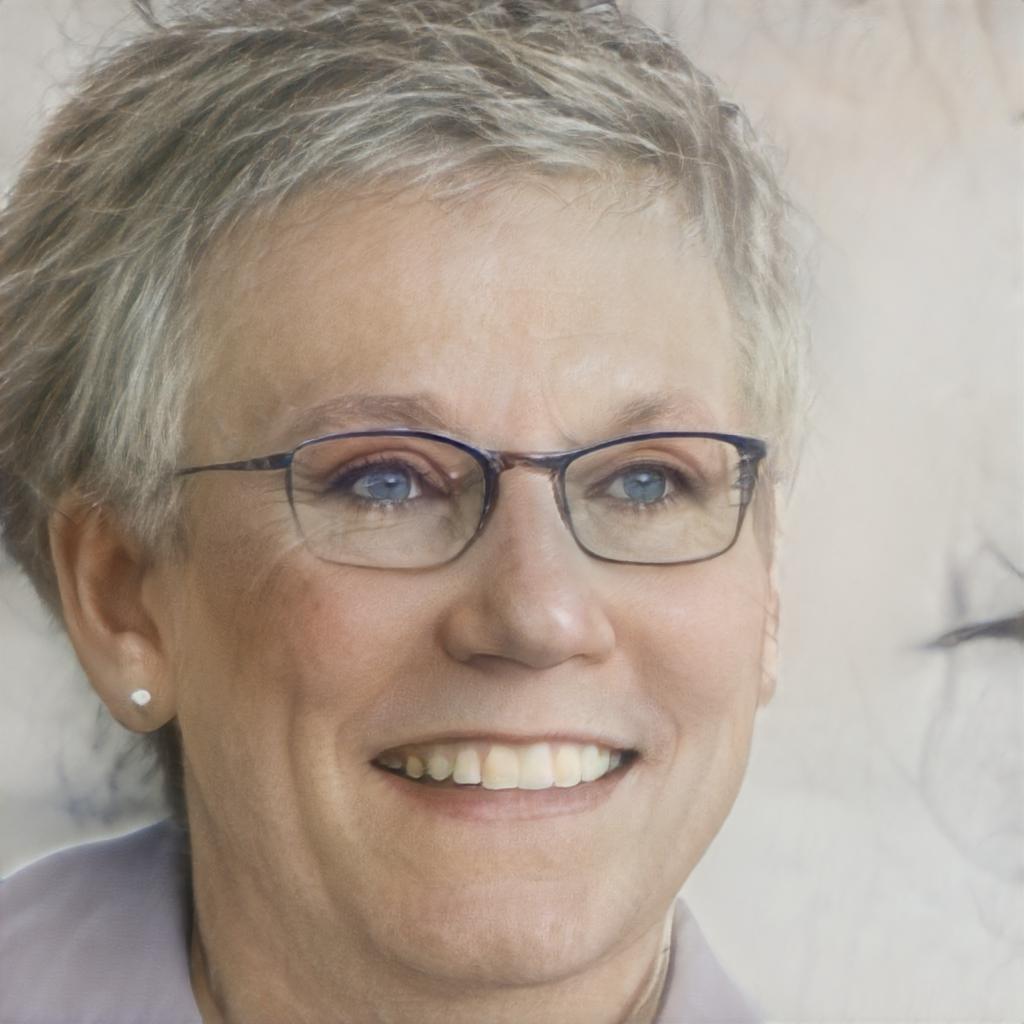} &
 \includegraphics[width=0.15\textwidth]{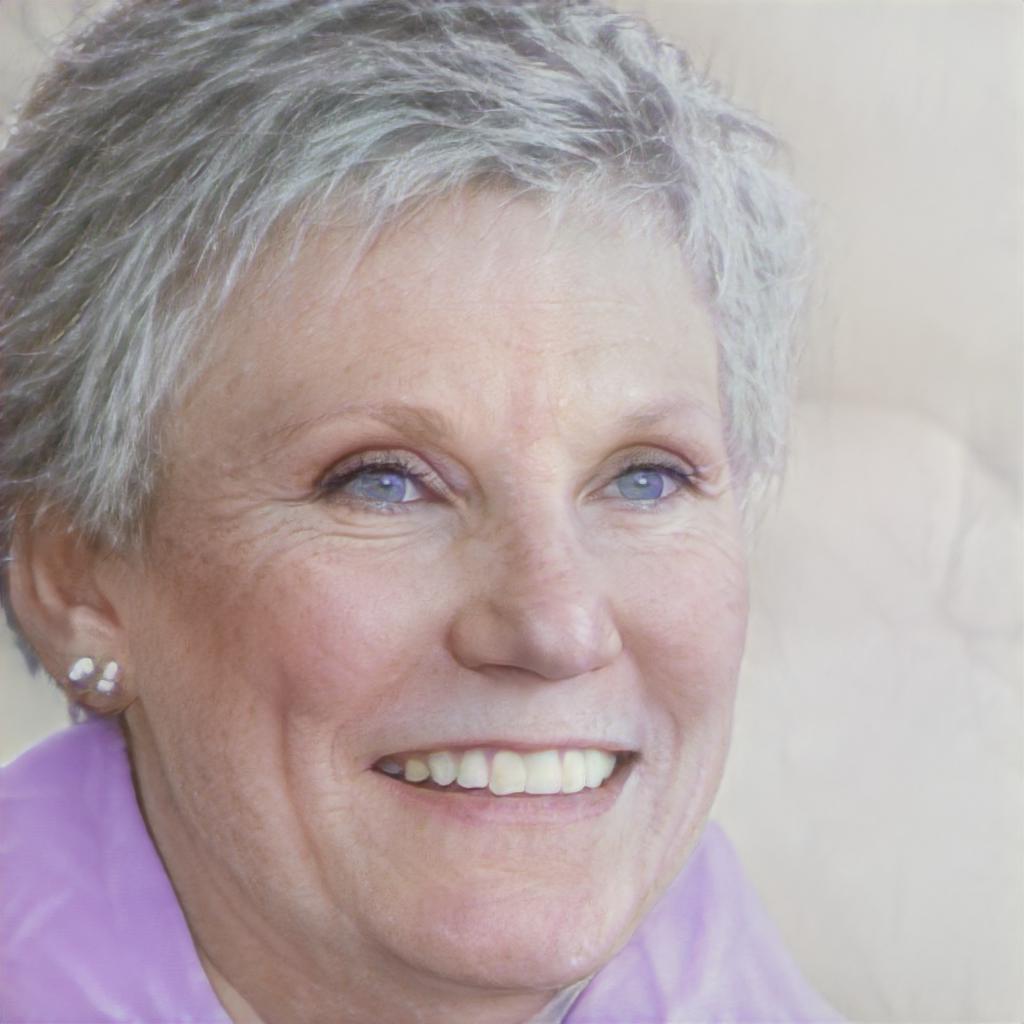} & 

 \includegraphics[width=0.15\textwidth]{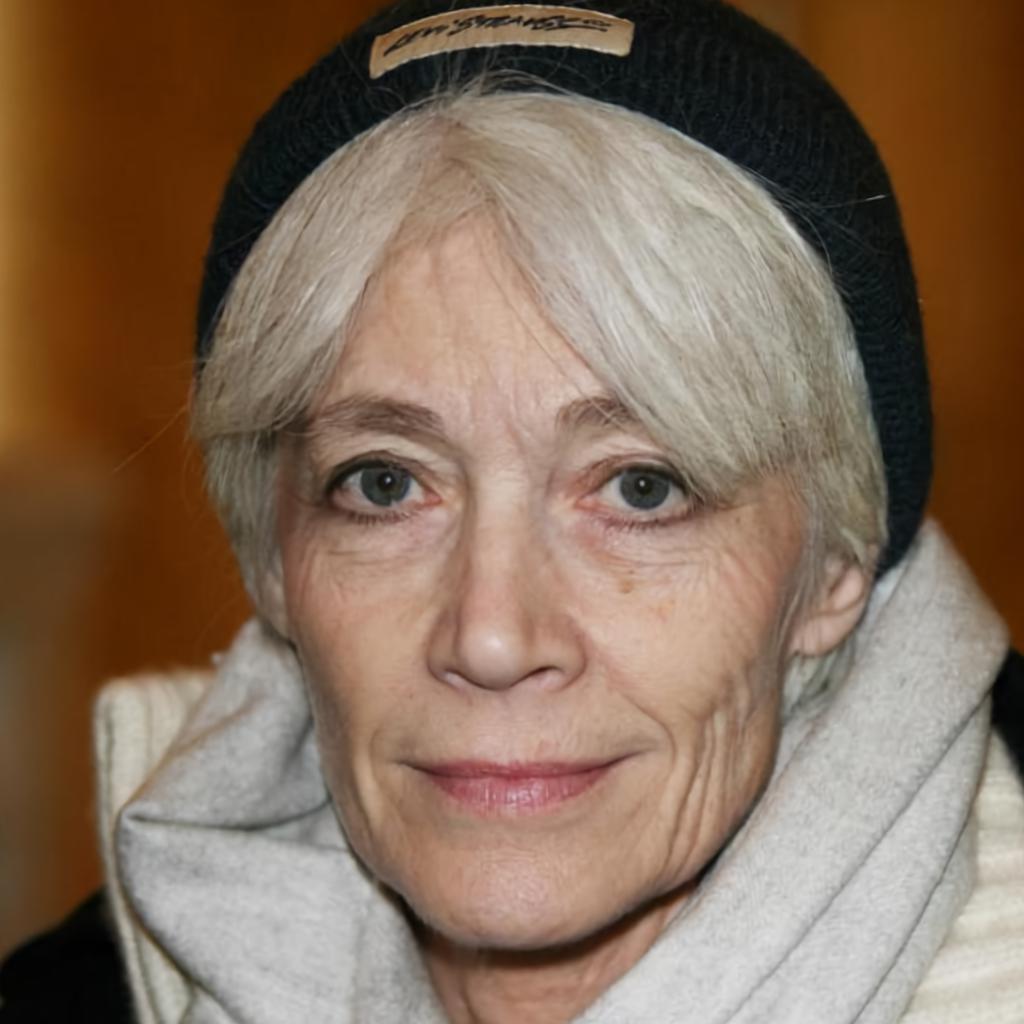} & 
 \includegraphics[width=0.15\textwidth]{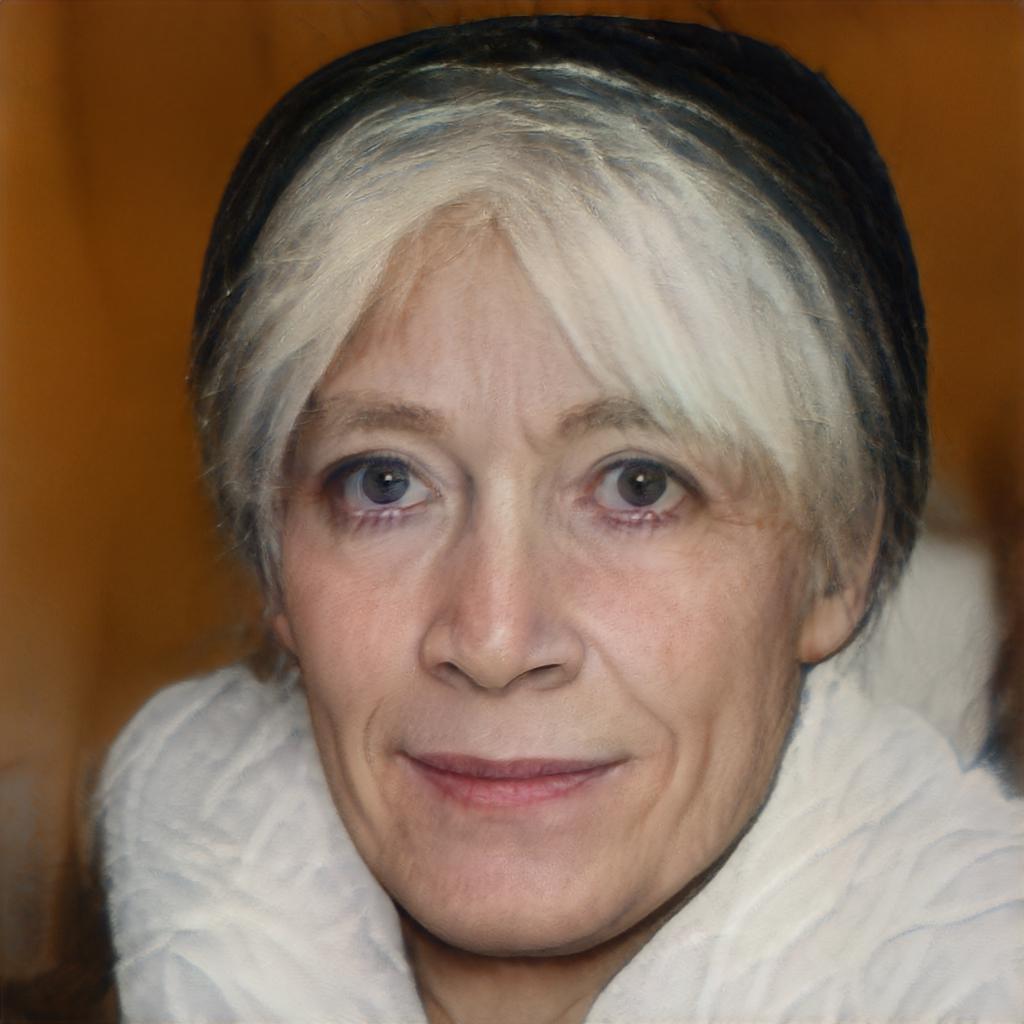} &
 \includegraphics[width=0.15\textwidth]{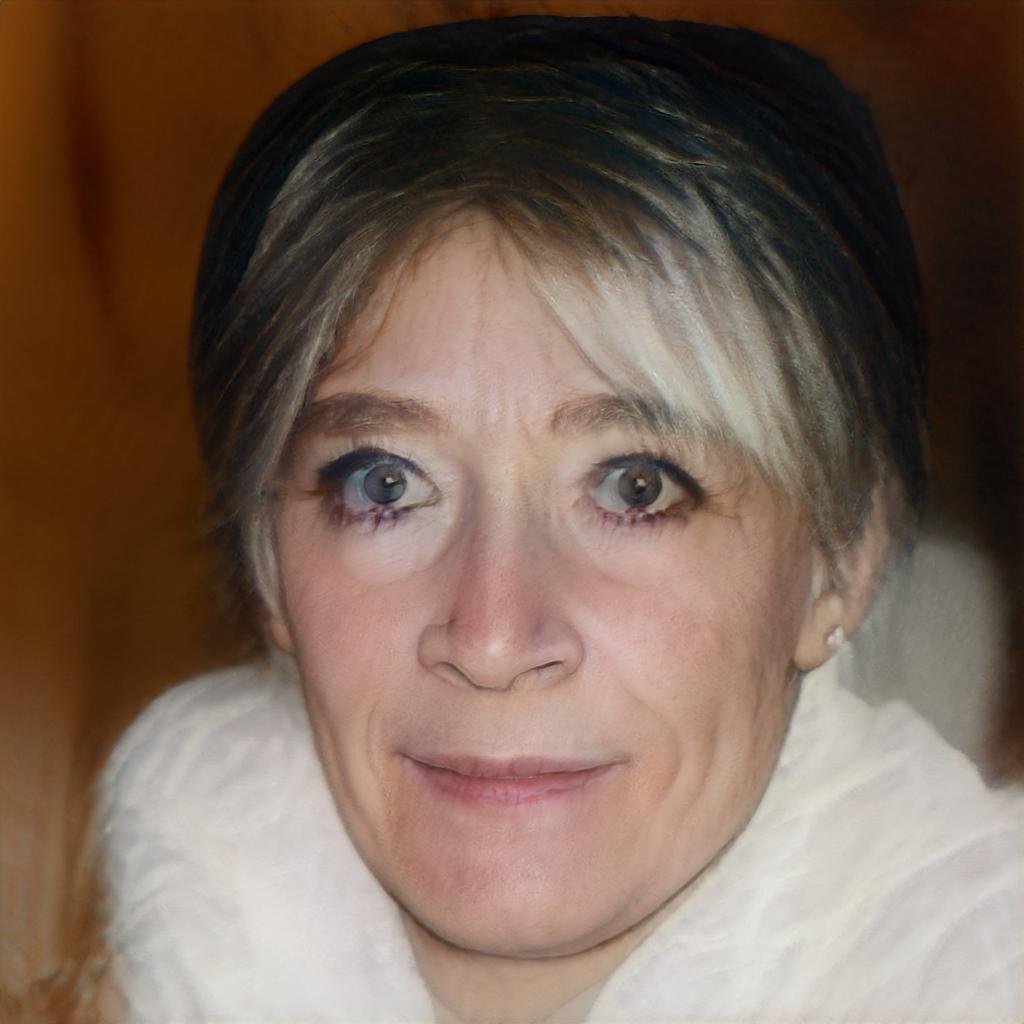} &
 \includegraphics[width=0.15\textwidth]{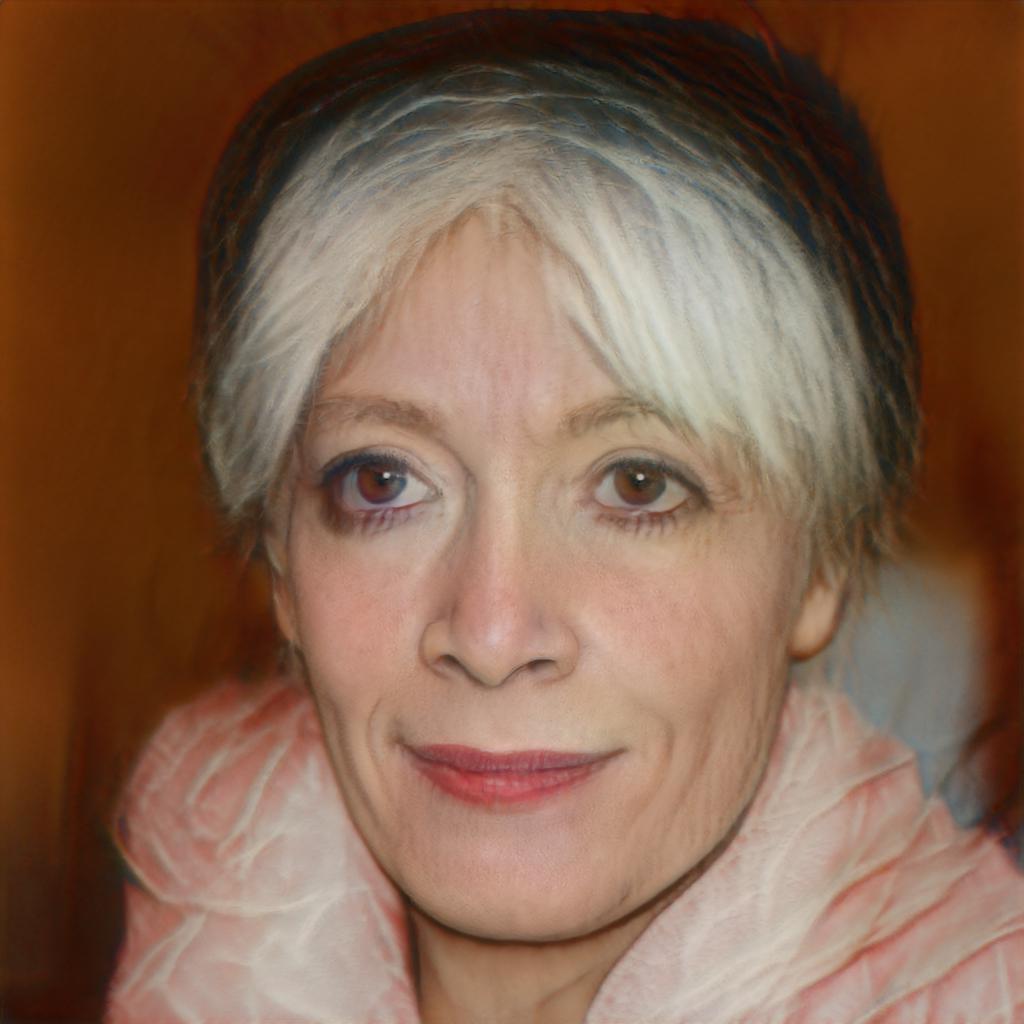} &
 \includegraphics[width=0.15\textwidth]{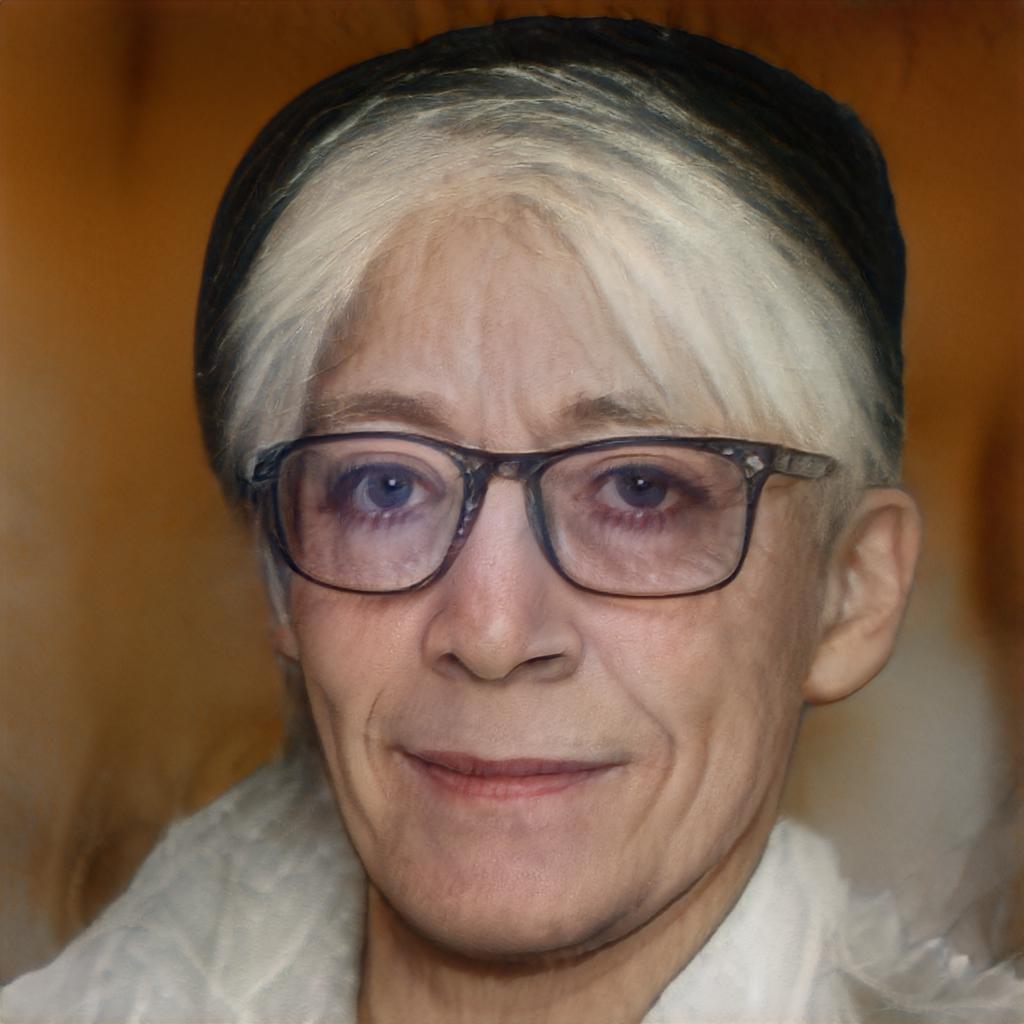} &
 \includegraphics[width=0.15\textwidth]{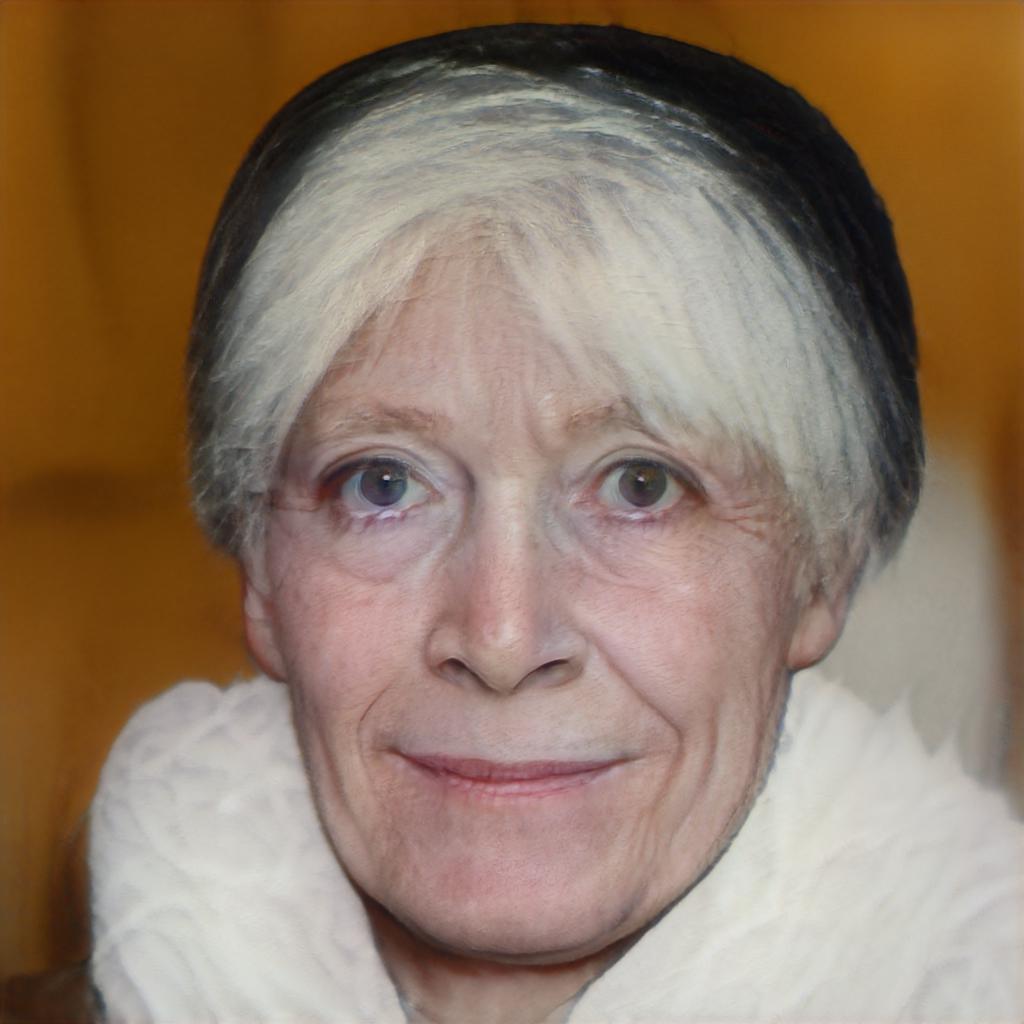} & 
  &
  &
 GANSpace &
 \wstar & 
 GANSpace &
 \wstar & 
 \\

\end{tabular}
\caption{Image editing at resolution $1024^2$ using InterFaceGAN in $\mathcal{W}^{\star}_{ID}$ and GANSpace: The editing is better in terms of attributes disentanglement in the new learned space ($\mathcal{W}^{\star}_{ID}$) compared to the editing done using GANspace.}
\label{fig:att_inter_results_vs_ganspace}
\end{figure*}\\

\begin{figure*}[h]
\setlength\tabcolsep{2pt}%
\centering
\begin{tabular}{p{0.25cm}ccccccc}
\centering
&
 \textbf{Original} &
 \textbf{Inverted} &
 \textbf{Makeup} &
 \textbf{Male} &
 \textbf{Mustache} &
 \textbf{Chubby} &
 \textbf{Lipstick} &
 \begin{turn}{90} \hspace{0.5cm} \wstara\end{turn} &
 \includegraphics[width=0.125\textwidth]{images/original/06002.jpg} & 
 \includegraphics[width=0.125\textwidth, ]{images/inversion/18_orig_img_1.jpg} &
 \includegraphics[width=0.125\textwidth, ]{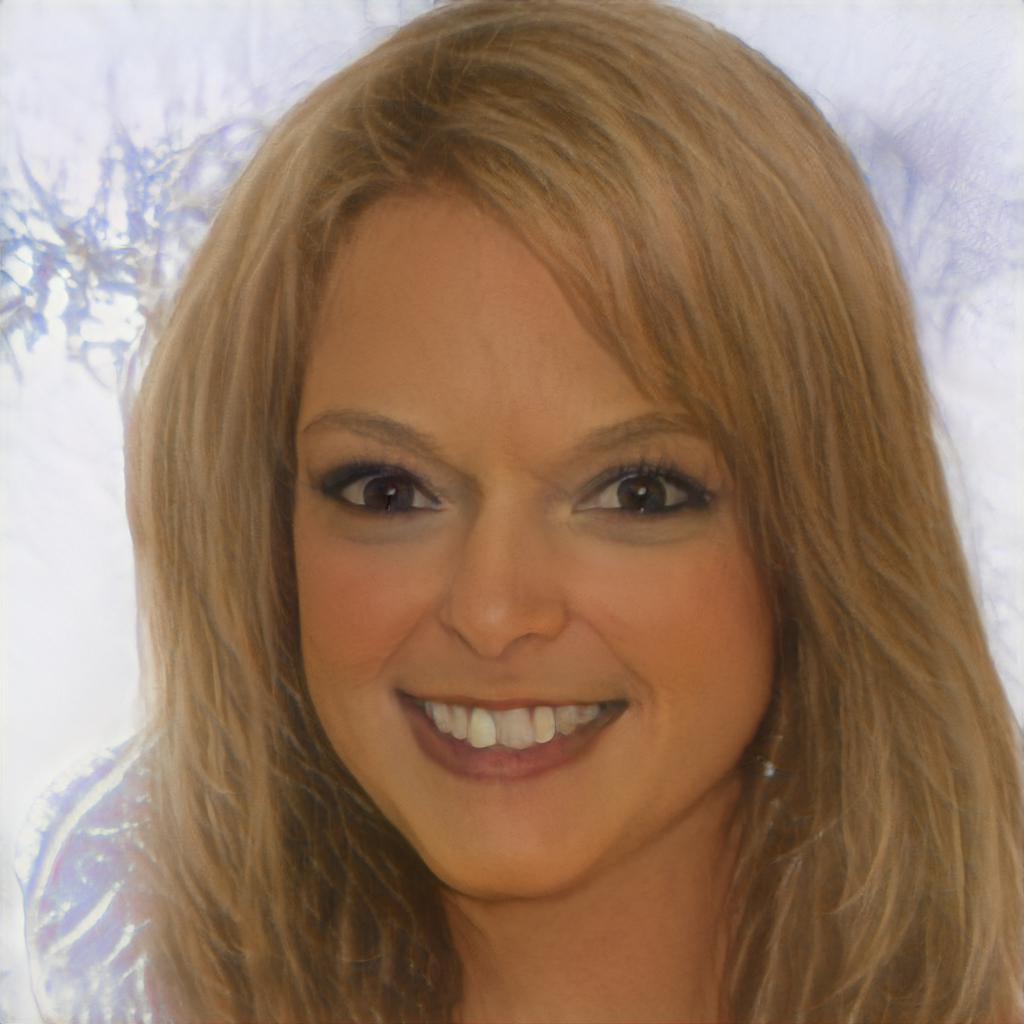} &
 \includegraphics[width=0.125\textwidth, ]{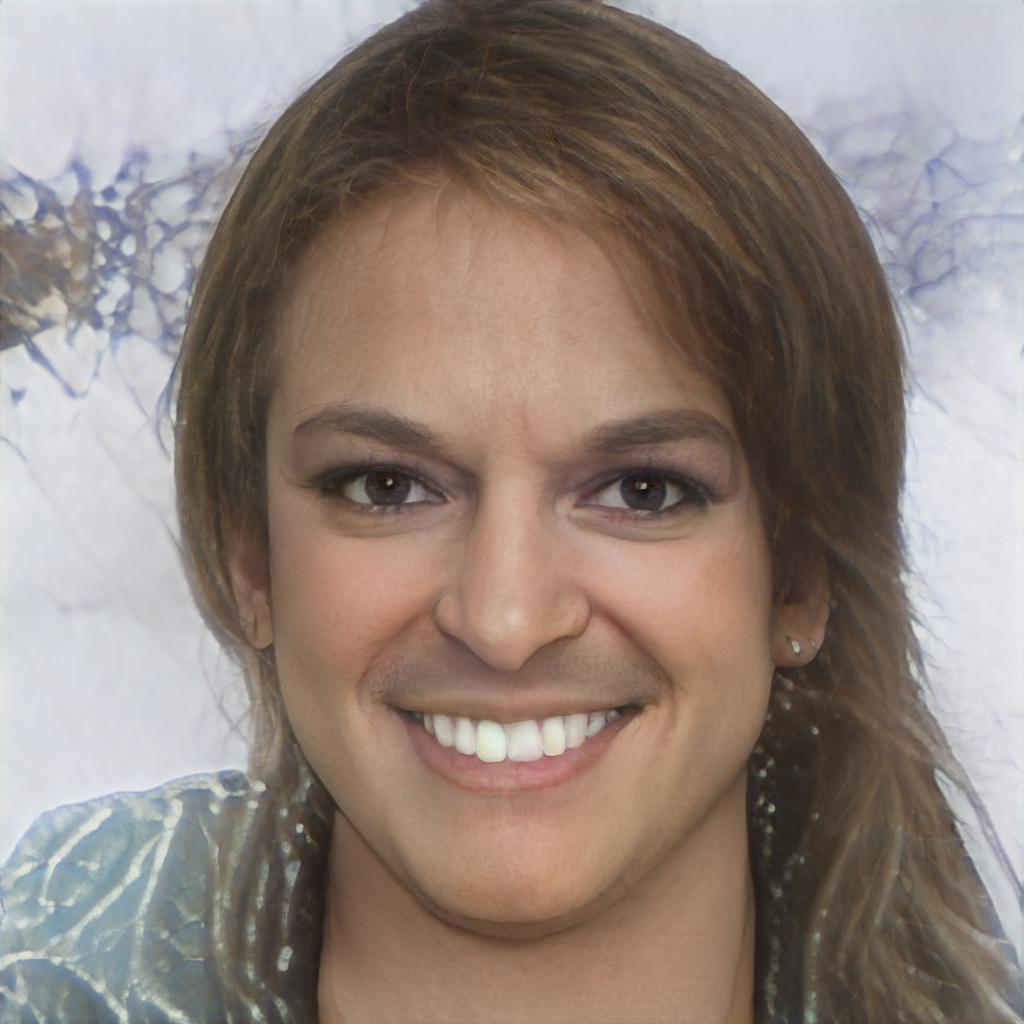} &
 \includegraphics[width=0.125\textwidth, ]{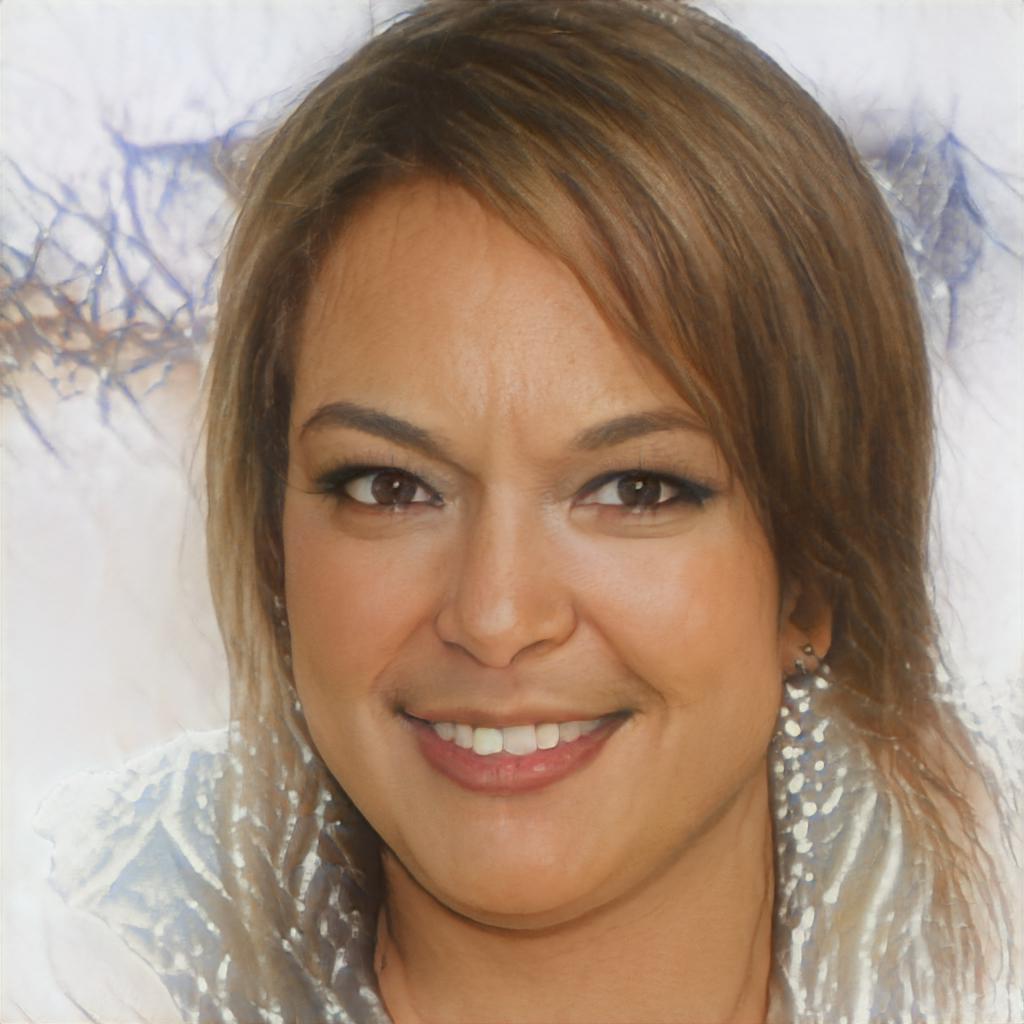} &
 \includegraphics[width=0.125\textwidth, ]{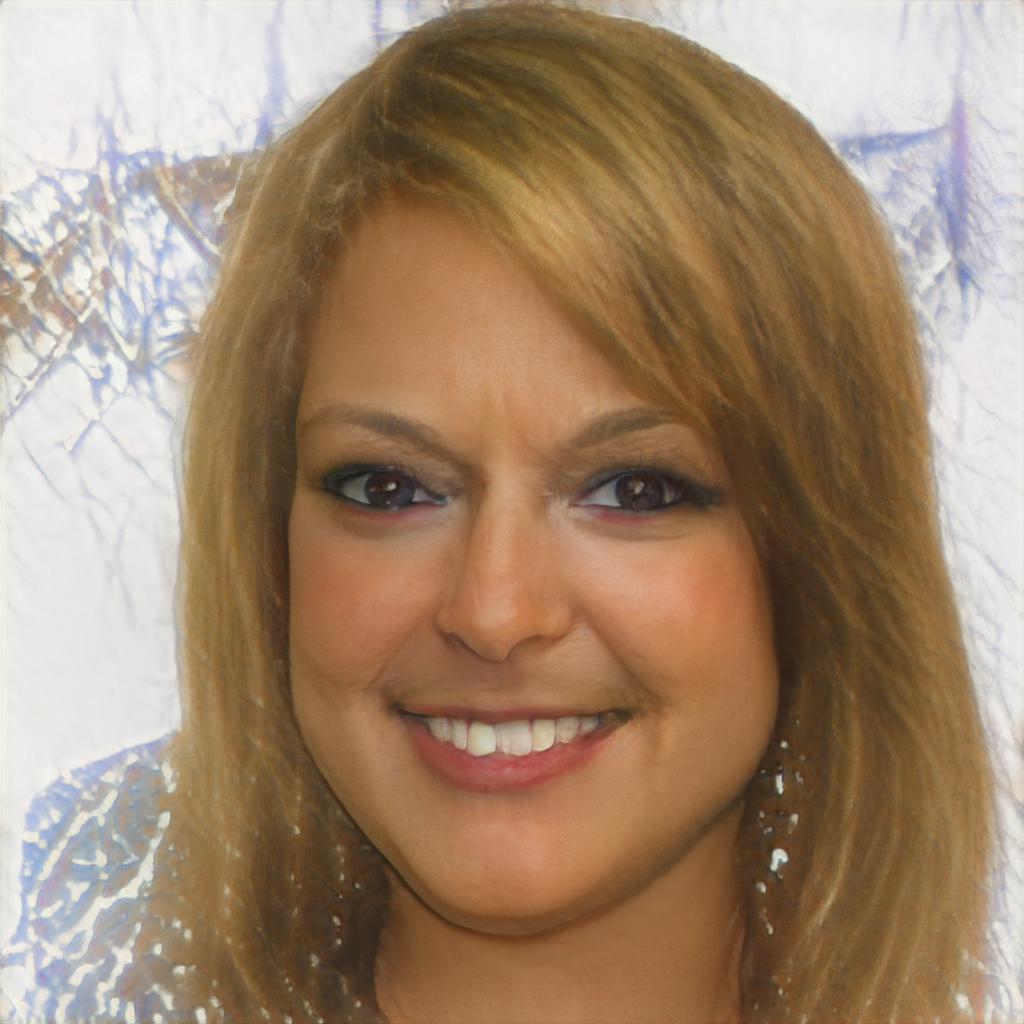} &
 \includegraphics[width=0.125\textwidth, ]{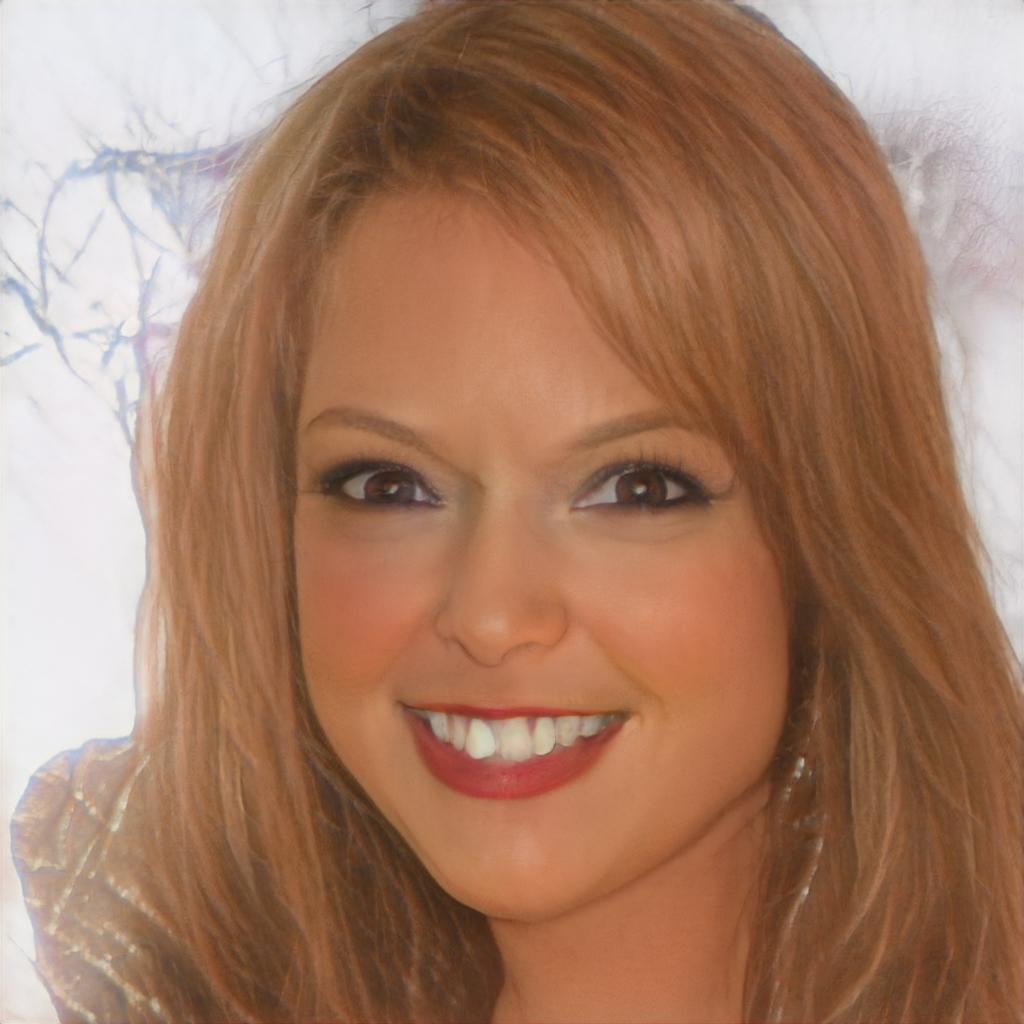} &
 \begin{turn}{90} \hspace{0.5cm} $\mathcal{W}^{\star}_{ID}$ \end{turn} &
 \includegraphics[width=0.125\textwidth]{images/original/06002.jpg} & 
 \includegraphics[width=0.125\textwidth, ]{images/inversion/18_orig_img_1.jpg} &
 \includegraphics[width=0.125\textwidth, ]{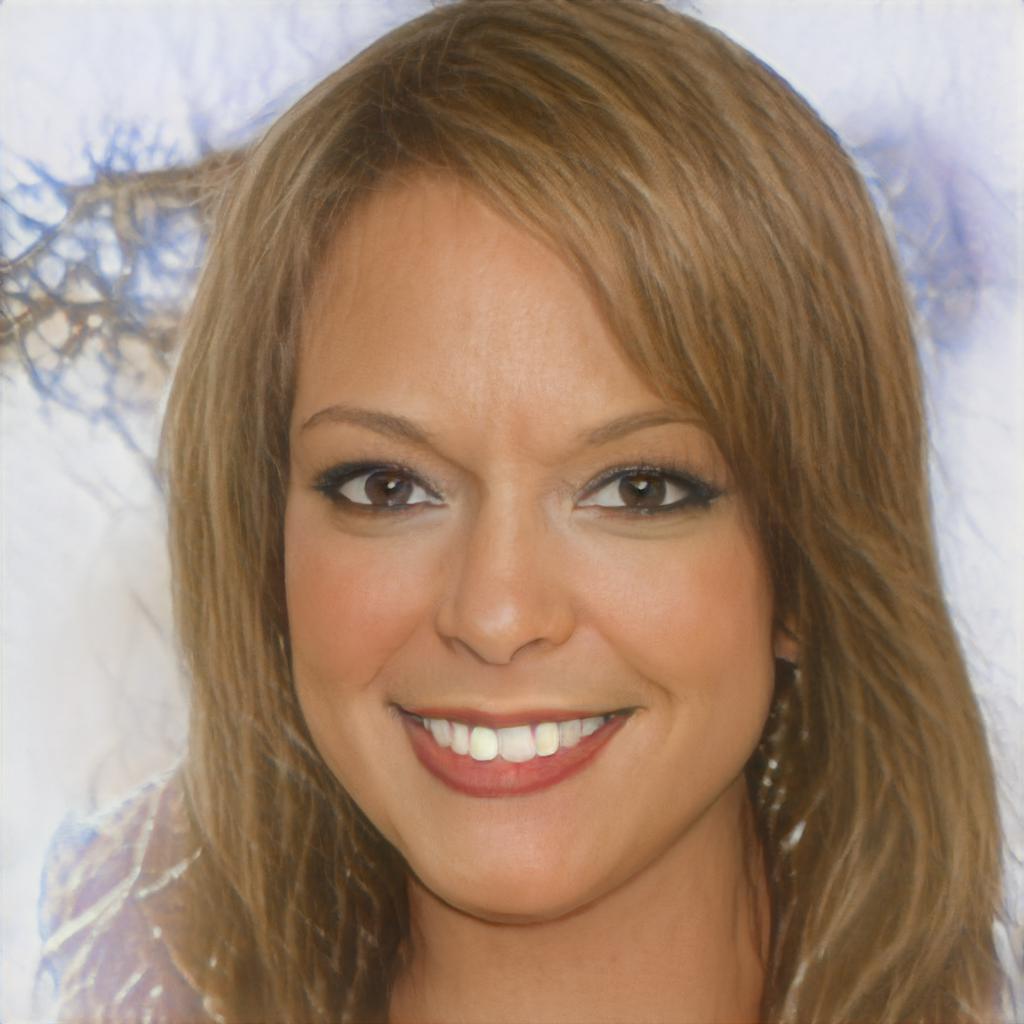} &
 \includegraphics[width=0.125\textwidth, ]{images/wstar_id_no_mag_3_coupl_dist_edit_train_ep5/20_img_1.jpg} &
 \includegraphics[width=0.125\textwidth, ]{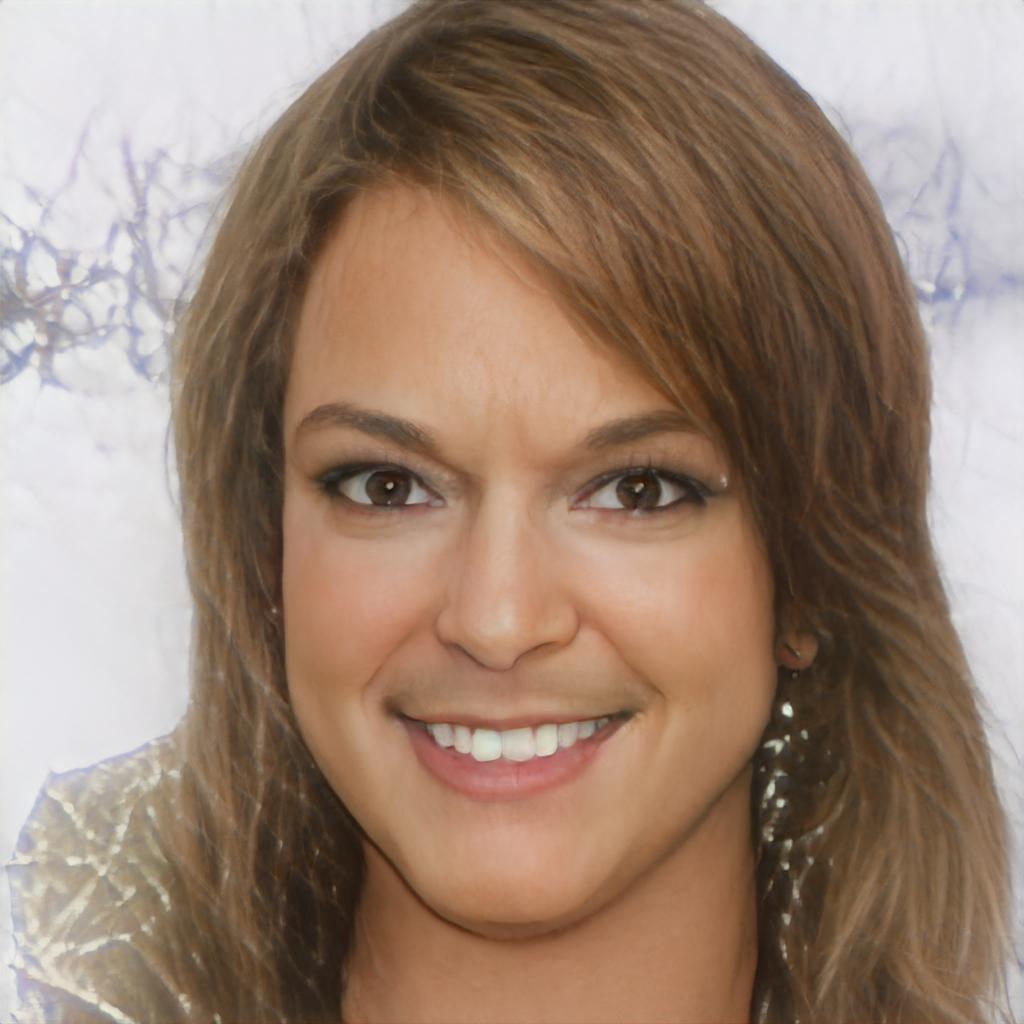} &
 \includegraphics[width=0.125\textwidth, ]{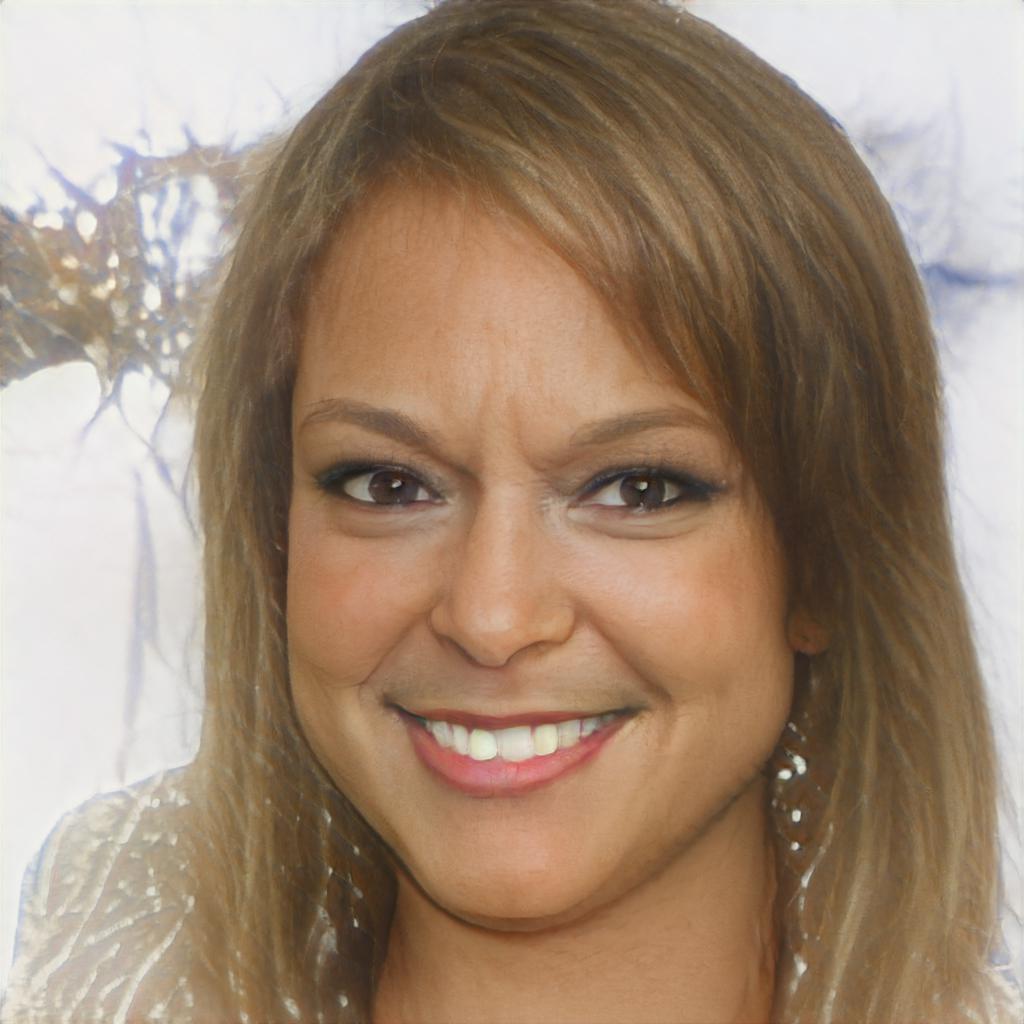} &
 \includegraphics[width=0.125\textwidth, ]{images/wstar_id_no_mag_3_coupl_dist_edit_train_ep5/36_img_1.jpg} &
 \\
\end{tabular}
\caption{Ablation study of Image editing using InterFaceGAN.Top row: model trained with only the attribute separation loss \wstarasp. The attribute separation loss alone is clearly not sufficient to obtain a controllable and identity-preserving editing.}
\label{fig:edit_abl_1}
\end{figure*}

\section{Linear Separation of the attributes}
\label{appendix:linear_sep}
\begin{figure*}[t]
\setlength\tabcolsep{2pt}%
\begin{center}
\begin{tabular}{p{0.25cm}ccc}

 &
 \textbf{Makeup} &
 \textbf{Male} &
 \textbf{Lipstick} &
 &
 Acc:0.90 &
 Acc:0.97 &
 Acc:0.94 &
 \begin{turn}{90} \hspace{1.5cm} \wplus\end{turn} &
 \includegraphics[width=0.3\textwidth]{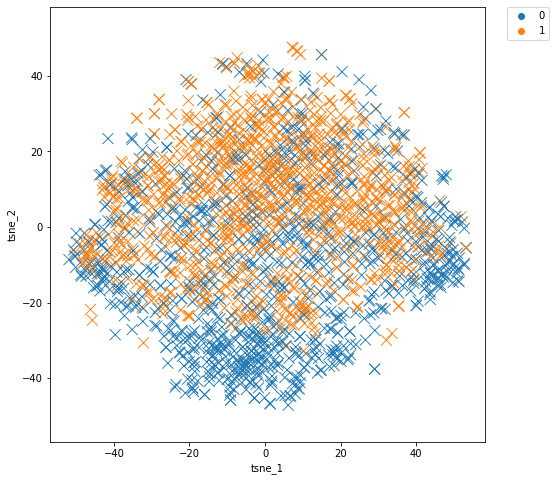} & 
 \includegraphics[width=0.3\textwidth]{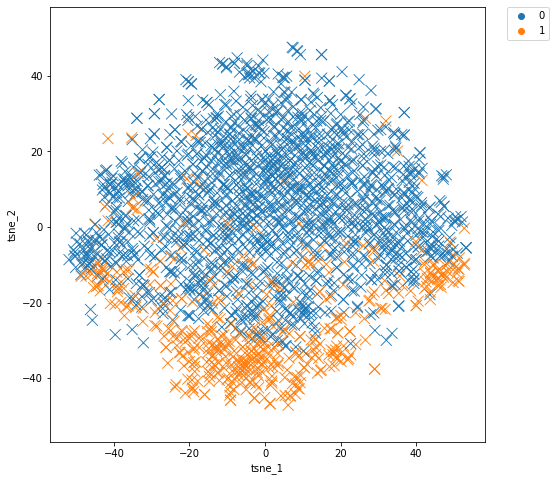} &
 \includegraphics[width=0.3\textwidth]{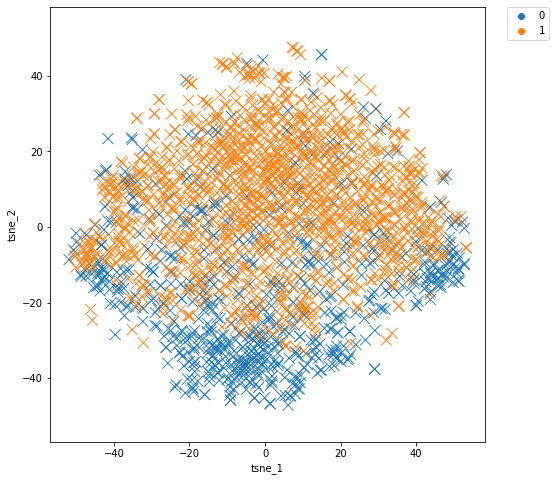} &
 &
 Acc:0.94 &
 Acc:0.98 &
 Acc:0.96 &
 \begin{turn}{90} \hspace{1.5cm} \wstar \end{turn} &
 \includegraphics[width=0.3\textwidth]{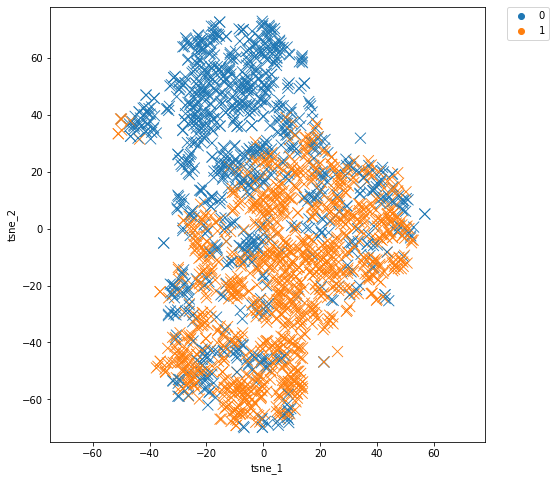} & 
 \includegraphics[width=0.3\textwidth]{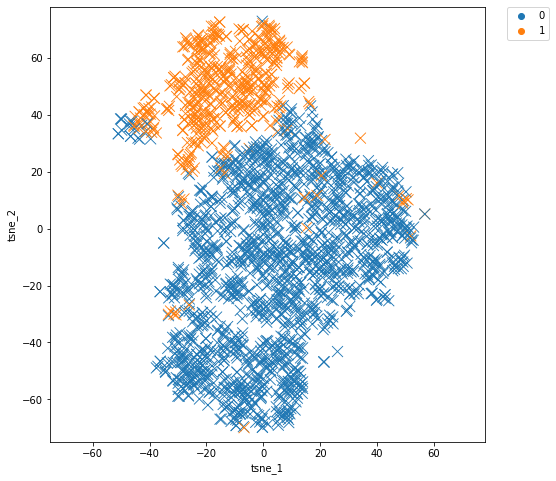} &
 \includegraphics[width=0.3\textwidth]{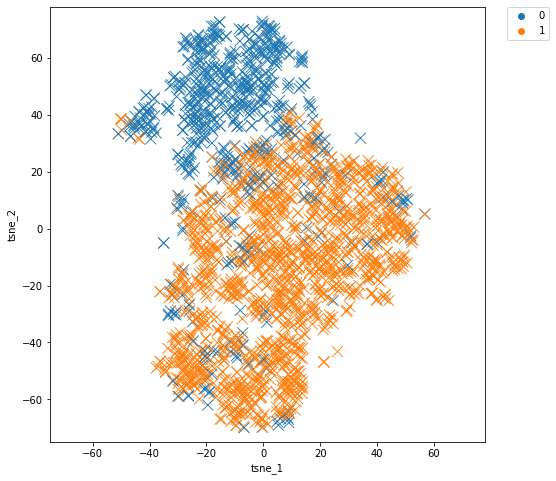} &
 
\end{tabular}
\end{center}
\caption{t-SNE visualization in \wplus and \wstarsp: the classification accuracy is shown at the top (Acc). the 2D embedding enables to visualize that the attributes are better linearly disentangled in \wstarsp.}
\label{fig:tsne}

\end{figure*}

In this section we use t-SNE \footnote{Used from scikit-learn library with default parameters: \href{https://scikit-learn.org/stable/modules/generated/sklearn.manifold.TSNE.html}{https://scikit-learn.org/stable/modules/generated/sklearn.manifold.TSNE.html}} \cite{tsne} to visualize the negative and positive examples for some attributes in both latent spaces (\ie \wstar and \wplussp), embedded onto the $2D$ plane. The training setup is the same as in Section 4.1 and 4.3.1. The dataset is composed of $2000$ latent codes averaged over the $18$ dimensions. \\
From Figure \ref{fig:tsne}, one can easily visualise that negative and positive attributes are more linearly separated in \wstar than in \wplus. 

\section{Ablation Study}
\label{appendix:ablation}

\subsection{Qualitative evaluation}

\begin{figure*}[h]
\setlength\tabcolsep{2pt}%
\centering
\begin{tabular}{p{0.25cm}ccccccc}
\centering
&
 \textbf{Original} &
 \textbf{Inverted} &
 \textbf{Makeup} &
 \textbf{Male} &
 \textbf{Mustache} &
 \textbf{Chubby} &
 \textbf{Lipstick} &
  \begin{turn}{90} \hspace{0.5cm} \wplus\end{turn} &
 \includegraphics[width=0.13\textwidth]{images/original/06002.jpg} & 
 \includegraphics[width=0.13\textwidth, ]{images/inversion/18_orig_img_1.jpg} &
 \includegraphics[width=0.13\textwidth, ]{images/wplus/18_img_3.jpg} &
 \includegraphics[width=0.13\textwidth, ]{images/wplus/20_img_3.jpg} &
 \includegraphics[width=0.13\textwidth, ]{images/wplus/22_img_3.jpg} &
 \includegraphics[width=0.13\textwidth, ]{images/wplus/13_img_3.jpg} &
 \includegraphics[width=0.13\textwidth]{images/wplus/36_img_3.jpg} &
 \begin{turn}{90} \hspace{0.5cm} \wstara\end{turn} &
 \includegraphics[width=0.13\textwidth]{images/original/06002.jpg} & 
 \includegraphics[width=0.13\textwidth, ]{images/inversion/18_orig_img_1.jpg} &
 \includegraphics[width=0.13\textwidth, ]{images/wstar_no_mag_low_step/18_img_1.jpg} &
 \includegraphics[width=0.13\textwidth, ]{images/wstar_no_mag_low_step/20_img_1.jpg} &
 \includegraphics[width=0.13\textwidth, ]{images/wstar_no_mag_low_step/22_img_1.jpg} &
 \includegraphics[width=0.13\textwidth, ]{images/wstar_no_mag_low_step/13_img_1.jpg} &
 \includegraphics[width=0.13\textwidth, ]{images/wstar_no_mag_low_step/36_img_1.jpg} &
\begin{turn}{90} \hspace{0.2cm} \wstara-ID\end{turn} &
 \includegraphics[width=0.13\textwidth]{images/original/06002.jpg} & 
 \includegraphics[width=0.13\textwidth, ]{images/inversion/18_orig_img_1.jpg} &
 \includegraphics[width=0.13\textwidth, ]{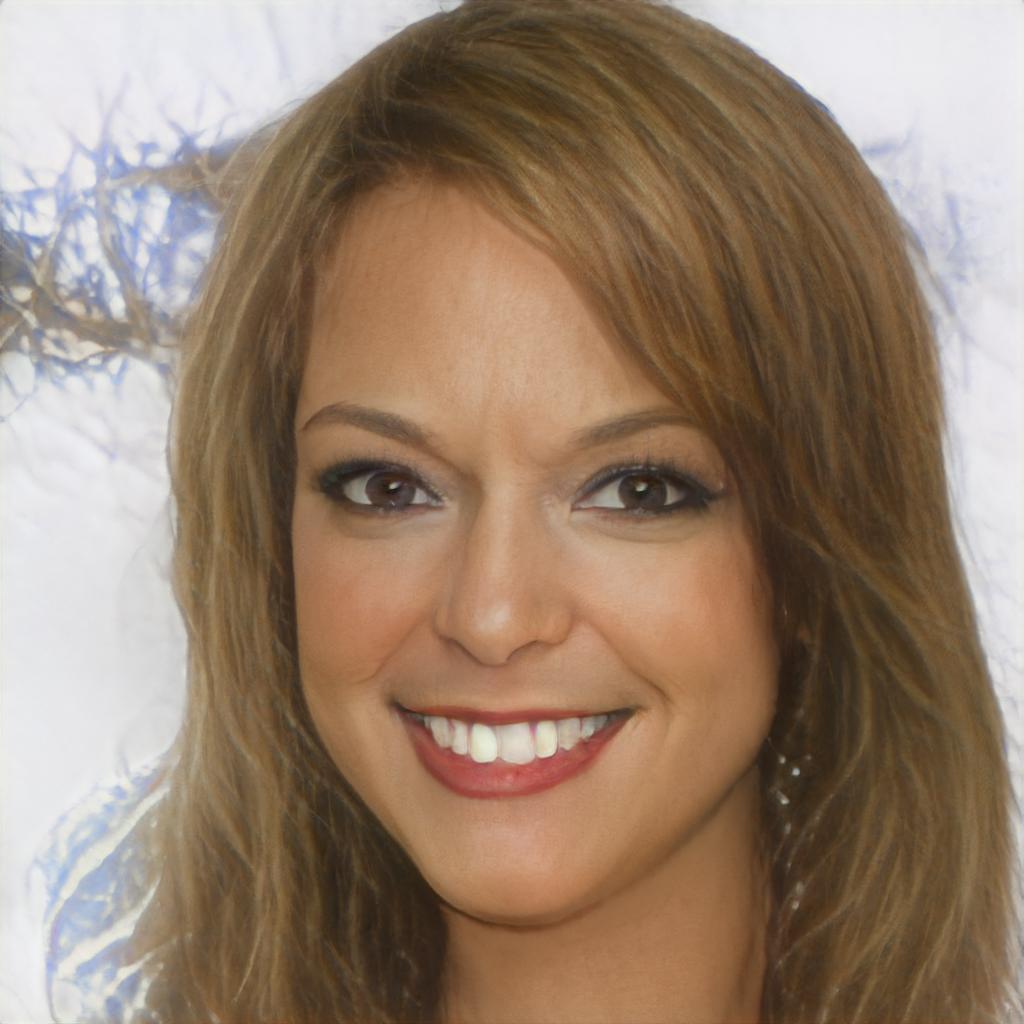} &
 \includegraphics[width=0.13\textwidth, ]{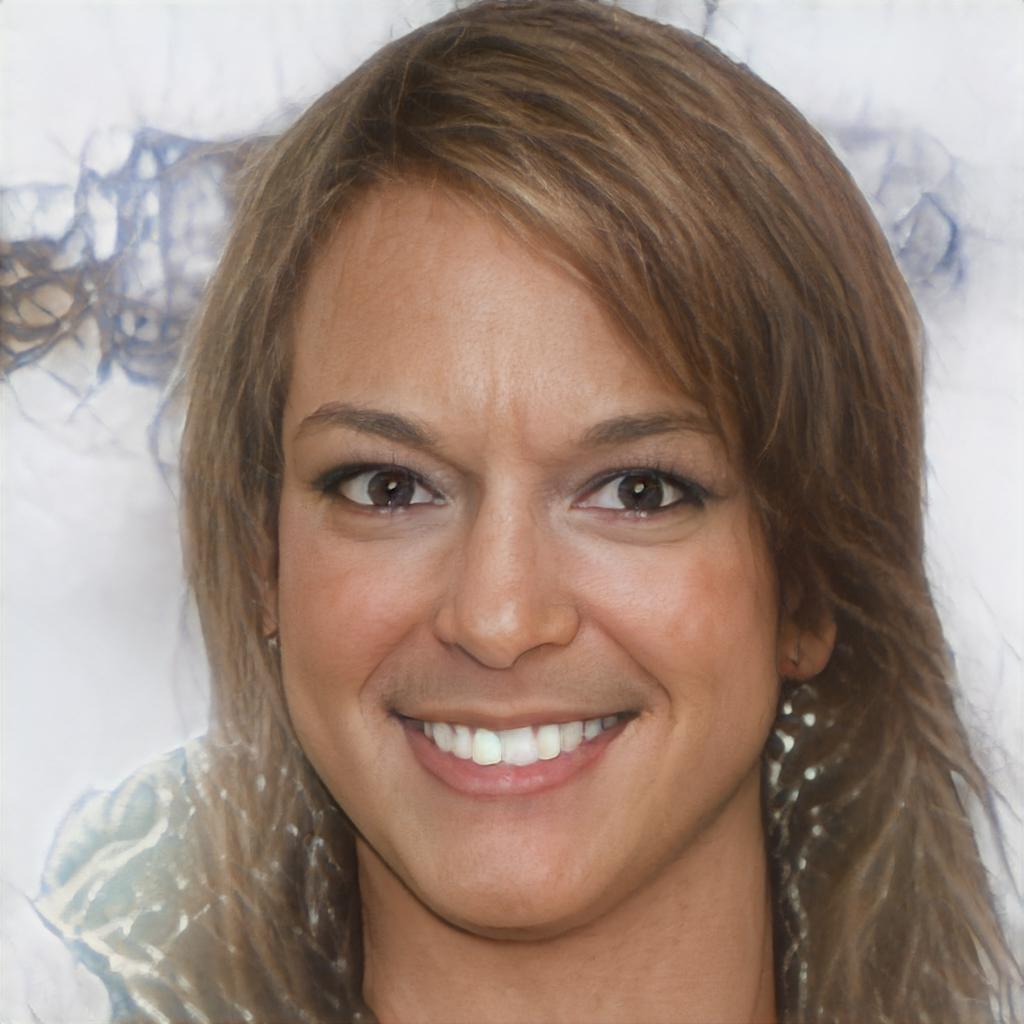} &
 \includegraphics[width=0.13\textwidth, ]{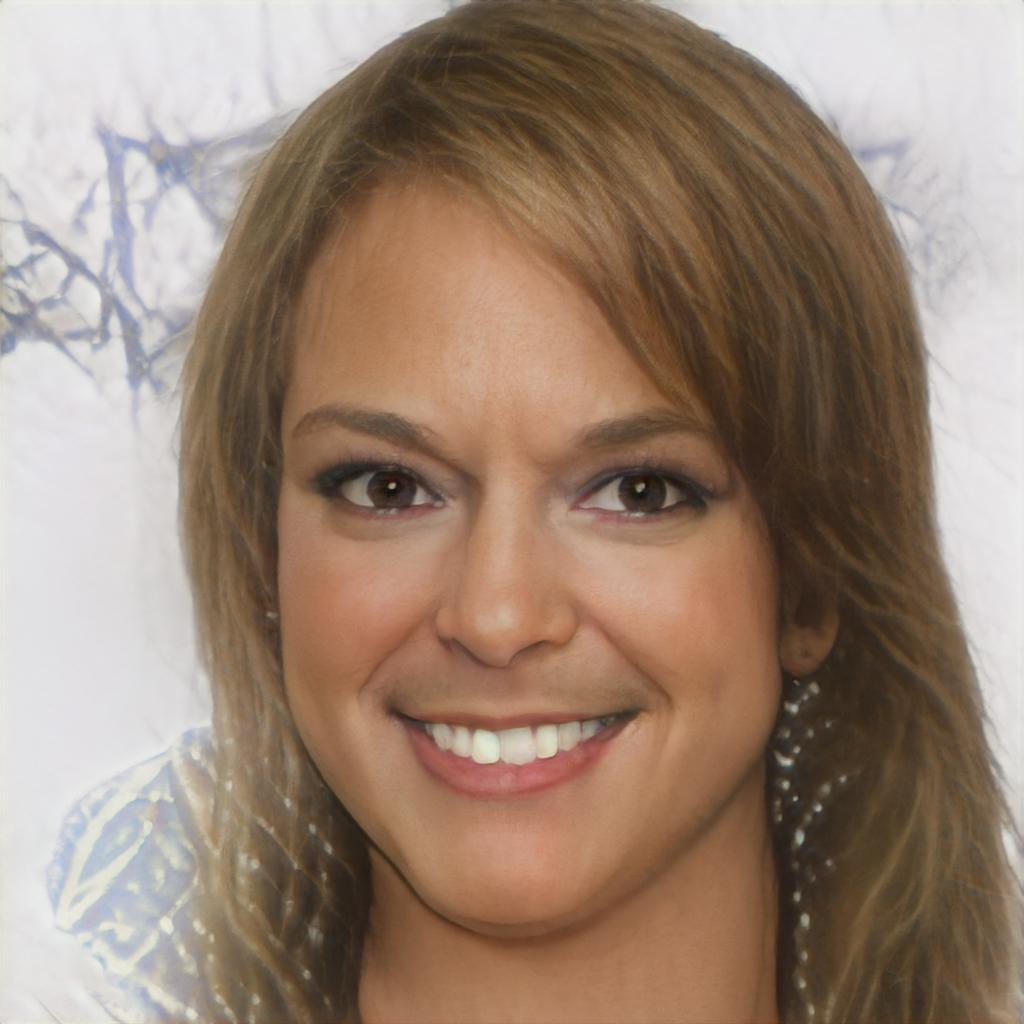} &
 \includegraphics[width=0.13\textwidth, ]{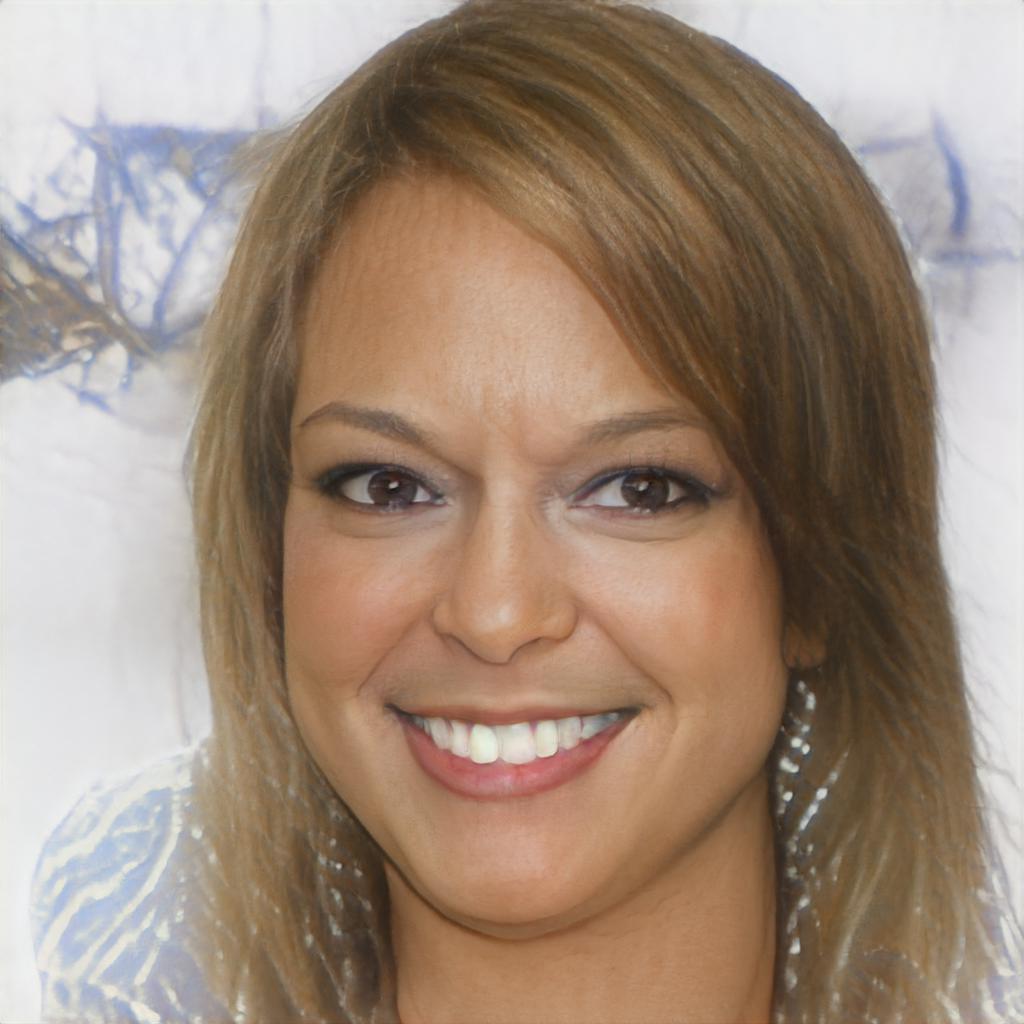} &
 \includegraphics[width=0.13\textwidth, ]{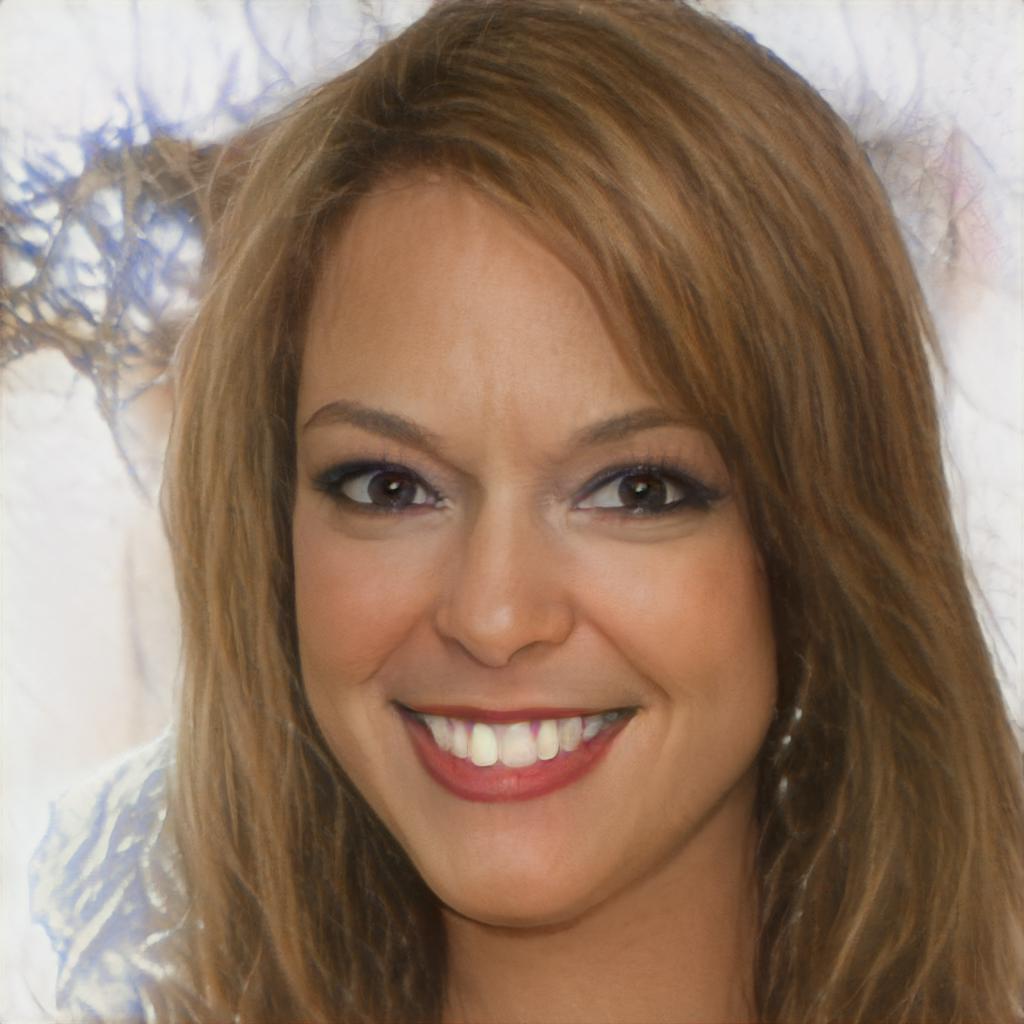} &
 \begin{turn}{90}  \wstara-ID^{\dagger}\end{turn} &
 \includegraphics[width=0.13\textwidth]{images/original/06002.jpg} & 
 \includegraphics[width=0.13\textwidth, ]{images/inversion/18_orig_img_1.jpg} &
 \includegraphics[width=0.13\textwidth, ]{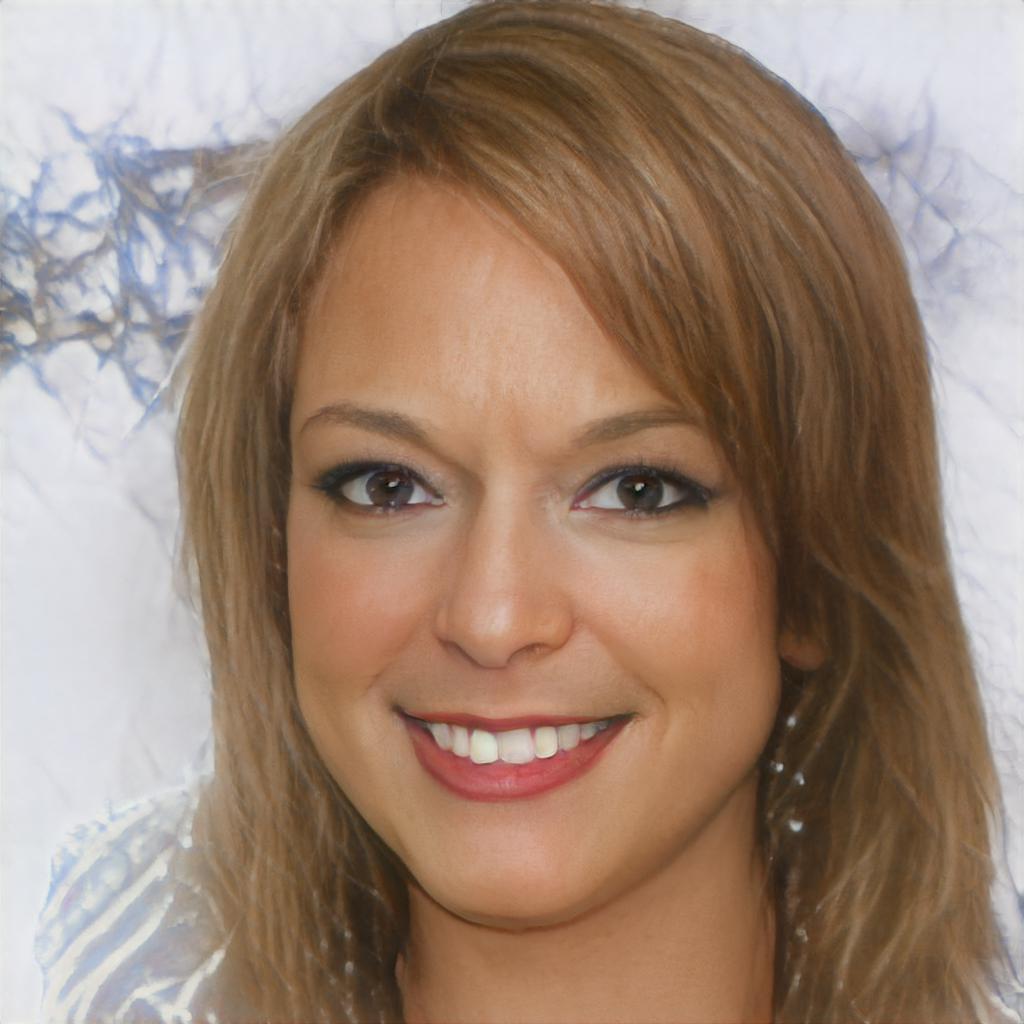} &
 \includegraphics[width=0.13\textwidth, ]{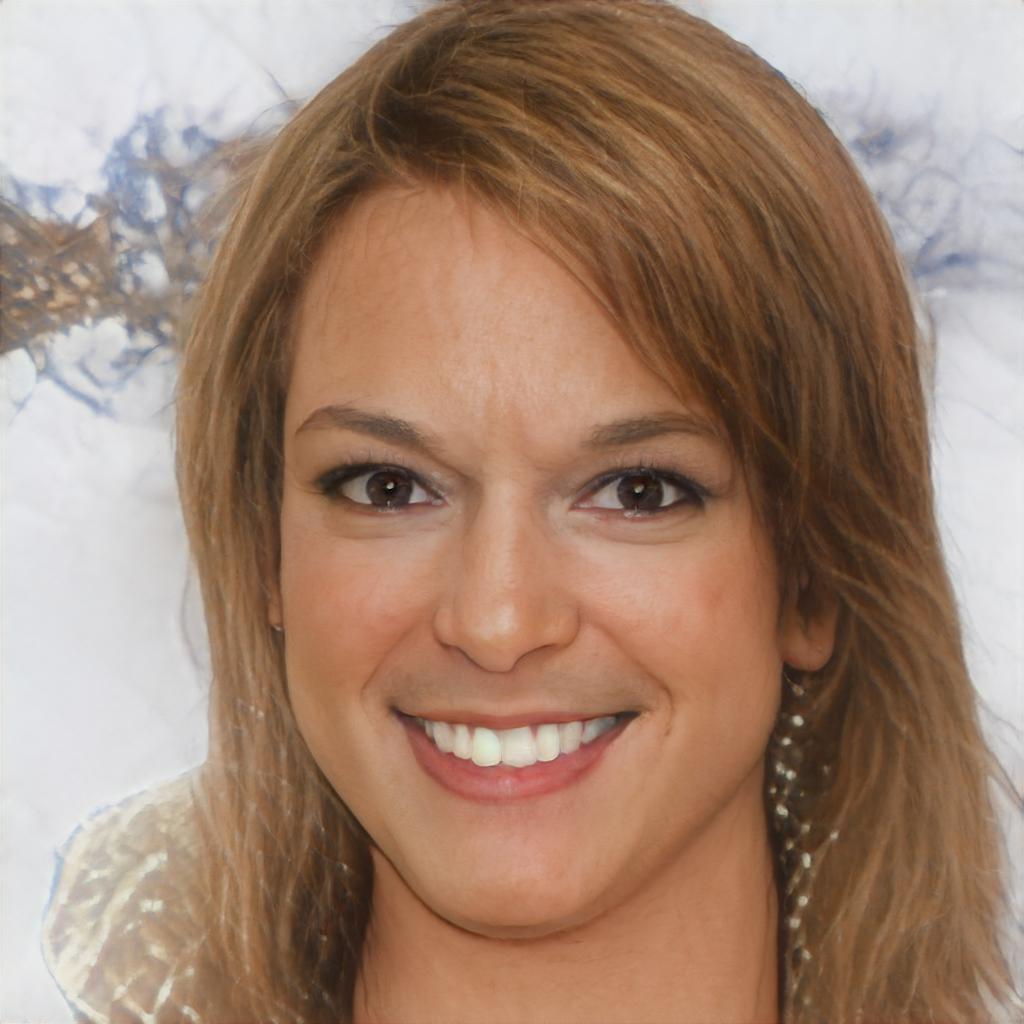} &
 \includegraphics[width=0.13\textwidth, ]{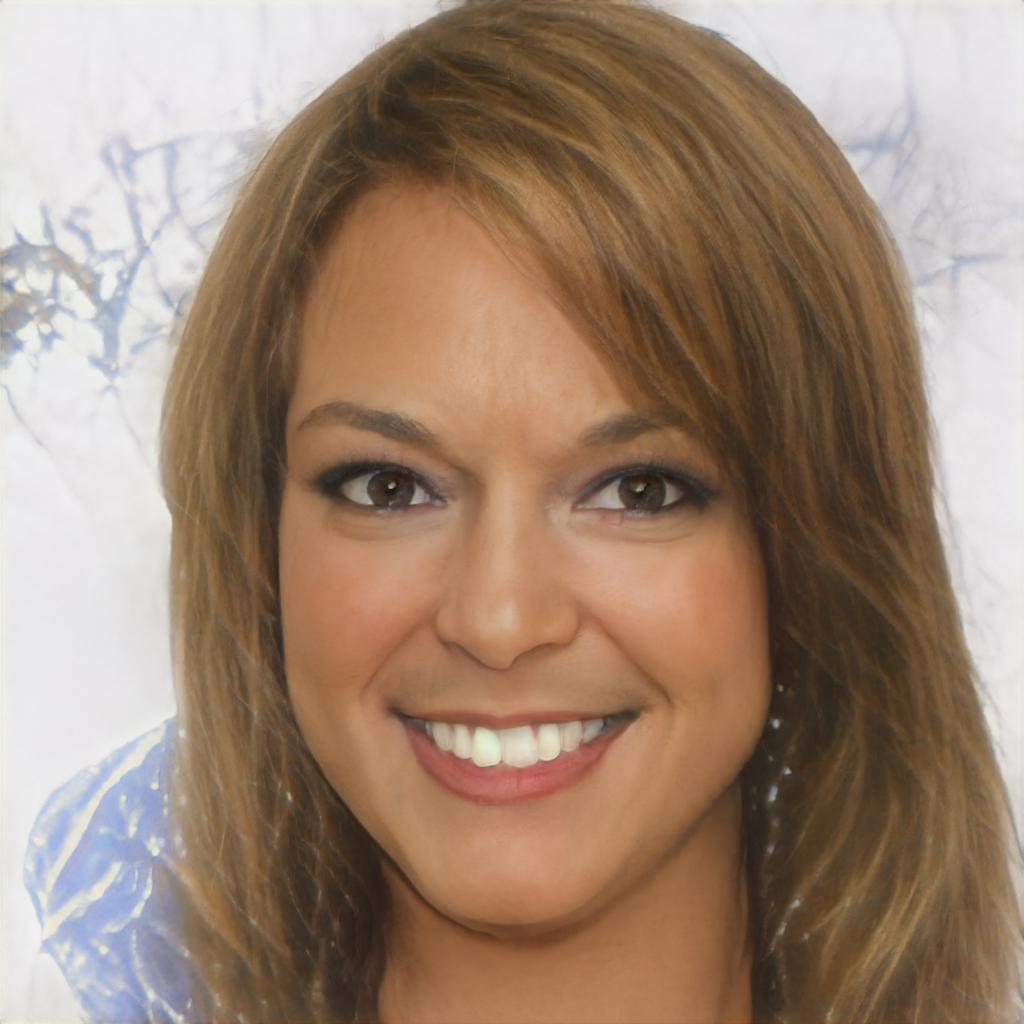} &
 \includegraphics[width=0.13\textwidth, ]{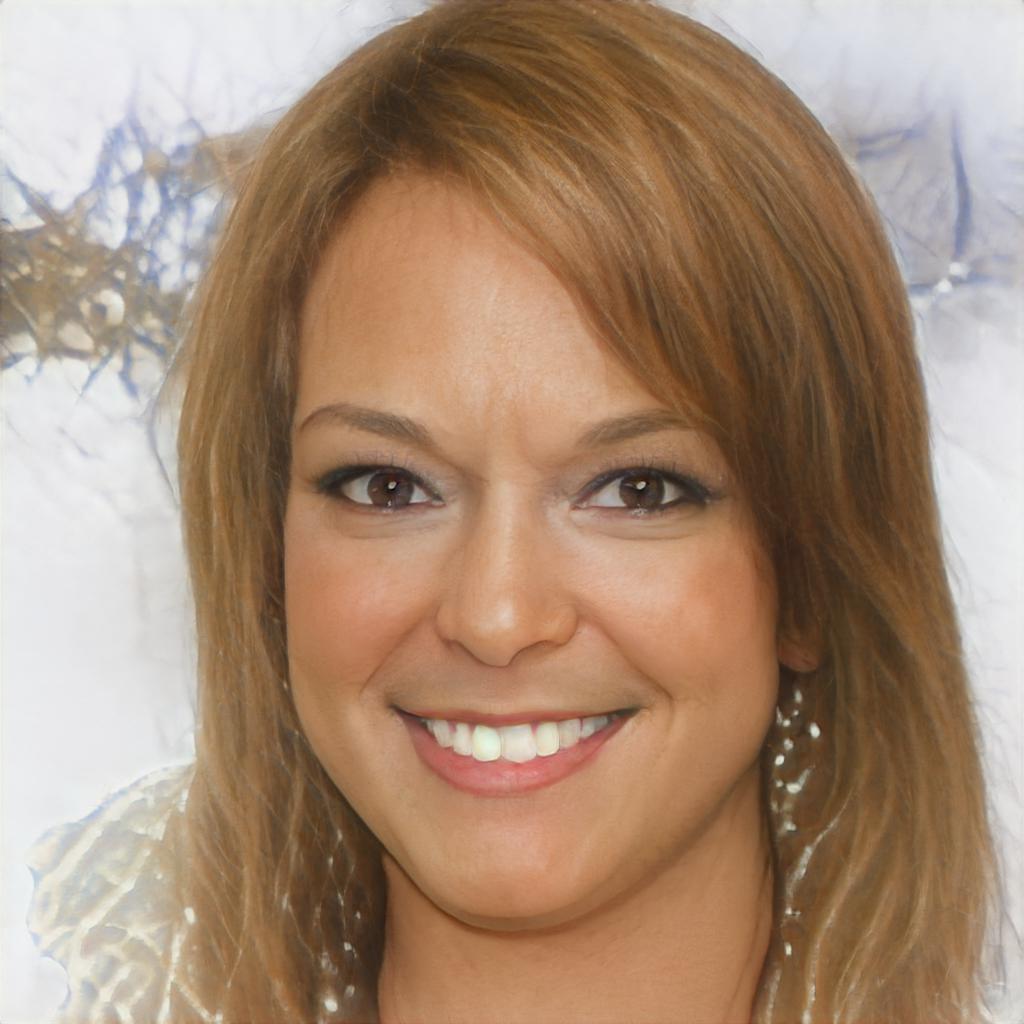} &
 \includegraphics[width=0.13\textwidth, ]{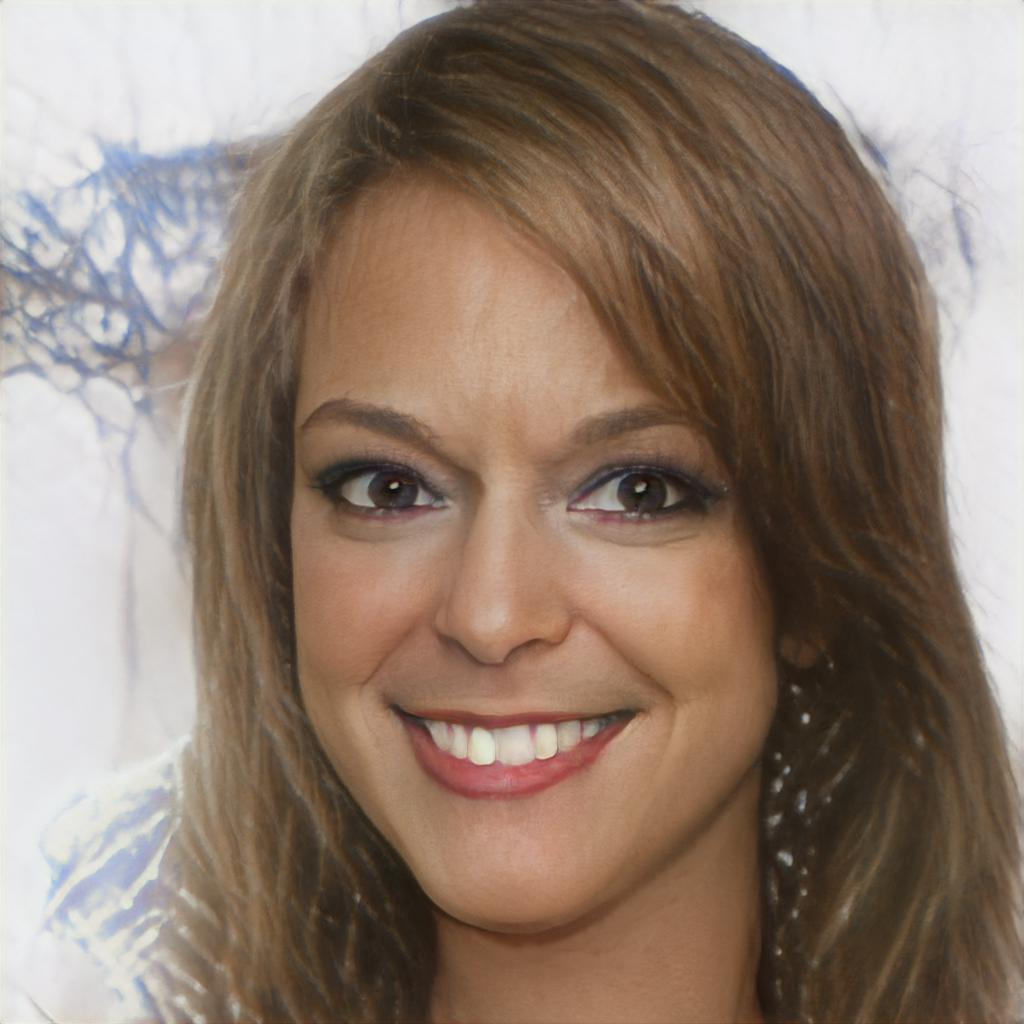} &
 \begin{turn}{90}  \wstara-ID^{\star}\end{turn} &
 \includegraphics[width=0.13\textwidth]{images/original/06002.jpg} & 
 \includegraphics[width=0.13\textwidth, ]{images/inversion/18_orig_img_1.jpg} &
 \includegraphics[width=0.13\textwidth, ]{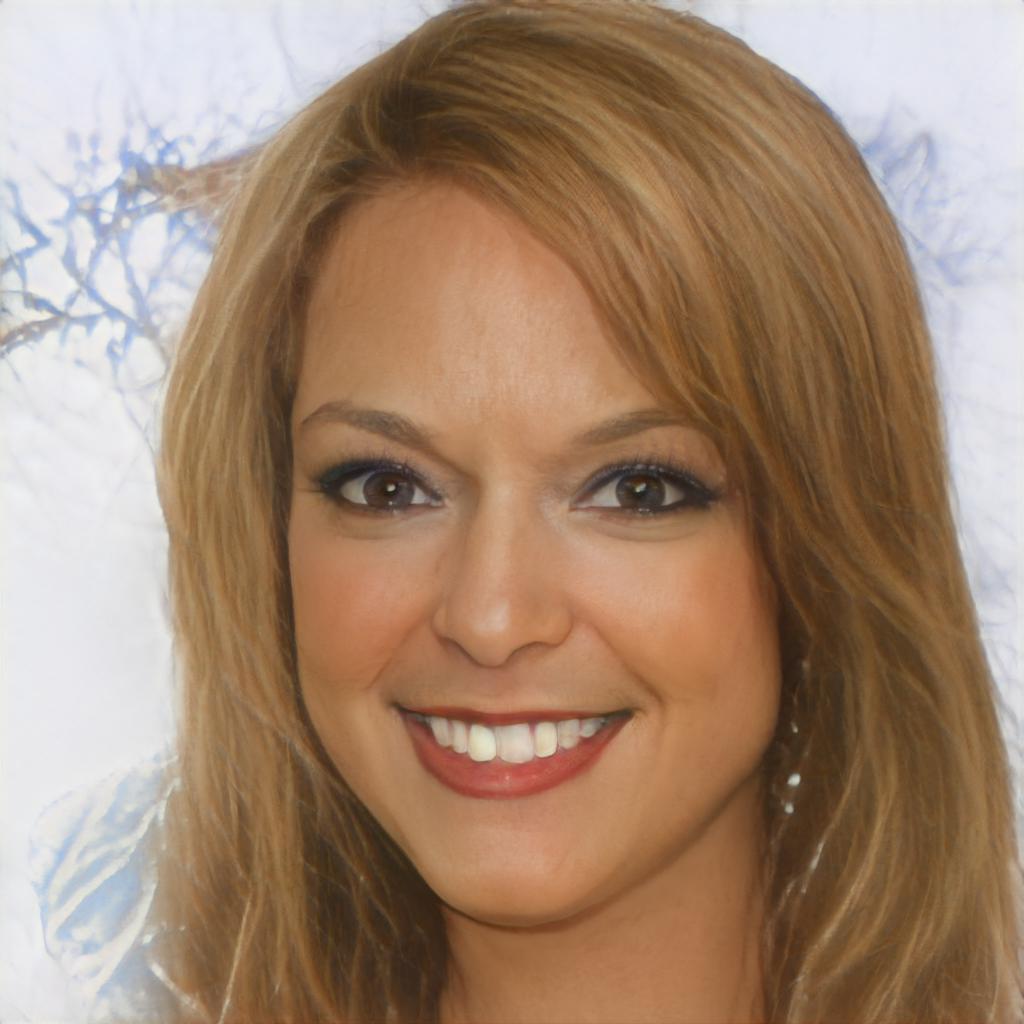} &
 \includegraphics[width=0.13\textwidth, ]{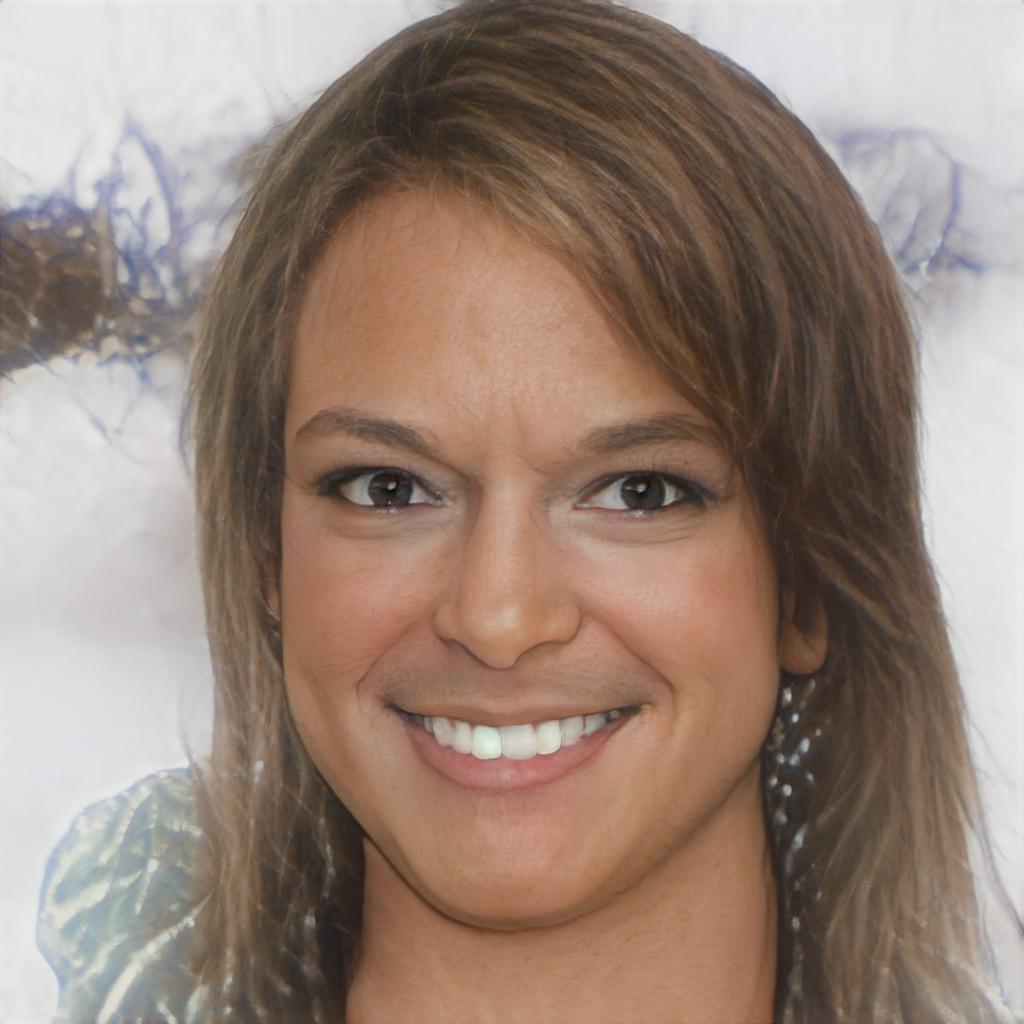} &
 \includegraphics[width=0.13\textwidth, ]{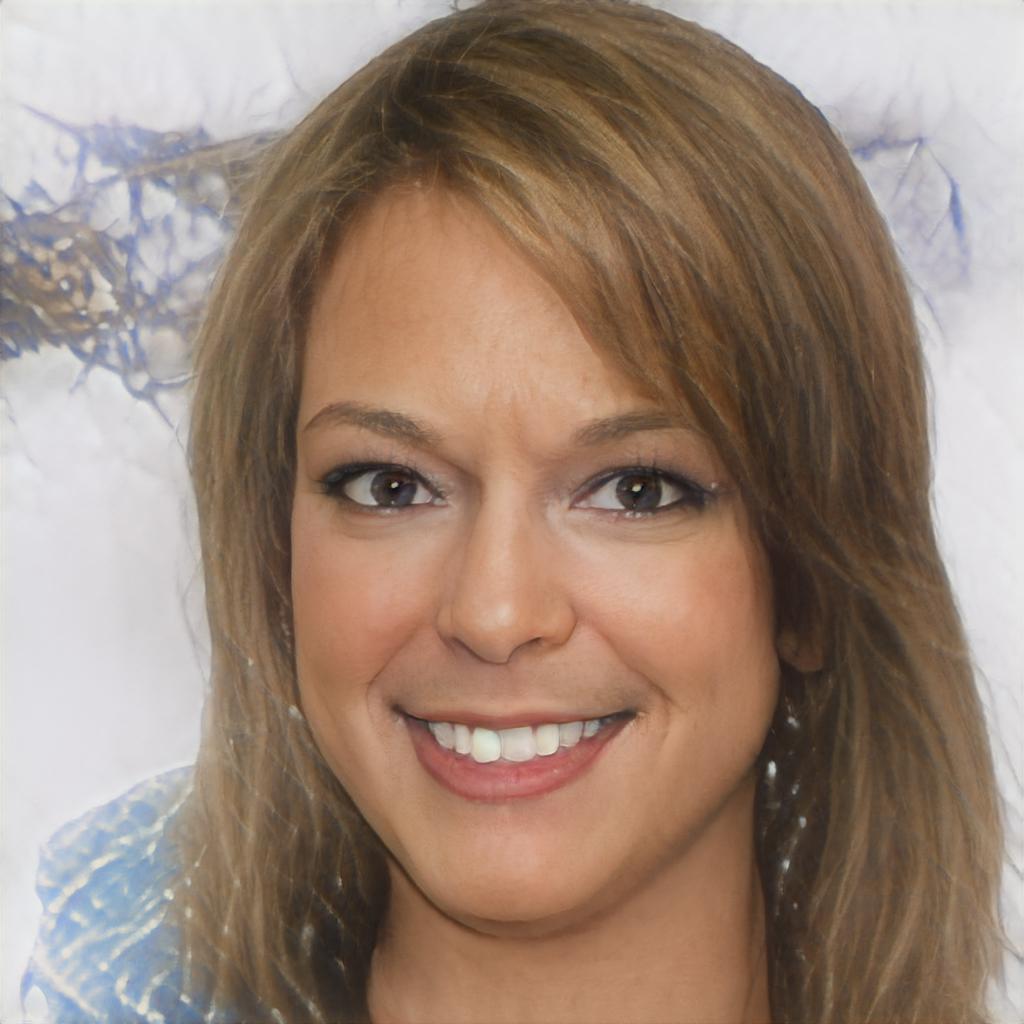} &
 \includegraphics[width=0.13\textwidth, ]{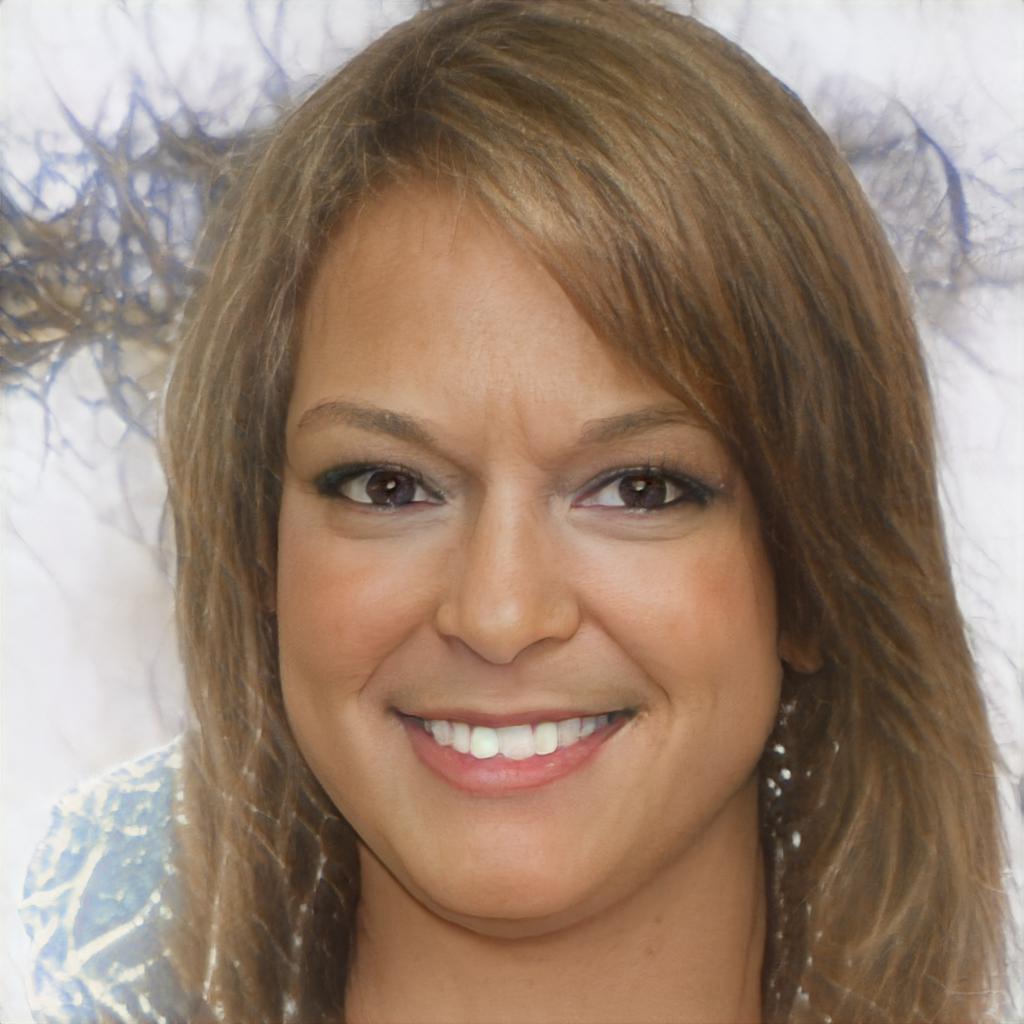} &
 \includegraphics[width=0.13\textwidth, ]{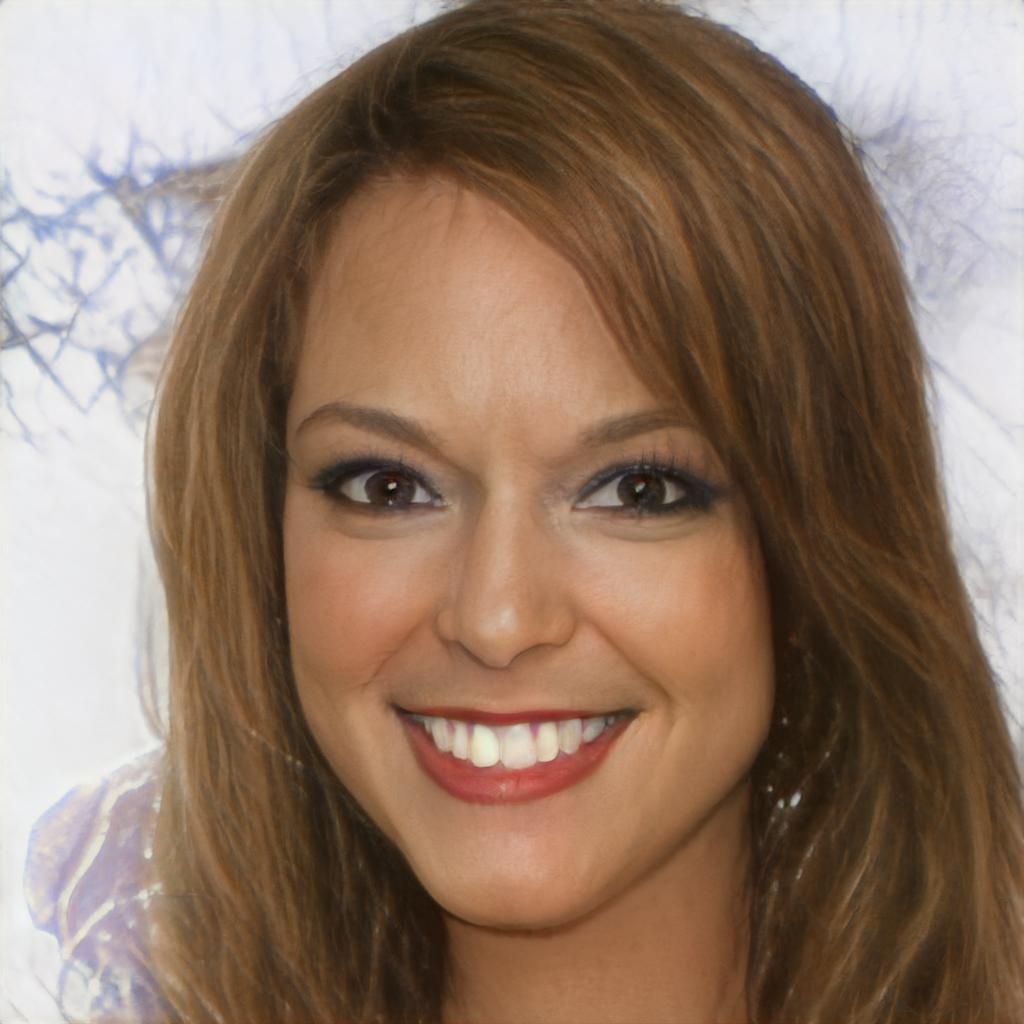} &
 \begin{turn}{90} \hspace{0.4cm} $\mathcal{W}^{\star}_{ID}$  \end{turn} &
 \includegraphics[width=0.13\textwidth]{images/original/06002.jpg} & 
 \includegraphics[width=0.13\textwidth, ]{images/inversion/18_orig_img_1.jpg} &
 \includegraphics[width=0.13\textwidth, ]{images/wstar_id_no_mag_3_coupl_dist_edit_train_ep5/18_img_1.jpg} &
 \includegraphics[width=0.13\textwidth, ]{images/wstar_id_no_mag_3_coupl_dist_edit_train_ep5/20_img_1.jpg} &
 \includegraphics[width=0.13\textwidth, ]{images/wstar_id_no_mag_3_coupl_dist_edit_train_ep5/22_img_1.jpg} &
 \includegraphics[width=0.13\textwidth, ]{images/wstar_id_no_mag_3_coupl_dist_edit_train_ep5/13_img_1.jpg} &
 \includegraphics[width=0.13\textwidth, ]{images/wstar_id_no_mag_3_coupl_dist_edit_train_ep5/36_img_1.jpg} &
 \\
\end{tabular}
\caption{Ablation study: Image editing using InterFaceGAN.}
\label{fig:edit_abl_2}
\end{figure*}

\begin{figure*}[h]
\setlength\tabcolsep{2pt}%
\centering
\begin{tabular}{p{0.25cm}ccccccc}
\centering
&
 \textbf{Original} &
 \textbf{Inverted} &
 \textbf{Makeup} &
 \textbf{Male} &
 \textbf{Mustache} &
 \textbf{Chubby} &
 \textbf{Lipstick} &
  \begin{turn}{90} \hspace{0.5cm} \wplus\end{turn} &
 \includegraphics[width=0.13\textwidth]{images/original/06011.jpg} & 
 \includegraphics[width=0.13\textwidth, ]{images/inversion/18_orig_img_10.jpg} &
 \includegraphics[width=0.13\textwidth, ]{images/wplus/18_img_12.jpg} &
 \includegraphics[width=0.13\textwidth, ]{images/wplus/20_img_12.jpg} &
 \includegraphics[width=0.13\textwidth, ]{images/wplus/22_img_12.jpg} &
 \includegraphics[width=0.13\textwidth, ]{images/wplus/13_img_12.jpg} &
 \includegraphics[width=0.13\textwidth]{images/wplus/36_img_12.jpg} &
 \begin{turn}{90} \hspace{0.5cm} \wstara\end{turn} &
 \includegraphics[width=0.13\textwidth]{images/original/06011.jpg} & 
 \includegraphics[width=0.13\textwidth, ]{images/inversion/18_orig_img_10.jpg} &
 \includegraphics[width=0.13\textwidth, ]{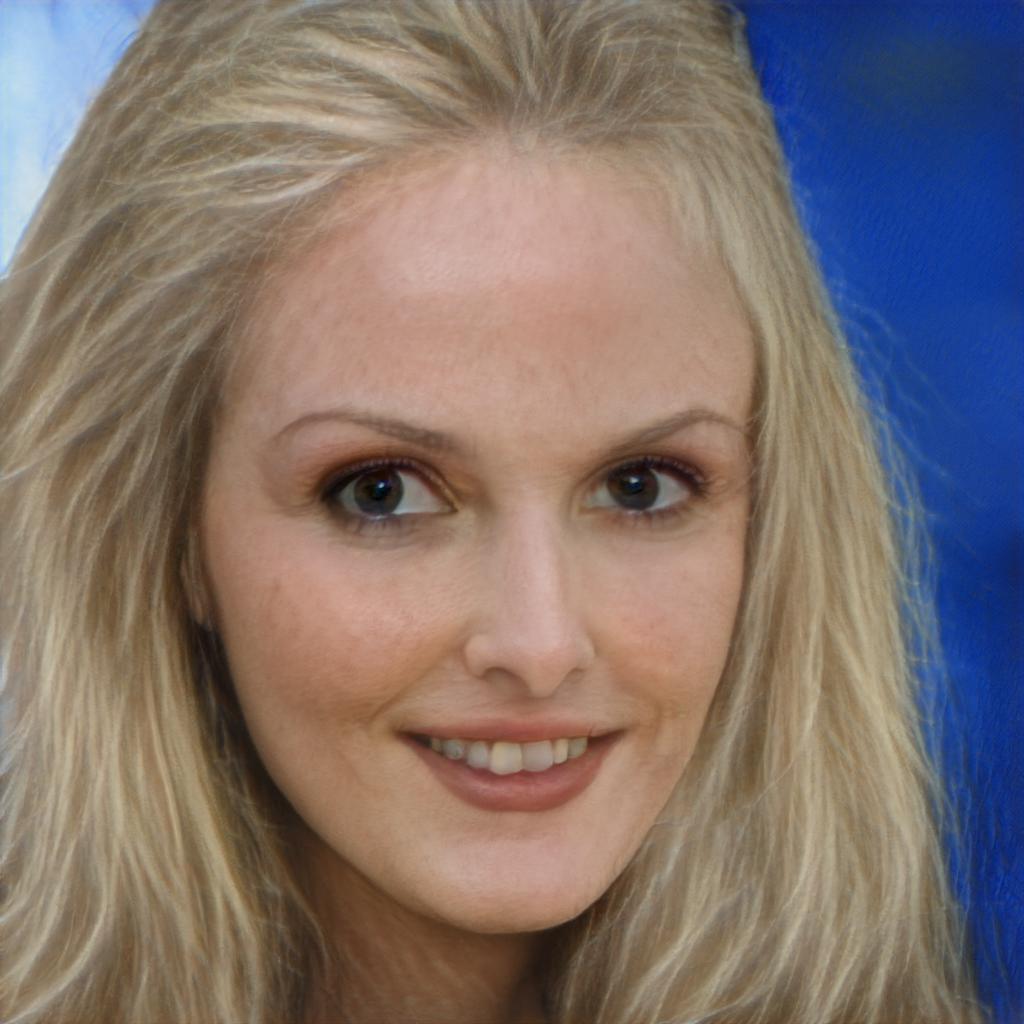} &
 \includegraphics[width=0.13\textwidth, ]{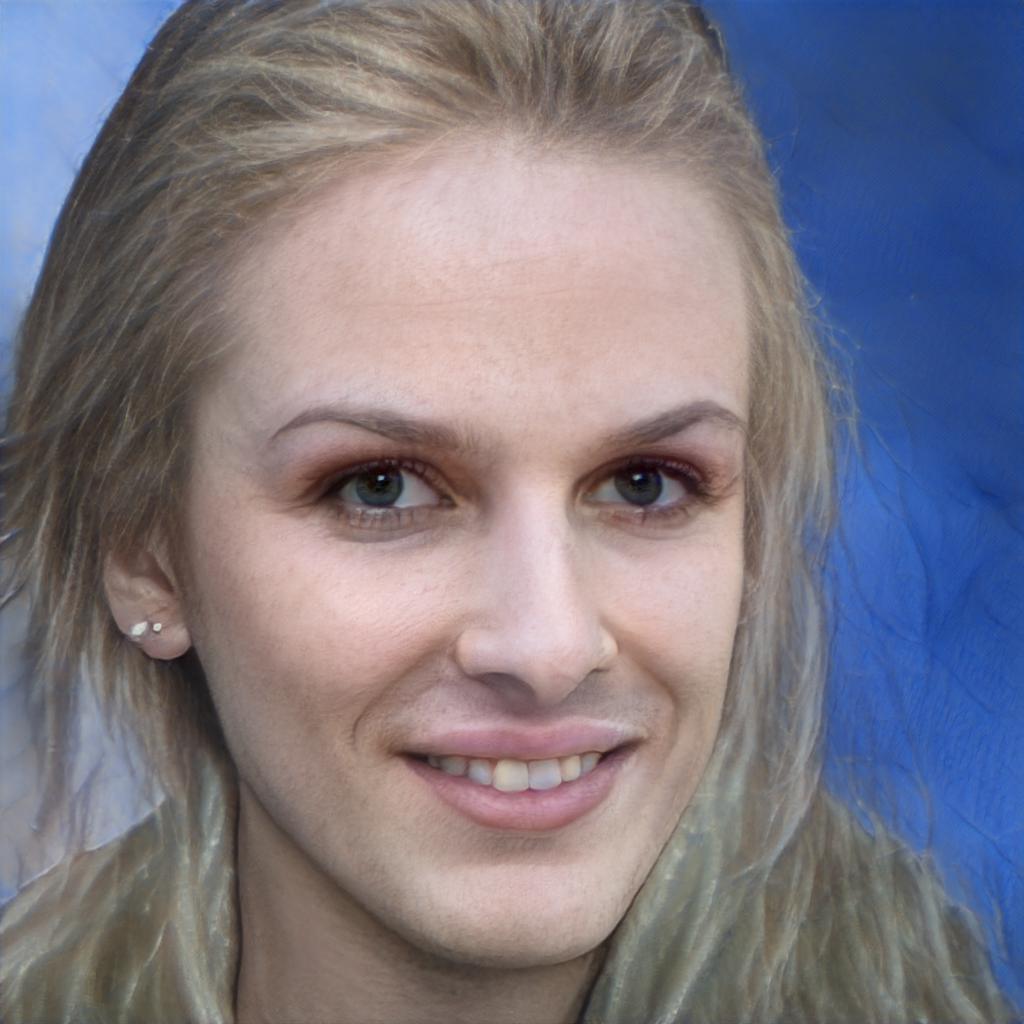} &
 \includegraphics[width=0.13\textwidth, ]{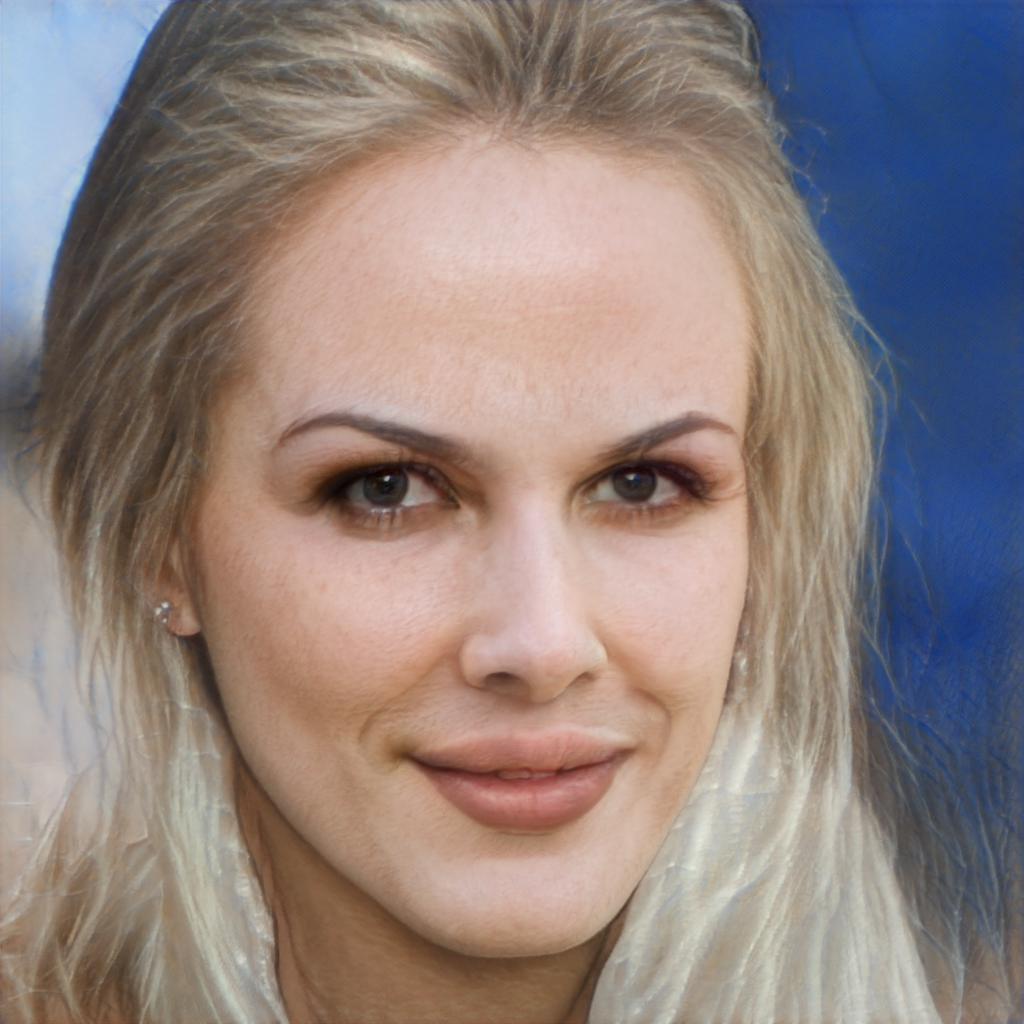} &
 \includegraphics[width=0.13\textwidth, ]{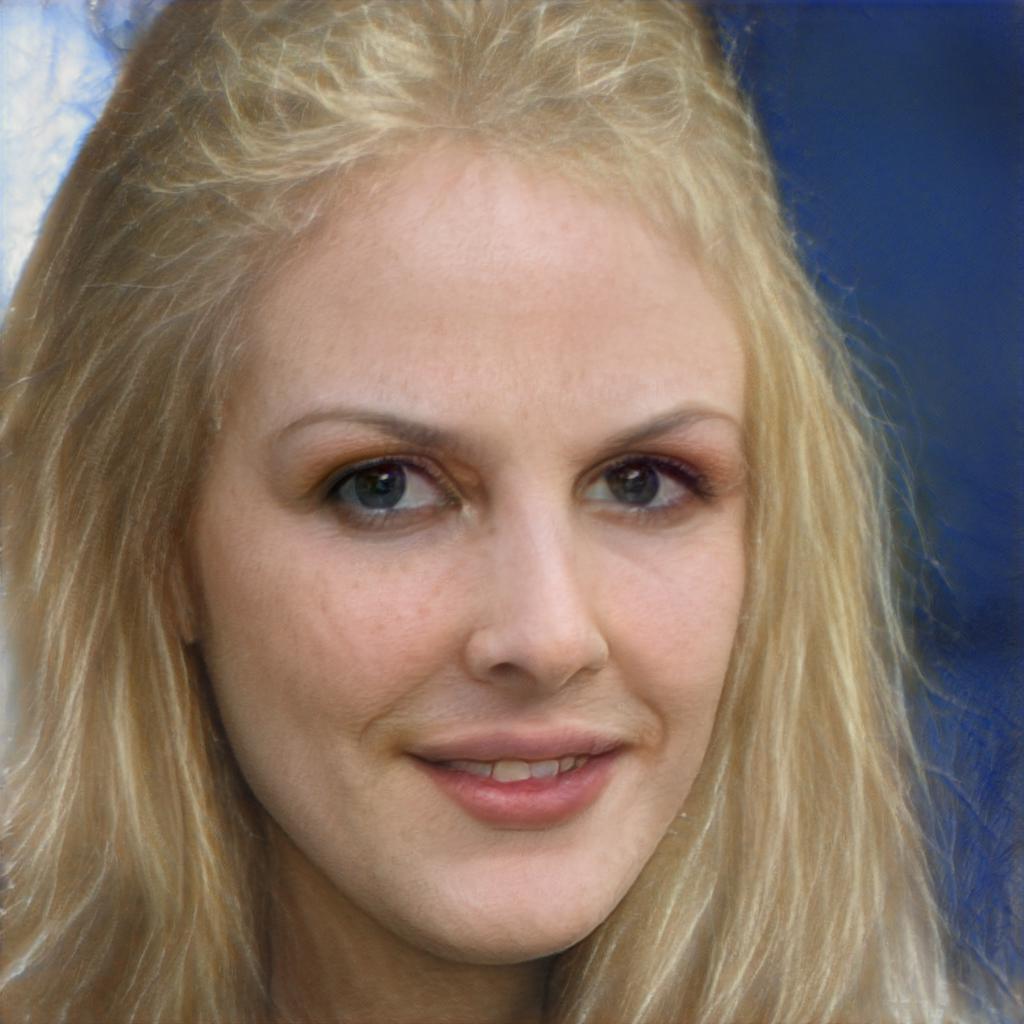} &
 \includegraphics[width=0.13\textwidth, ]{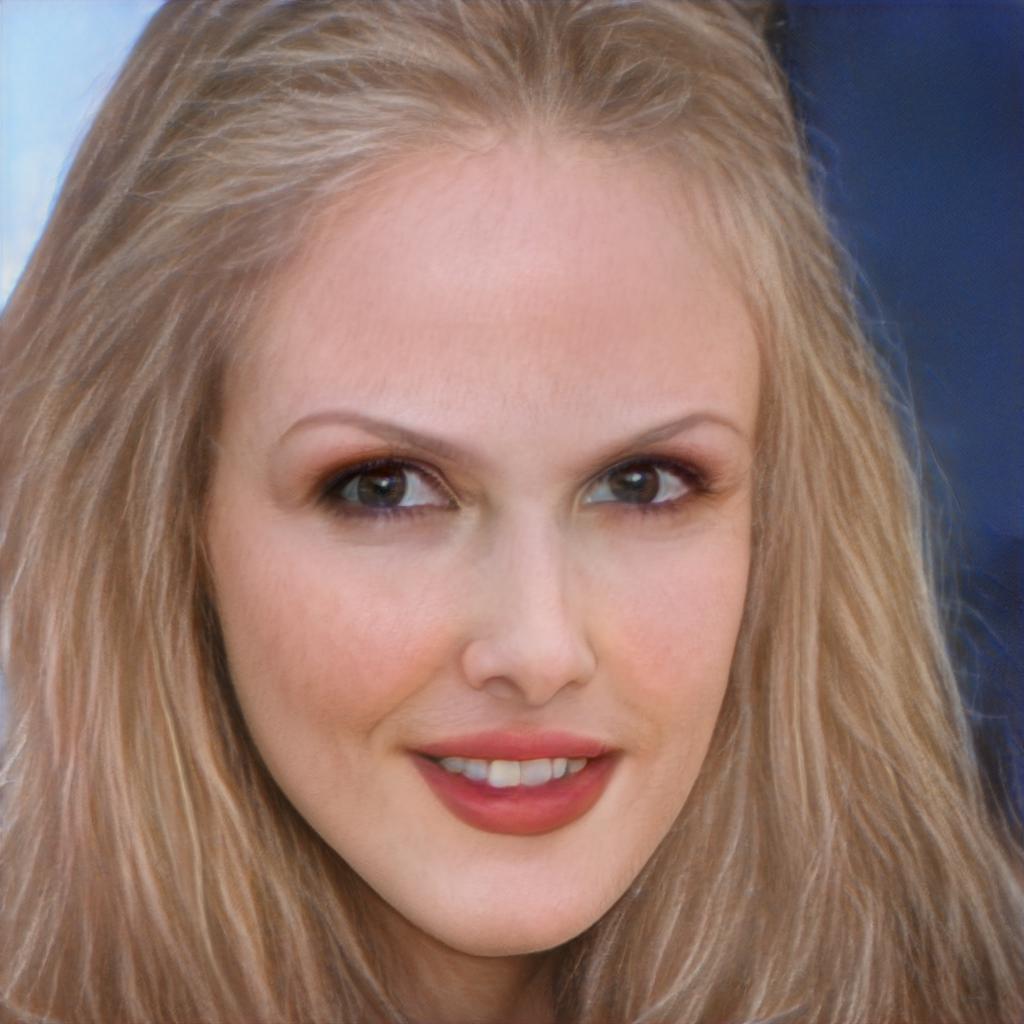} &
 \begin{turn}{90} \hspace{0.2cm} \wstara-ID\end{turn} &
 \includegraphics[width=0.13\textwidth]{images/original/06011.jpg} & 
 \includegraphics[width=0.13\textwidth, ]{images/inversion/18_orig_img_10.jpg} &
 \includegraphics[width=0.13\textwidth, ]{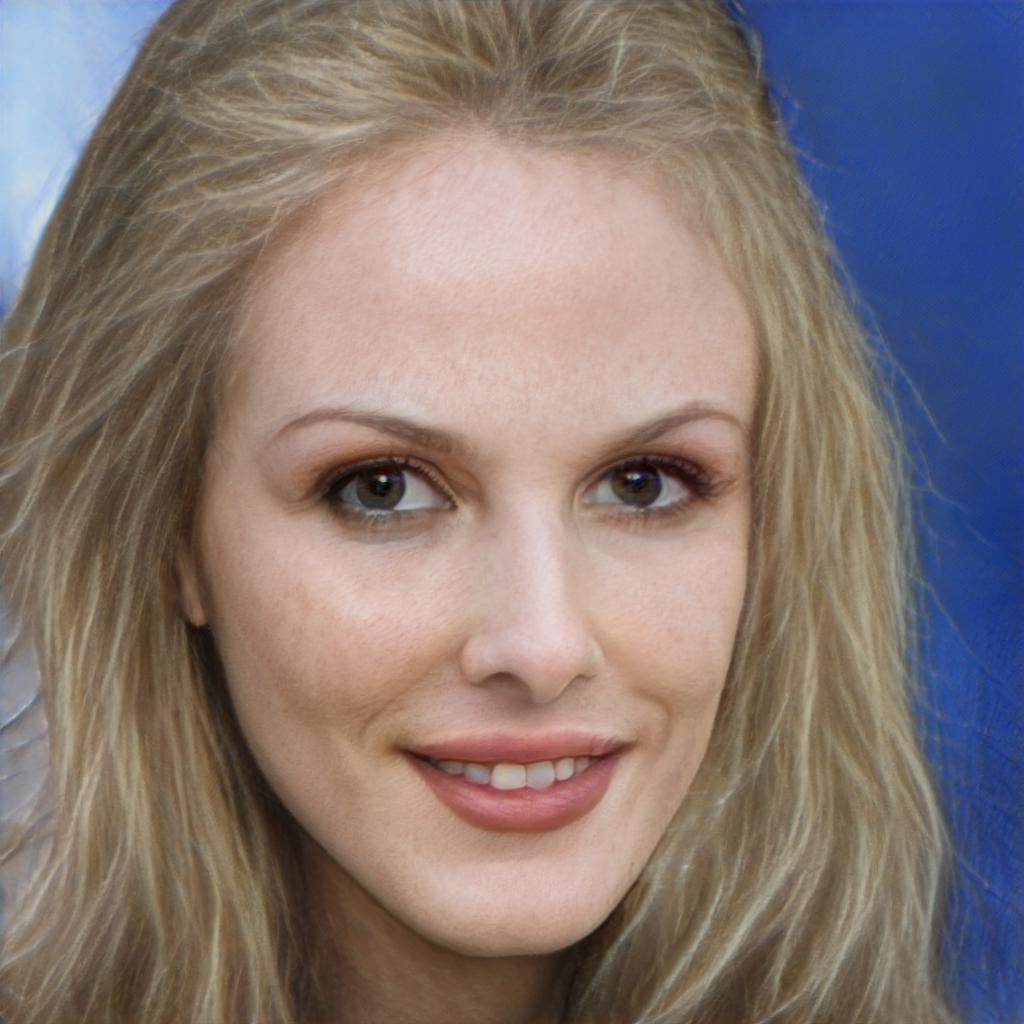} &
 \includegraphics[width=0.13\textwidth, ]{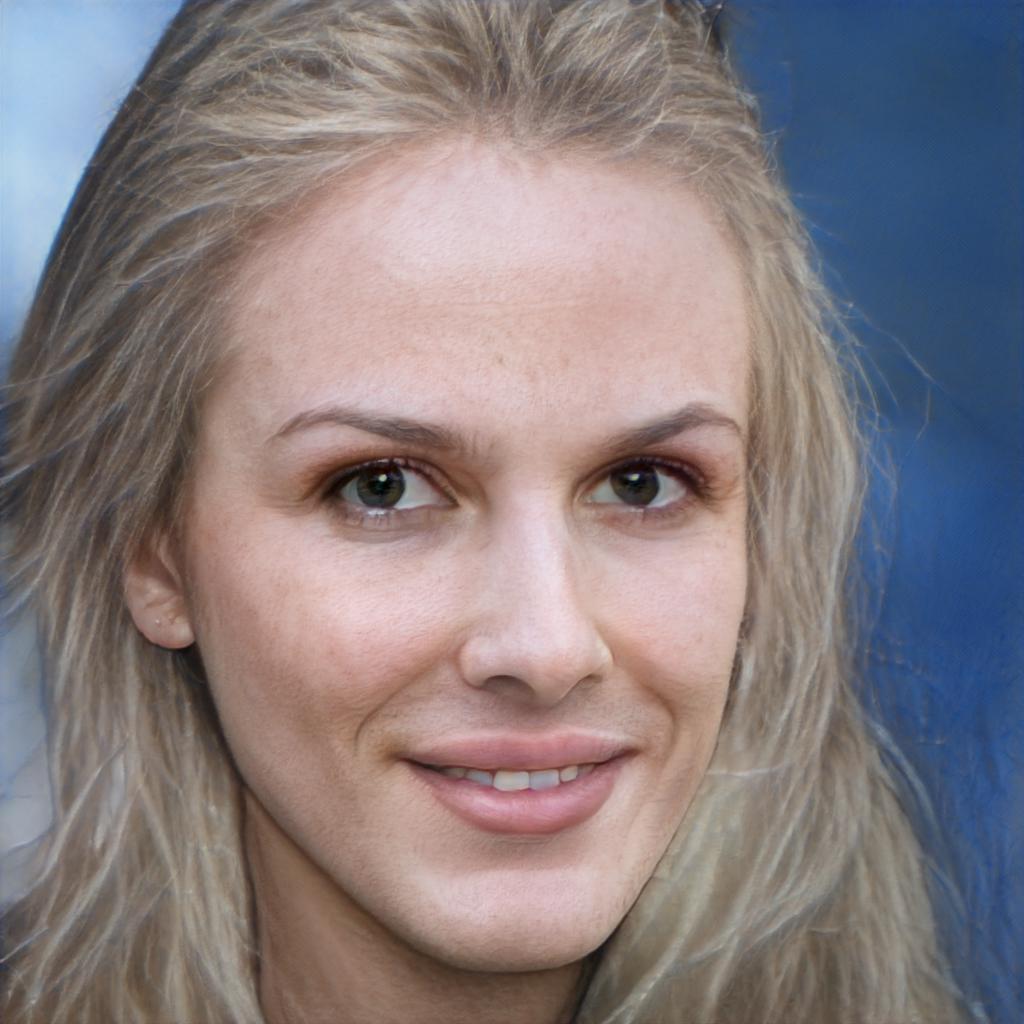} &
 \includegraphics[width=0.13\textwidth, ]{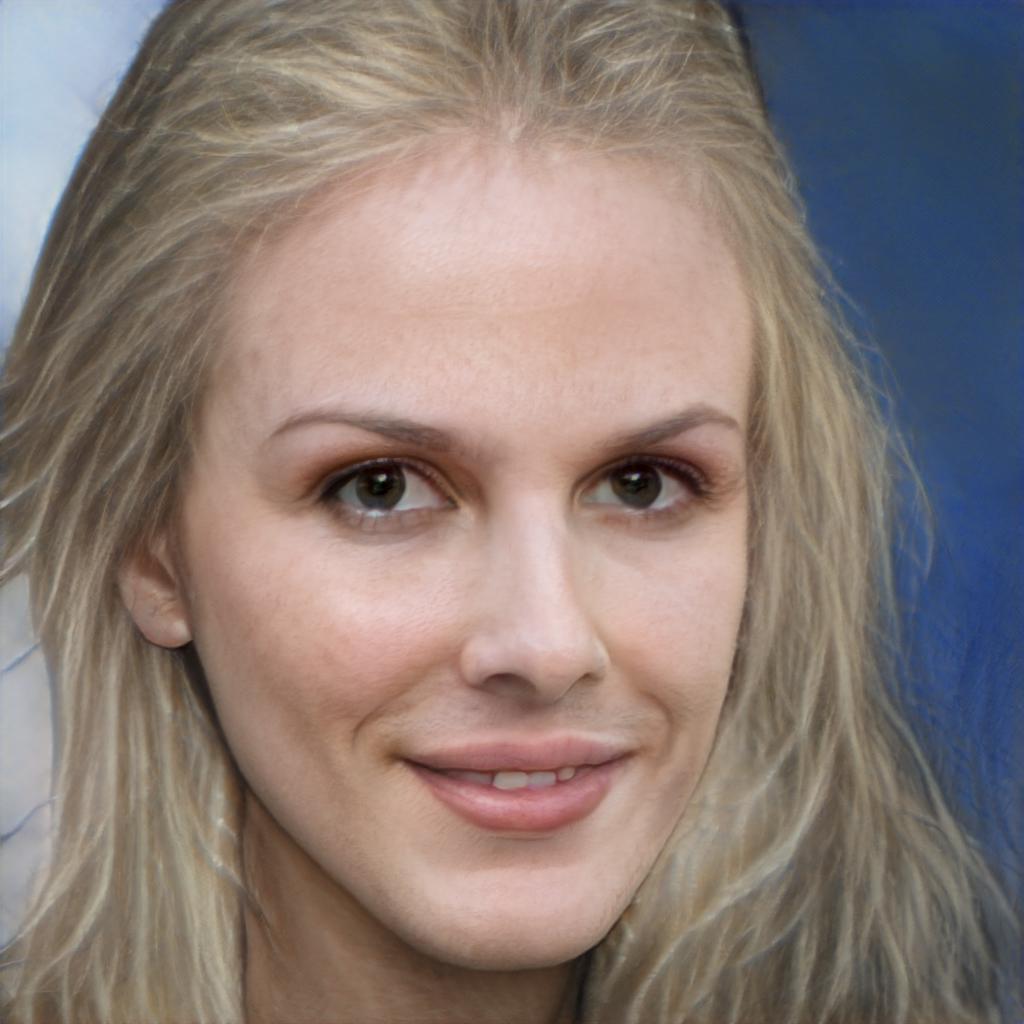} &
 \includegraphics[width=0.13\textwidth, ]{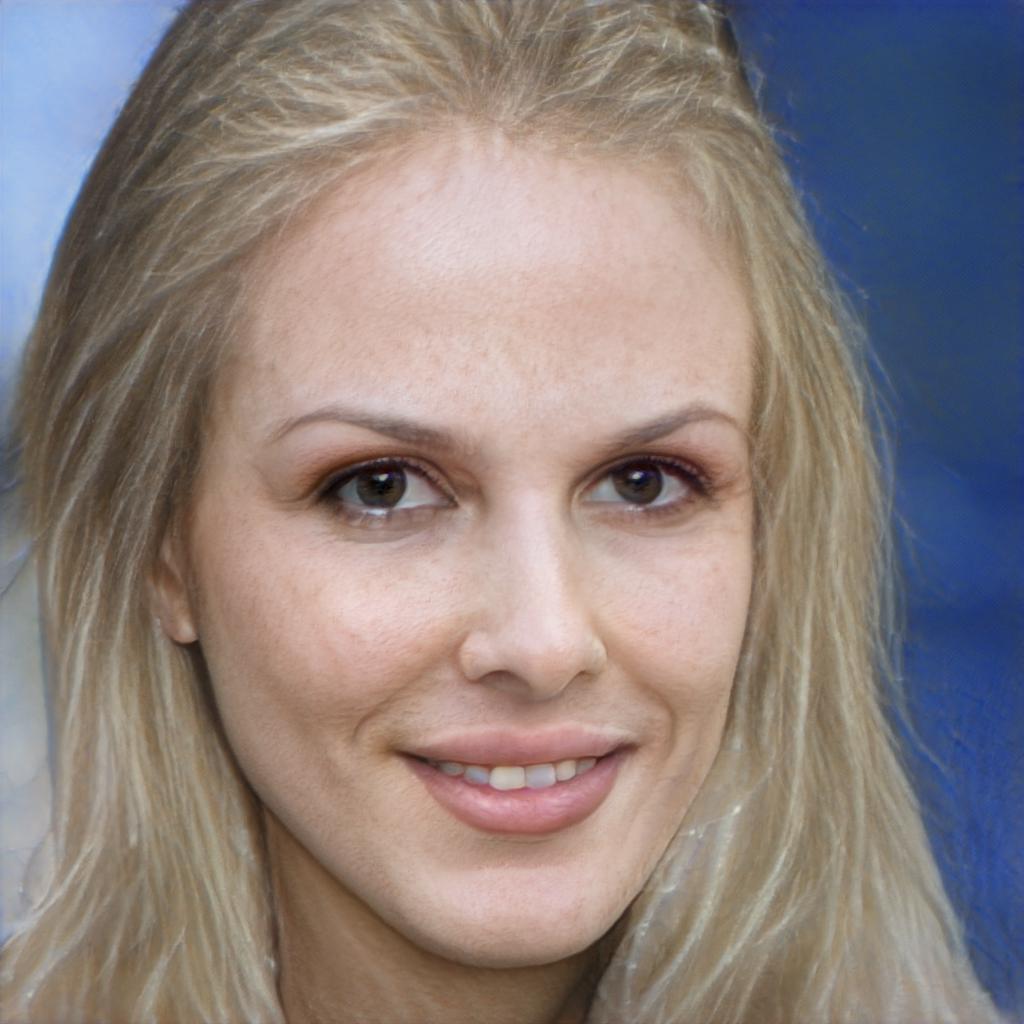} &
 \includegraphics[width=0.13\textwidth, ]{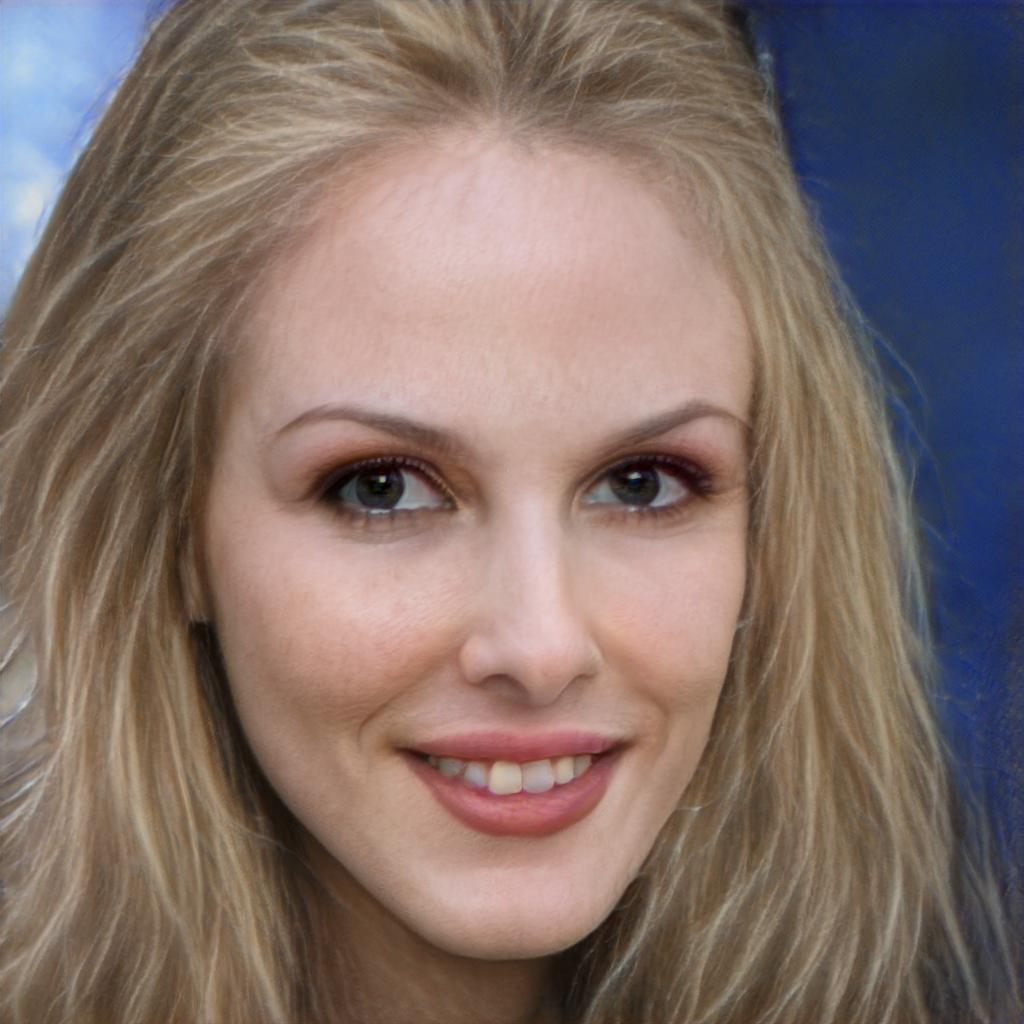} &
 \begin{turn}{90}  \wstara-ID^{\dagger}\end{turn} &
 \includegraphics[width=0.13\textwidth]{images/original/06011.jpg} & 
 \includegraphics[width=0.13\textwidth, ]{images/inversion/18_orig_img_10.jpg} &
 \includegraphics[width=0.13\textwidth, ]{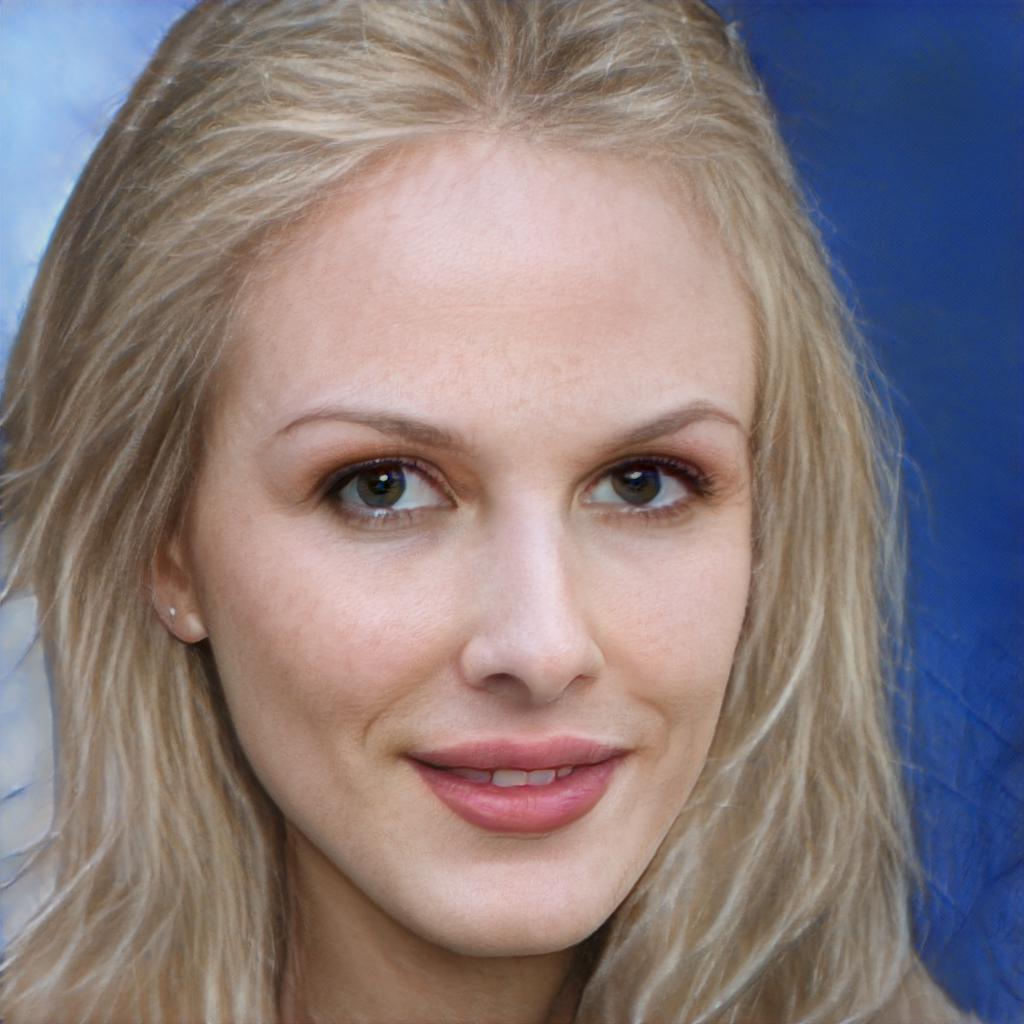} &
 \includegraphics[width=0.13\textwidth, ]{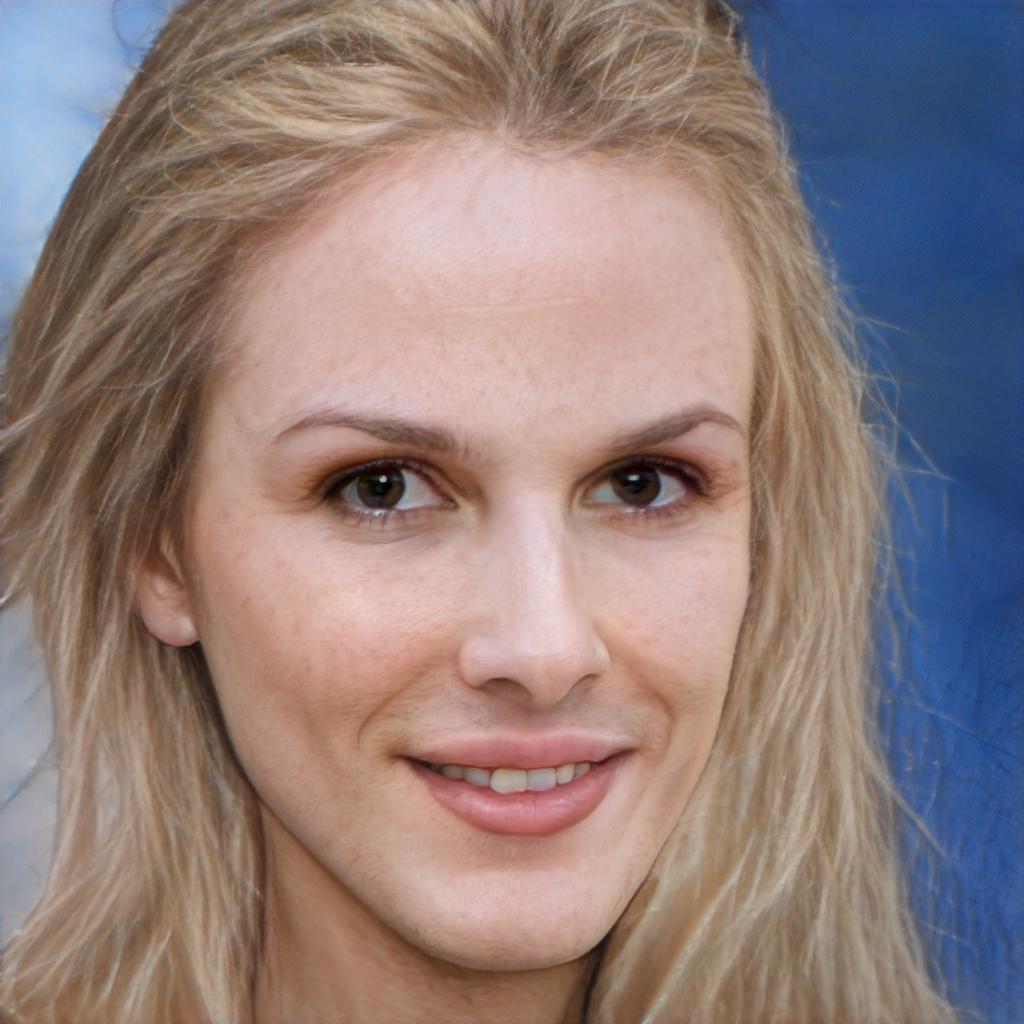} &
 \includegraphics[width=0.13\textwidth, ]{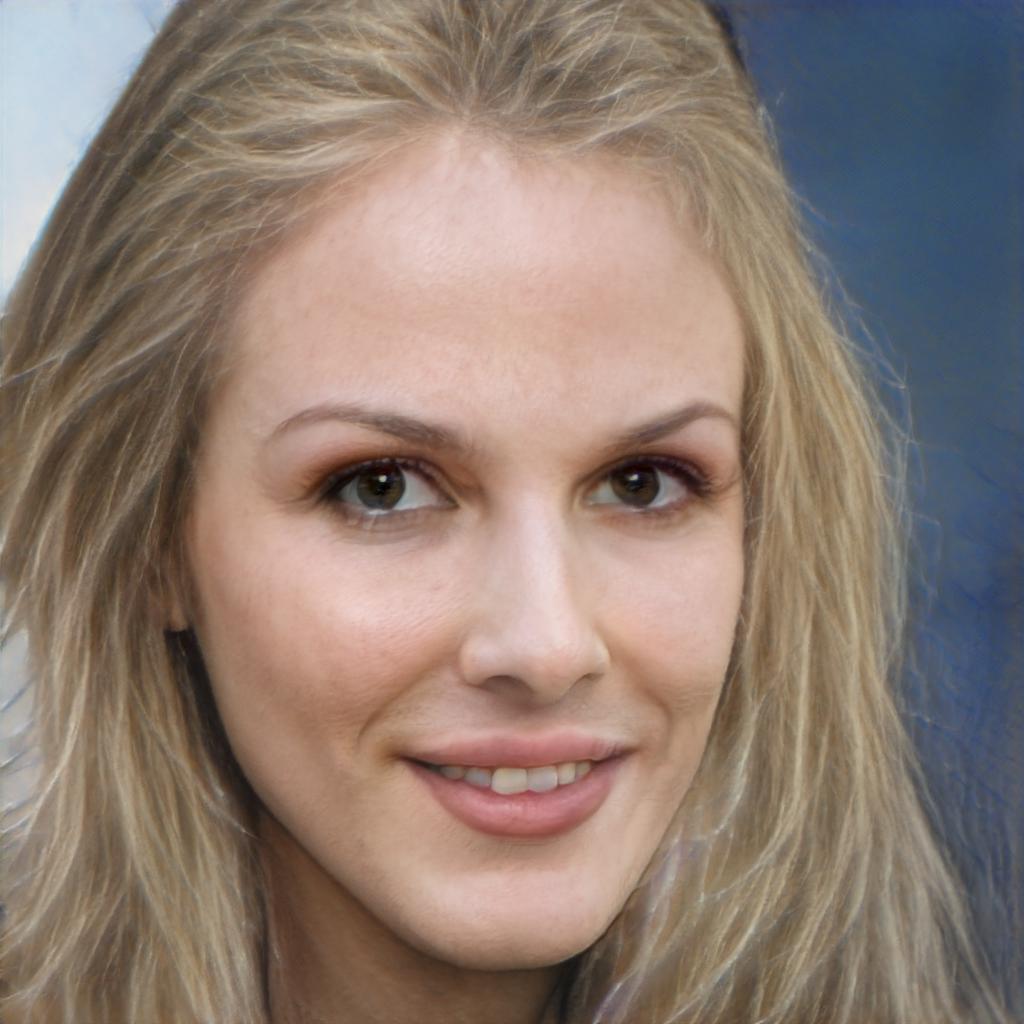} &
 \includegraphics[width=0.13\textwidth, ]{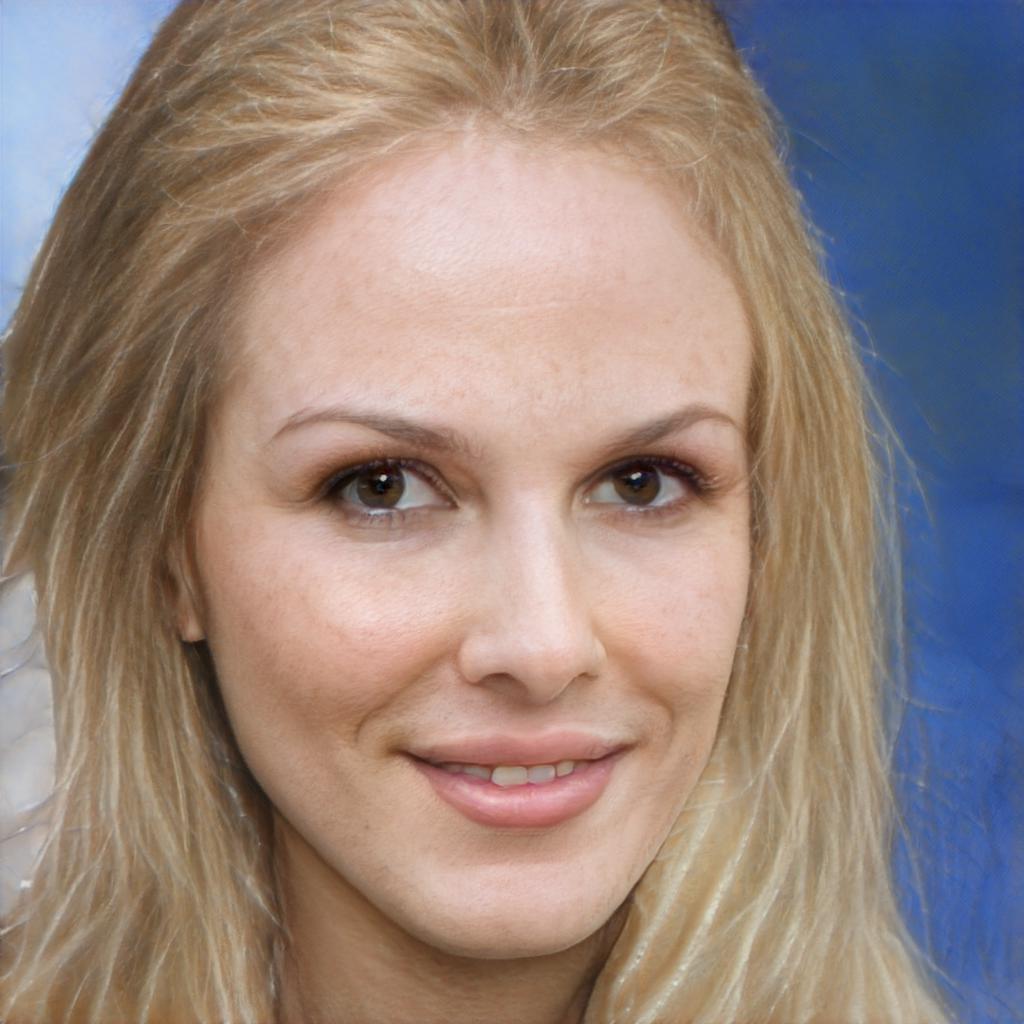} &
 \includegraphics[width=0.13\textwidth, ]{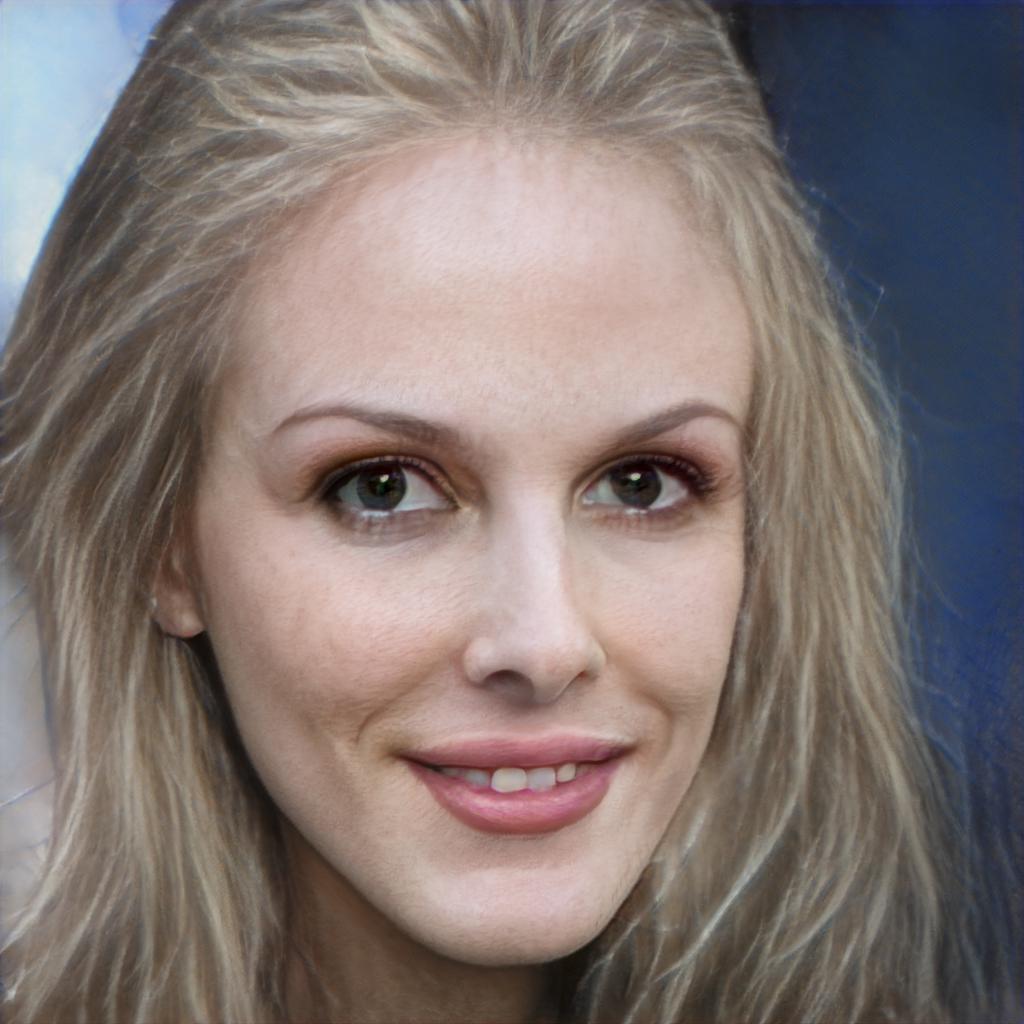} &
 \begin{turn}{90}  \wstara-ID^{\star}\end{turn} &
 \includegraphics[width=0.13\textwidth]{images/original/06011.jpg} & 
 \includegraphics[width=0.13\textwidth, ]{images/inversion/18_orig_img_10.jpg} &
 \includegraphics[width=0.13\textwidth, ]{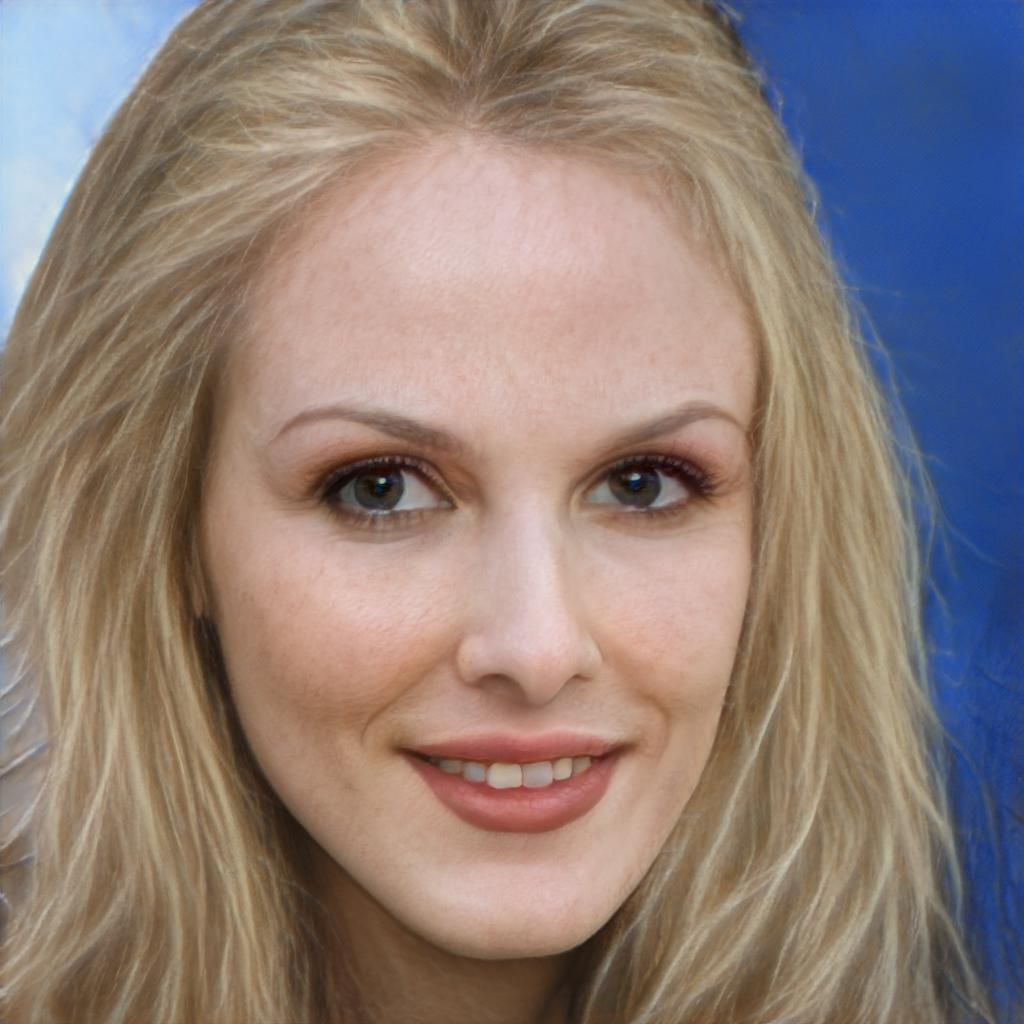} &
 \includegraphics[width=0.13\textwidth, ]{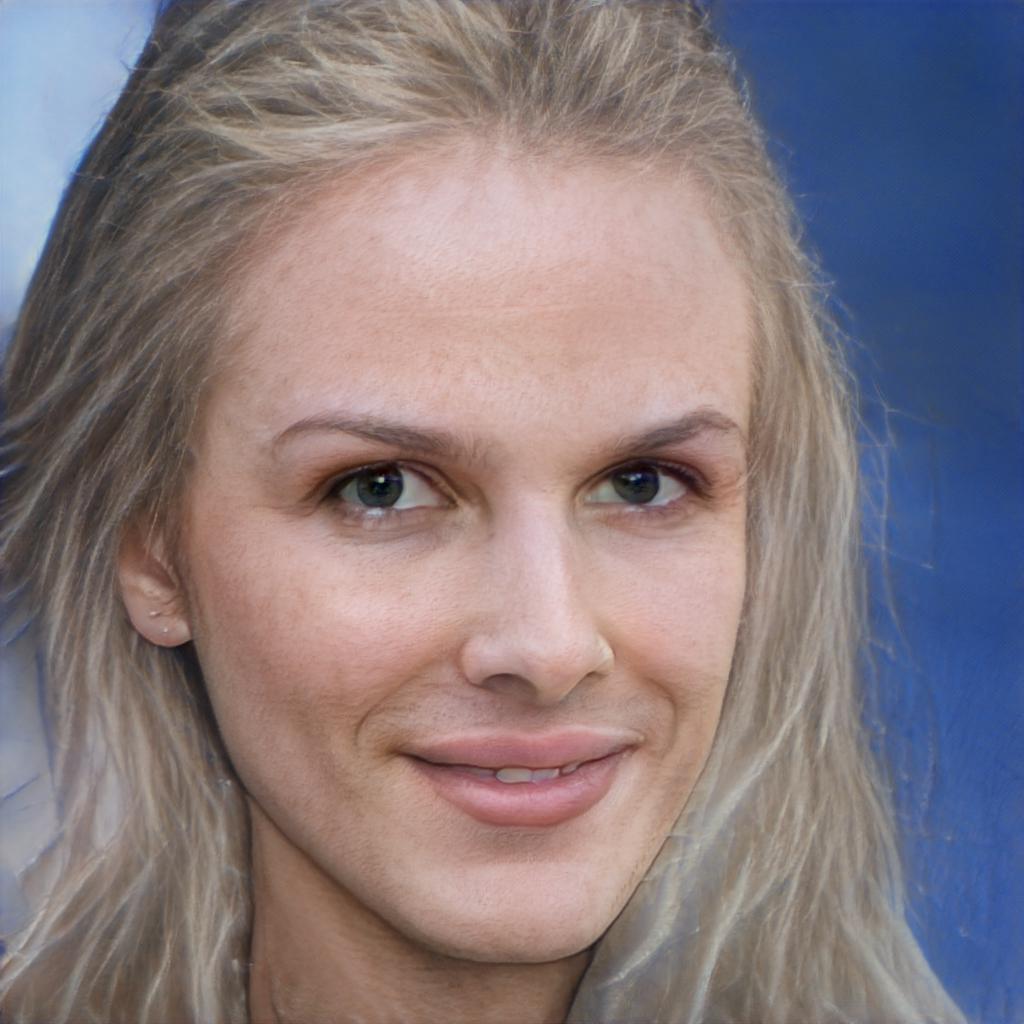} &
 \includegraphics[width=0.13\textwidth, ]{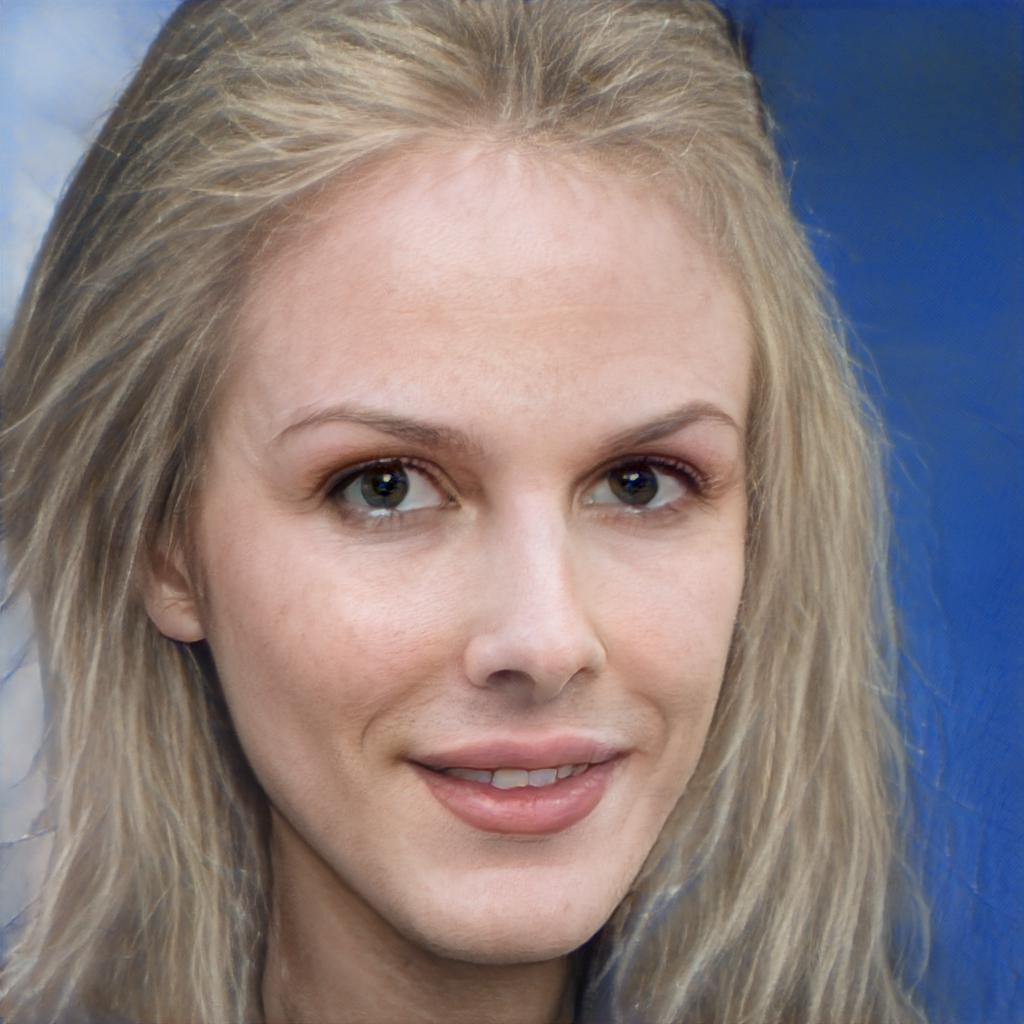} &
 \includegraphics[width=0.13\textwidth, ]{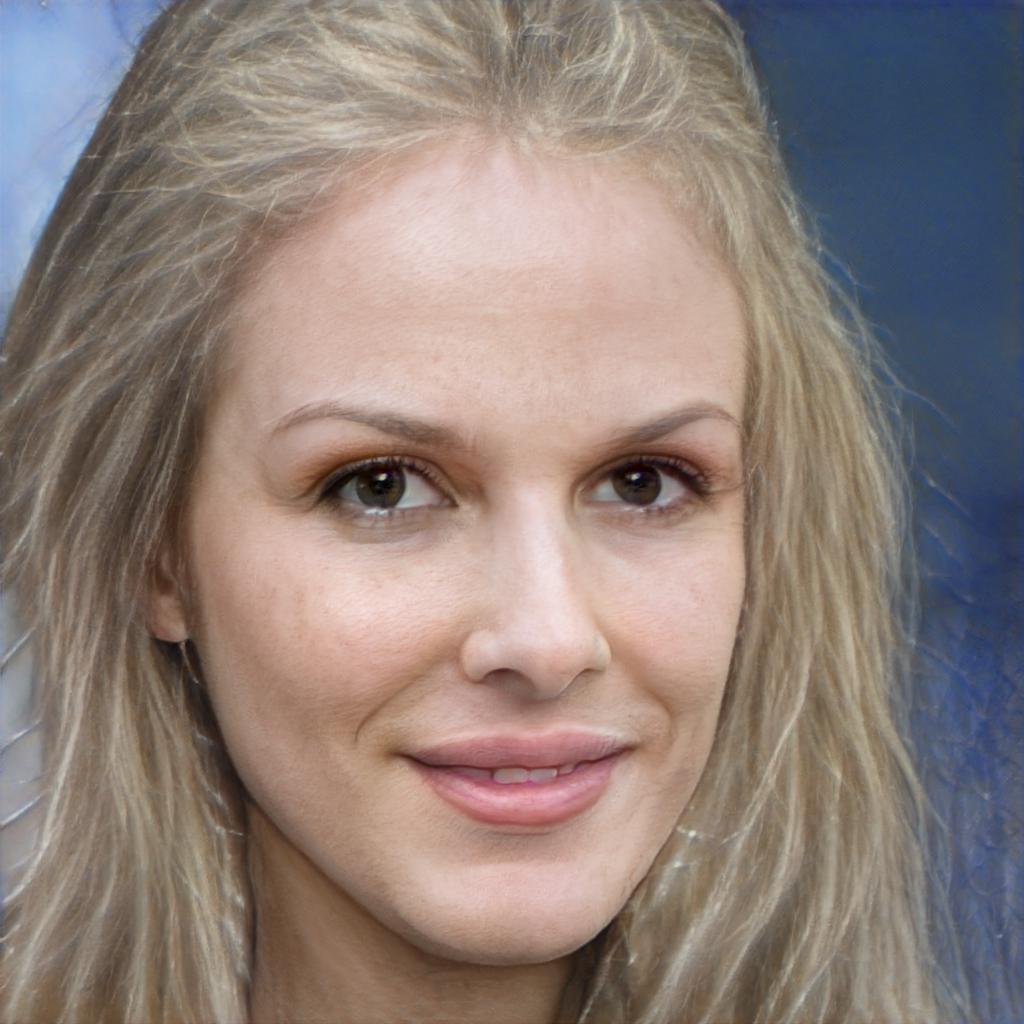} &
 \includegraphics[width=0.13\textwidth, ]{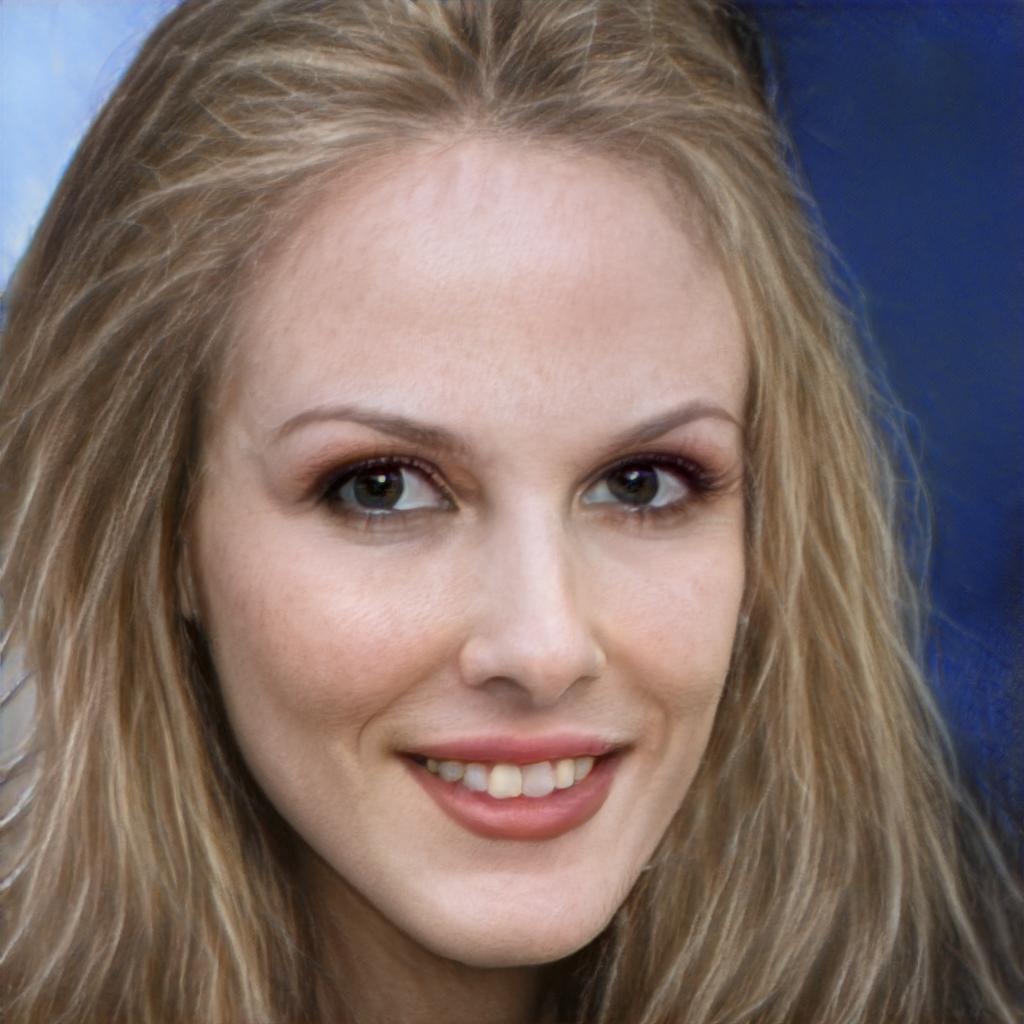} &
 \begin{turn}{90} \hspace{0.4cm}  $\mathcal{W}^{\star}_{ID}$  \end{turn} &
 \includegraphics[width=0.13\textwidth]{images/original/06011.jpg} & 
 \includegraphics[width=0.13\textwidth, ]{images/inversion/18_orig_img_10.jpg} &
 \includegraphics[width=0.13\textwidth, ]{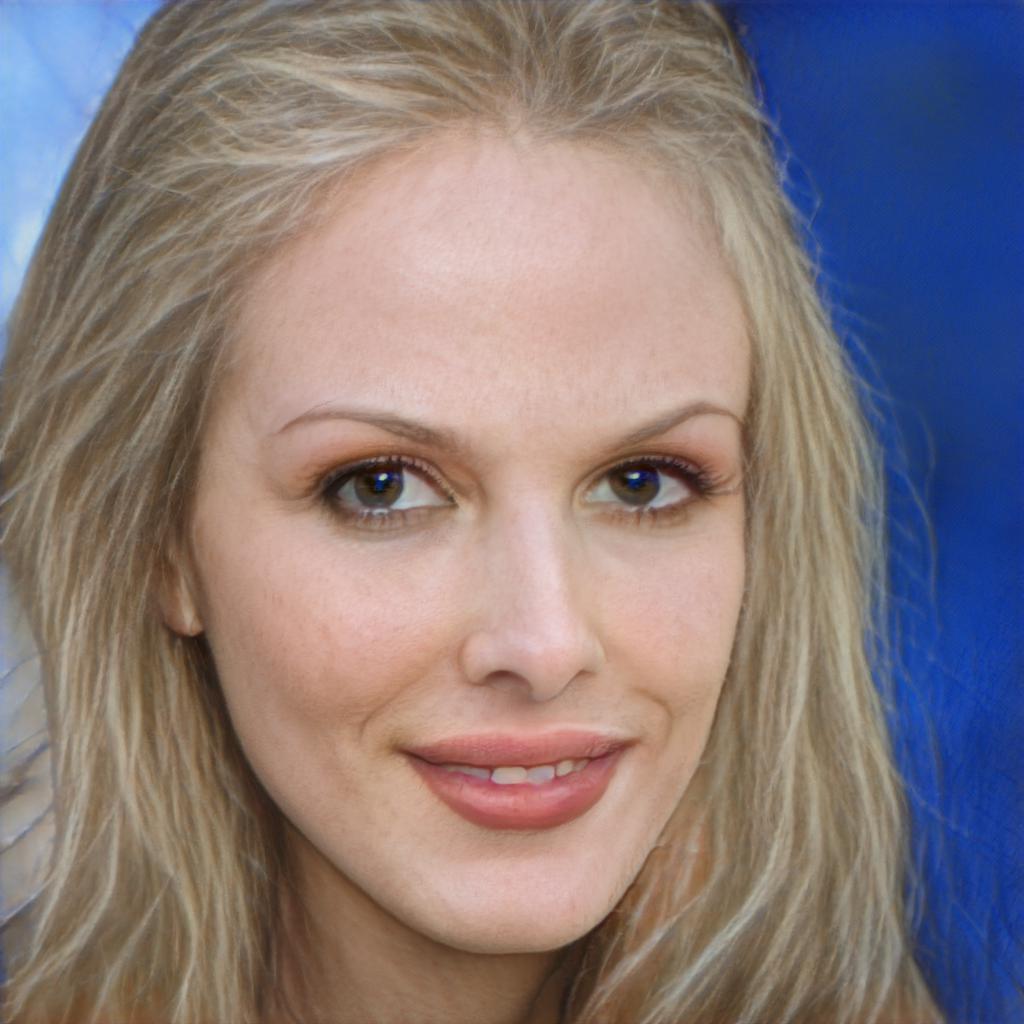} &
 \includegraphics[width=0.13\textwidth, ]{images/wstar_id_no_mag_3_coupl_dist_edit_train_ep5/20_img_10.jpg} &
 \includegraphics[width=0.13\textwidth, ]{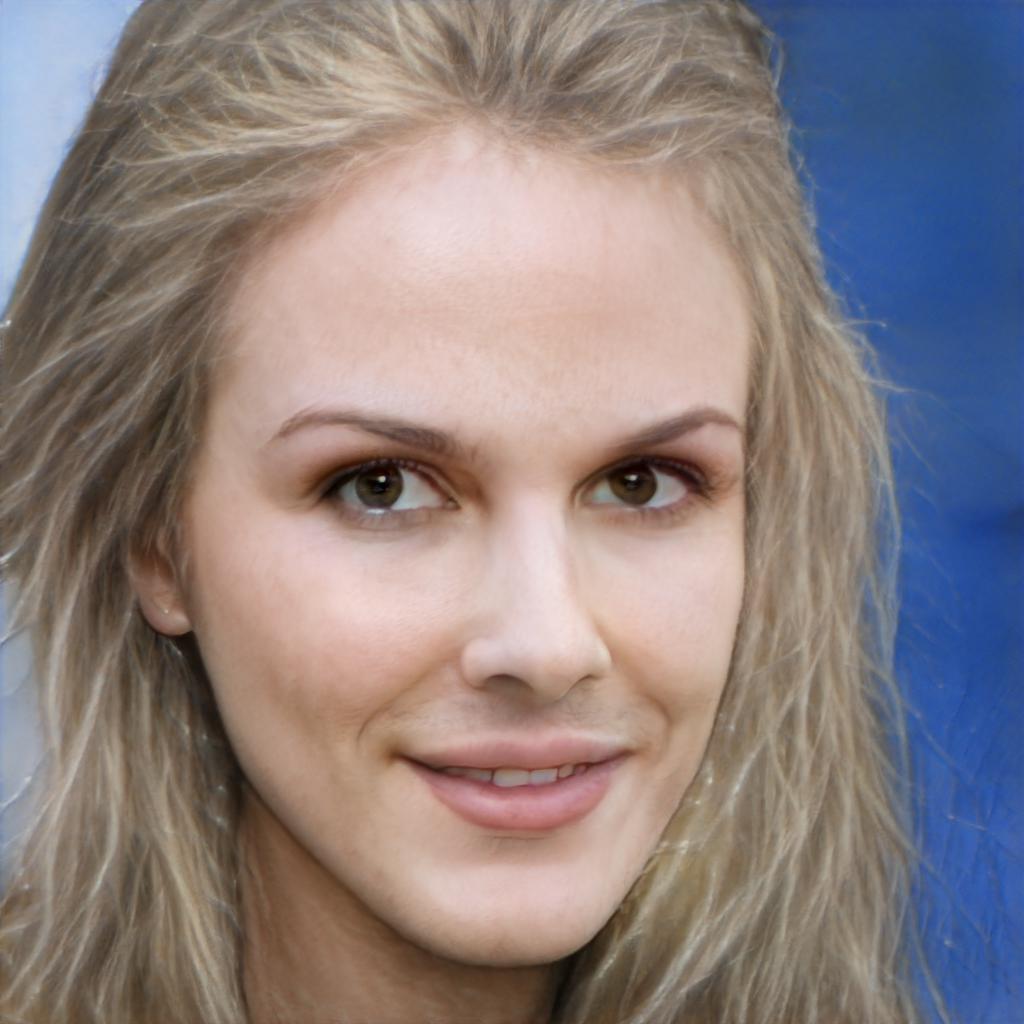} &
 \includegraphics[width=0.13\textwidth, ]{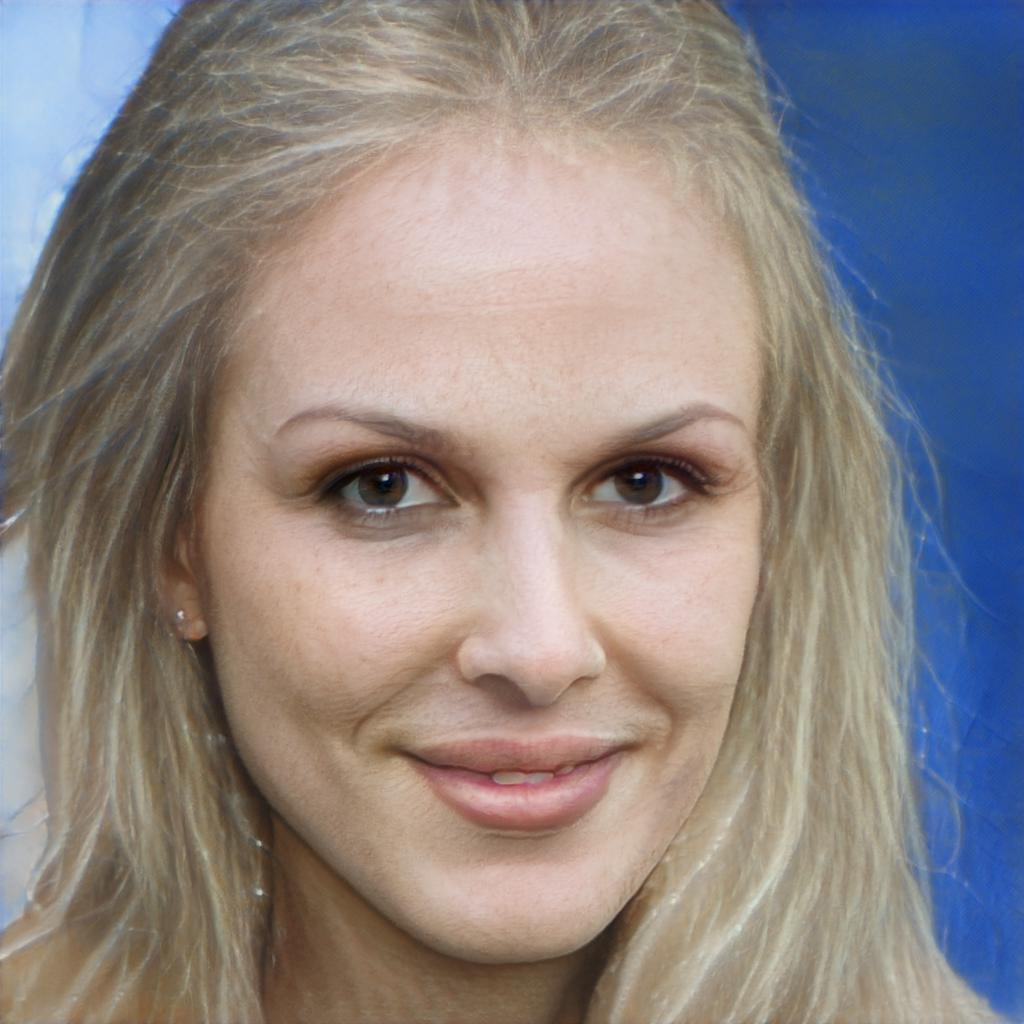} &
 \includegraphics[width=0.13\textwidth, ]{images/wstar_id_no_mag_3_coupl_dist_edit_train_ep5/36_img_10.jpg} &
 \\
\end{tabular}
\caption{Ablation study: Image editing using InterFaceGAN. }
\label{fig:edit_abl_3}
\end{figure*}

In this section, we will investigate the effect of different losses used for image editing. The same setup is used as in Section 4.3.2 except that we fix the editing step to $10$ in \wstar (except we put it $6$ for \wstarasp). We compare the following choices:
\begin{itemize}
    \item \wstara, where $T$ is trained only with the attribute separation loss eq. (2),
    \item \wstara-ID: \wstara with the identity preservation loss,
    \item \wstara-ID*: \wstara with lower weight for the identity loss $\lambda_{ID}=0.1$,
    \item \wstara-ID$^{\dagger}$: \wstara-ID with random classifiers. 
\end{itemize}  
From Figure \ref{fig:edit_abl_1} , \ref{fig:edit_abl_2}, \ref{fig:edit_abl_3}, we can conclude that:
\begin{enumerate}
    \item The identity loss helps to preserve the identity after editing (Row 2 vs 3, 4 and 5). In addition, there is a trade-off between preserving the identity and the amount of editing effect. Specifically, increasing $\lambda_{ID}$ lead to lower editing effect (Row 3 vs 5). 
    \item The latent distance unfolding loss has negligible effect on editing (Row 3 vs 6).
    \item Pretraining the classifiers does not seem to improve the editing results.
    
\end{enumerate}

\begin{figure*}[h]
\setlength\tabcolsep{2pt}%
\centering
\begin{tabular}{p{0.25cm}ccccccc}
\centering
&
 \textbf{Original} &
 \textbf{Inverted} &
 \textbf{Makeup} &
 \textbf{Male} &
 \textbf{Mustache} &
 \textbf{Chubby} &
 \textbf{Lipstick} &
 \begin{turn}{90} \hspace{0.5cm} \wplus\end{turn} &
 \includegraphics[width=0.13\textwidth]{images/original/06004.jpg} & 
 \includegraphics[width=0.13\textwidth, ]{images/inversion/18_orig_img_3.jpg} &
 \includegraphics[width=0.13\textwidth, ]{images/wplus/18_img_5.jpg} &
 \includegraphics[width=0.13\textwidth, ]{images/wplus/20_img_5.jpg} &
 \includegraphics[width=0.13\textwidth, ]{images/wplus/22_img_5.jpg} &
 \includegraphics[width=0.13\textwidth, ]{images/wplus/13_img_5.jpg} &
 \includegraphics[width=0.13\textwidth]{images/wplus/36_img_5.jpg} &
 \begin{turn}{90} \hspace{0.5cm} \wstara\end{turn} &
 \includegraphics[width=0.13\textwidth]{images/original/06004.jpg} & 
 \includegraphics[width=0.13\textwidth, ]{images/inversion/18_orig_img_3.jpg} &
 \includegraphics[width=0.13\textwidth, ]{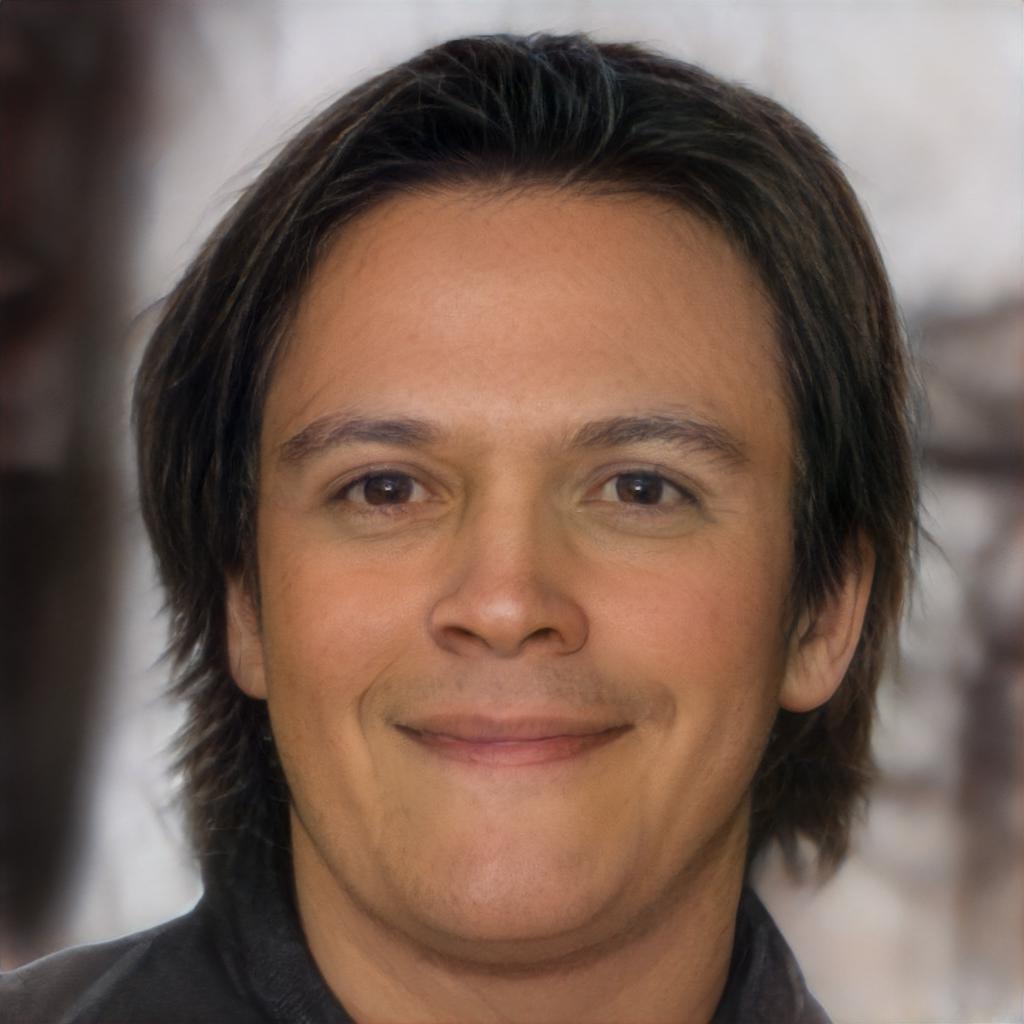} &
 \includegraphics[width=0.13\textwidth, ]{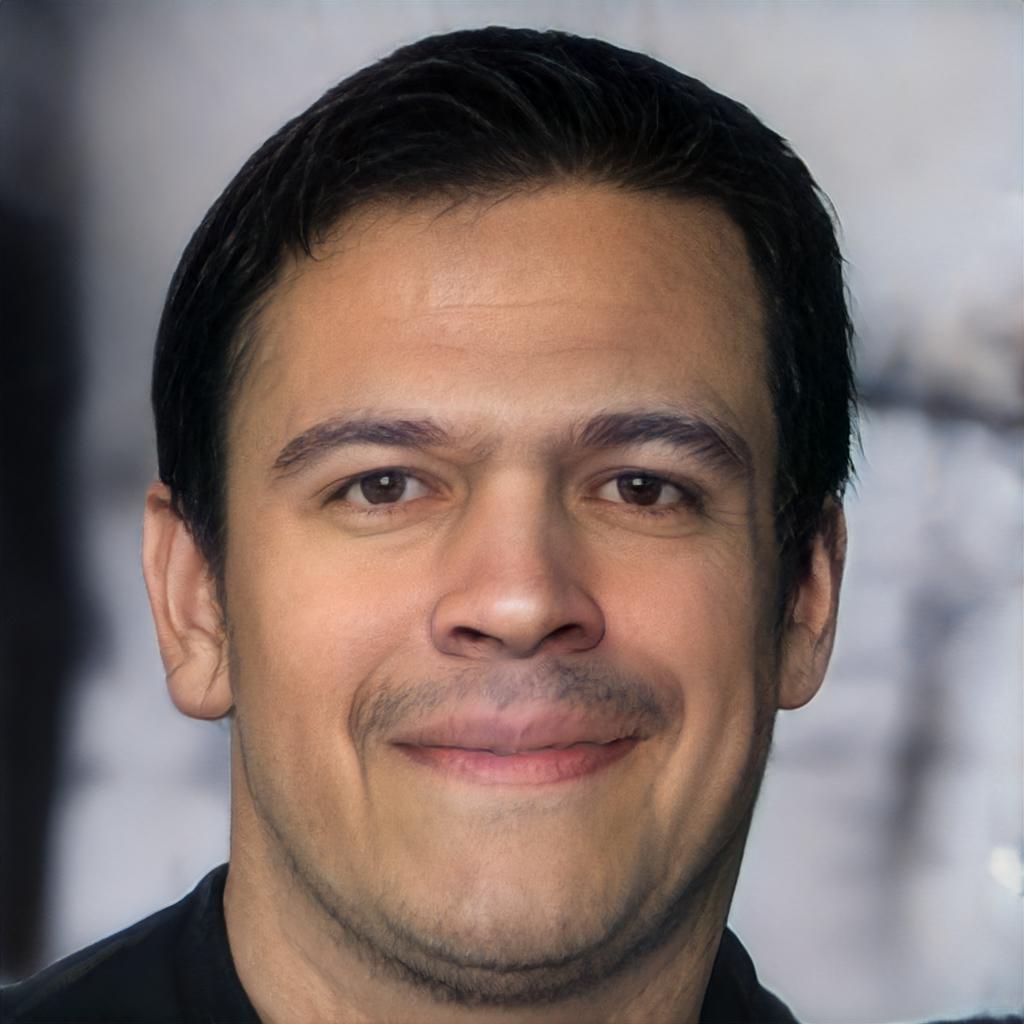} &
 \includegraphics[width=0.13\textwidth, ]{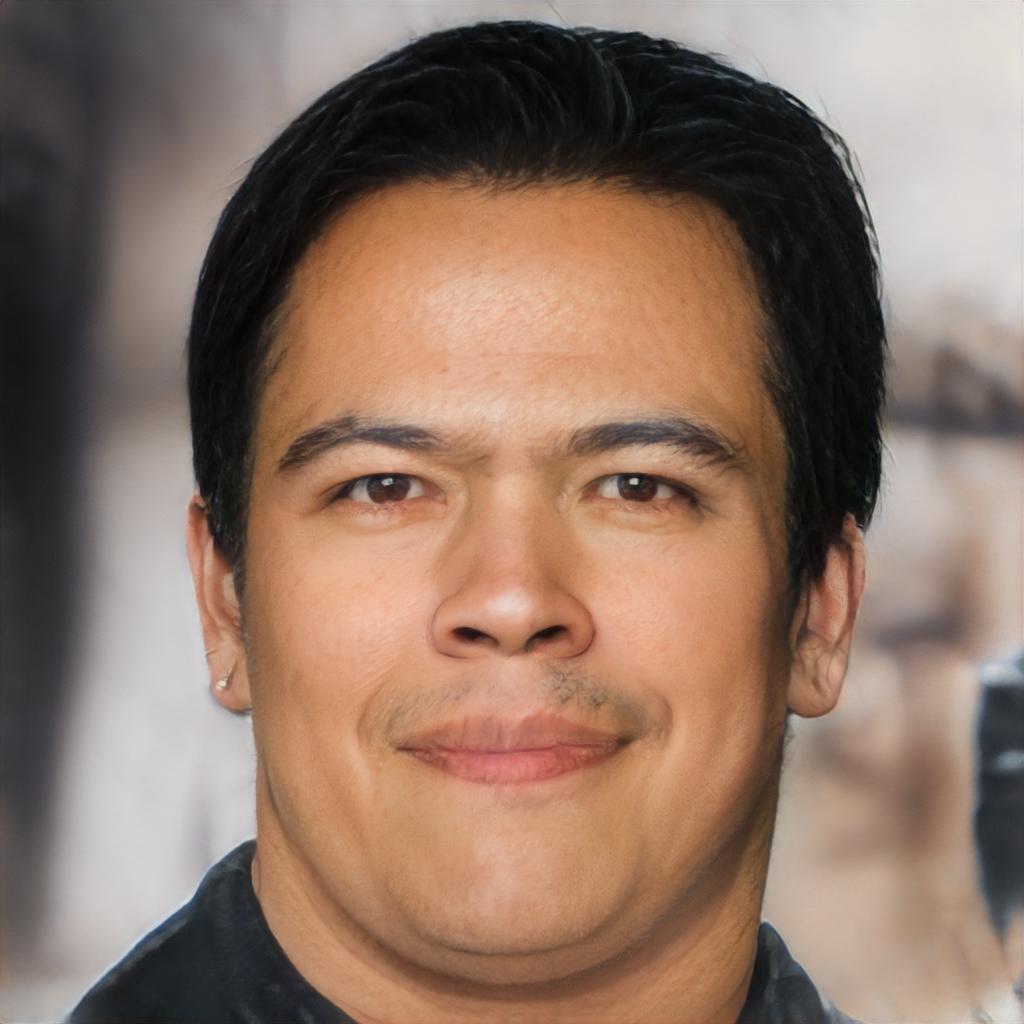} &
 \includegraphics[width=0.13\textwidth, ]{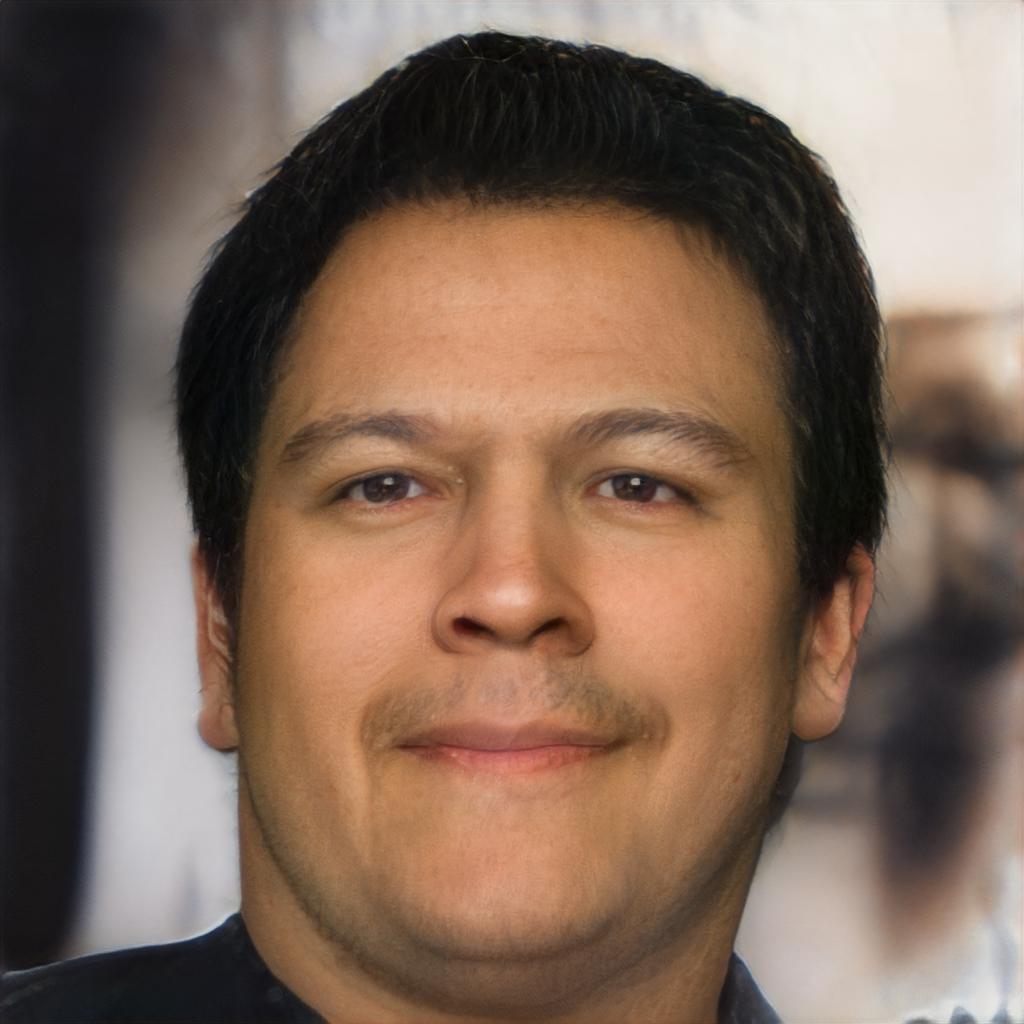} &
 \includegraphics[width=0.13\textwidth, ]{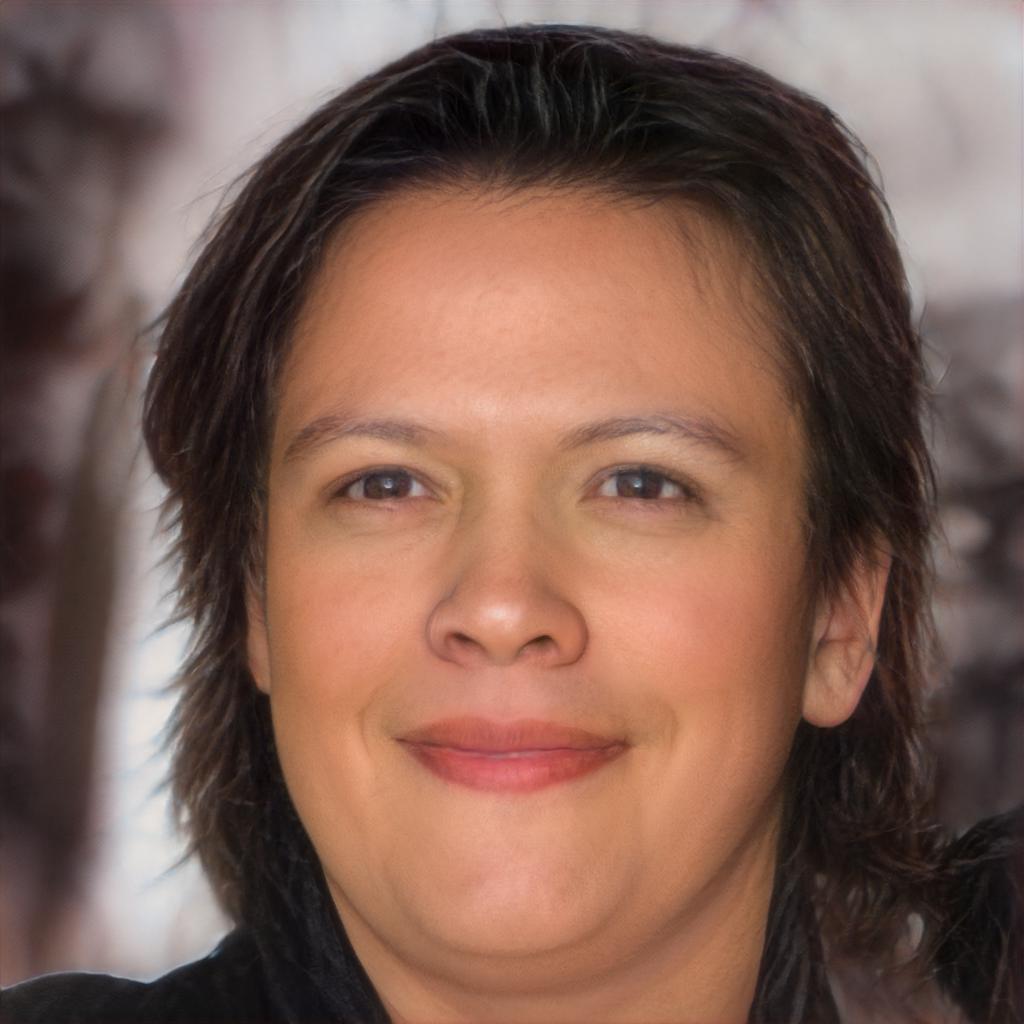} &
 \begin{turn}{90} \hspace{0.2cm} \wstara-ID\end{turn} &
 \includegraphics[width=0.13\textwidth]{images/original/06004.jpg} & 
 \includegraphics[width=0.13\textwidth, ]{images/inversion/18_orig_img_3.jpg} &
 \includegraphics[width=0.13\textwidth, ]{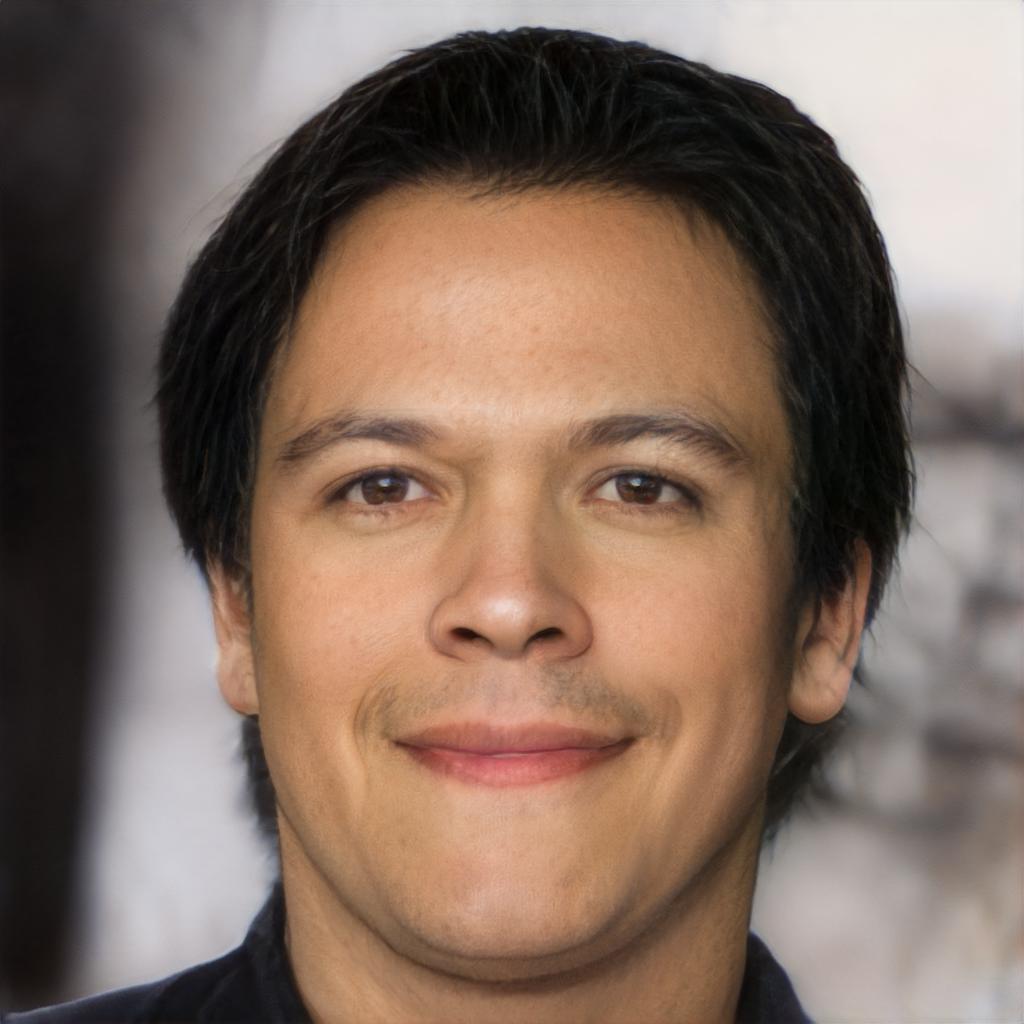} &
 \includegraphics[width=0.13\textwidth, ]{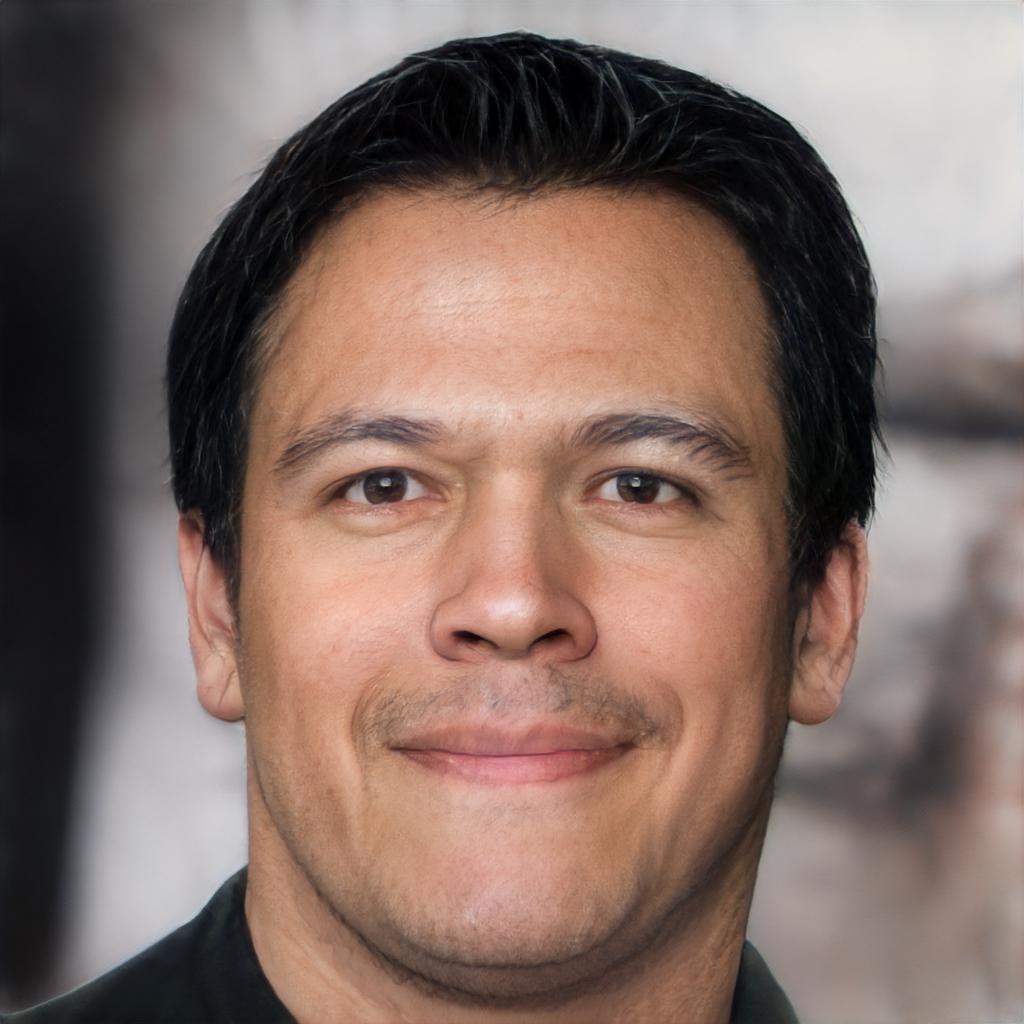} &
 \includegraphics[width=0.13\textwidth, ]{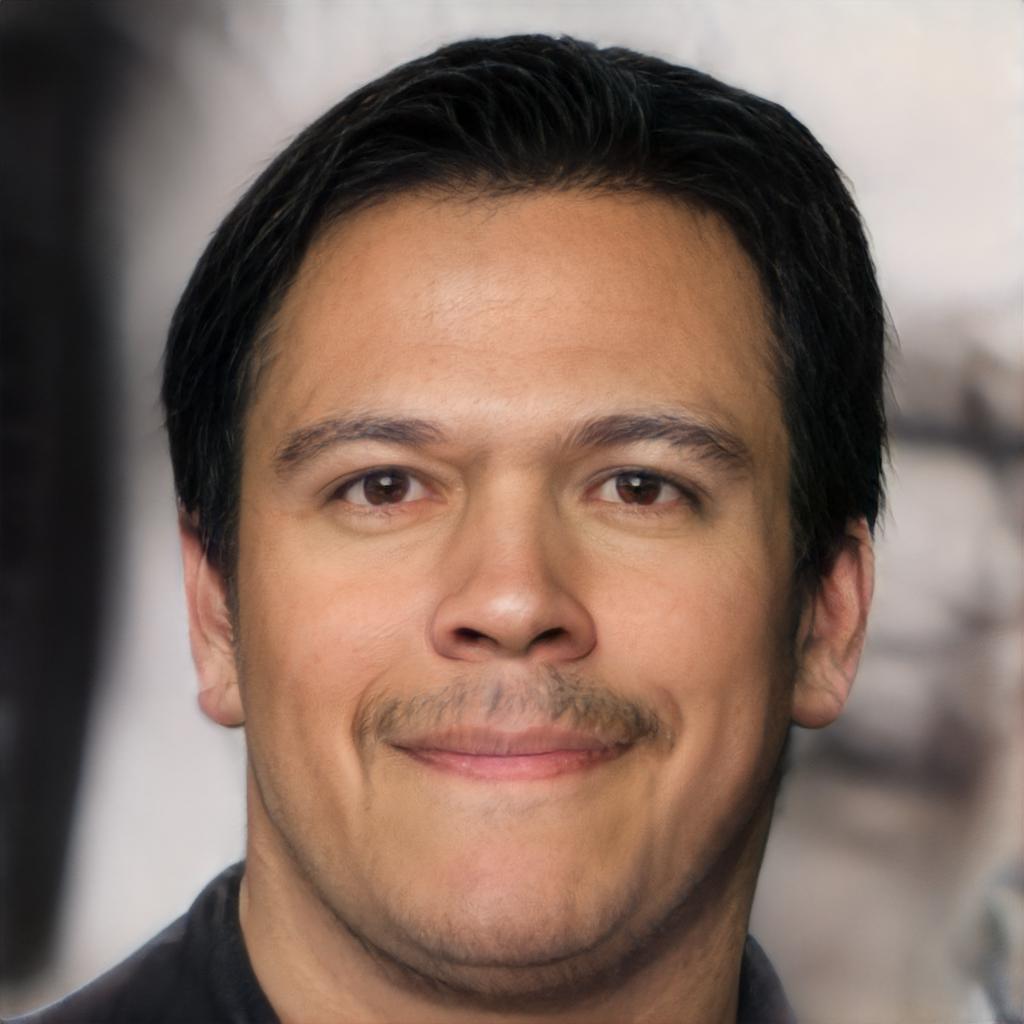} &
 \includegraphics[width=0.13\textwidth, ]{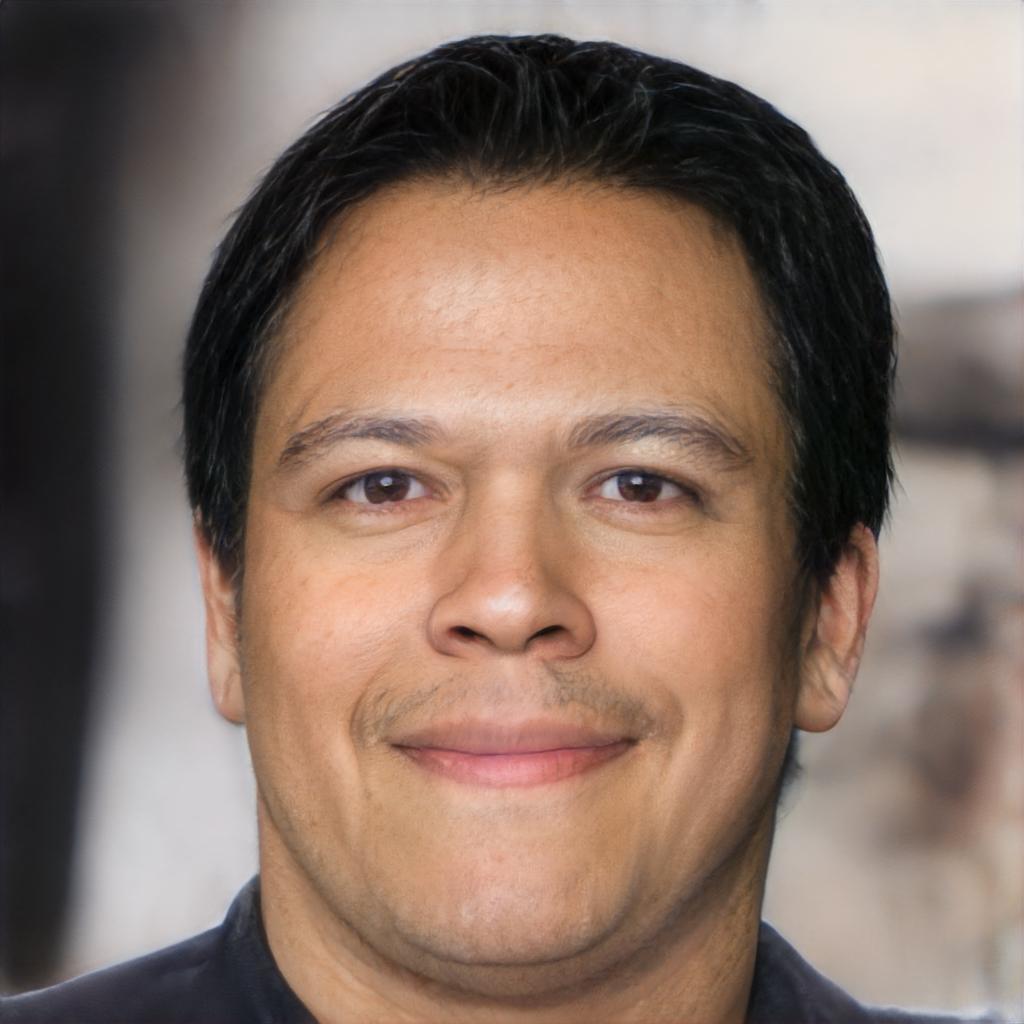} &
 \includegraphics[width=0.13\textwidth, ]{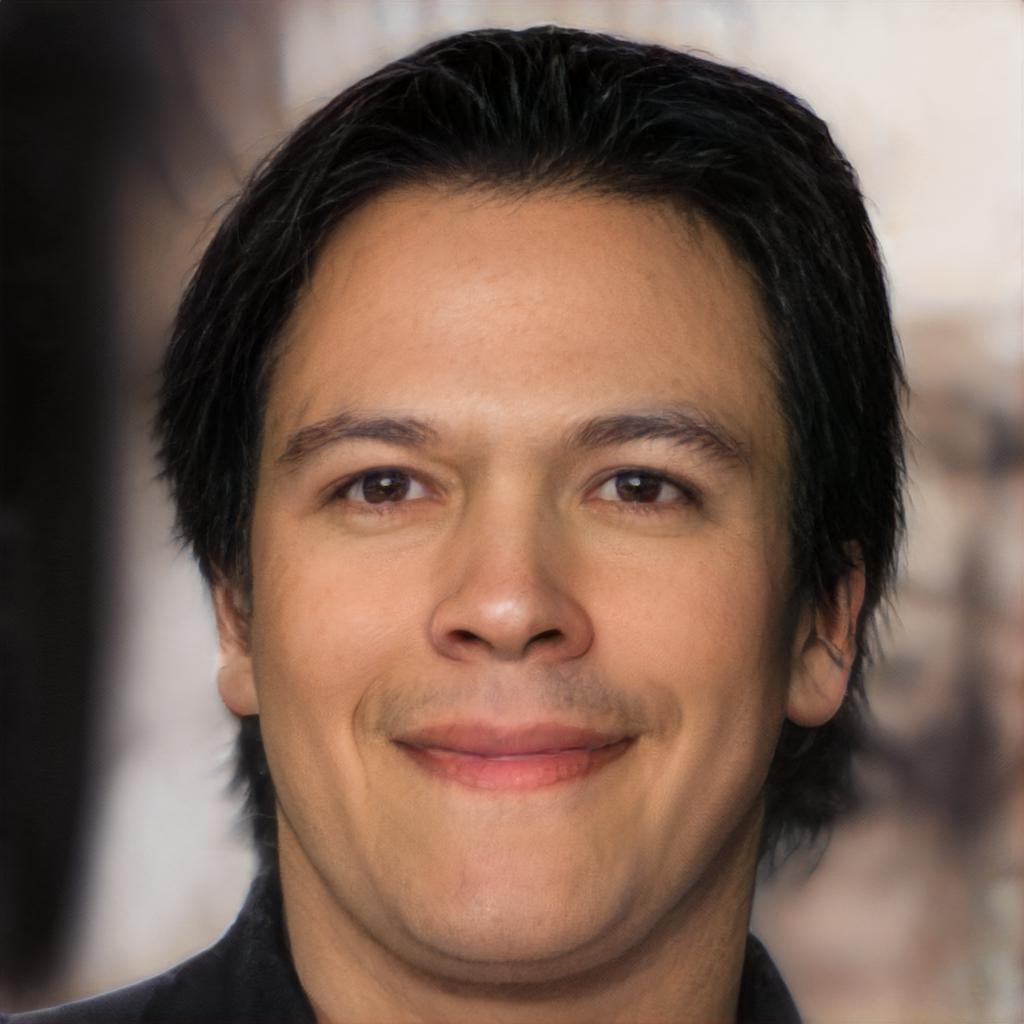} &
 \begin{turn}{90} \wstara-ID^{\dagger}\end{turn} &
 \includegraphics[width=0.13\textwidth]{images/original/06004.jpg} & 
 \includegraphics[width=0.13\textwidth, ]{images/inversion/18_orig_img_3.jpg} &
 \includegraphics[width=0.13\textwidth, ]{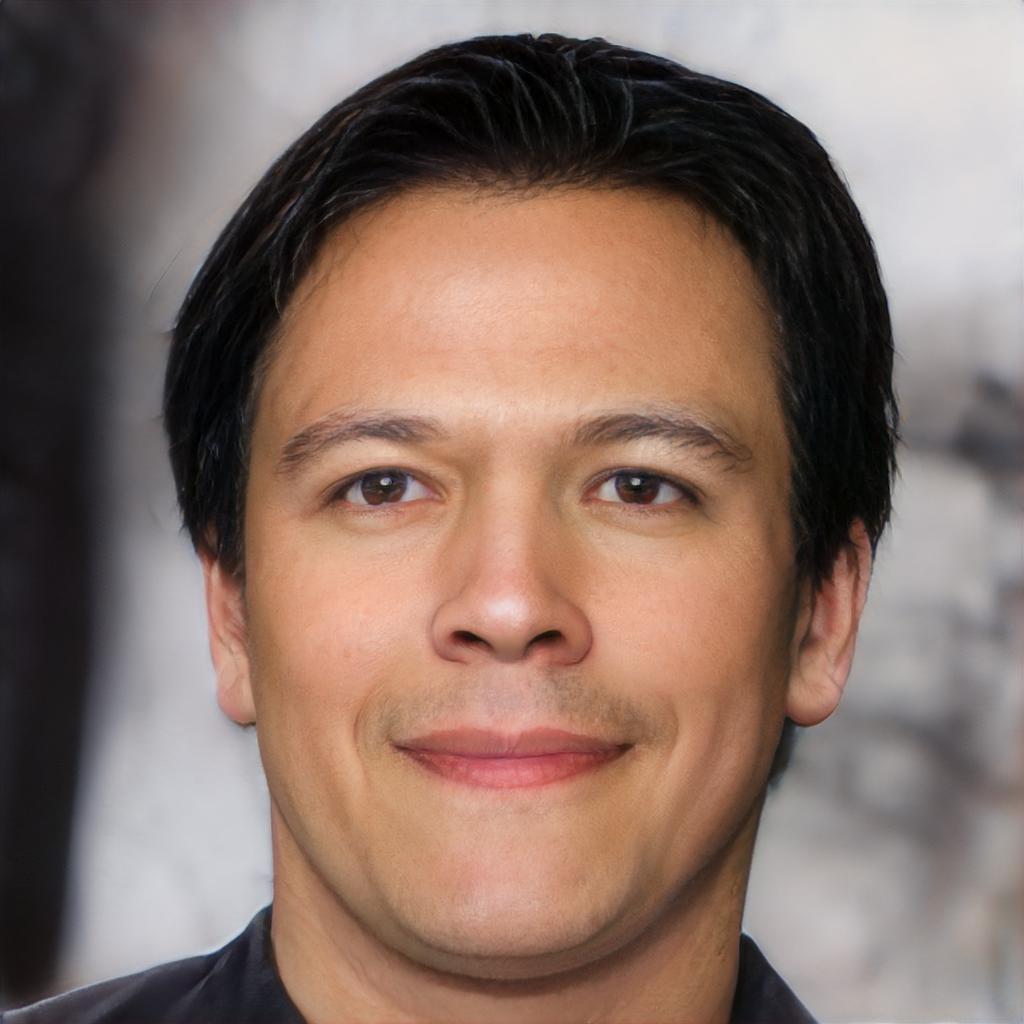} &
 \includegraphics[width=0.13\textwidth, ]{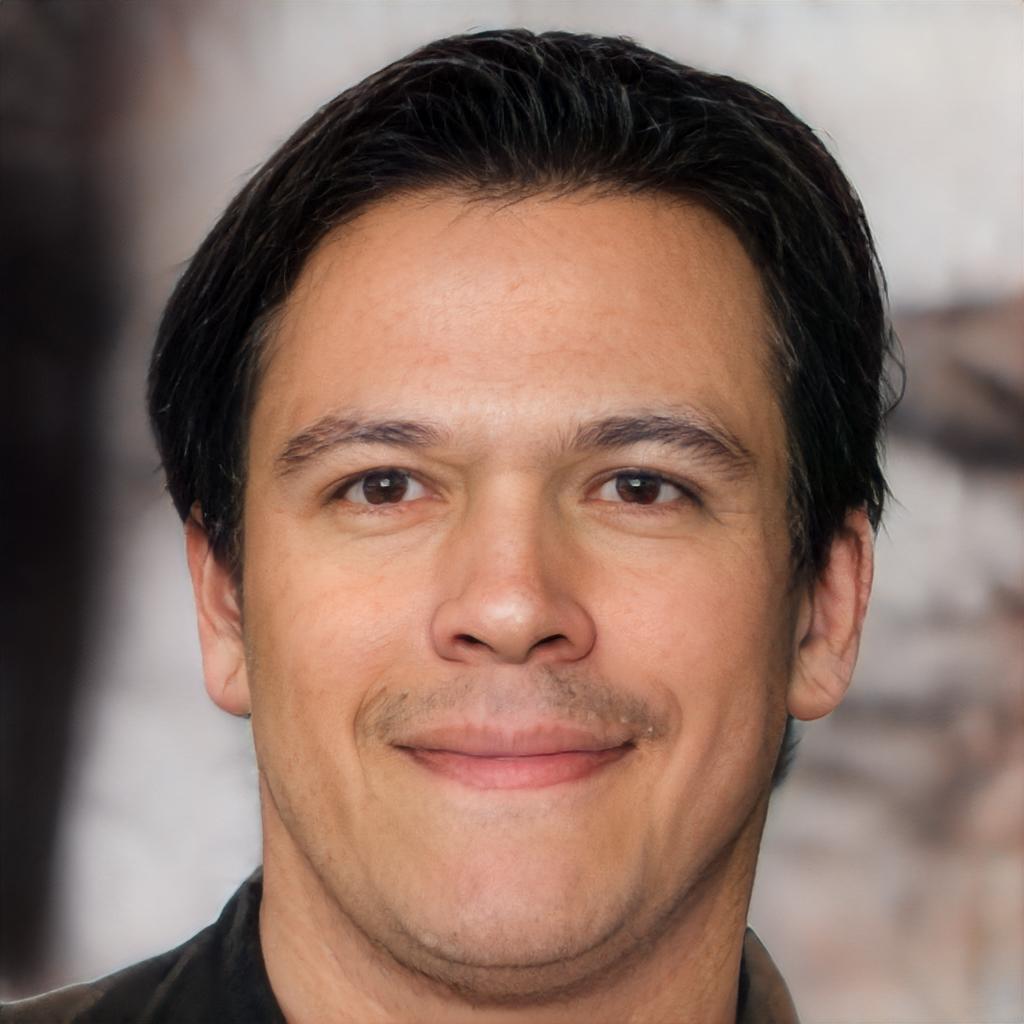} &
 \includegraphics[width=0.13\textwidth, ]{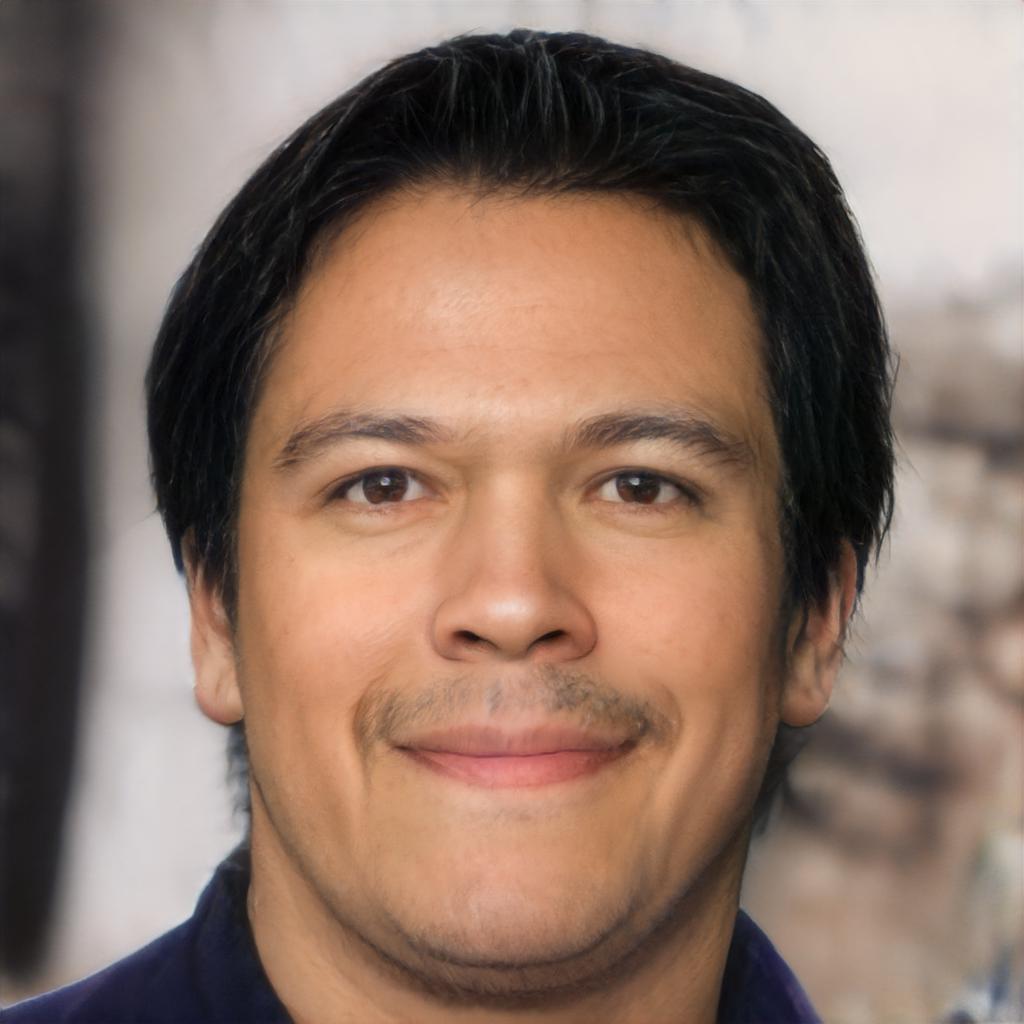} &
 \includegraphics[width=0.13\textwidth, ]{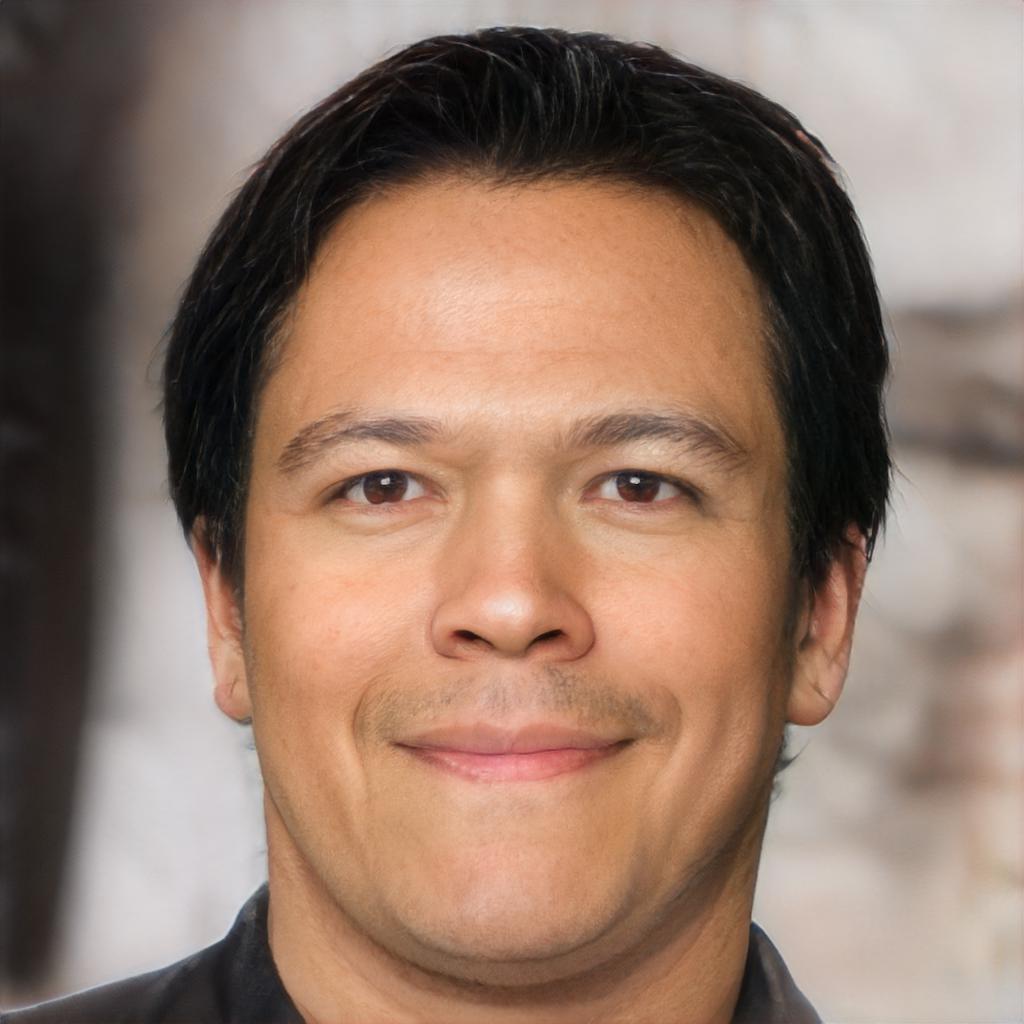} &
 \includegraphics[width=0.13\textwidth, ]{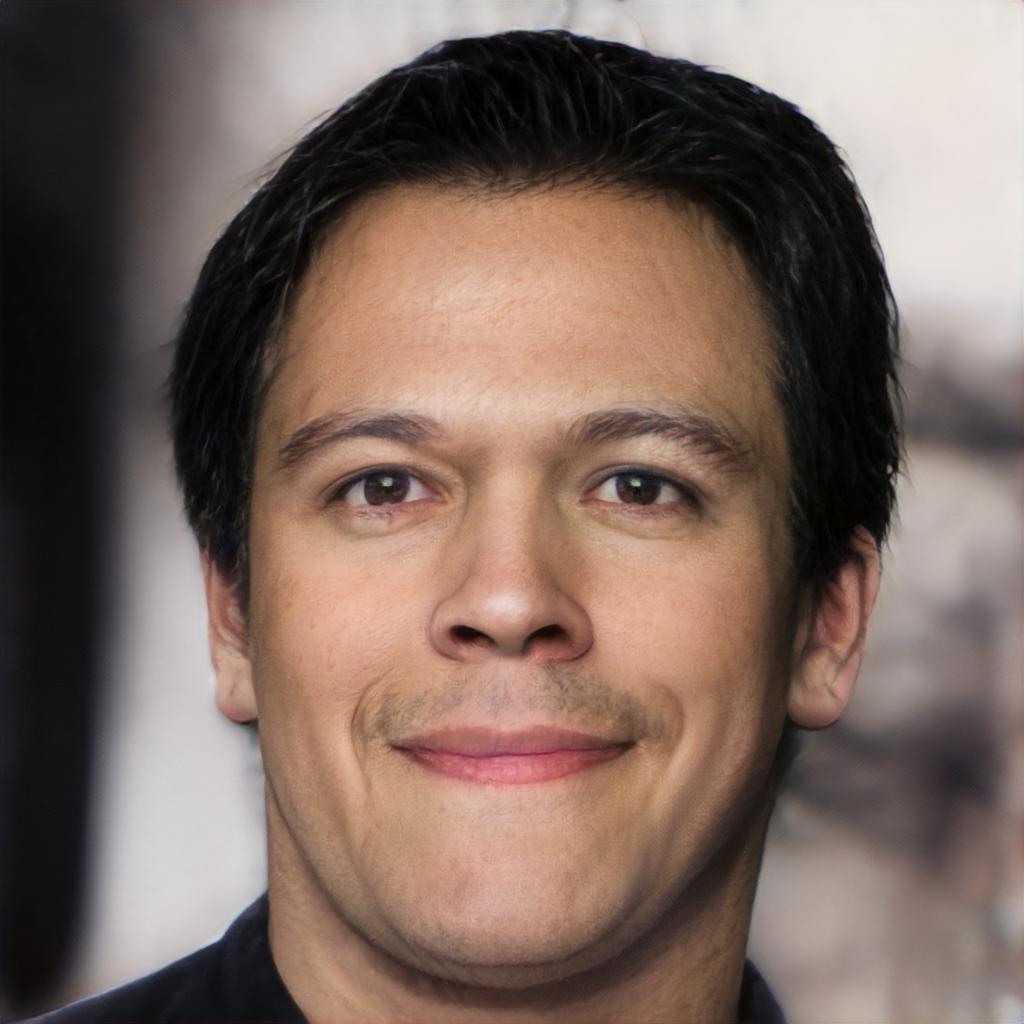} &
 \begin{turn}{90} \wstara-ID^{\star}\end{turn} &
 \includegraphics[width=0.13\textwidth]{images/original/06004.jpg} & 
 \includegraphics[width=0.13\textwidth, ]{images/inversion/18_orig_img_3.jpg} &
 \includegraphics[width=0.13\textwidth, ]{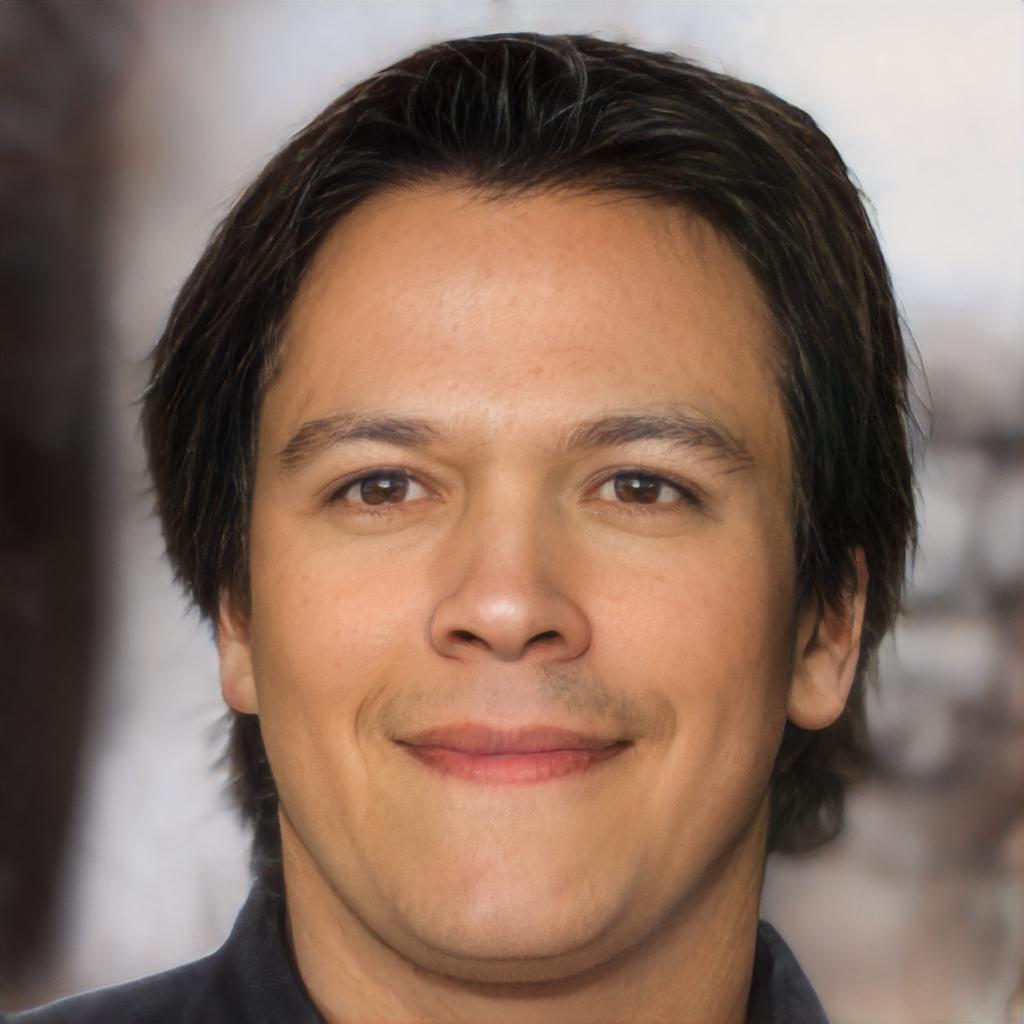} &
 \includegraphics[width=0.13\textwidth, ]{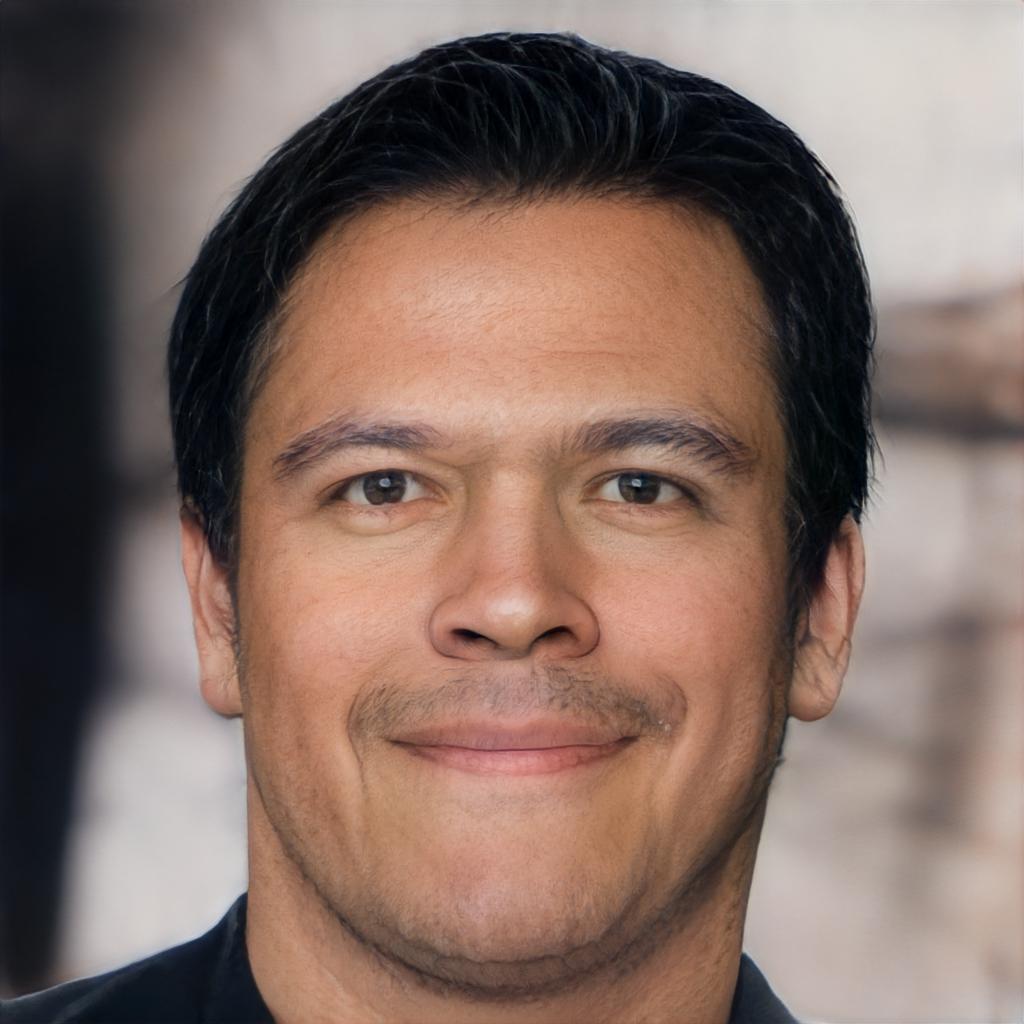} &
 \includegraphics[width=0.13\textwidth, ]{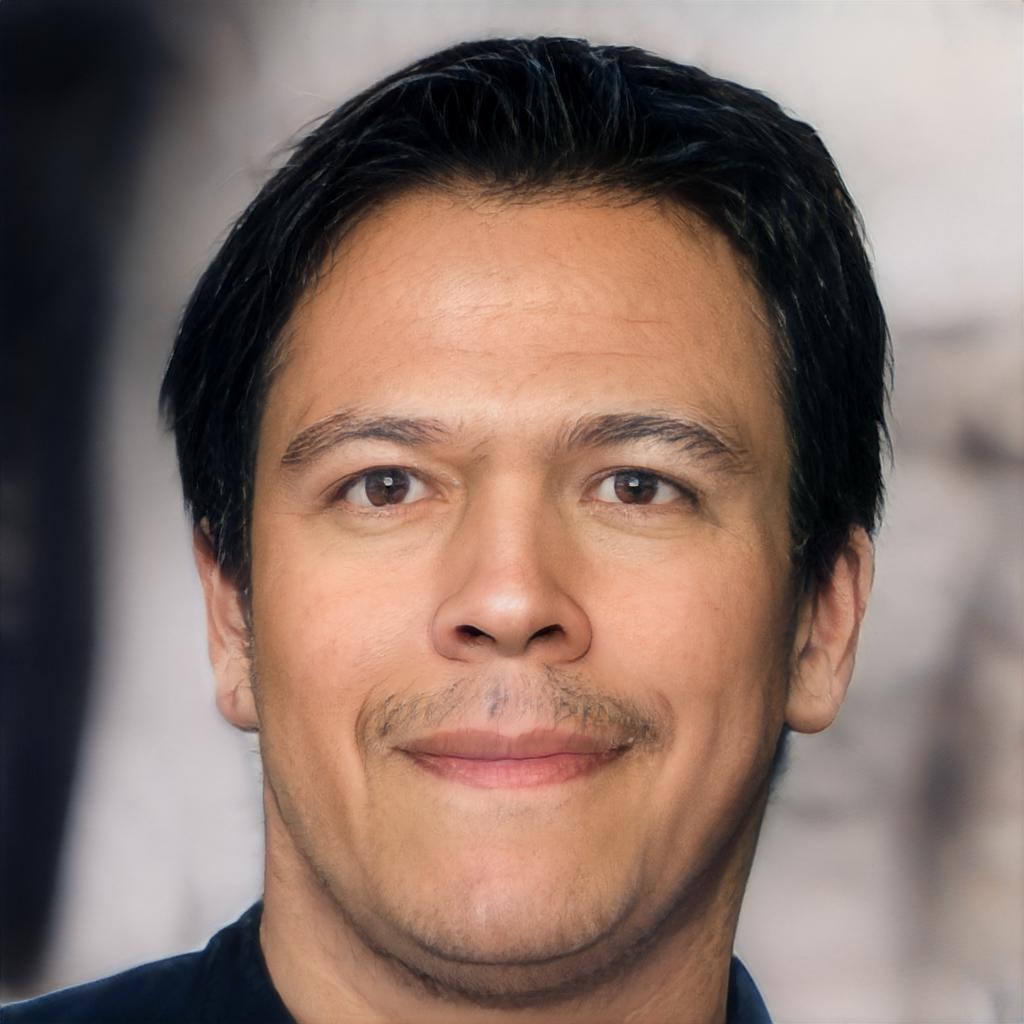} &
 \includegraphics[width=0.13\textwidth, ]{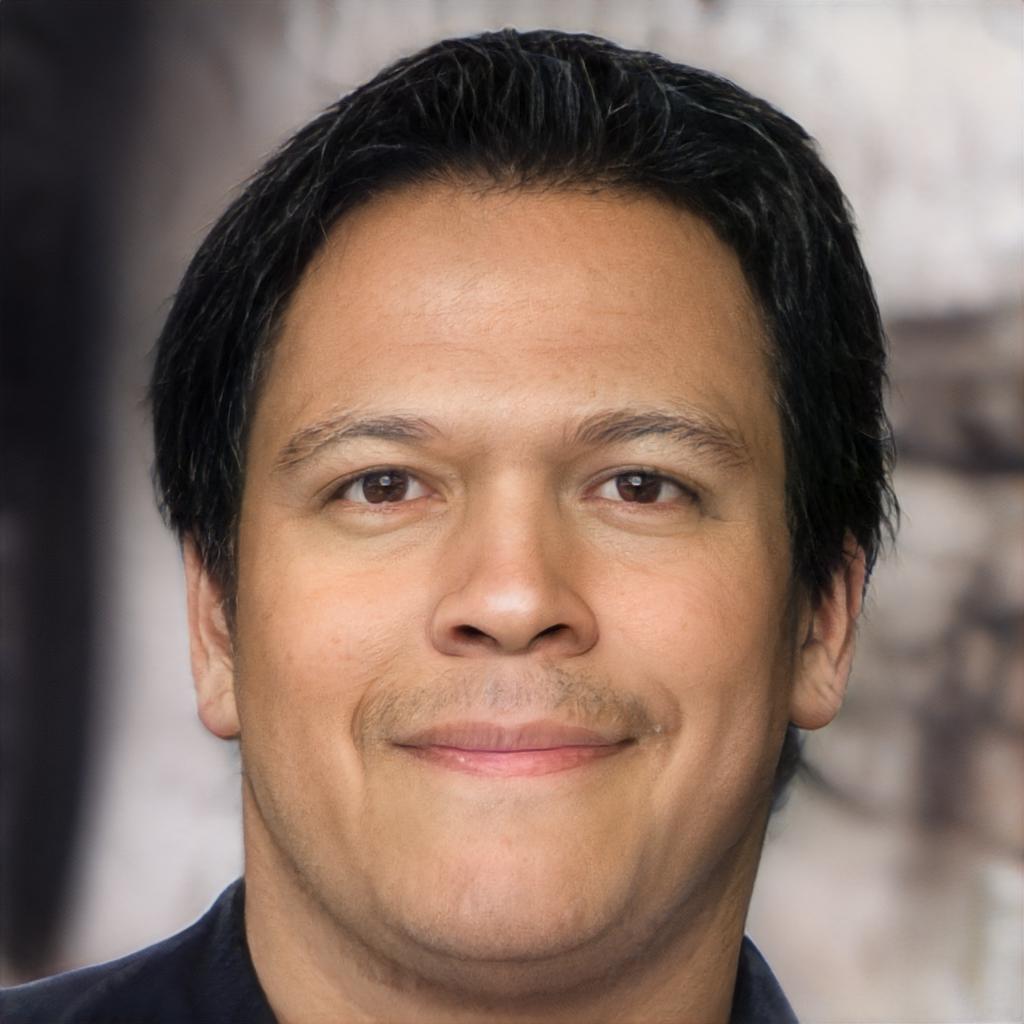} &
 \includegraphics[width=0.13\textwidth, ]{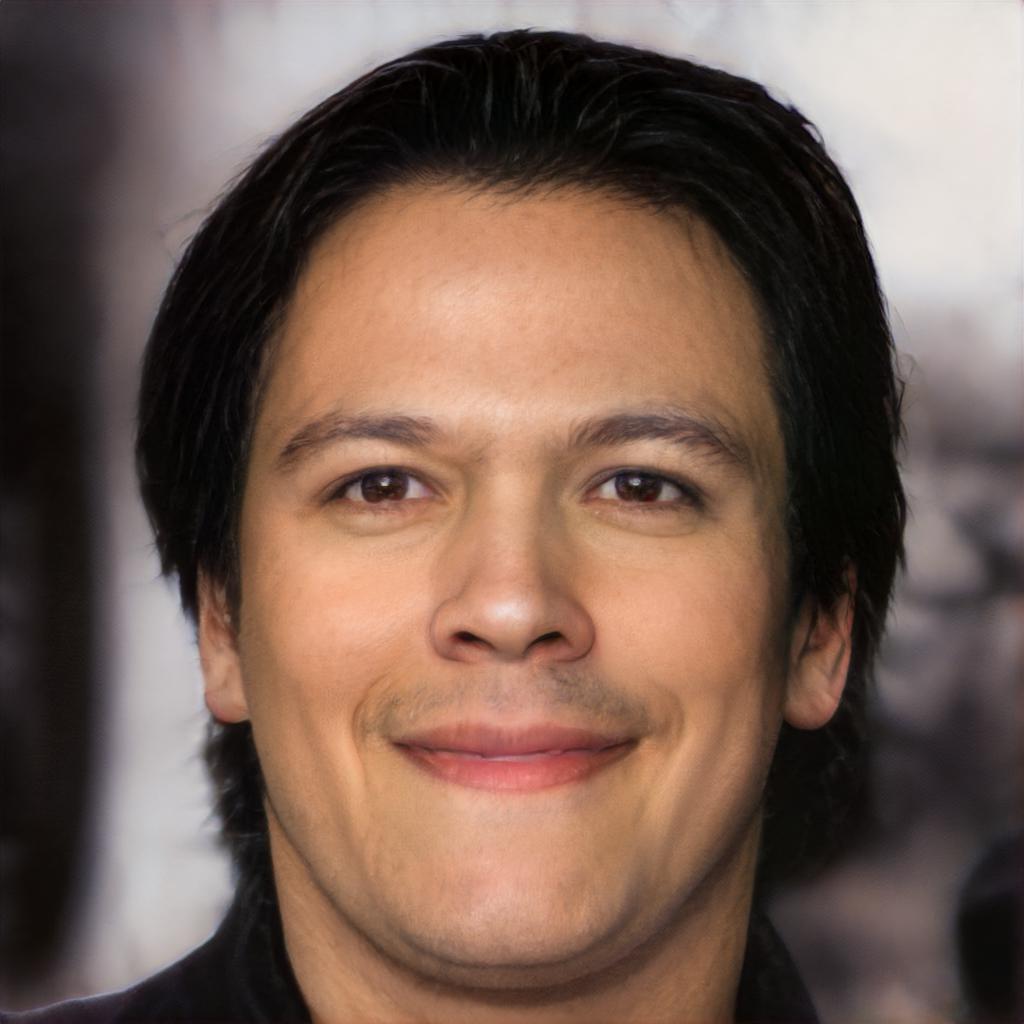} &
 \begin{turn}{90} \hspace{0.4cm} $\mathcal{W}^{\star}_{ID}$ \end{turn} &
 \includegraphics[width=0.13\textwidth]{images/original/06004.jpg} & 
 \includegraphics[width=0.13\textwidth, ]{images/inversion/18_orig_img_3.jpg} &
 \includegraphics[width=0.13\textwidth, ]{images/wstar_id_no_mag_3_coupl_dist_edit_train_ep5/18_img_3.jpg} &
 \includegraphics[width=0.13\textwidth, ]{images/wstar_id_no_mag_3_coupl_dist_edit_train_ep5/20_img_3.jpg} &
 \includegraphics[width=0.13\textwidth, ]{images/wstar_id_no_mag_3_coupl_dist_edit_train_ep5/22_img_3.jpg} &
 \includegraphics[width=0.13\textwidth, ]{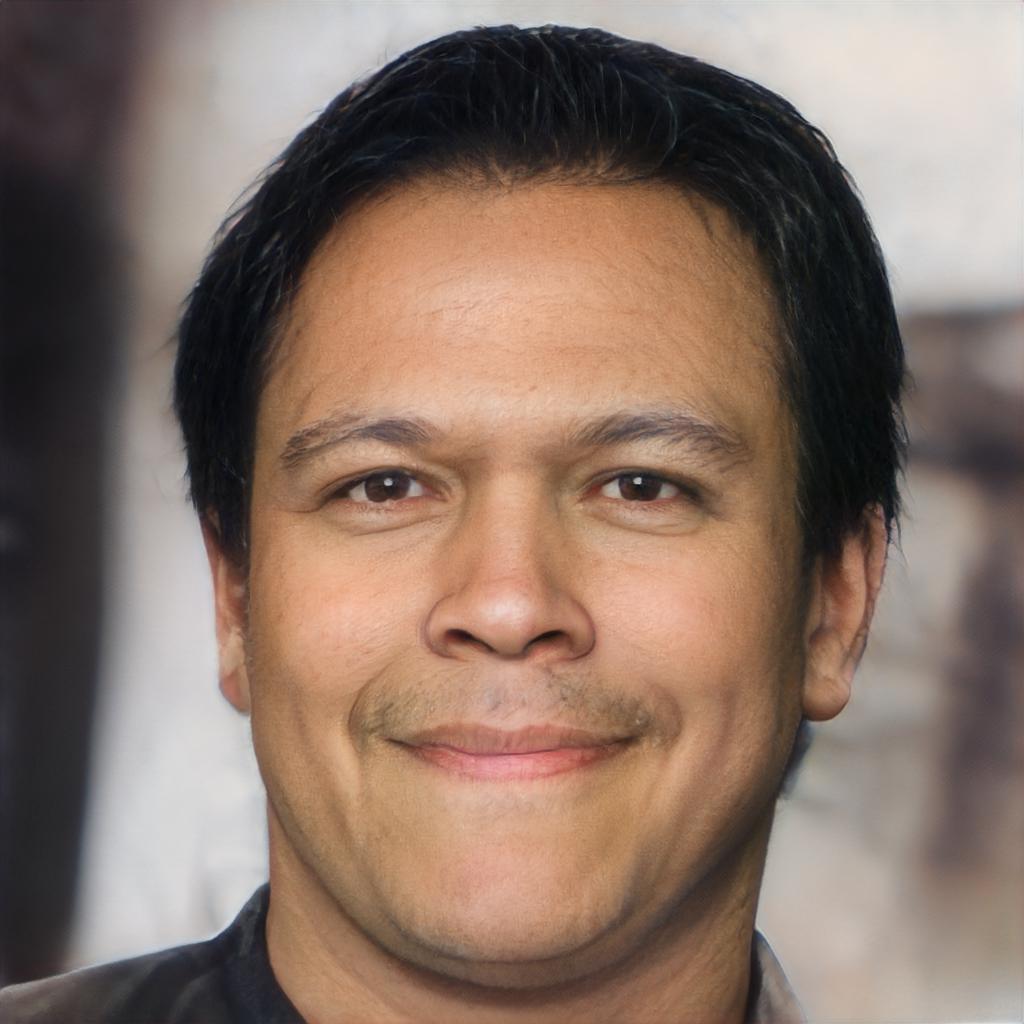} &
 \includegraphics[width=0.13\textwidth, ]{images/wstar_id_no_mag_3_coupl_dist_edit_train_ep5/36_img_3.jpg} &
 
 \\
\end{tabular}
\caption{Ablation study: Image editing using InterFaceGAN.}
\label{fig:edit_abl_1}

\end{figure*}

\begin{figure*}[h]
\setlength\tabcolsep{2pt}%
\centering
\begin{tabular}{p{0.25cm}ccccccc}
\centering
&
 \textbf{Original} &
 \textbf{Inverted} &
 \textbf{Makeup} &
 \textbf{Male} &
 \textbf{Mustache} &
 \textbf{Chubby} &
 \textbf{Lipstick} &
 \begin{turn}{90} \hspace{0.5cm} \wplus\end{turn} &
 \includegraphics[width=0.13\textwidth]{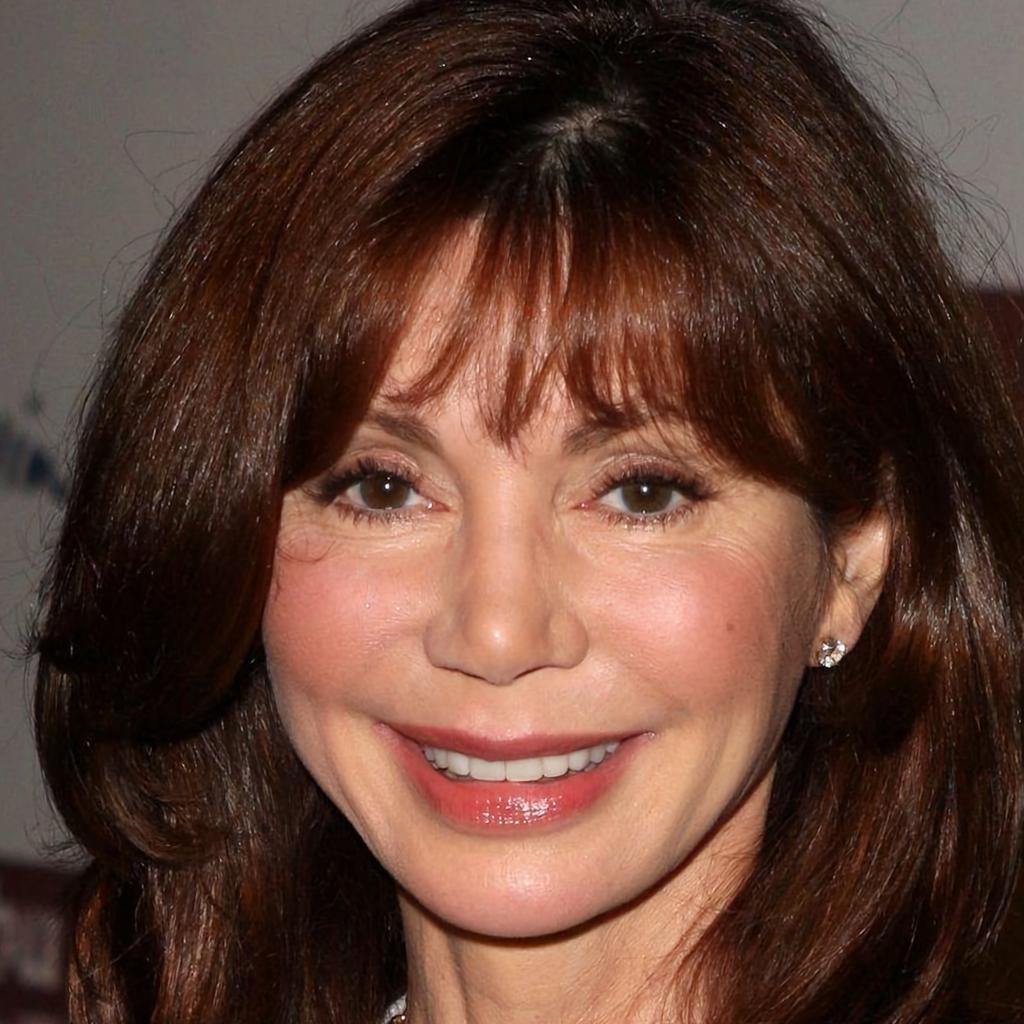} & 
 \includegraphics[width=0.13\textwidth]{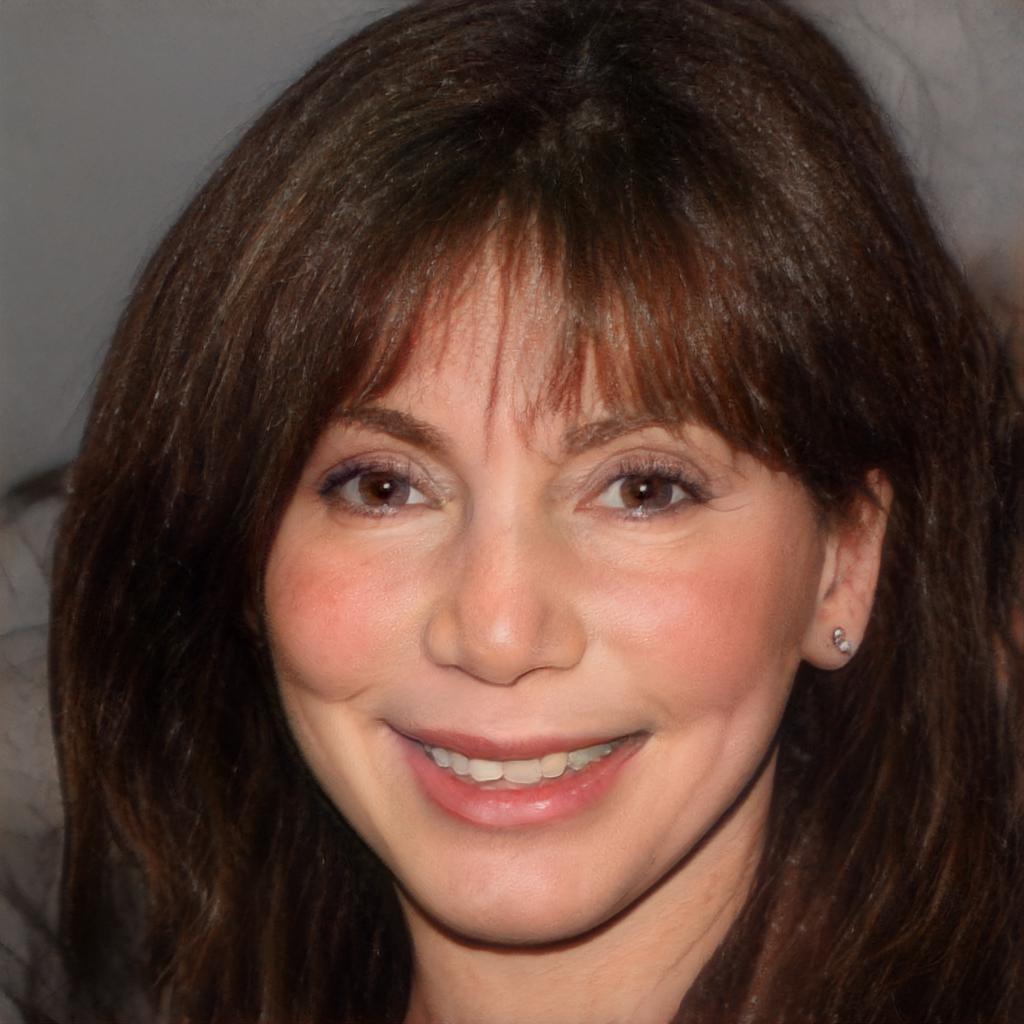} &
 \includegraphics[width=0.13\textwidth]{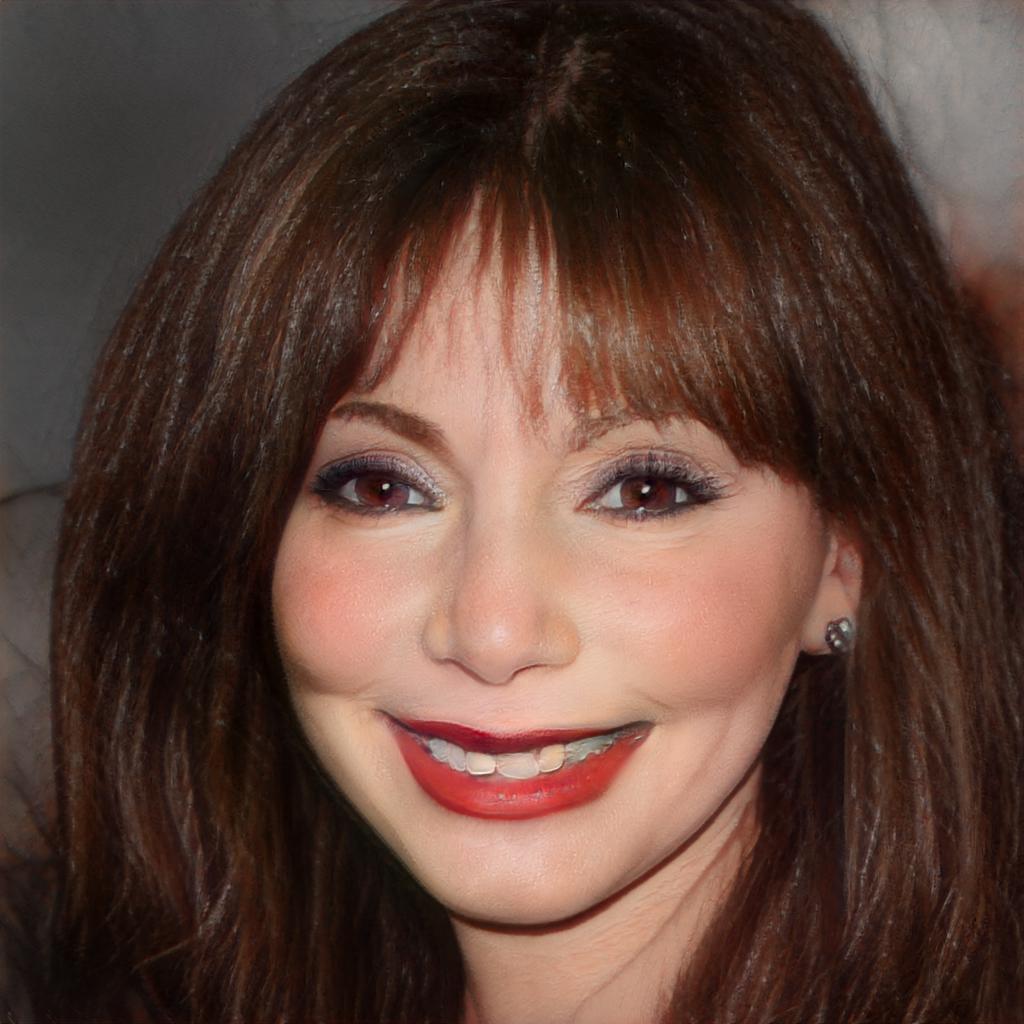} &
 \includegraphics[width=0.13\textwidth]{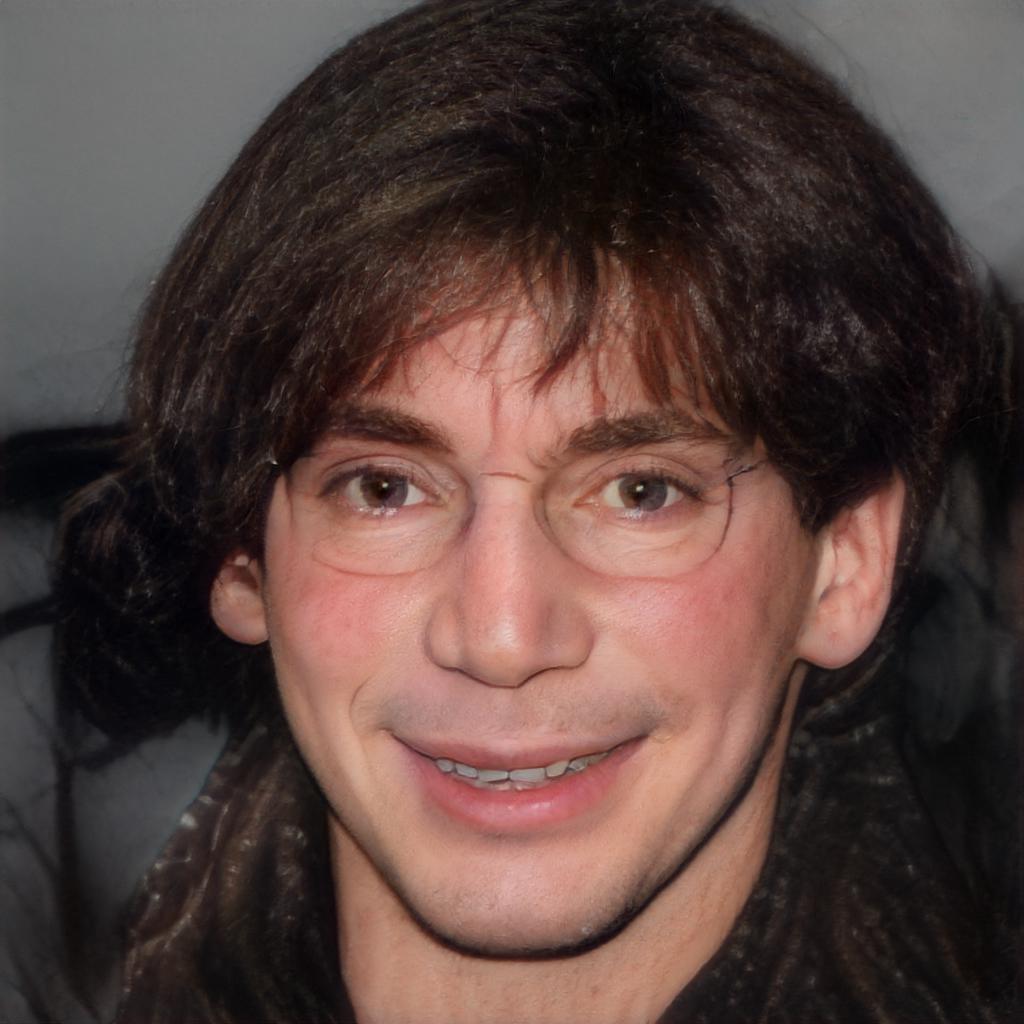} &
 \includegraphics[width=0.13\textwidth]{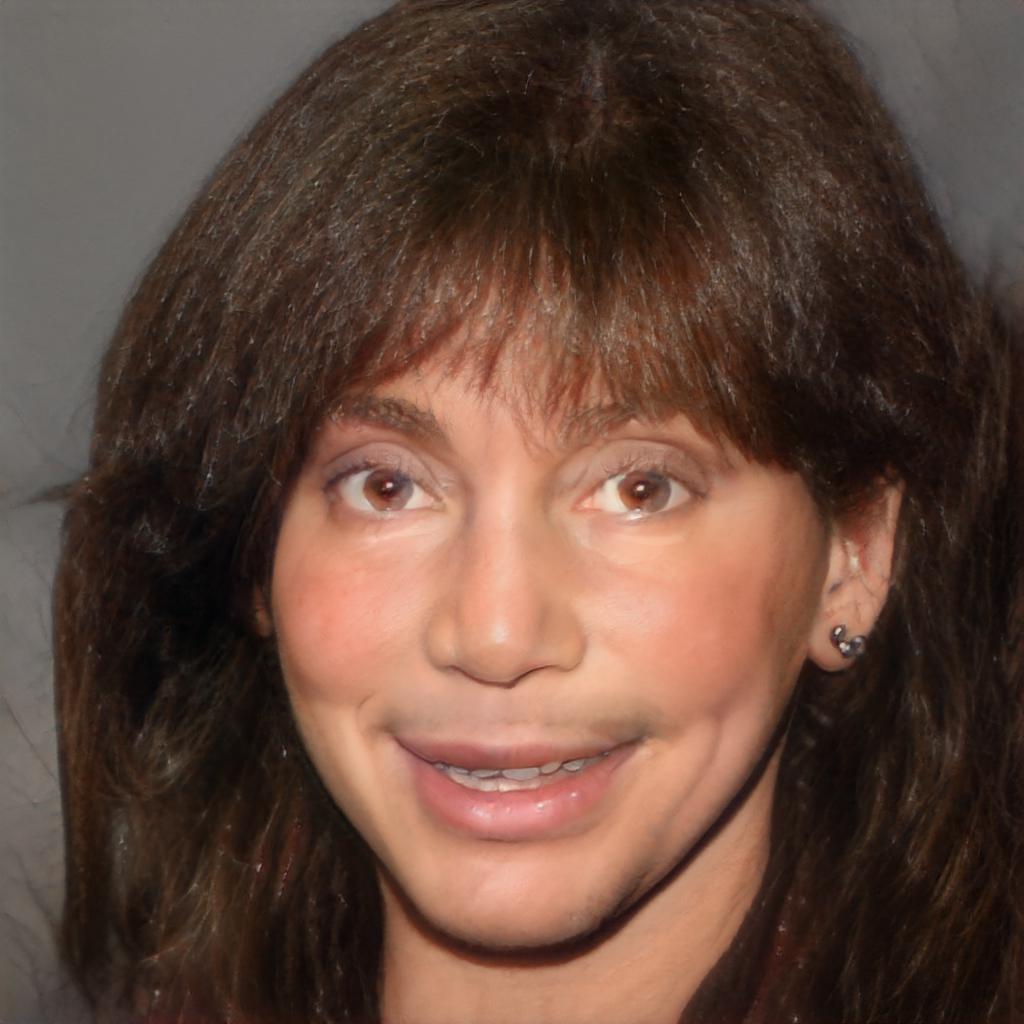} &
 \includegraphics[width=0.13\textwidth]{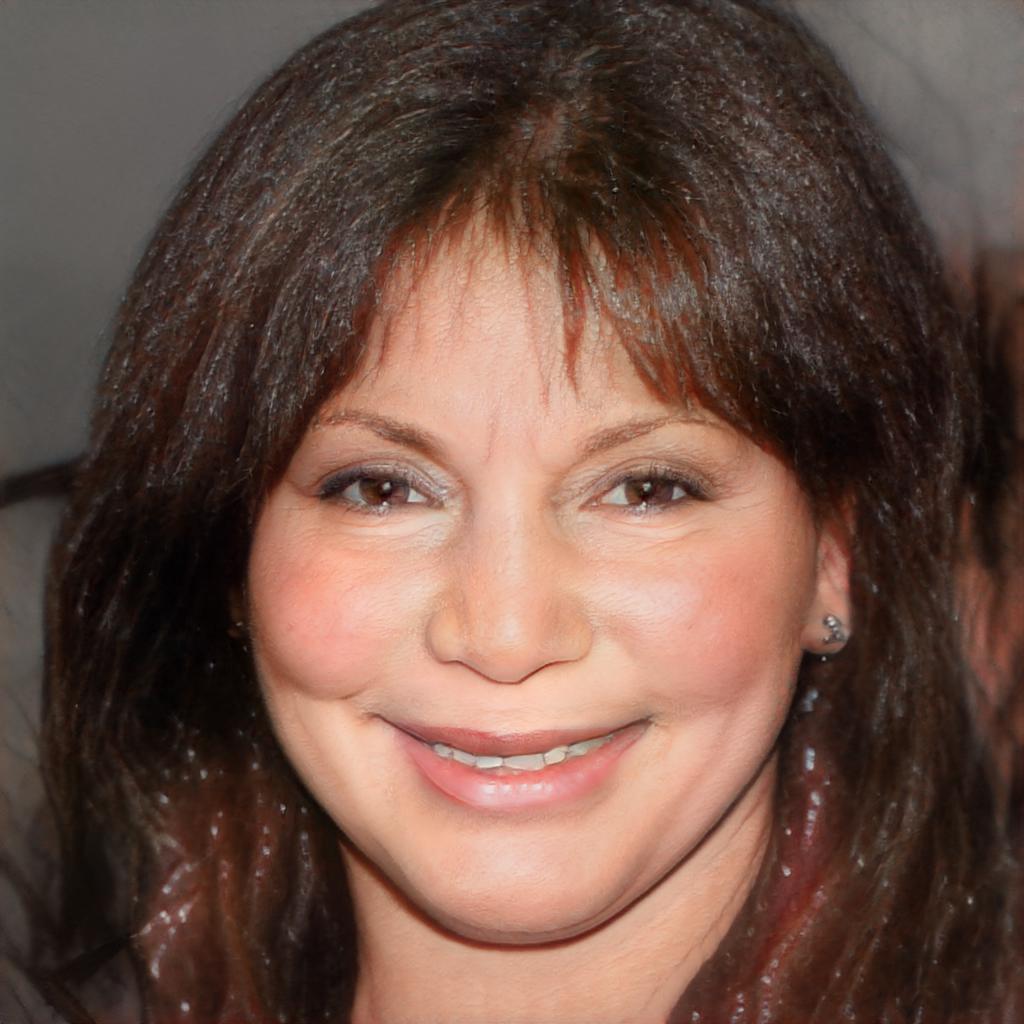} &
 \includegraphics[width=0.13\textwidth]{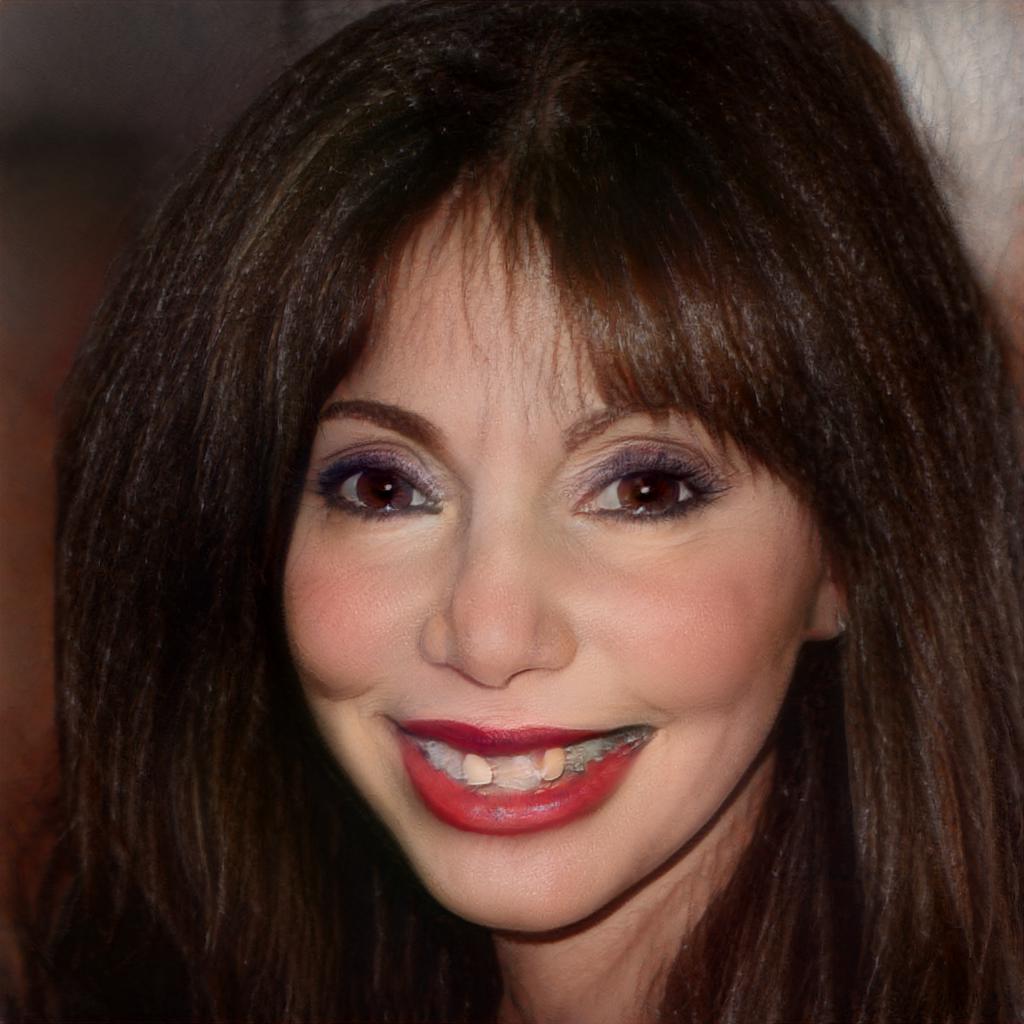} &
\begin{turn}{90} \hspace{0.5cm} $\mathcal{W}^{\star}_{ID}$ \end{turn} &
 \includegraphics[width=0.13\textwidth]{images/original/06001.jpg} & 
 \includegraphics[width=0.13\textwidth]{images/inversion/13_orig_img_2.jpg} &
 \includegraphics[width=0.13\textwidth, ]{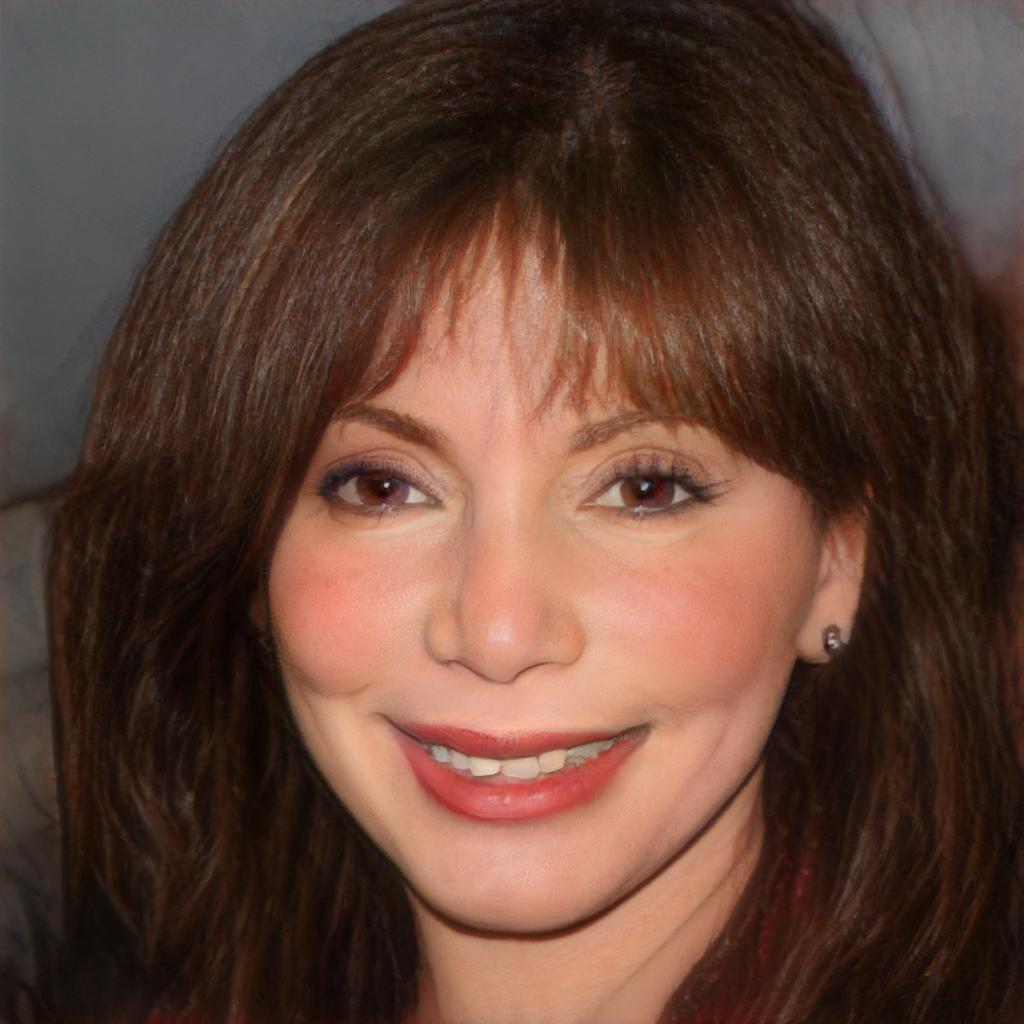} &
 \includegraphics[width=0.13\textwidth, ]{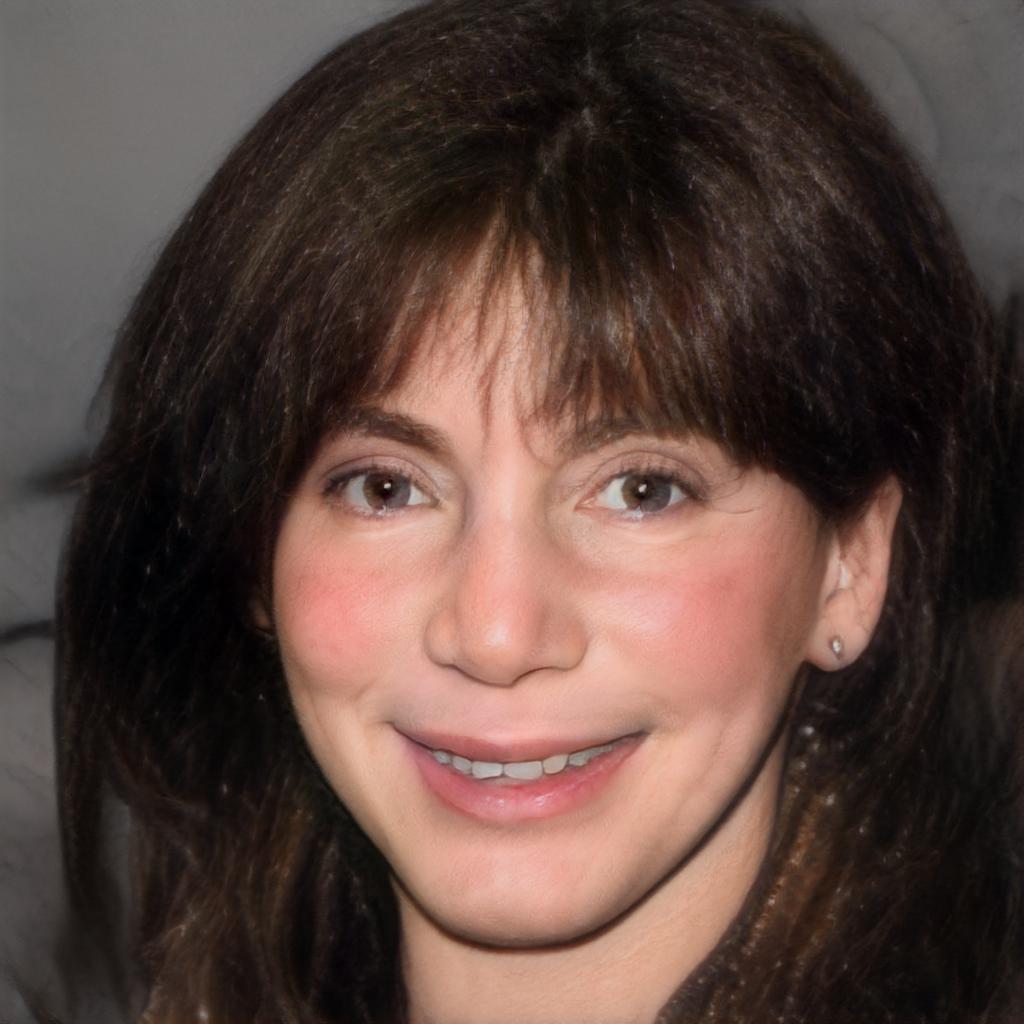} &
 \includegraphics[width=0.13\textwidth, ]{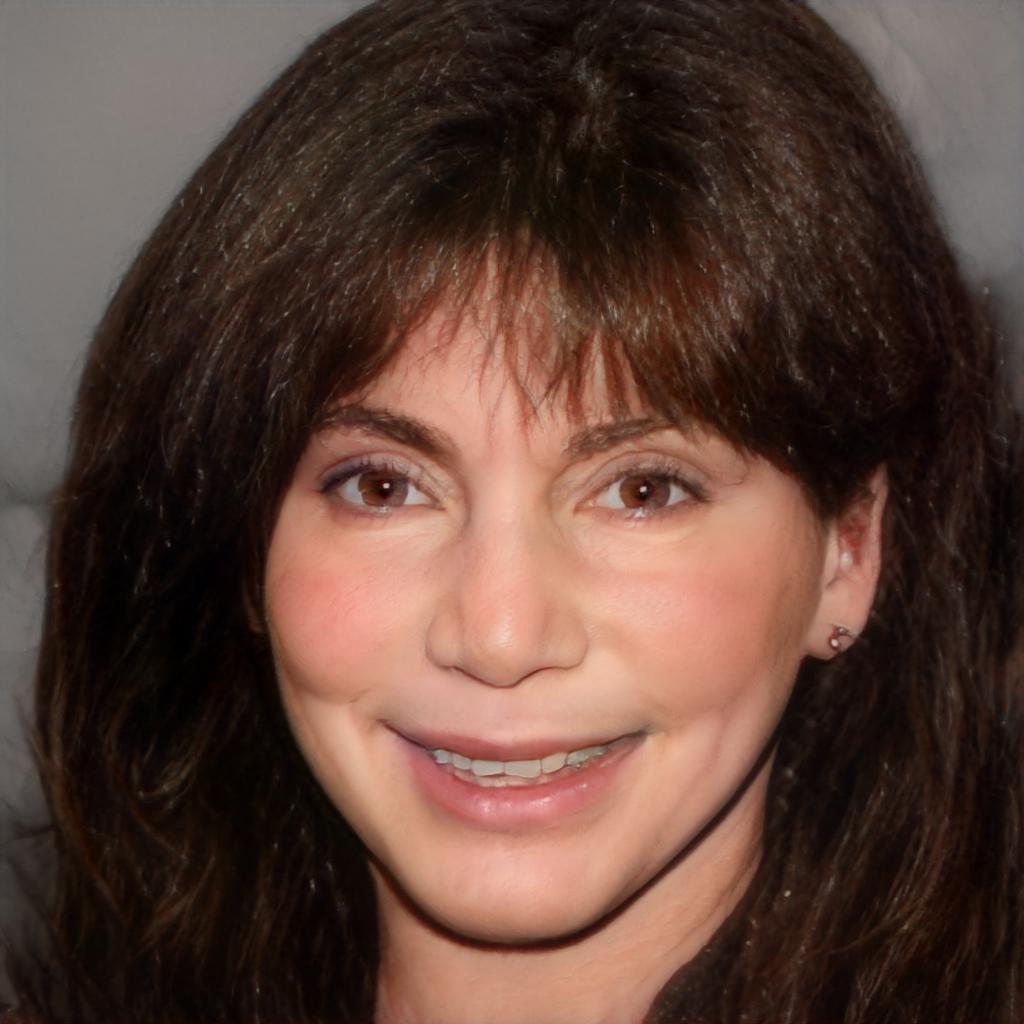} &
 \includegraphics[width=0.13\textwidth, ]{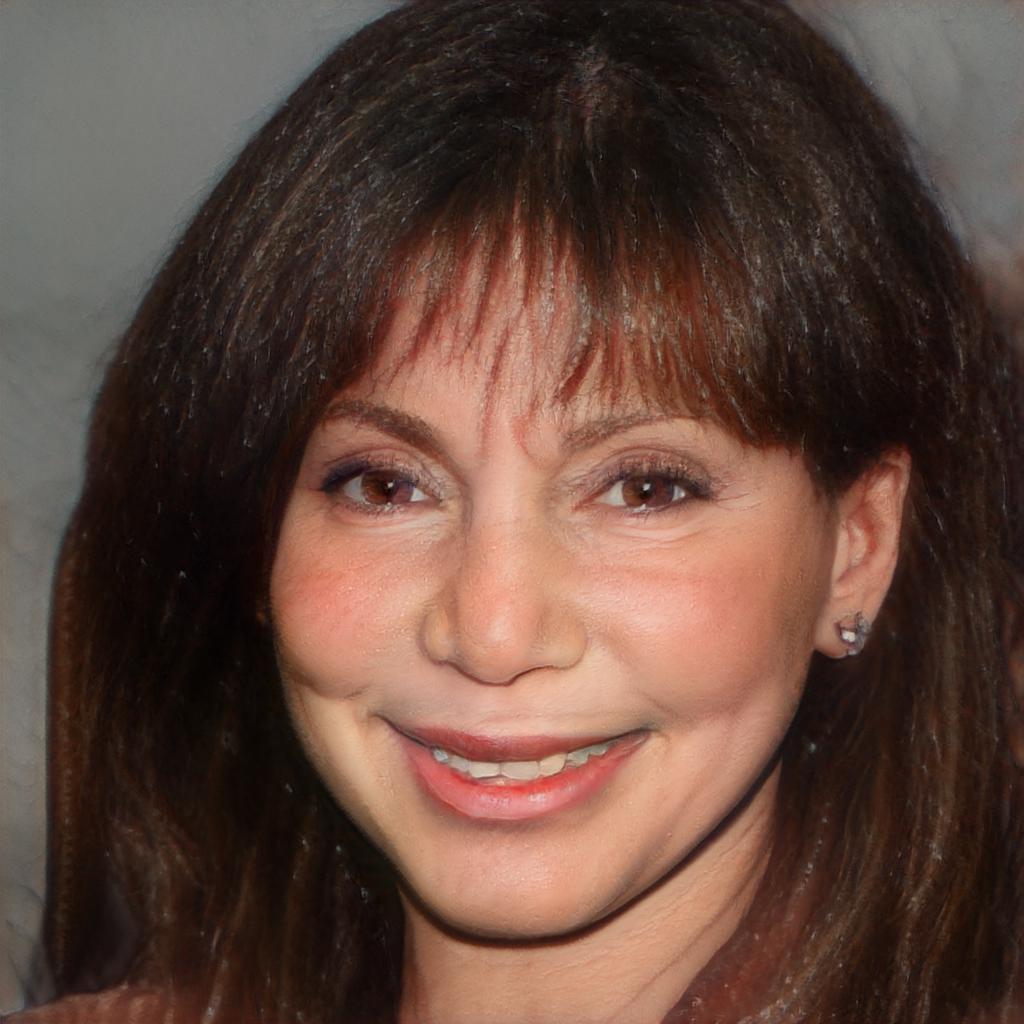} &
 \includegraphics[width=0.13\textwidth, ]{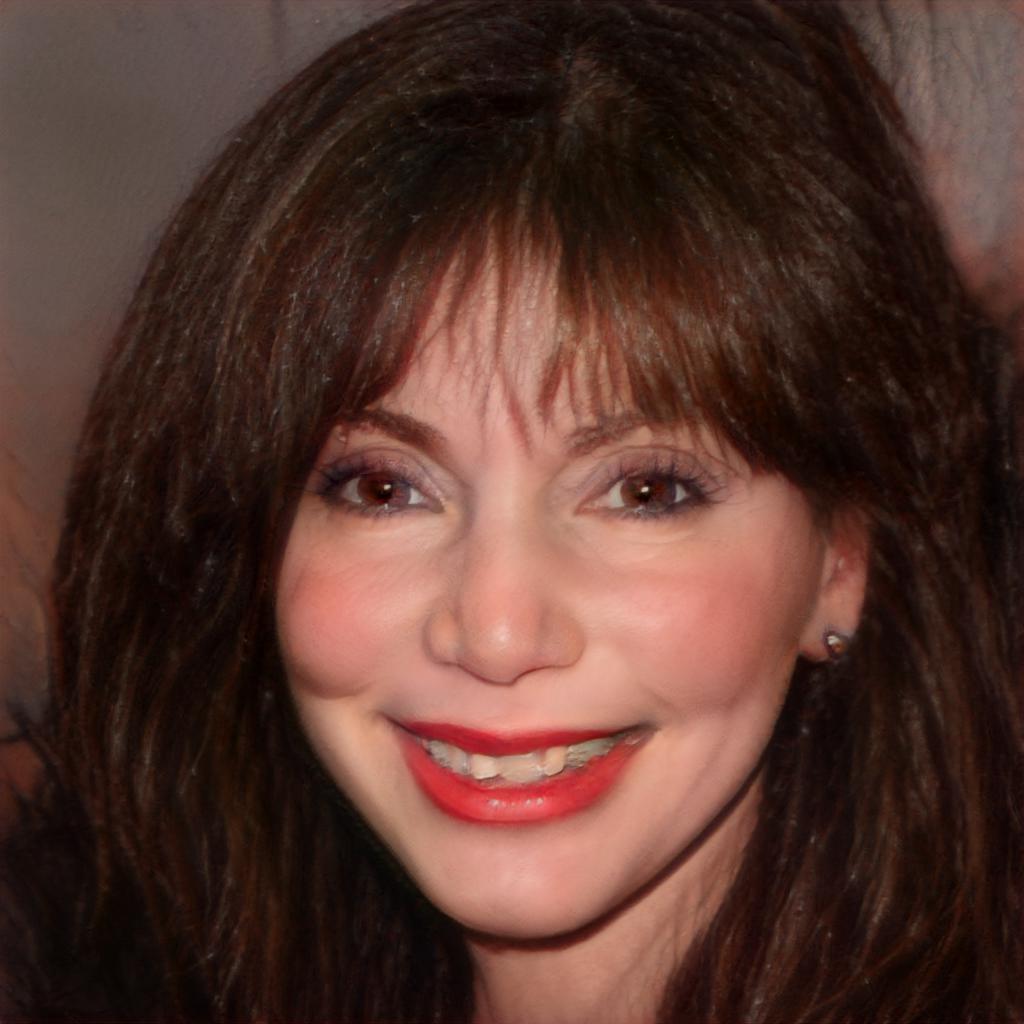} &
 \midrule
 \begin{turn}{90} \hspace{0.5cm} \wplus\end{turn} &
 \includegraphics[width=0.13\textwidth]{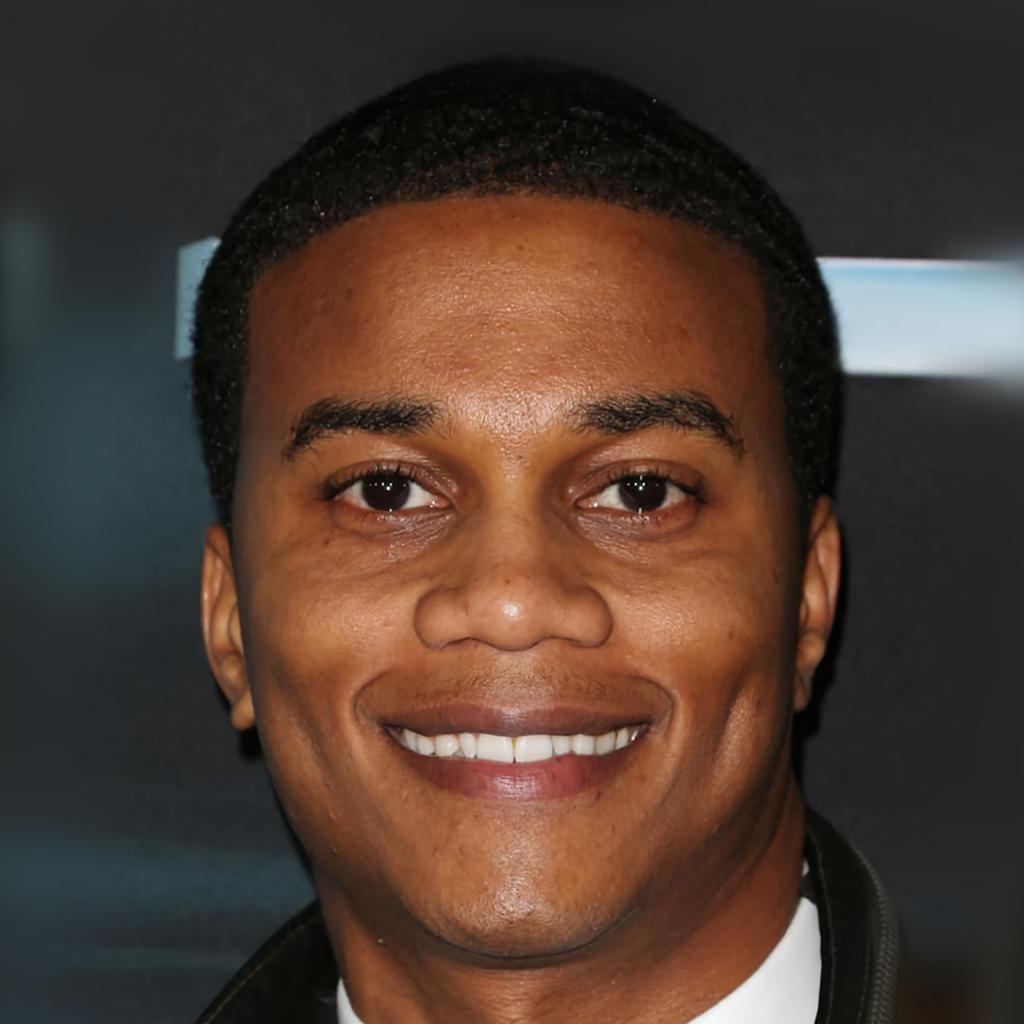} & 
 \includegraphics[width=0.13\textwidth]{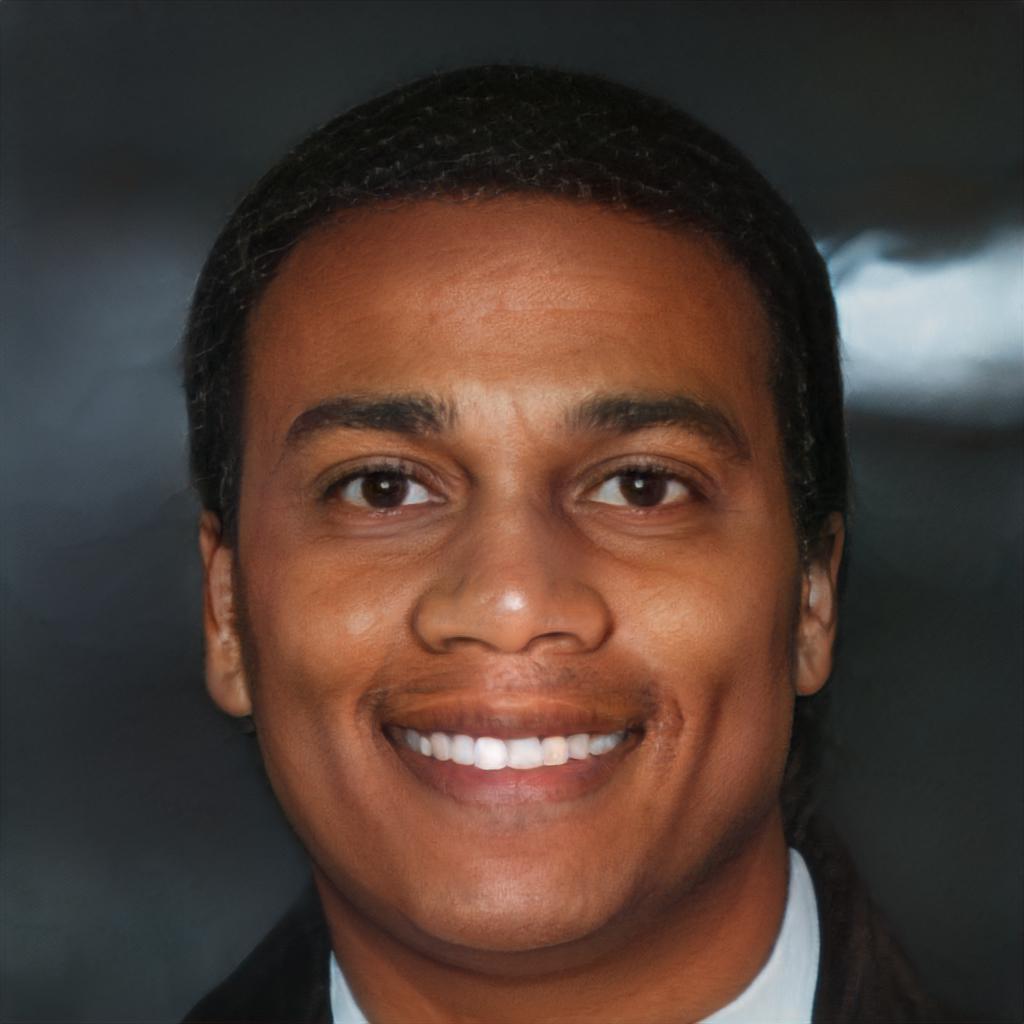} &
 \includegraphics[width=0.13\textwidth, ]{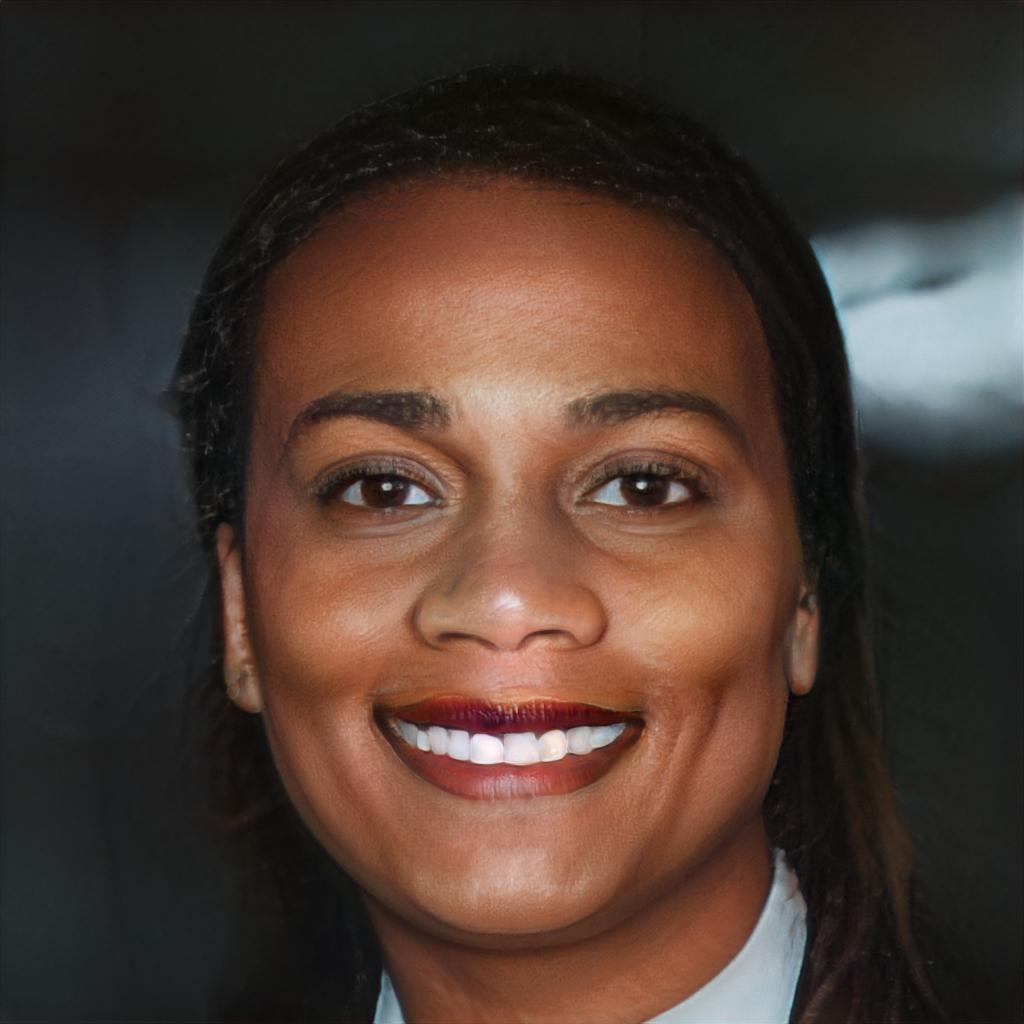} &
 \includegraphics[width=0.13\textwidth, ]{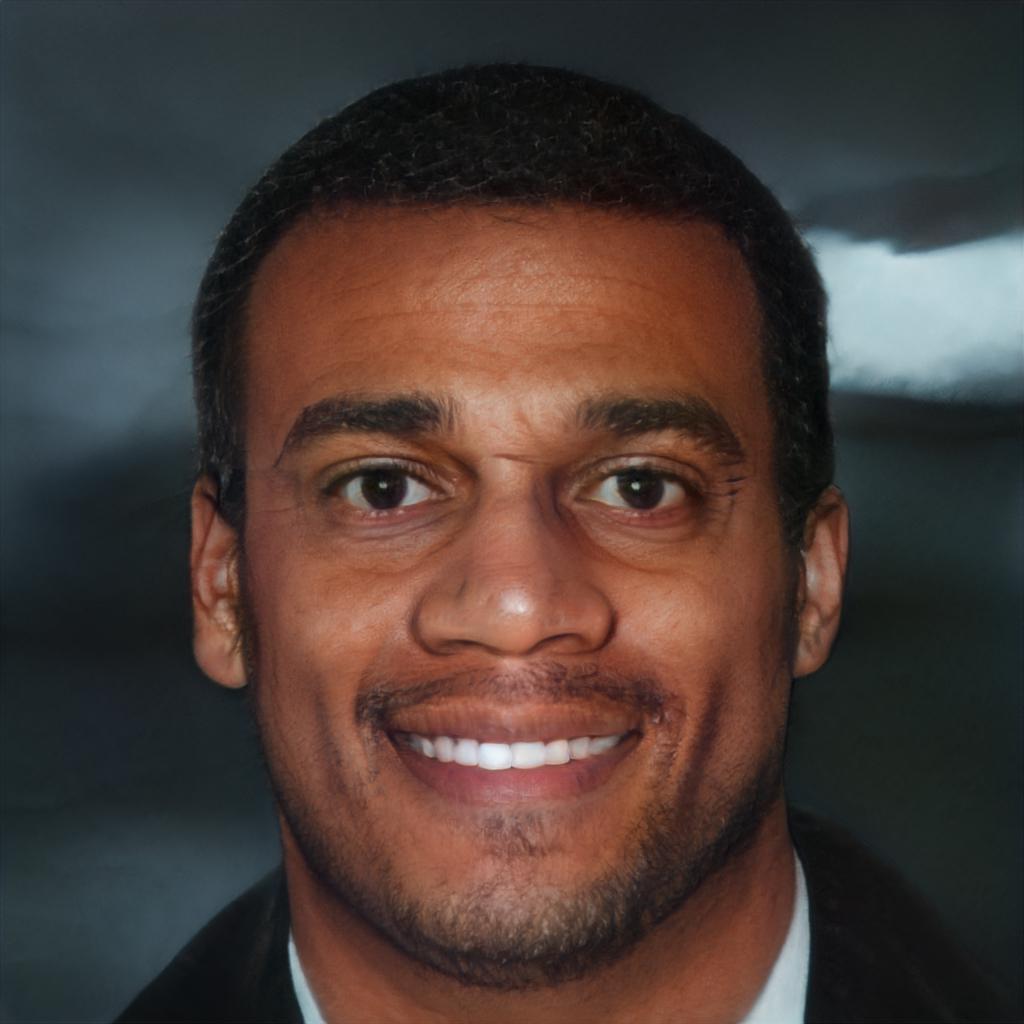} &
 \includegraphics[width=0.13\textwidth, ]{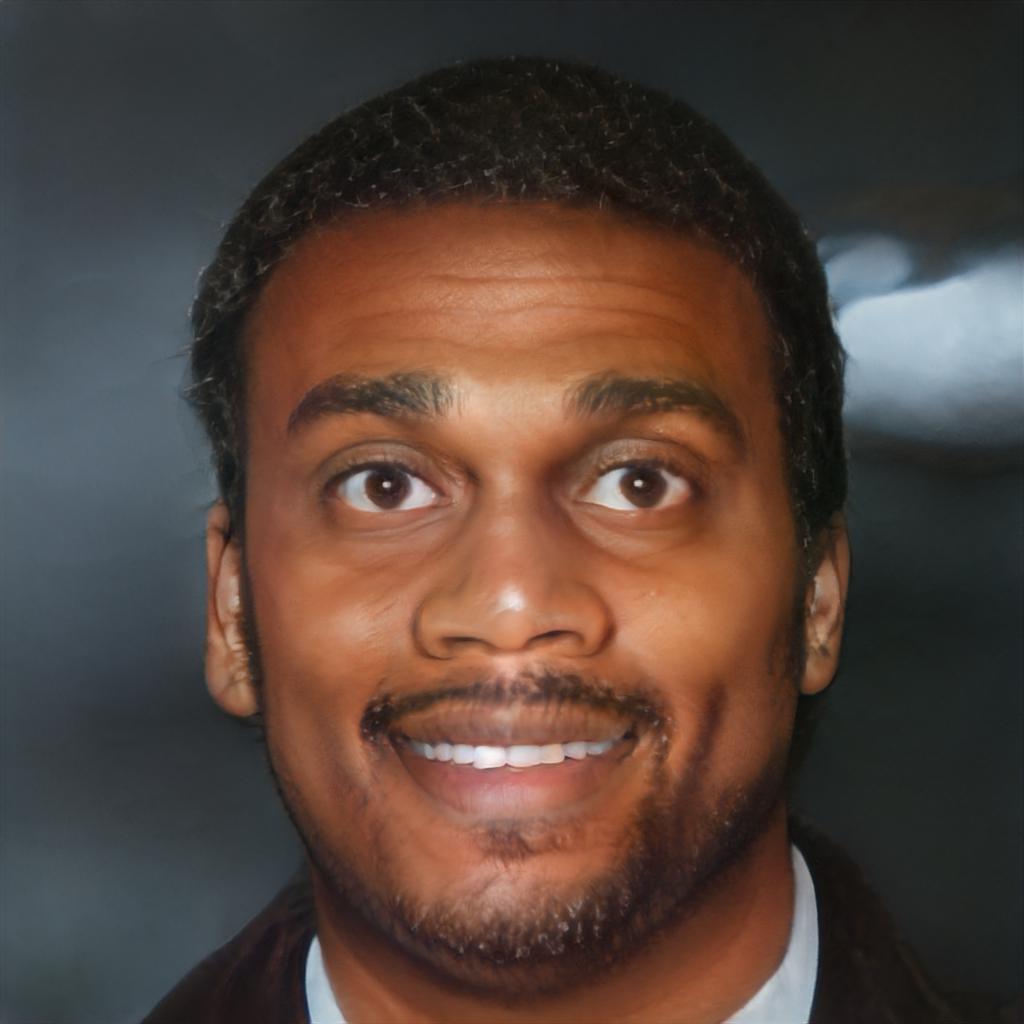} &
 \includegraphics[width=0.13\textwidth, ]{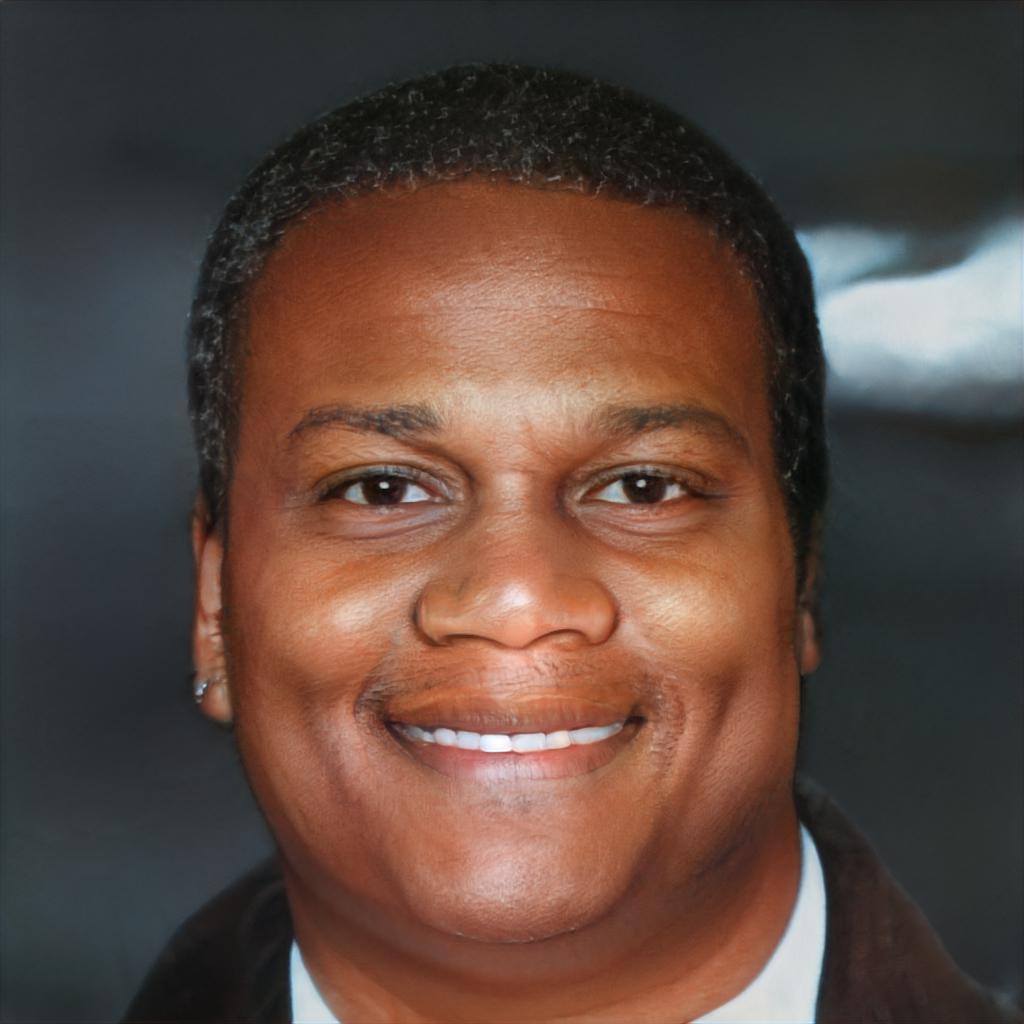} &
 \includegraphics[width=0.13\textwidth]{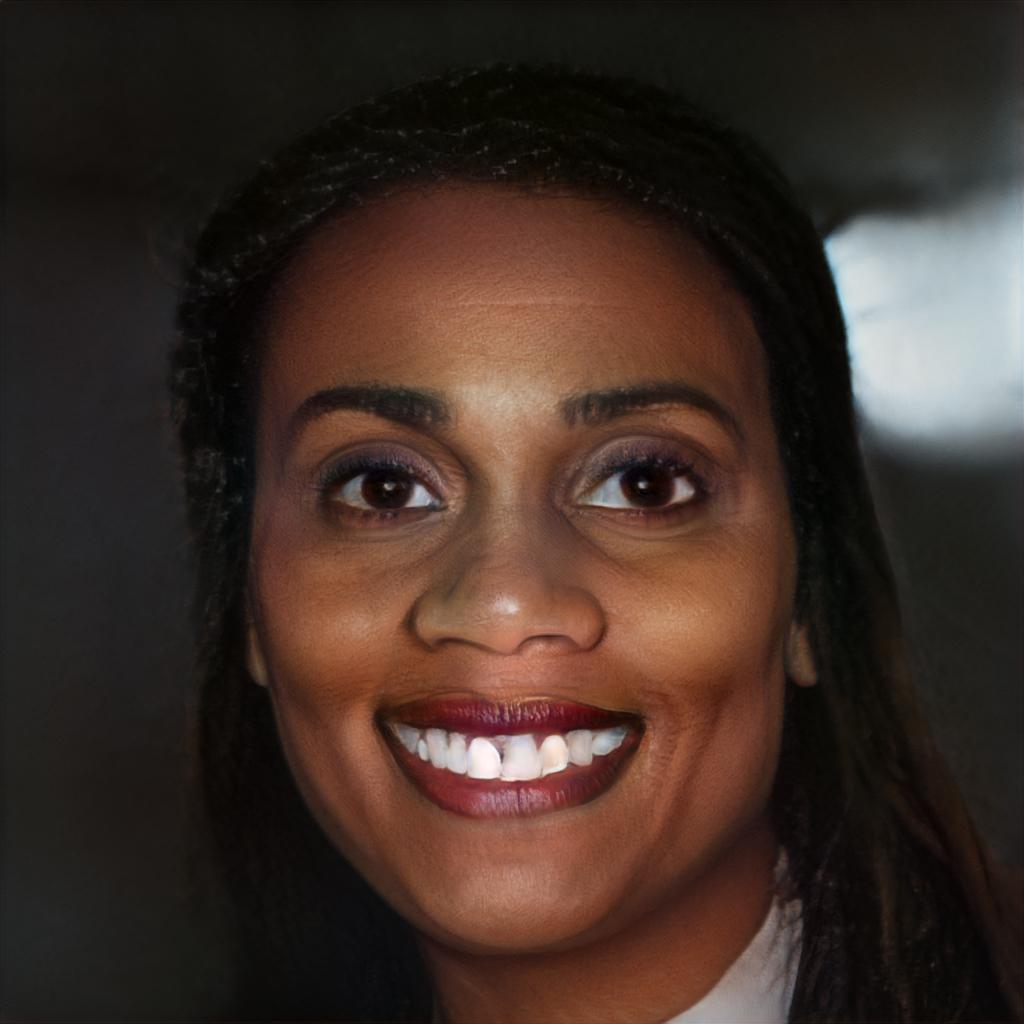} &
\begin{turn}{90} \hspace{0.5cm} $\mathcal{W}^{\star}_{ID}$ \end{turn} &
 \includegraphics[width=0.13\textwidth]{images/original/06019.jpg} & 
 \includegraphics[width=0.13\textwidth]{images/inversion/13_orig_img_20.jpg} &
 \includegraphics[width=0.13\textwidth, ]{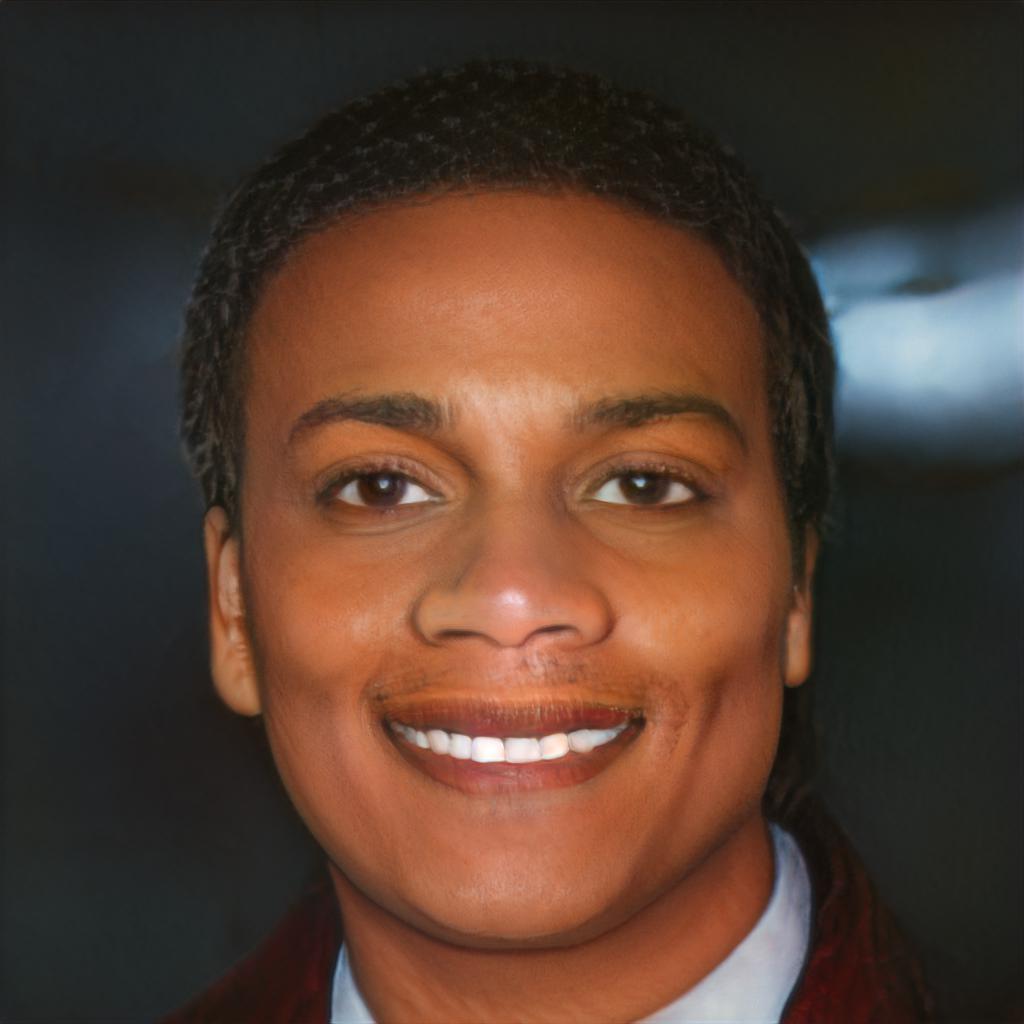} &
 \includegraphics[width=0.13\textwidth, ]{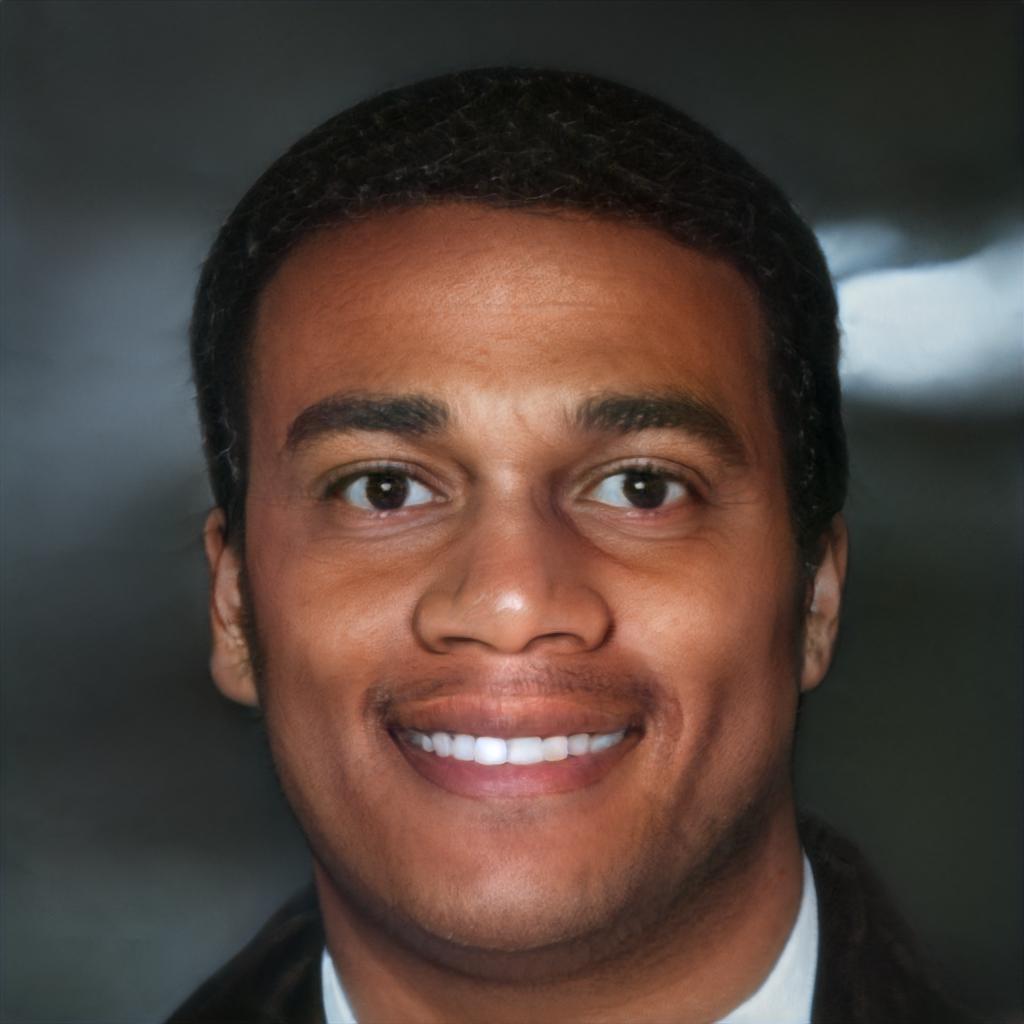} &
 \includegraphics[width=0.13\textwidth, ]{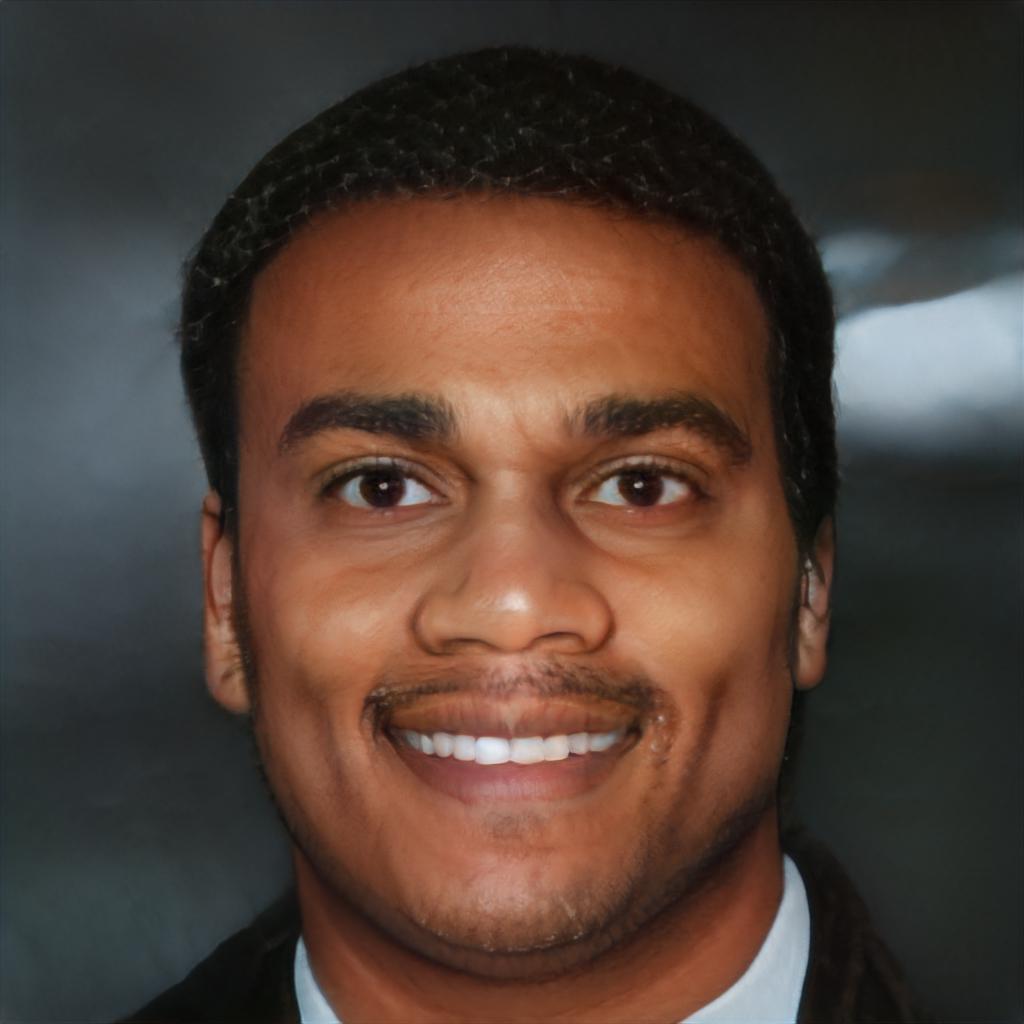} &
 \includegraphics[width=0.13\textwidth, ]{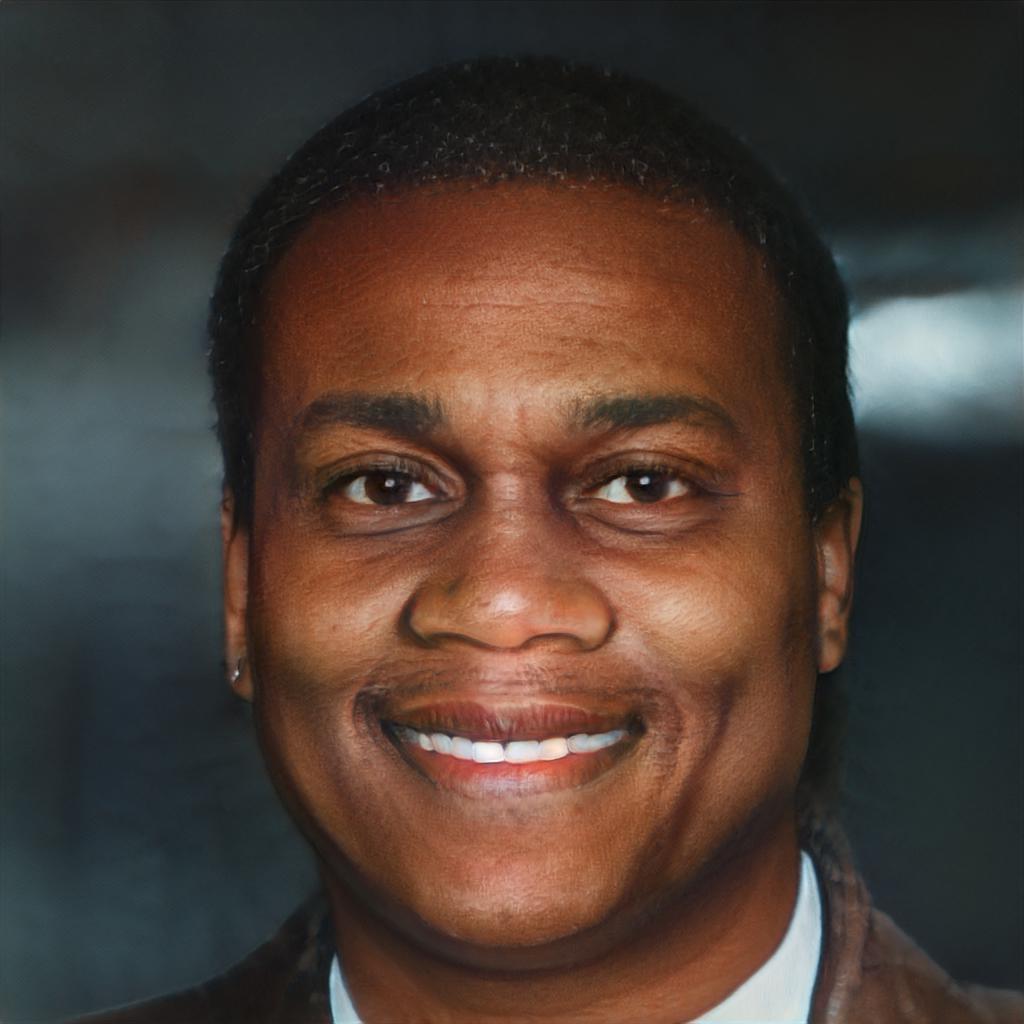} &
 \includegraphics[width=0.13\textwidth, ]{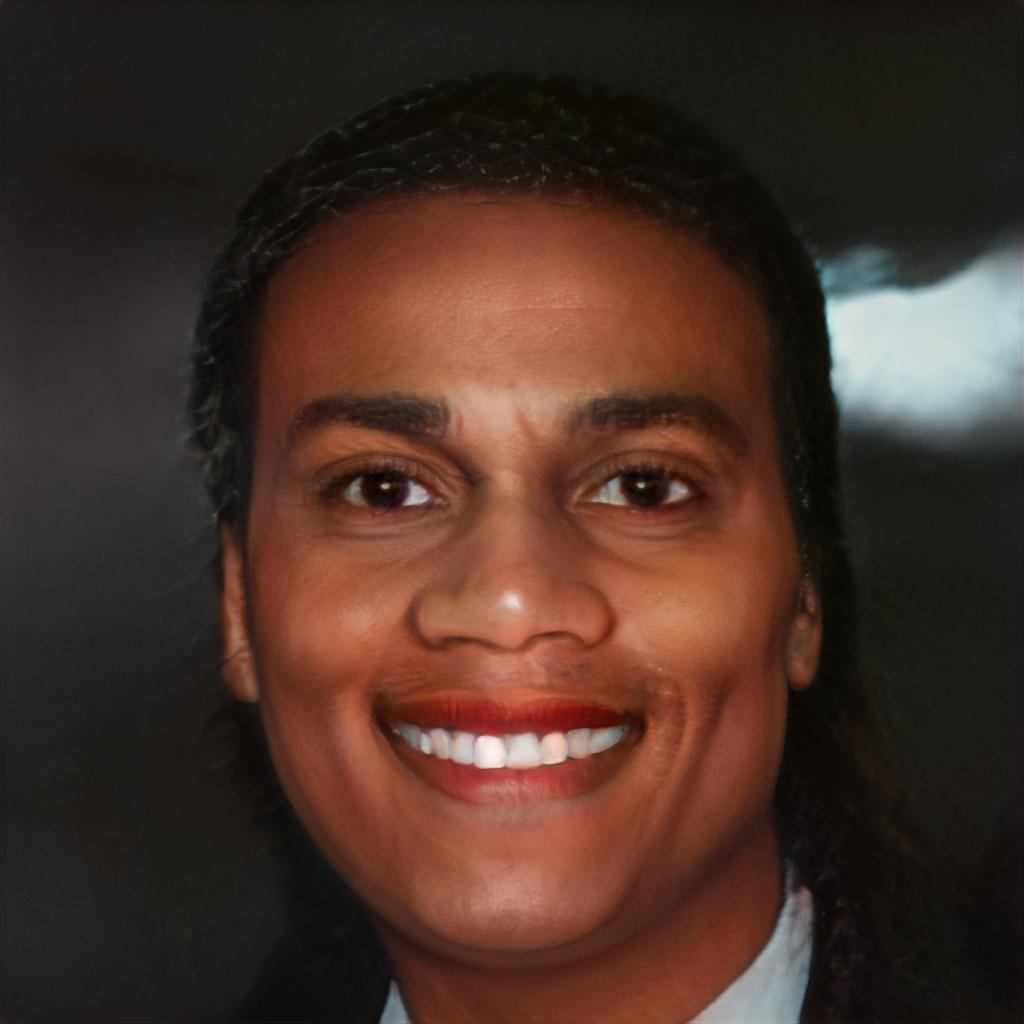} &
 \midrule
 \begin{turn}{90} \hspace{0.5cm} \wplus\end{turn} &
 \includegraphics[width=0.13\textwidth, ]{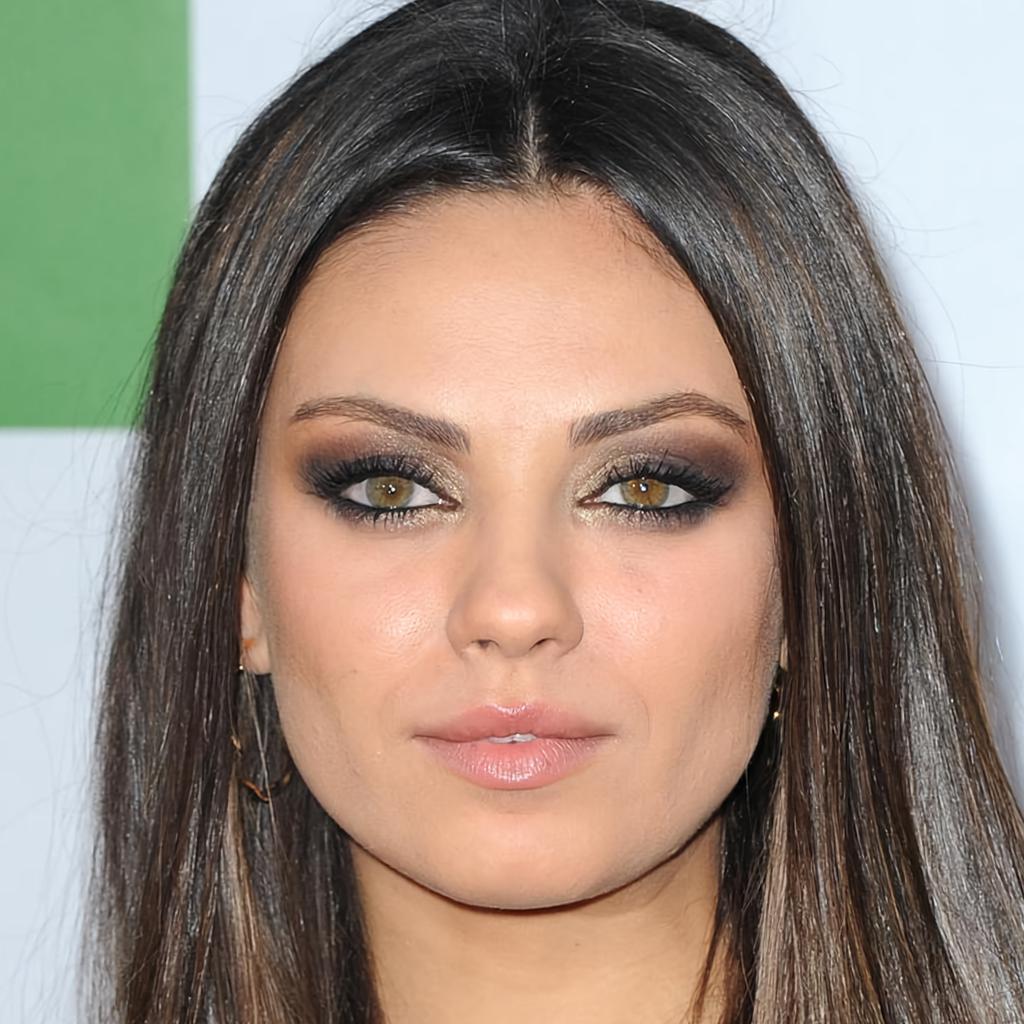} & 
 \includegraphics[width=0.13\textwidth, ]{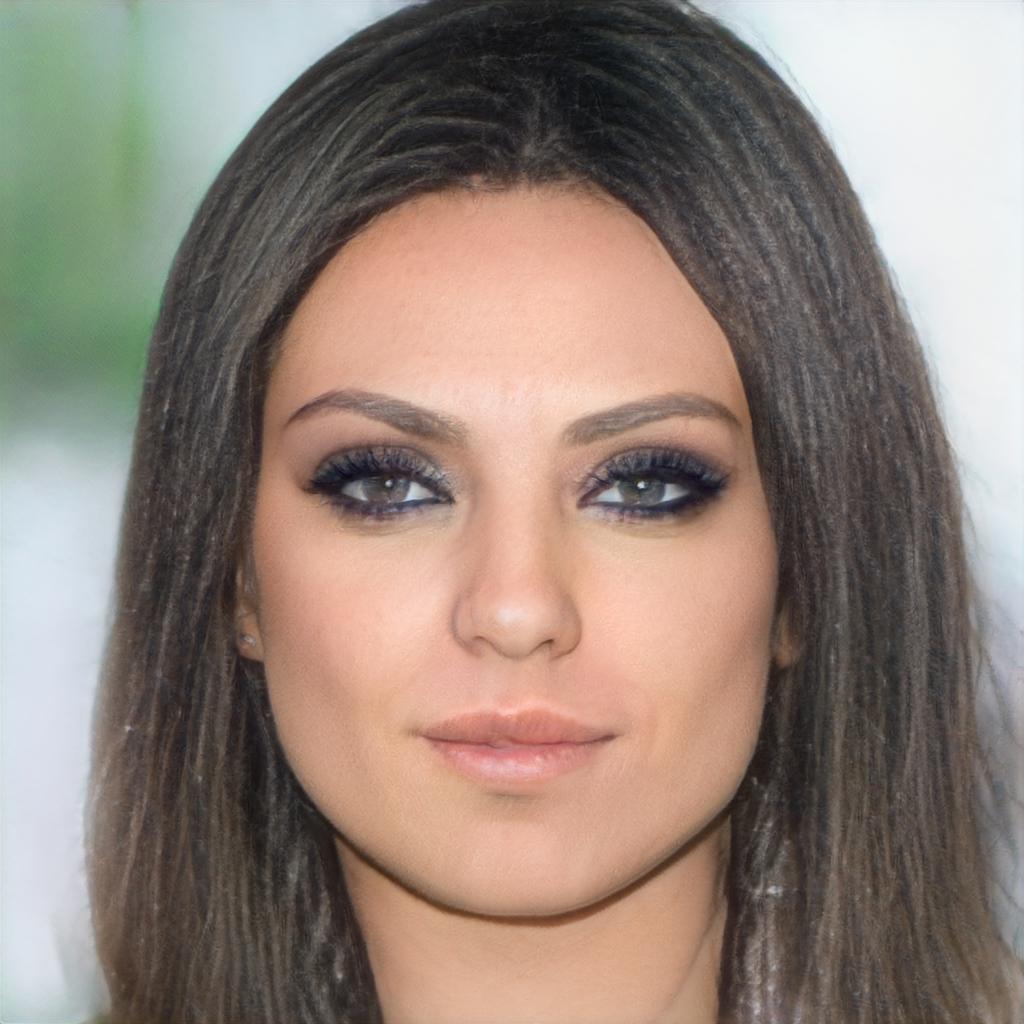} &
 \includegraphics[width=0.13\textwidth, ]{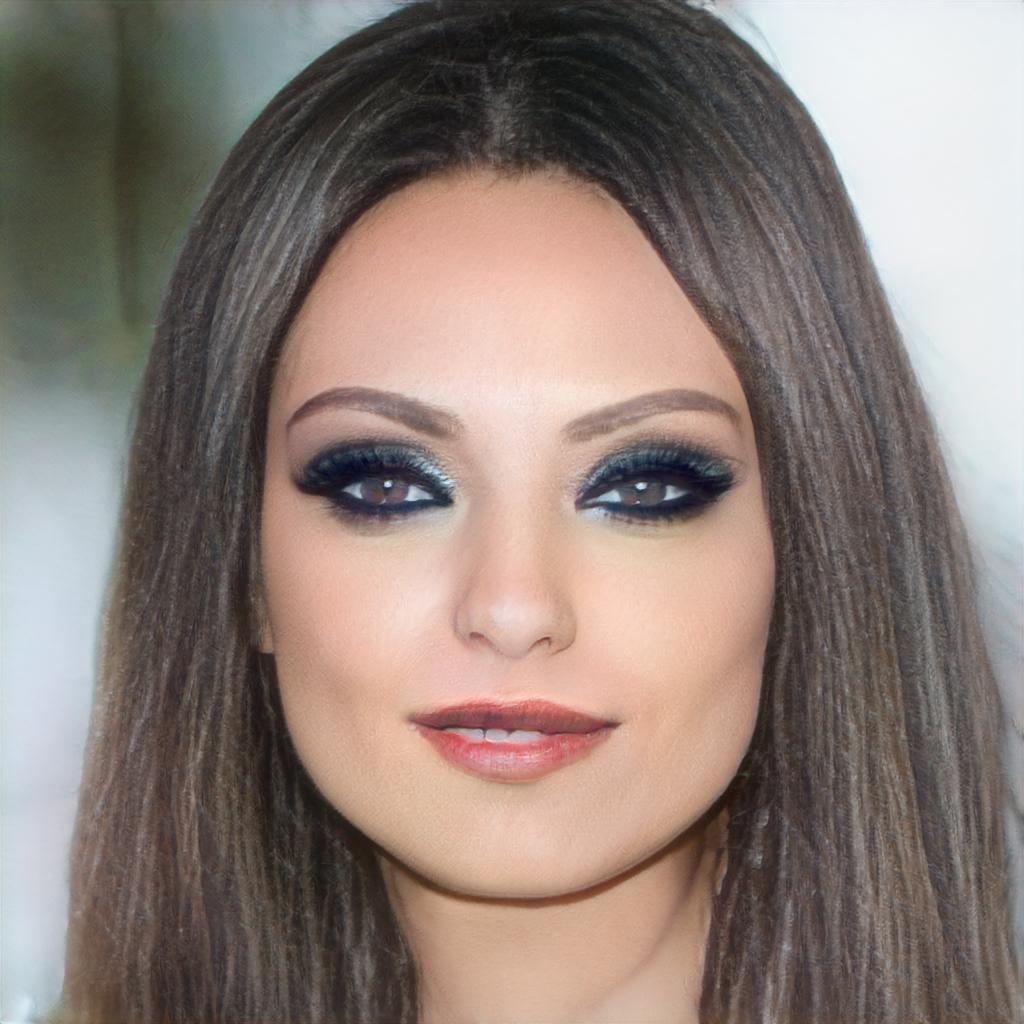} &
 \includegraphics[width=0.13\textwidth, ]{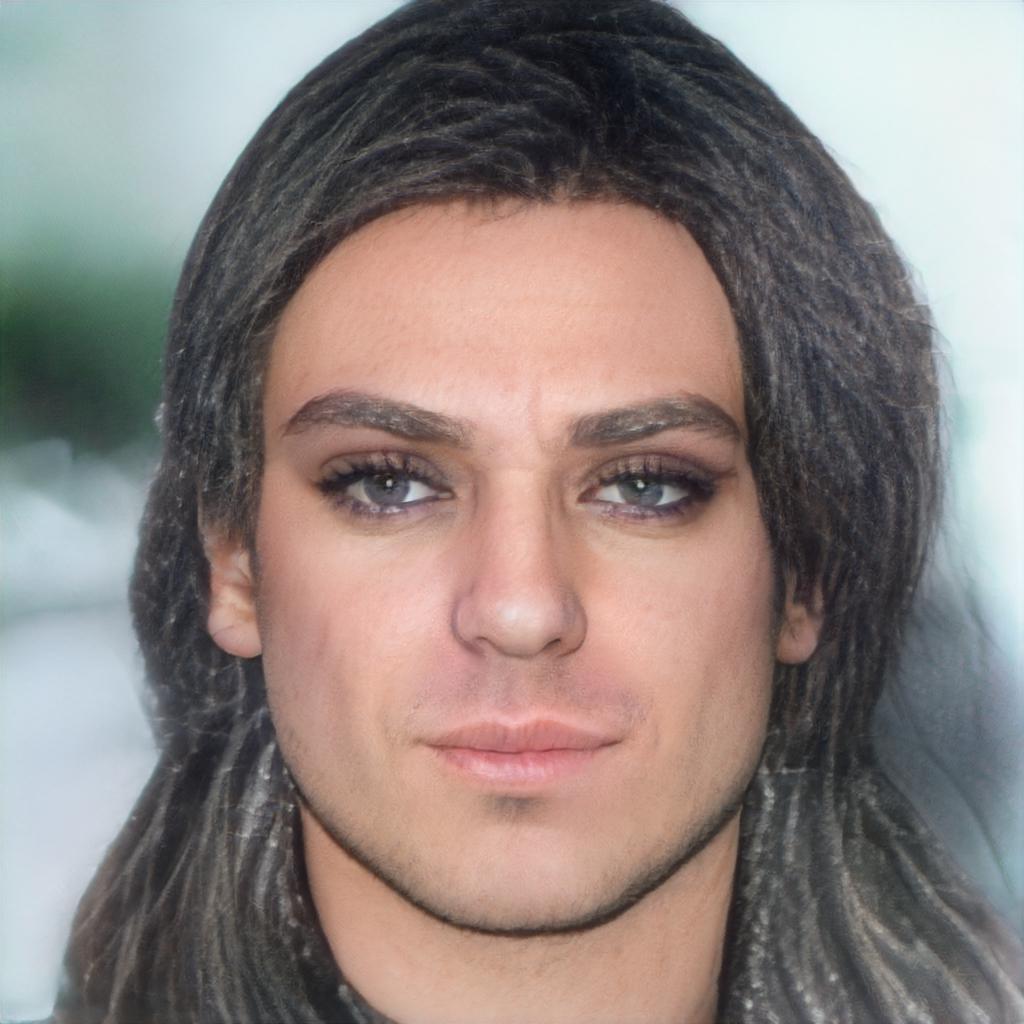} &
 \includegraphics[width=0.13\textwidth, ]{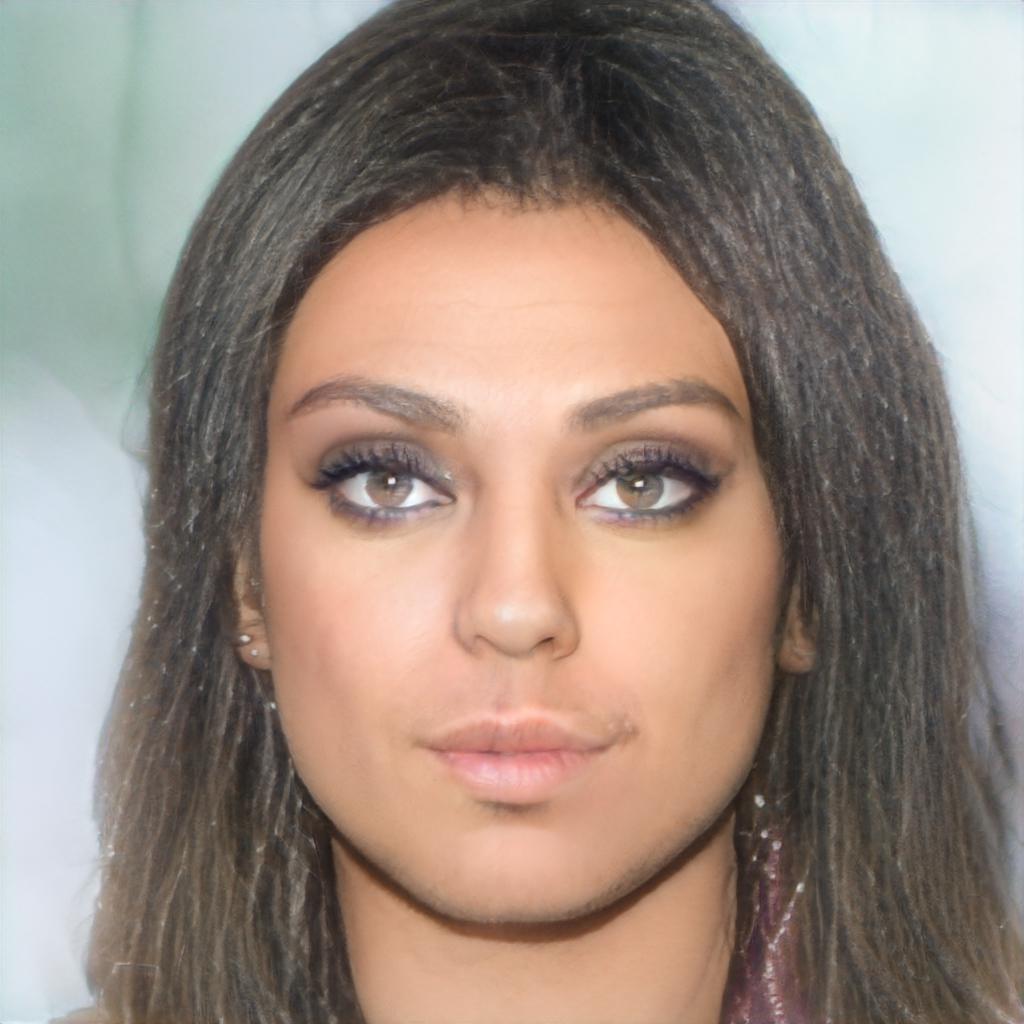} &
 \includegraphics[width=0.13\textwidth, ]{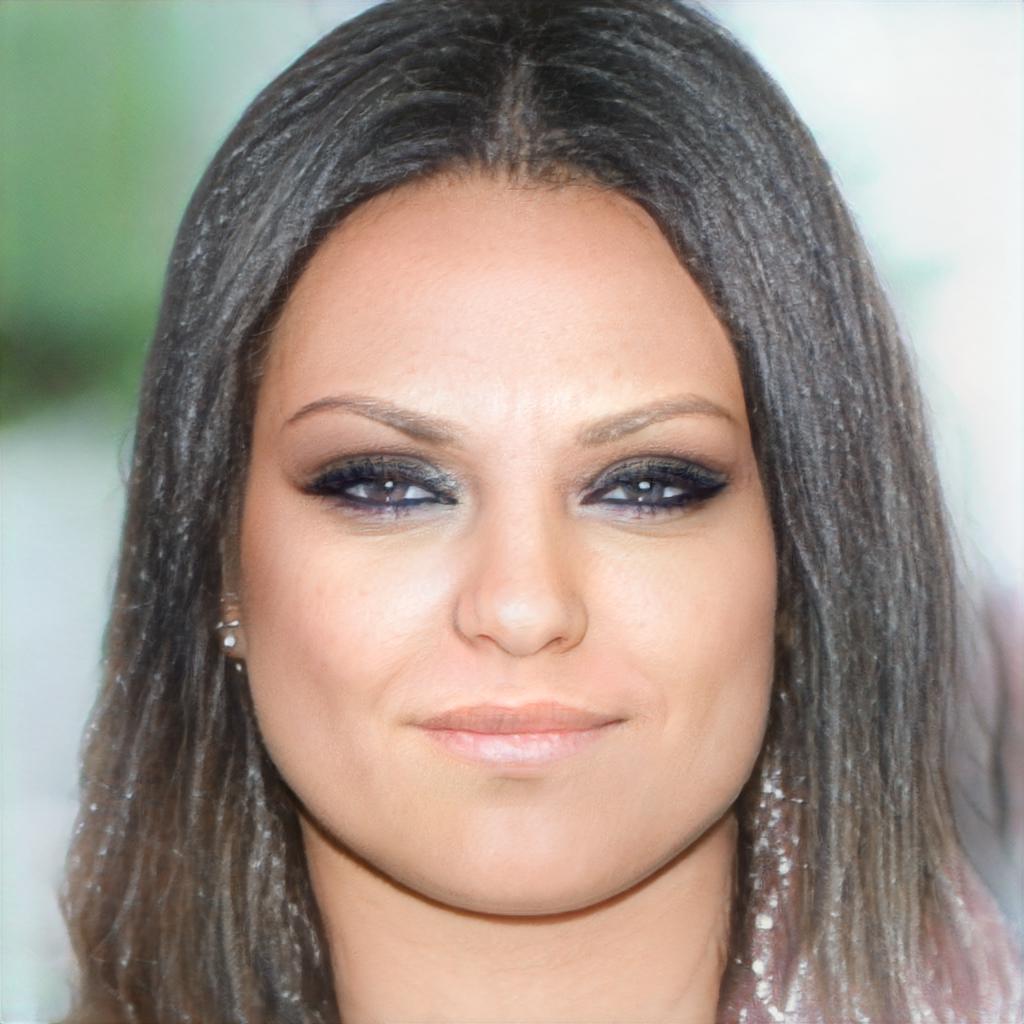} &
 \includegraphics[width=0.13\textwidth]{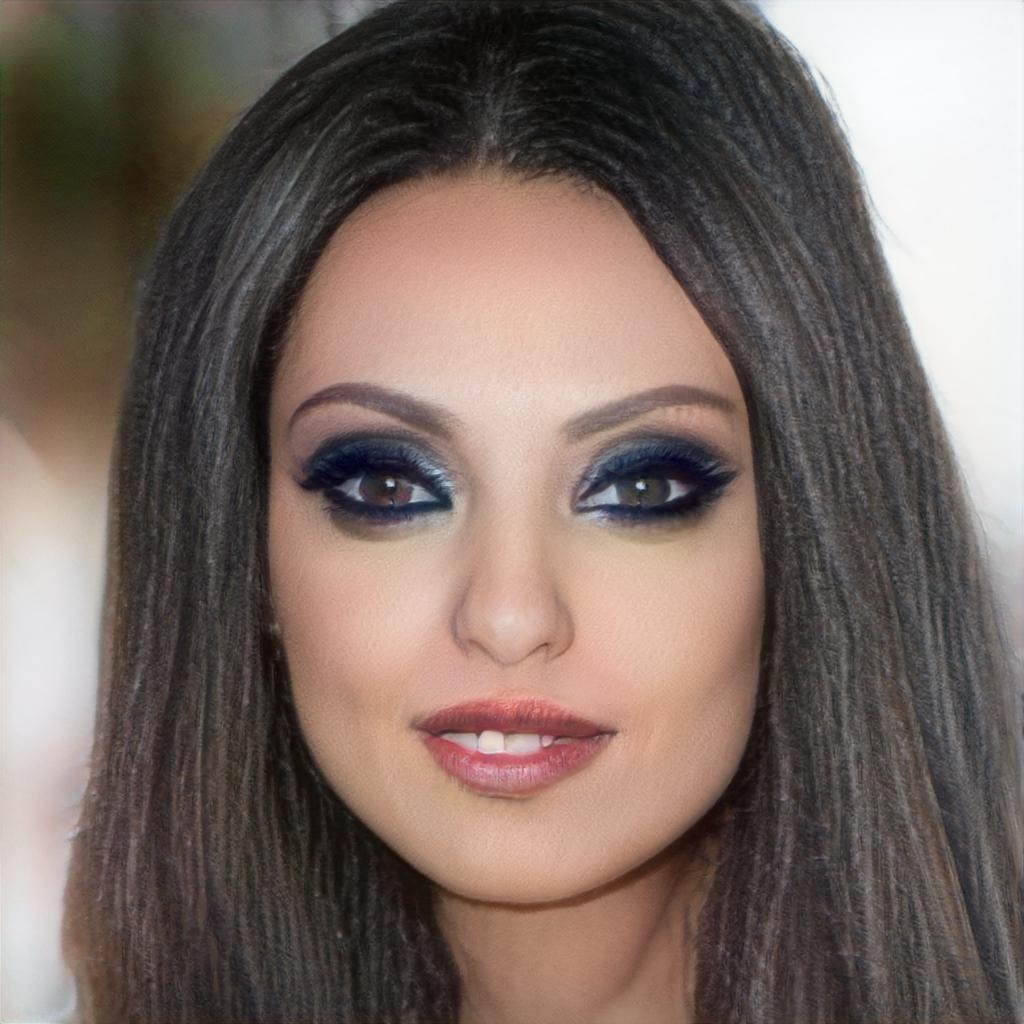} &
 \begin{turn}{90} \hspace{0.5cm} $\mathcal{W}^{\star}_{ID}$ \end{turn} &
 \includegraphics[width=0.13\textwidth, ]{images/original/06026.jpg} & 
 \includegraphics[width=0.13\textwidth, ]{images/inversion/13_orig_img_27.jpg} &
 \includegraphics[width=0.13\textwidth,  ]{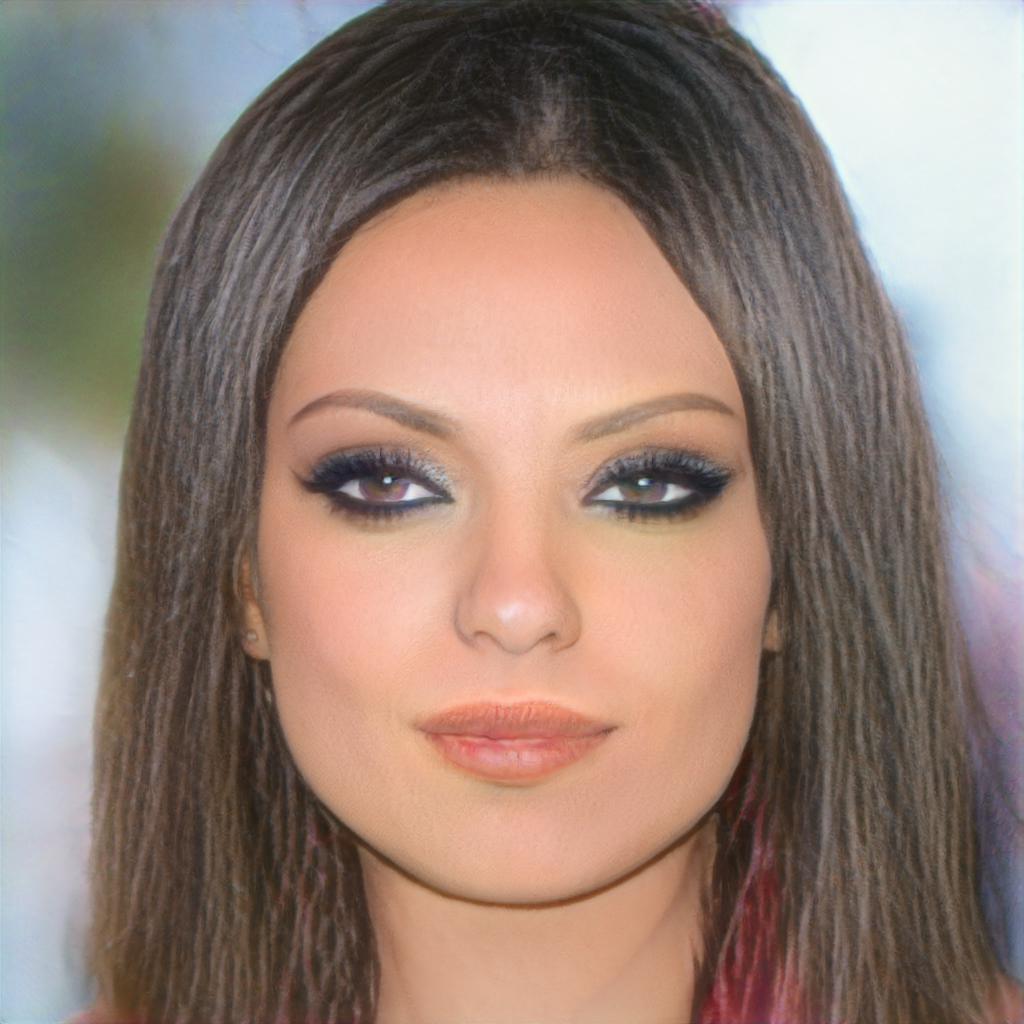} &
 \includegraphics[width=0.13\textwidth, ]{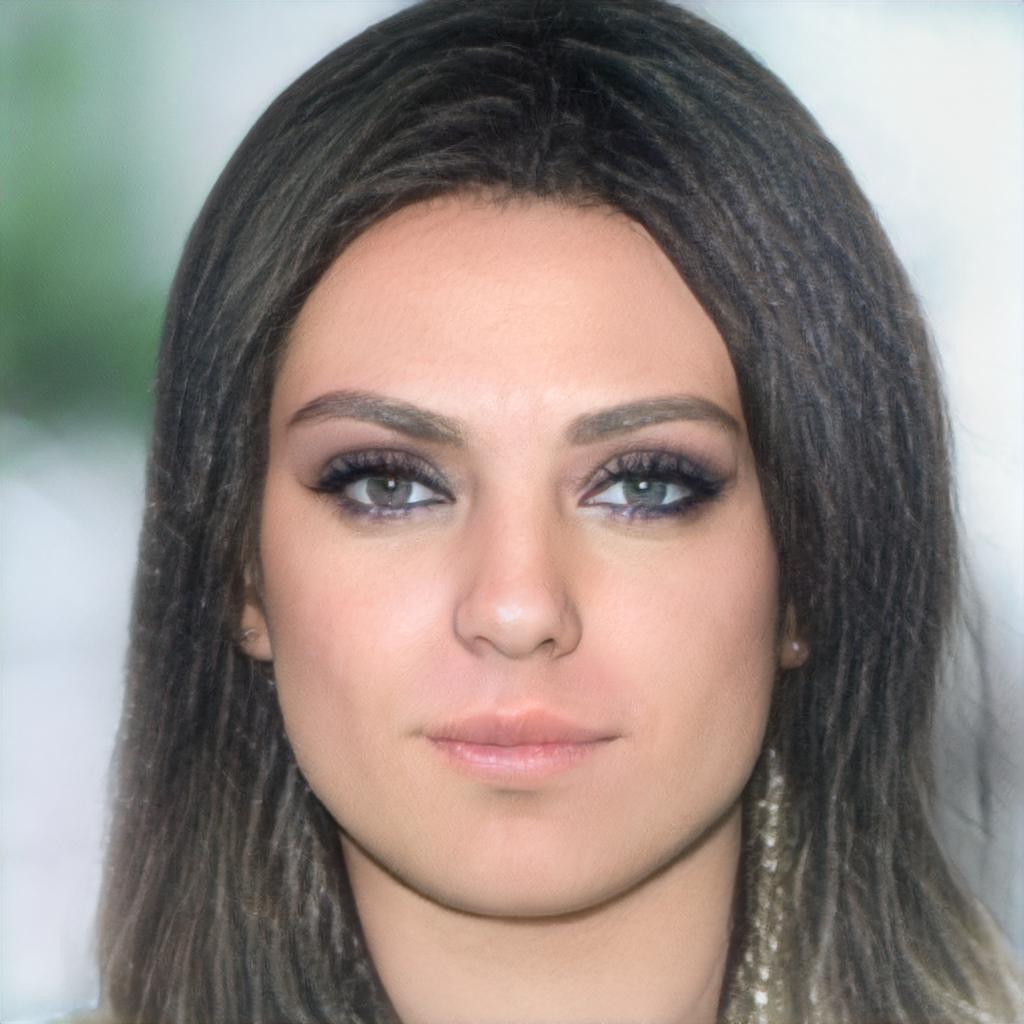} &
 \includegraphics[width=0.13\textwidth, ]{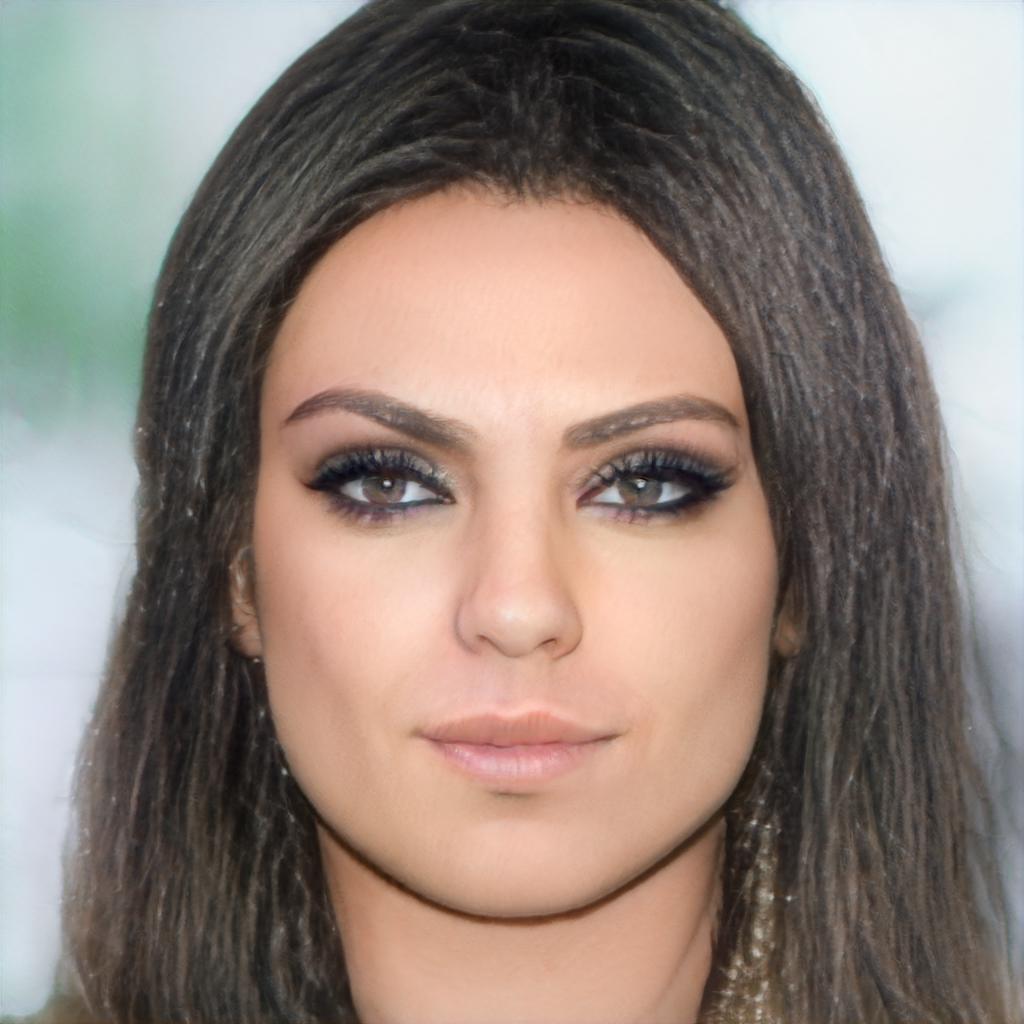} &
 \includegraphics[width=0.13\textwidth, ]{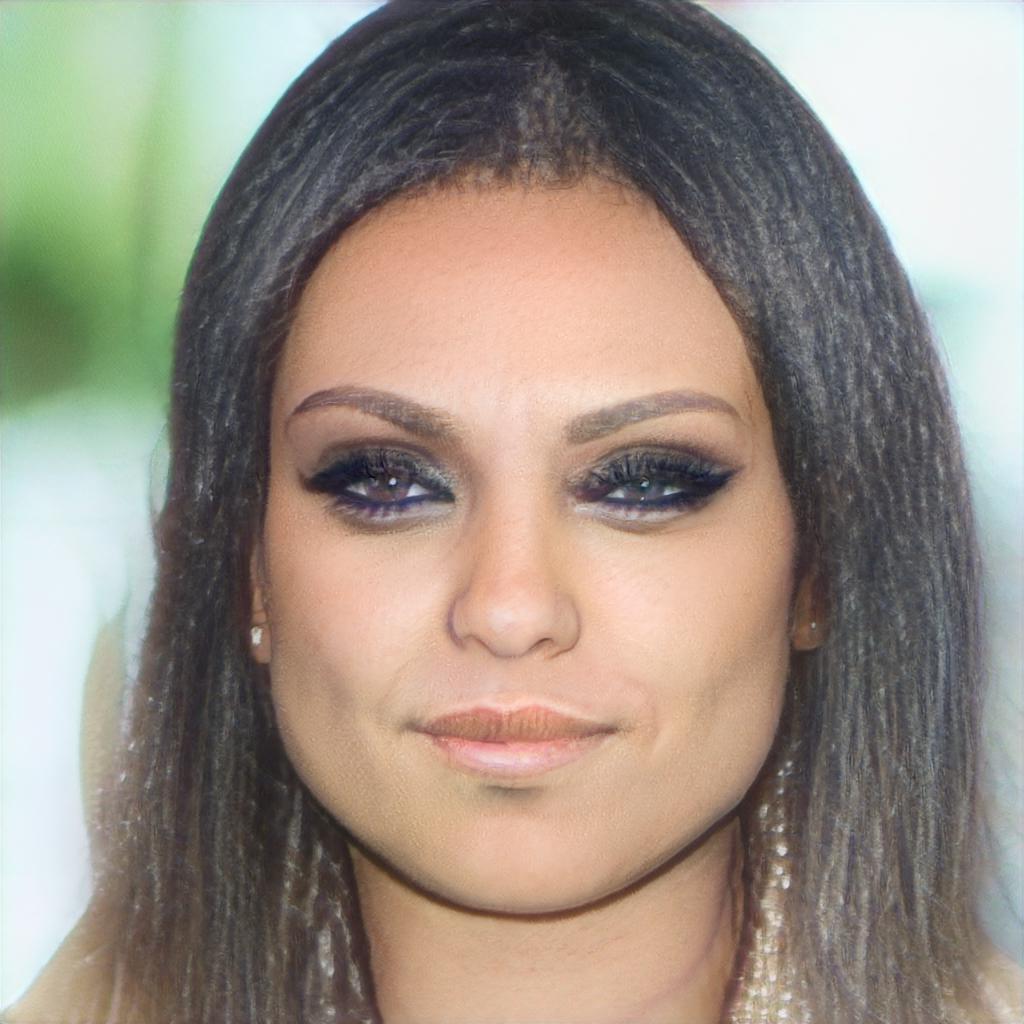} &
 \includegraphics[width=0.13\textwidth, ]{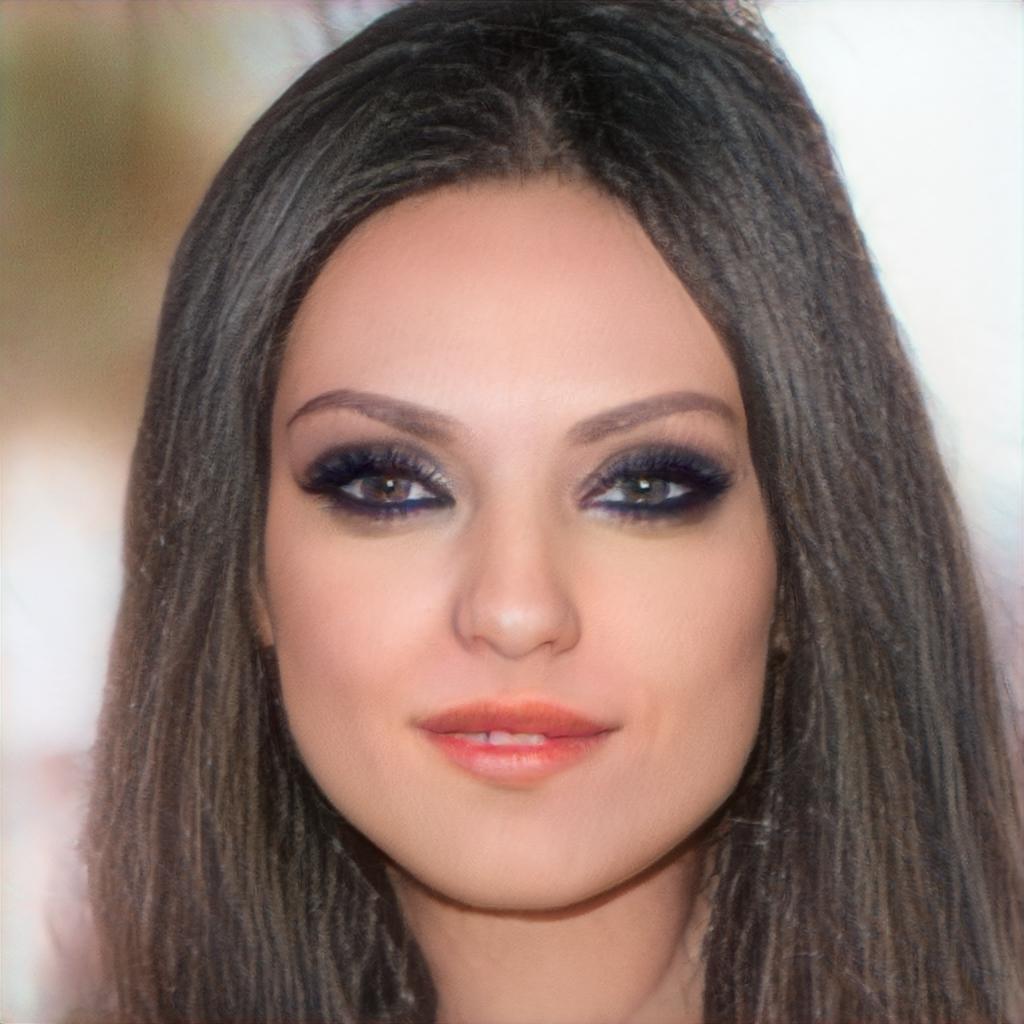} &
\\
\end{tabular}
\caption{Image editing at resolution $1024^2$ using InterFaceGAN in \wplus  and $\mathcal{W}^{\star}_{ID}$: The images are projected in the latent space of StyleGAN2, then the latent codes are moved in the direction that corresponds to changing one facial attribute. The editing is better in terms of attributes disentanglement in the new learned space ($\mathcal{W}^{\star}_{ID}$) compared to the editing done in the original latent space (\wplussp). Our method also leads to a better identity preservation.}
\label{fig:att_inter_results2}

\end{figure*}

\bibliographystyle{IEEEbib}
\bibliography{egbib}